\documentclass[12pt]{article}
\usepackage{graphicx}
\usepackage{setspace}
\usepackage{bigints}
\usepackage{graphics} 
\usepackage{epstopdf}
\usepackage{epsfig} 
\usepackage{mathptmx} 
\usepackage{times} 
\usepackage{amsmath} 
\usepackage{amsthm}
\usepackage{amssymb}  
\usepackage{algorithmicx}
\usepackage{algorithm}
\usepackage[noend]{algpseudocode}
\usepackage{cite}
\usepackage{subfig}
\usepackage{gensymb}
\usepackage{booktabs}
\usepackage{url}

\usepackage{tikz,pgfplots}
\usetikzlibrary{external}
\usepgfplotslibrary{external} 
\pgfplotsset{compat=newest} 
\pgfplotsset{plot coordinates/math parser=false} 

\DeclareMathOperator*{\minimize}{minimize}
\DeclareMathOperator*{\argmin}{\arg\min}
\DeclareMathOperator*{\subjectto}{subject\:to}
\DeclareMathOperator*{\sign}{sign}

\newtheorem{theorem}{Theorem}
\newtheorem{assumption}{Assumption}
\newtheorem{definition}{Definition}

\newtheorem{corollary}{Corollary}
\newtheorem{lemma}{Lemma}

\DeclareMathAlphabet{\pazocal}{OMS}{zplm}{m}{n}

\newlength\figureheight 
\newlength\figurewidth  

\usepackage{algcompatible}
\usepackage{algpseudocode}

\makeatletter
\renewcommand*{\ALG@name}{Workflow}
\makeatother

\title{A path planning and path-following control framework for a general 2-trailer with a car-like tractor}

\author
{\small Oskar Ljungqvist$^{\dagger*}$, Niclas Evestedt$^{\ddagger}$, Daniel Axehill$^{\dagger}$, Marcello Cirillo$^{\diamond}$ and Henrik Pettersson$^{\diamond}$\\
\\
\small{$^{\dagger}$Department of Automatic Control, Link\"oping University, Link\"oping, Sweden}\\
\small{$^{\ddagger}$Embark Trucks Inc. San Francisco, USA} \\
\small{$^{\diamond}$Autonomous Transport Solutions, Scania CV, S\"odert\"alje, Sweden}\\
\small{$^*$E-mail: \texttt{oskar.ljungqvist@liu.se}}
}

\date{}

\begin{document}

\baselineskip 16pt

\maketitle 

\begin{abstract}
Maneuvering a general 2-trailer with a car-like tractor in backward motion is a task that requires significant skill to master and is unarguably one of the most complicated tasks a truck driver has to perform. This paper presents a path planning and path-following control solution that can be used to automatically plan and execute difficult parking and obstacle avoidance maneuvers by combining backward and forward motion. A lattice-based path planning framework is developed in order to generate kinematically feasible and collision-free paths and a path-following controller is designed to stabilize the lateral and angular path-following error states during path execution. To estimate the vehicle state needed for control, a nonlinear observer is developed which only utilizes information from sensors that are mounted on the car-like tractor, making the system independent of additional trailer sensors. The proposed path planning and path-following control framework is implemented on a full-scale test vehicle and results from simulations and real-world experiments are presented.  
\end{abstract}

\section{Introduction}
\begin{figure}[t]
	\centering
	\includegraphics[width=1\linewidth]{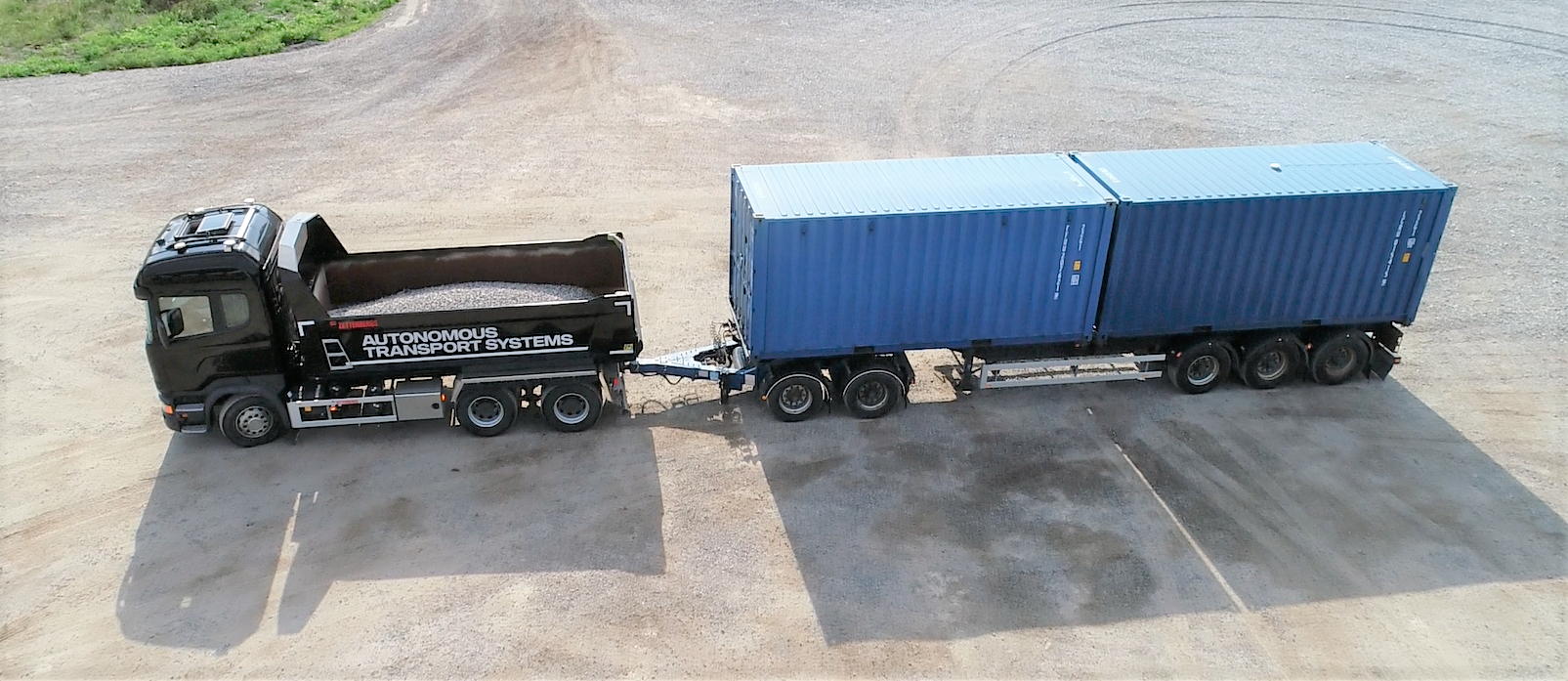}   
	\caption{The full-scale test vehicle that is used as a research platform. The car-like tractor is a modified version of a Scania R580 6x4 tractor.} 
	\label{j1:fig:truck_scania} 
\end{figure}

A massive interest for intelligent and fully autonomous transport solutions has been seen from industry over the past years as technology in this area has advanced. 
The predicted productivity gains and the relatively simple implementation have made controlled environments such as mines, harbors, airports, etc., interesting areas for commercial launch of such systems. 
In many of these applications, tractor-trailer systems are used for transportation and therefore require fully automated control.    
Reversing a semitrailer with a car-like tractor is known to be a task that require lots of training to perfect and an inexperienced driver usually encounter problems already when performing simple tasks, such as reversing straight backwards. 
To help the driver in such situations, trailer assist systems have been developed and released to the passenger car market~\cite{werling2014reversing, hafner2017control}. 
These systems enable the driver to easily control the semitrailer's curvature though a control knob. An even greater challenge arise when reversing a general 2-trailer (G2T) with a car-like tractor.  As seen in Figure~\ref{j1:fig:truck_scania}, this system is composed of three interconnected vehicle segments; a front-wheel steered tractor, an off-axle hitched dolly and an on-axle hitched semitrailer. The word general refers to that the connection between the vehicle segments are of mixed hitching types~\cite{altafini1998general}.  

Compared to a single semitrailer, the dolly introduces an additional degree of freedom into the system, making it very difficult to stabilize the semitrailer and the joint angles in backward motion. A daily challenge that many truck drivers encounter is to perform a reverse maneuver in, \textit{e.g.}, a parking lot or a loading/off-loading site. 
In such scenarios, the vehicle is said to operate in an unstructured environment because no clear driving path is available. 
To perform a parking maneuver, the driver typically needs to plan the maneuver multiple steps ahead, which often involves a combination of driving forwards and backwards. 
For an inexperienced driver, these maneuvers can be both time-consuming and mentally exhausting. 
To aid the driver in such situations, this work presents a motion planning and path-following control framework for a G2T with a car-like tractor that is targeting unstructured environments. 
It is shown through several experiments that the framework can be used to automatically perform advanced maneuvers in different environments.  

The framework can be used as a driver assist system to relieve the driver from performing complex tasks or as part of a motion planning and feedback control layer within an autonomous system architecture. 
The motion planner is based on the state-lattice motion planning framework~\cite{Cirillo2017, CirilloIROS2014, pivtoraiko2009differentially} which has been tailored for this specific application in our previous work in~\cite{LjungqvistIV2017}. 
The lattice planner efficiently computes kinematically feasible and collision-free motion plans by combining a finite number of precomputed motion segments. During online planning, challenging parking and obstacle avoidance maneuvers can be constructed by deploying efficient graph search algorithms~\cite{arastar}. 
To execute the motion plan, a path-following controller based on our previous work in~\cite{Ljungqvist2016CDC} is used to stabilize the lateral and angular path-following error states during the execution of the planned maneuver. 
Finally, a nonlinear observer based on an extended Kalman filter (EKF) is proposed to obtain full state information of the system. 
Based upon request from our commercial partner and since multiple trailers are usually switched between during daily operation, the observer is developed so that it only uses information from sensors that are mounted on the tractor.  

The proposed path planning and path-following control framework summarizes and extends our previous work in~\cite{Ljungqvist2016CDC,LjungqvistIV2017,LjungqvistACC2018}. 
Here, the complete system is implemented on a full-scale test vehicle and results from both simulations and real-world experiments are presented to demonstrate its performance. 
To the best of the author's knowledge, this paper presents the first path planning and path-following control framework for a G2T with a car-like tractor that is implemented on a full-scale test vehicle.

The remainder of the paper is structured as follows. 
In Section~\ref{j1:sec:systemArchitecture}, the responsibility of each module in the path planning and path-following control framework is briefly explained and an overview of related work is provided. 
In Section~\ref{j1:sec:Modeling}, the kinematic vehicle model of the G2T with a car-like tractor and the problem formulations are presented. 
The lattice-based path planner is presented in Section~\ref{j1:sec:MotionPlanner} and the hybrid path-following controller in Section~\ref{j1:sec:Controller}. In Section~\ref{j1:sec:stateEstimation}, the nonlinear observer that is used for state estimation is presented. Implementation details are covered in Section~\ref{j1:sec:implementation} and simulation results as well as results from real-world experiments are presented in Section~\ref{j1:sec:Results}. 
The paper is concluded in Section~\ref{j1:sec:conclusions} by summarizing the contributions and discusses directions for future work.   
\section{Background and related work}
\label{j1:sec:systemArchitecture}
\begin{figure}[t!]
	\begin{center}
		\includegraphics[width=0.6\linewidth]{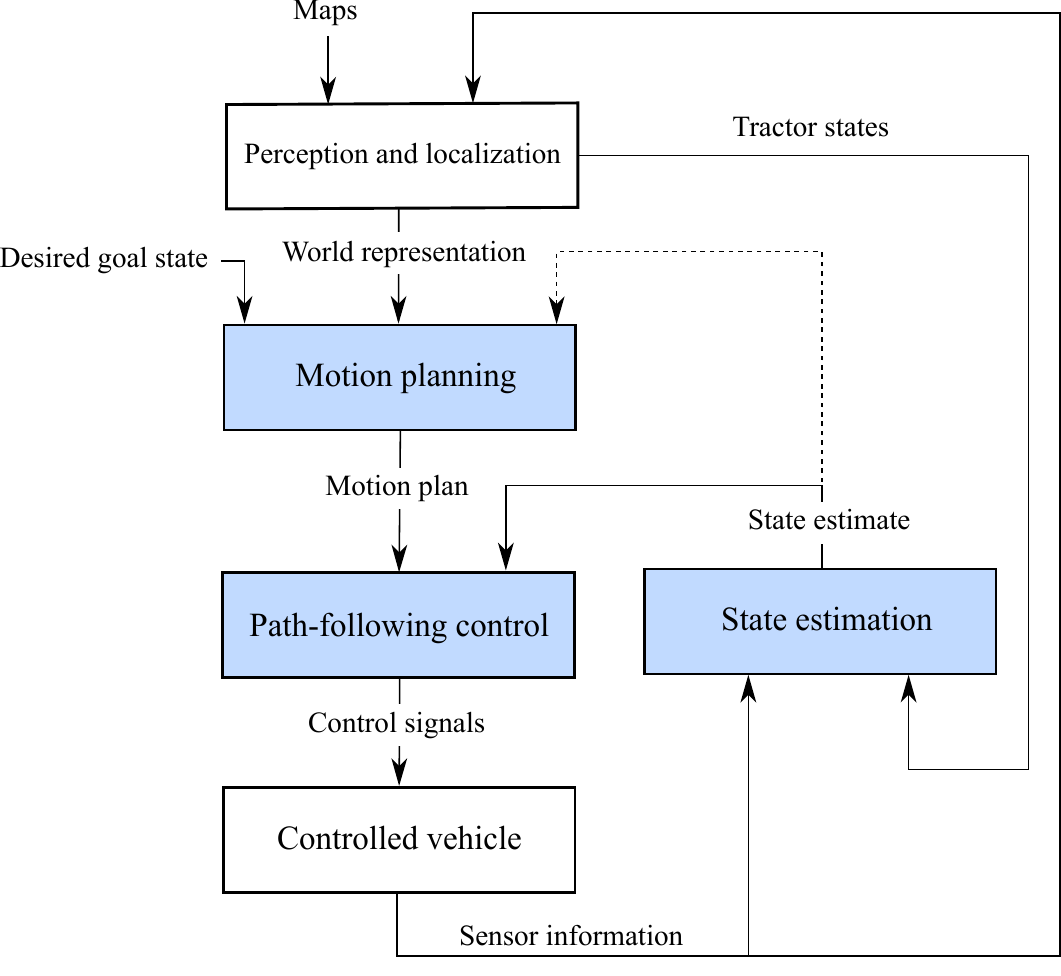}   
		\caption{A schematic illustration of the proposed system architecture where the blue subsystems; path planning, path-following control and state estimation, are considered in this work.} 
		\label{j1:fig:sys_arch}
	\end{center} 
\end{figure}

The full system is built from several modules and a simplified system architecture is illustrated in Figure~\ref{j1:fig:sys_arch}, where the integration and design of state estimation, path planning and path-following control are considered as the main contributions of this work. 
Below, the task of each module is briefly explained and for clarity, related work for each module is given individually. 

\subsection{Perception and localization}
\label{j1:sec:loc}
The objective of the perception and localization layer is to provide the planning and control layer with a consistent representation of the surrounding environment and an accurate estimation of where the tractor is located in the world. 
A detailed description of the perception layer is outside the scope of this paper, but a brief introduction is given for clarity. 

Precomputed maps and onboard sensors on the car-like tractor (RADARs, LIDARs, a global positioning system (GPS), inertial measurement units (IMUs) and cameras) are used to construct an occupancy grid map~\cite{occupancyGridMap} that gives a probabilistic representation of drivable and non-drivable areas. 
Dynamic objects are also detected and tracked but they are not considered in this work. Standard localization techniques are then used to obtain an accurate position and orientation estimate of the car-like tractor within the map~\cite{skog2009,levinson2011towards,montemerlo2008junior}. 
Together, the occupancy grid map and the tractor's position and orientation provide the environmental representation in which motion planning and control is performed. 

\subsection{State estimation}
To control the G2T with car-like tractor, accurate and reliable state estimation of the semitrailer's position and orientation as well as the two joint angles of the system need to be obtained.
An ideal approach would be to place sensors at each hitch connection to directly measure each joint angle~\cite{hafner2017control,evestedtLjungqvist2016,michalek2014highly} and equip the semitrailer with a similar localization system as the tractor (\textit{e.g.}, IMU and a high precision GPS). 
However, commercial trailers are often exchanged between tractors and a high-performance navigation system is very expensive, making it an undesirable solution for general applications. Furthermore, no standardized communication protocol between different trailer and tractor manufacturers exists. 

Different techniques for estimating the joint angle for a tractor with a semi-trailer and for a car with a trailer using wide-angle cameras are reported in~\cite{CameraSolSaxe} and~\cite{caup2013video}, respectively.
In~\cite{CameraSolSaxe}, an image bank with images taken at different joint angles is first generated and during execution used to compare and match against the current camera image.
Once a match is found, the corresponding joint angle is given from the matched image in the image bank. The work in~\cite{caup2013video} exploits symmetry of the trailer's drawbar in images to estimate the joint angle between a car and the trailer.
In~\cite{Fuchs2016trailerEst}, markers with known locations are placed on the trailer's body and then tracked with a camera to estimate the joint angles of a G2T with car-like tractor. The proposed solution is tested on a small-scale vehicle in a lab environment.

Even though camera-based joint angle estimation would be possible to utilize in practice, it is unclear how it would perform in different lighting conditions, \textit{e.g.}, during nighttime. 
The concept for angle estimation used in this work was first implemented on a full-scale test vehicle as part of the master's thesis~\cite{Patrik2016} supervised by the authors of this work. 
Instead of using a rear-view camera, a LIDAR sensor is mounted in the rear of the tractor. 
The LIDAR sensor is mounted such that the body of the semitrailer is visible in the generated point cloud for a wide range of joint angles. 
The semitrailer's body is assumed to be rectangular and by iteratively running the random sample consensus (RANSAC) algorithm~\cite{fischler1981random}, the visible edges of the semitrailer's body can be extracted from the point cloud.
Virtual measurements of the orientation of the semitrailer and the lateral position of the midpoint of its front with respect to the tractor are then constructed utilizing known geometric properties of the vehicle. 
These virtual measurements together with information of the position and orientation of the tractor are used as observations to an EKF for state estimation.

In~\cite{Daniel2018}, the proposed iterative RANSAC algorithm is benchmarked against deep-learning techniques to compute the estimated joint angles directly from the LIDAR's point cloud or from camera images. That work concludes that for trailers with rectangular bodies, the LIDAR and iterative RANSAC solution outperforms the other tested methods in terms of accuracy and robustness which makes it a natural choice for state estimation in this work.   

\subsection{Path planning}
Motion planning for car-like vehicles is a difficult problem due to the vehicle's nonholonomic constraints and the non-convex environment the vehicle is operating in~\cite{lavalle2006planning}. 
Motion planning for tractor-trailer systems is even more challenging due to the vehicle's complex kinematics, its relatively large dimensional state-space and its structurally unstable joint angle kinematics in backward motion. 
The standard $N$-trailer (SNT) which only allows on-axle hitching, is differentially flat and can be converted into chained form when the position of the axle of the last trailer is used as the flat output~\cite{sordalen1993conversion}. 
This property of the SNT is explored in~\cite{Murray1991,tilbury1995trajectory} to develop efficient techniques for local trajectory generation. 
In~\cite{tilbury1995trajectory}, simulation results for the one and two trailer cases are presented but obstacles as well as state and input constraints are omitted. 
A well-known issue with flatness-based trajectory generation is that it is hard to incorporate constraints, as well as minimizing a general performance measure while computing the motion plan. 
Some of these issues are handled in~\cite{sekhavat1997multi} where a motion planner for unstructured environments with obstacles for the S2T is proposed. 
In that work, the motion planning problem is split into two phases where a holonomic path that violates the vehicle's nonholonomic constraints is first generated and then iteratively replaced with a kinematically feasible trajectory by converting the system into chained form. A similar hierarchical motion planning scheme is proposed in~\cite{hillary} for a G1T robot which is also experimentally validated on a small scale platform. 

An important contribution in this work is that most of the approaches presented above only consider the SNT-case with on-axle hitching, despite that most practical applications have both on-axle and off-axle hitching. The off-axle hitching makes the kinematics for the general $N$-trailer (GNT) more complicated~\cite{altafini1998general}. 
To include the G2T with car-like tractor, we presented a probabilistic motion planner in~\cite{evestedtLjungqvist2016planning}. 
Even though the proposed motion planner is capable of solving several hard problems, the framework lacks all completeness and optimality guarantees that are given by the approach developed in this work. 

The family of motion planning algorithms that belong to the lattice-based motion planning family, can guarantee resolution optimality and completeness~\cite{pivtoraiko2009differentially}. 
In contrast to probabilistic methods, a lattice-based motion planner requires a regular discretization of the vehicle's state-space and is constrained to a precomputed set of feasible motions which, combined, can connect two discrete vehicle states. The precomputed motions are called motion primitives and can be generated offline by solving several optimal control problems (OCPs).
This implies that the vehicle's nonholonomic constraints already have been considered offline and what remains during online planning is a search over the set of precomputed motions. Due to its deterministic nature and real-time capabilities, lattice-based motion planning has been used with great success on various robotic platforms~\cite{pivtoraiko2009differentially,BOSSDarpa,CirilloIROS2014,LjungqvistCDC2018,oliveira2018combining} and is therefore the chosen motion planning strategy for this work.  

Other deterministic motion planning algorithms rely on input-space discretization~\cite{dolgov2010path,Beyersdorfer2013tractortrailer} in contrast to state-space discretization. A model of the vehicle is used during online planning to simulate the system for certain time durations, using constant or parametrized control signals. 
In general, the constructed motions do not end up at specified final states. This implies that the search graph becomes irregular and results in an exponentially exploding frontier during online planning~\cite{pivtoraiko2009differentially}. 
To resolve this, the state-space is often divided into cells where a cell is only allowed to be explored once. A motion planning algorithm that uses input-space discretization is the hybrid A$^*$~\cite{dolgov2010path}. In~\cite{Beyersdorfer2013tractortrailer}, a similar motion planner is proposed to generate feasible paths for a G1T with a car-like tractor with active trailer steering. A drawback 
with motion planning algorithms that rely on input-space discretization, is that they lack completeness and optimality guarantees. 
Moreover, input-space discretization is in general not applicable for unstable systems, unless the online simulations are performed in closed-loop with a stabilizing feedback controller~\cite{evestedtLjungqvist2016planning}.

A problem with lattice-based approaches is the curse of dimensionality, \textit{i.e.}, exponential complexity in the dimension of the state-space and in the number of precomputed motions. 
In~\cite{LjungqvistIV2017}, we circumvented this problem and developed a real-time capable lattice-based motion planner for a G2T with a car-like tractor. 
By discretizing the state-space of the vehicle such that the precomputed motions always move the vehicle from and to a circular equilibrium configuration, the dimension of the state lattice remained sufficiently low and made real-time use of classical graph search algorithms tractable. 
Even though the dimension of the discretized state-space is limited, the motion planner was shown to efficiently solve difficult and practically relevant motion planning problems. 

In this work, the work in~\cite{LjungqvistIV2017} is extended by better connecting the cost functional in the motion primitive generation and the cost function in the online motion planning problem. Additionally, the objective functional in backward motion is adjusted such that it reflects the difficulty of executing a maneuver. To avoid maneuvers in backward motion that in practice have a large risk of leading to a jack-knife state, a quadratic penalty on the two joint angles is included in the cost functional.    

\subsection{Path-following control}
During the past decades, an extensive amount of feedback control techniques for different tractor-trailer systems 
for both forward and backward motion have been proposed. 
The different control tasks include path-following control (see \textit{e.g.}, \cite{sampei1995arbitrary,hybridcontrol2001,astolfi2004path,Cascade-nSNT,bolzern1998path}), trajectory-tracking and set-point control (see \textit{e.g.},~\cite{CascadeNtrailer,divelbiss1997trajectory,michalek2018forward,SamsonChainedform1995}). 
Here, the focus will be on related path-following control solutions.

For the SNT, its flatness property can be used to design path-following controllers based on feedback linearization~\cite{sampei1995arbitrary} or by converting the system into chained form~\cite{SamsonChainedform1995}. 
The G1T with a car-like tractor is still differentially flat using a certain choice of flat outputs~\cite{rouchon1993flatness}. 
However, the flatness property does not hold when two consecutive trailers are off-axle hitched~\cite{CascadeNtrailernonmin,rouchon1993flatness}.
In~\cite{bolzern1998path}, this issue is circumvented by introducing a simplified reference vehicle which has equivalent stationary behavior but different transient behavior. Similar concepts have also been proposed in~\cite{virtualMorales2013,pushing2010}. 
Input-output linearization techniques are used in~\cite{altafini2003path} to stabilize the GNT around paths with constant curvature, where the path-following controller minimizes the sum of the lateral offsets to the path. 
The proposed approach is however limited to forward motion since the introduced zero-dynamics become unstable in backward motion. 
A closely related approach is presented in~\cite{minimumSweep}, where the objective of the path-following controller is to minimize the swept path of a G1T with a car-like tractor along paths in backward and forward motion.

Tractor-trailer vehicles that have pure off-axle hitched trailers, are referred to as non-standard N-trailers (nSNT)~\cite{CascadeNtrailernonmin,chung2011backward}. 
For these systems, scalable cascade-like path-following control techniques are presented in~\cite{michalek2014highly,Cascade-nSNT}. 
Compared to many other path-following control approaches, these controllers do not need to find the closest distance to the nominal path and the complexity of the feedback controllers scales well with increasing number of trailers. 
By introducing artificial off-axle hitches, the proposed controller can also be used for the GNT-case~\cite{Cascade-nSNT}. 
However, as experimental results illustrate, the path-following controller becomes sensitive to measurement noise when an off-axle distance approaches zero. 

A hybrid linear quadratic (LQ) controller is proposed in~\cite{hybridcontrol2001} to stabilize the G2T with car-like tractor around different equilibrium configurations corresponding to straight lines and circles, and a survey in the area of control techniques for tractor-trailer systems can be found in~\cite{david2014control}. 
Inspired by~\cite{hybridcontrol2001}, a cascade control approach for stabilizing the G2T with car-like tractor in backward motion around piecewise linear reference paths is proposed in~\cite{evestedtLjungqvist2016}. 
An advantage of this approach is that the controller can track reference paths that are not necessarily kinematically feasible. 
However, if a more detailed reference path with full state information is available, this method is only using a subset of the available information and the control accuracy might be reduced. 
A similar approach for path tracking is also proposed in~\cite{rimmer2017implementation} for reversing a G2T with a car-like tractor which has been successfully demonstrated in practice.

Most of the path-following approaches presented above consider the problem of following a path defined in the position and orientation of the last trailer's axle. 
In this work, the nominal path obtained from the path planner is composed of full state information as well as nominal control signals. 
Furthermore, in a motion planning and path-following control architecture, it is crucial that all nominal vehicle states are followed to avoid collision with surrounding obstacles.
To utilize all information in the nominal path, we presented a state-feedback controller with feedforward action in~\cite{Ljungqvist2016CDC}. The proposed path-following controller is proven to stabilize the path-following error kinematics for the G2T with a car-like tractor in backward motion around a set of admissible paths. 
The advantage of this approach is that the nominal path satisfies the vehicle kinematics making it, in theory, possible to follow exactly.
However, the developed stability result in~\cite{Ljungqvist2016CDC} fails to guarantee stability in continuous-time for motion plans that are combining forward and backward motion segments~\cite{LjungqvistACC2018}. 
In~\cite{LjungqvistACC2018}, we proposed a solution to this problem and presented a framework that is exploiting the fact that a lattice planner is combining a finite number of precomputed motion segments. 
Based on this, a framework is proposed for analyzing the behavior of the path-following error, how to design the path-following controller and how to potentially impose restrictions on the lattice planner to guarantee that the path-following error is bounded and decays towards zero. 
Based on this, the same framework is used in this work, where results from real-world experiments on a full-scale test vehicle are also presented.

\begin{figure}[b!]
	\begin{center}
		\includegraphics[width=0.65\linewidth]{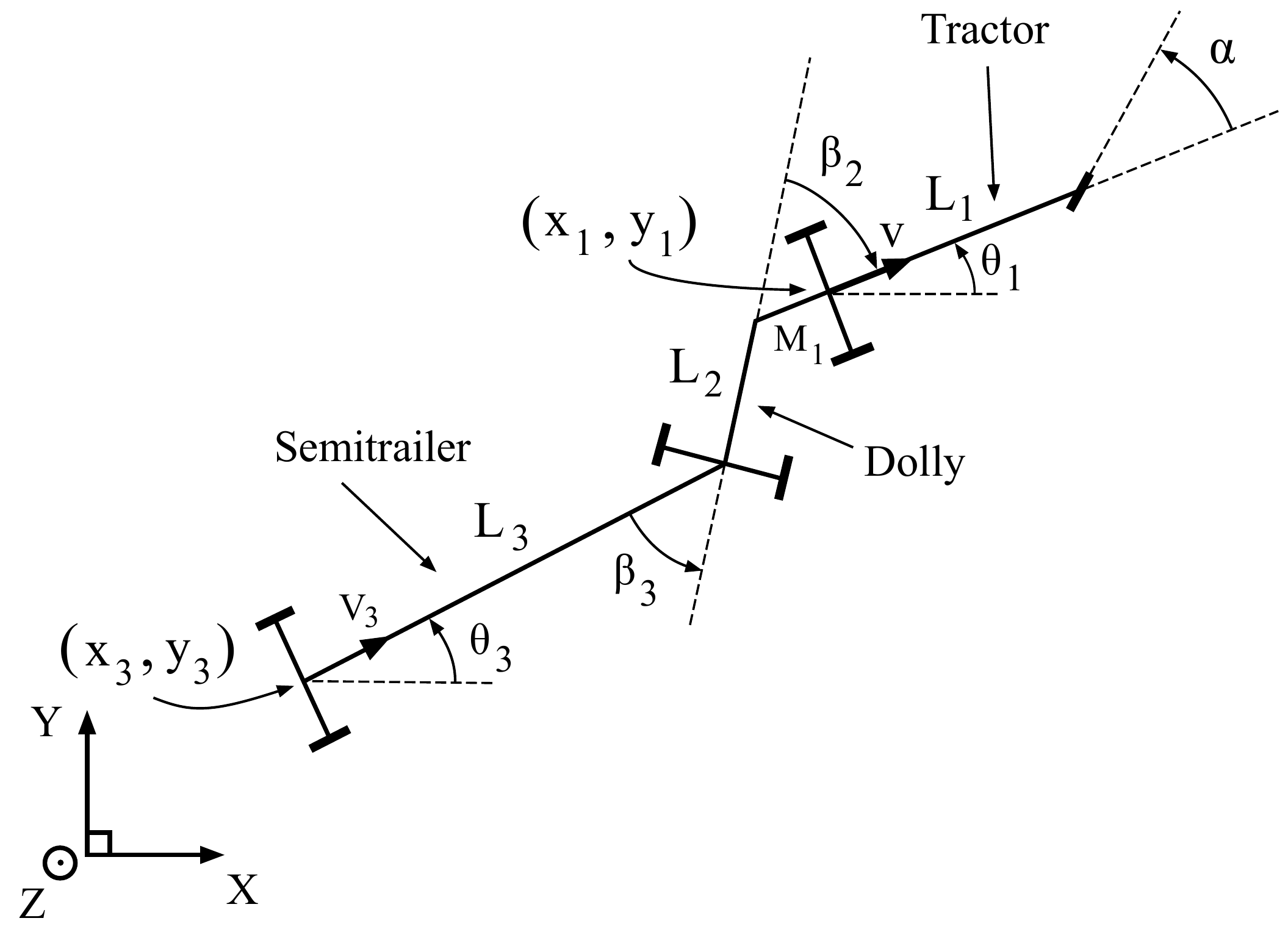}   
		\caption{Definition of the geometric lengths, states and control signals that are of relevance for modeling the general 2-trailer with a car-like tractor.} 
		\label{j1:fig:schematic_model_description}
	\end{center} 
\end{figure}

\section{Kinematic vehicle model and problem formulations}
\label{j1:sec:Modeling}
The G2T with a car-like tractor considered in this work is schematically illustrated in Figure~\ref{j1:fig:schematic_model_description}. 
This system has a positive off-axle connection between the car-like tractor and the dolly and an on-axle connection between the dolly and the semitrailer. 
The state vector $x=\begin{bmatrix} x_3 & y_3 & \theta_3 & \beta_3 & \beta_2\end{bmatrix}^T\in\mathbb R^5$ is used to represent a configuration of the vehicle, where $(x_3,y_3)$ is the position of the center of the semitrailer's axle, $\theta_3$ is the orientation of the semitrailer, $\beta_3$ is the joint angle between the semitrailer and the dolly and $\beta_2$ is the joint angle between the dolly and the car-like tractor\footnote{All angles are defined positive counter clockwise.}. 
The length $L_3$ represent the distance between the axle of the semitrailer and the axle of the dolly, $L_2$ is the distance between the axle of the dolly and the off-axle hitching connection at the car-like tractor, $M_1>0$ is the length of the positive off-axle hitching, and $L_1$ denotes the wheelbase of the car-like tractor. 
The car-like tractor is front-wheeled steered and assumed to have perfect Ackerman geometry. 
The control signals to the system are the steering angle $\alpha$ and the longitudinal velocity $v$ of the rear axle of the car-like tractor. 
A recursive formula derived from nonholonomic and holonomic constraints for the GNT vehicle is presented in~\cite{altafini1998general}. 
Applying the formula for this specific G2T with a car-like tractor results in the following vehicle model~\cite{altafini2002hybrid}:
\begin{subequations}
	\label{j1:eq:model_global_coord}
	\begin{align} 
	\dot{x}_3 &= v \cos \beta_3 C_1(\beta_2,\tan\alpha/L_1) \cos \theta_3, \label{eq:model1}\\
	\dot{y}_3 & =  v \cos \beta_3 C_1(\beta_2,\tan\alpha/L_1) \sin \theta_3,  \label{eq:model2}\\
	\dot{\theta}_3 & = v \frac{\sin \beta_3 }{L_3} C_1(\beta_2,\tan\alpha/L_1), \label{eq:model3}\\
	\dot{\beta}_3 & =v \left( \frac{1}{L_2}\left(\sin\beta_2 - \frac{M_1}{L_1}\cos\beta_2\tan \alpha \right) - \frac{\sin\beta_3}{L_3}C_1(\beta_2,\tan\alpha/L_1)\right), \label{eq:model4}\\
	\dot{\beta}_2 &= v \left(\frac{\tan\alpha}{L_1} - \frac{\sin \beta_2}{L_2} + \frac{M_1}{L_1 L_2}\cos\beta_2\tan\alpha  \right), \label{eq:model5}
	\end{align}
\end{subequations}
where $C_1(\beta_2,\kappa)$ is defined as 
\begin{align}
C_1(\beta_2,\kappa) = \cos{\beta_2} + M_1\sin\beta_2\kappa.
\label{j1:eq:C1}
\end{align}
By performing the input substitution $\kappa = \frac{\tan \alpha}{L_1}$, the model in~\eqref{j1:eq:model_global_coord} can be written on the form $\dot{ x} = vf(x,\kappa)$. Define
\begin{align}\label{j1:relation_v_v3}
g_v(\beta_2,\beta_3,\kappa) = \cos\beta_3 C_1(\beta_2,\kappa),
\end{align}
which describes the relationship, $v_3 = vg_v(\beta_2,\beta_3,\kappa)$, between the longitudinal velocity of the axle of the semitrailer, $v_3$ and the longitudinal velocity of the rear axle of the car-like tractor, $v$. 
When $g_v(\beta_2,\beta_3,\kappa)=0$, the system in~\eqref{j1:eq:model_global_coord} is uncontrollable which practically implies that the position of the axle of the dolly or the semitrailer remain in stationarity even though the tractor moves. 
To avoid these vehicle configurations, it is assumed that $g_v(\beta_2,\beta_3,\kappa)>0$, which implies that the joint angles has to satisfy {$|\beta_3| < \pi/2$} and {$|\beta_2| < \pi/2$}, respectively, and that $C_1(\beta_2,\kappa)>0$. 
These imposed restrictions are closely related to the segment-platooning assumption defined in~\cite{michalek2014highly} and does not limit the practical usage of the model since structural damage could occur on the semitrailer or the tractor, if these limits are exceeded. 

The model in~\eqref{j1:eq:model_global_coord} is derived based on no-slip assumptions and the vehicle is assumed to operate on a flat surface. 
Since the intended operational speed is quite low for our use case, these assumptions are expected to hold. 
The direction of motion is essential for the stability of the system~\eqref{j1:eq:model_global_coord}, where the joint angle kinematics are structurally unstable in backward motion ($v < 0$), where it risks to fold and enter what is called a jack-knife state~\cite{altafini2002hybrid}. 
In forward motion ($v > 0$), these modes are stable. 

Since the longitudinal velocity $v$ enters linearly into the model in~\eqref{j1:eq:model_global_coord}, time-scaling~\cite{sampei1986time} can be applied to eliminate the dependence on the longitudinal speed $|v|$. 
Define $s(t)$ as the distance traveled by the rear axle of the tractor, \textit{i.e.}, $s(t)=\int_0^t|v(\tau)|\mathrm{d}\tau$. By substituting time with $s(t)$, the differential equation in~\eqref{j1:eq:model_global_coord} can be written as 
\begin{align}
\frac{\mathrm{d} x}{\mathrm{d} s}  = \sign{(v(s))} f(x(s), \kappa(s)).
\label{j1:eq:time_scaling}
\end{align}
Since only the sign of $v$ enters into the state equation, it implies that the traveled path is independent of the tractor's speed $|v|$ and the motion planning problem can be formulated as a path planning problem~\cite{lavalle2006planning}, where the speed is omitted. 
Therefore, the longitudinal velocity $v$ is, without loss of generality, assumed to take on the values $v = 1$ for forward motion and $v = -1$ for backward motion, when path planning is considered. 

In practice, the vehicle has limitations on the maximum steering angle $|\alpha|\leq\alpha_{\text{max}}<\pi/2$, the maximum steering angle rate $|\omega|\leq\omega_{\text{max}}$ and the maximum steering angle acceleration $|u_\omega|\leq u_{\omega,\text{max}}$.
These constraints have to be considered in the path planning layer in order to generate feasible paths that the physical vehicle can execute. 
\subsection{Problem formulations} \label{j1:sec:problemformulation}
In this section, the path planning and the path-following control problems are defined. 
To make sure the planned path avoid uncontrollable regions and the nominal steering angle does not violate any of its physical constraints, an augmented state-vector $z = \begin{bmatrix}
x^T & \alpha & \omega\end{bmatrix}^T\in\mathbb R^{7}$ is used during path planning. The augmented model of the G2T with a car-like tractor~\eqref{j1:eq:model_global_coord} can be expressed in the following form 
\begin{align}  \label{j1:driftless_system}
\frac{\text{d}z}{\text{d}s} = f_z(z(s),u_p(s)) = \begin{bmatrix}
v(s)f(x(s),\tan\alpha(s)/L_1) \\ \omega(s) \\ u_\omega(s) \end{bmatrix},
\end{align}
where its state-space ${\mathbb Z}\subset\mathbb R^7$ is defined as follows
\begin{align}
\mathbb Z = \left\{ z\in\mathbb R^7 \mid |\beta_3| < \pi/2, \hspace{2pt} \hspace{2pt} |\beta_2| < \pi/2,\hspace{2pt} |\alpha|\leq\alpha_{\text{max}} ,\hspace{2pt} |\omega|\leq\omega_{\text{max}} ,\hspace{2pt} C_1(\beta_2,\tan\alpha/L_1)>0  \right\},
\end{align}
where $C_1(\beta_2,\tan\alpha/L_1)$ is defined in~\eqref{j1:eq:C1}. During path planning, the control signals are $u_p = \begin{bmatrix} v & u_\omega \end{bmatrix}^T \in {\mathbb  U}_p$, where ${\mathbb  U}_p=\{-1,1\}\times [-u_{\omega,\text{max}},u_{\omega,\text{max}}]$. 
Here, $u_\omega$ denotes the steering angle acceleration and the longitudinal velocity $v$ is constrained to $\pm1$ and determines the direction of motion. 
It is assumed that the perception layer provides the path planner with a representation of the surrounding obstacles $\mathbb Z_{\text{obs}}$. 
In the formulation of the path planning problem, it is assumed that $\mathbb Z_{\text{obs}}$ can be described analytically ($e.g.$, circles, ellipsoids, polytopes or other bounding regions~\cite{lavalle2006planning}). 
Therefore, the free-space where the vehicle is not in collision with any obstacles can be defined as \mbox{$\mathbb Z_{\text{free}} = {\mathbb Z} \setminus \mathbb Z_{\text{obs}}$}. 

Given an initial state $z_I = \begin{bmatrix} x_I^T & \alpha_I & 0 \end{bmatrix}^T \in\mathbb Z_{\text{free}}$ and a desired goal state $z_G= \begin{bmatrix} x_G^T & \alpha_G & 0 \end{bmatrix}^T \in\mathbb Z_{\text{free}}$, a feasible solution to the path planning problem is an arc-length parametrized control signal $u_{p}(s)\in {\mathbb  U}_p$, $s\in[0,s_G]$ which results in a nominal path in $z(s)$, $s\in[0,s_G]$ that is feasible, collision-free and moves the vehicle from its initial state $z_I$ to the desired goal state $z_G$. 
Among all feasible solutions to this problem, the optimal solution is the one that minimizes a specified cost functional $J$. The optimal path planning problem is defined as follows.
\begin{definition}[The optimal path planning problem] \label{j1:pathplanningproblem} 
	Given the 5-tuple ($z_I,z_G, \mathbb Z_{\text{free}},{\mathbb  U}_p, J$), find the path length $s_G\in\mathbb R_+$ and an arc-length parametrized control signal $u_{p}(s)= \begin{bmatrix} v(s) & u_{\omega}(s)	\end{bmatrix}^T$, $s\in[0,s_G]$ that minimizes the following OCP:
	\begin{subequations}
		\label{j1:eq:MotionPlanningOCP}
		\begin{align} 
		\minimize_{u_{p}(\cdot), \hspace{0.5ex}s_{G} }\hspace{3.7ex}
		& J = \int_{0}^{s_{G}}L(x(s),\alpha(s), \omega(s), u_\omega(s))\,\mathrm{d}s \label{j1:eq:MotionPlanningOCP_obj}	\\
		\subjectto\hspace{3ex}
		& \frac{\mathrm{d}z}{\mathrm{d}s} = f_z(z(s),u_p(s)), \label{j1:eq:MotionPlanningOCP_syseq} \\
		& z(0) = z_I, \quad z(s_{G}) = z_G, \label{j1:eq:MotionPlanningOCP_initfinal} \\ 
		& z(s) \in \mathbb Z_{\text{free}}, \quad
		u_{p}(s) \in {\mathbb  U}_p, \label{j1:eq:MotionPlanningOCP_constraints}
		\end{align}
	\end{subequations}
	where $L:\mathbb R^5\times\mathbb R\times\mathbb R\times\mathbb R\rightarrow \mathbb R_+$ is the cost function.  
\end{definition}
The optimal path planning problem in~\eqref{j1:eq:MotionPlanningOCP} is a nonlinear OCP which is often, depending on the shape of $\mathbb Z_{\text{free}}$, highly non-convex. 
Thus, the OCP in~\eqref{j1:eq:MotionPlanningOCP} is in general hard to solve by directly invoking a numerical optimal control solver~\cite{bergman2018combining,zhang2018optimization} and sampling-based path planning algorithms are commonly employed to obtain an approximate solution~\cite{lavalle2006planning,reviewFrazzoli2016}. 
In this work, a lattice-based path planner~\cite{pivtoraiko2009differentially,CirilloIROS2014} is used and the framework is presented in Section~\ref{j1:sec:MotionPlanner}. 

For the path-following control design, a nominal path that the vehicle is expected to follow is defined as $(x_r(s),u_r(s)), s\in[0,s_{G}]$, where $x_r(s)$ is the nominal vehicle state and $u_r(s)=\begin{bmatrix}
v_r(s) & \kappa_r(s) \end{bmatrix}^T$ is the nominal velocity and curvature control signals.
The objective of the path-following controller is to locally stabilize the vehicle around this path in the presence of disturbances and model errors. 
When path-following control is considered, it is not crucial that the vehicle is located at a specific nominal state in time, rather that the nominal path is executed with a small and bounded path-following error \mbox{$\tilde x(t)=x(t)-x_r(s(t))$}. The path-following control problem is formally defined as follows.
\begin{definition}[The path-following control problem]\label{j1:pathfollowingproblem}
	Given a controlled G2T with a car-like tractor~\eqref{j1:eq:model_global_coord} and a feasible nominal path $(x_r(s),u_r(s))$, $s\in[0,s_{G}]$. 
	Find a control-law $\kappa(t)=g(s(t),x(t))$ with $v(t)=v_r(s(t))$, such that the solution to the closed-loop system $\dot x(t)=v_r(s(t))f(x(t),g(s(t), x(t)))$ satisfies the following locally around the nominal path: For all $t \in\{t\in\mathbb{R}_+ \mid 0 \leq s(t) \leq s_G \}$, there exist positive constants $r$, $\rho$ and $\epsilon$ such that  
	\begin{enumerate}
		\item $||\tilde x(t)||\leq \rho ||\tilde x(t_0)||e^{-\epsilon (t-t_0)}, \quad \forall ||\tilde x(t_0)||<r$,
		\item $\dot{s}(t)>0$.
	\end{enumerate}
\end{definition}
If the nominal path would be infinitely long ($s_G\rightarrow \infty$), Definition~\ref{j1:pathfollowingproblem} coincides with the definition of local exponential stability of the path-following error model around the origin~\cite{khalil}. 
In this work, the path-following controller is designed by first deriving a path-following error model. 
This derivation as well as the design of the path-following controller are presented in Section~\ref{j1:sec:Controller}. 
\subsection{System properties}
Some relevant and important properties of the model in~\eqref{j1:eq:model_global_coord} that will be exploited for path planning are presented below.
\begin{figure}[!t]
	\begin{center}
		\includegraphics[width=0.6\linewidth]{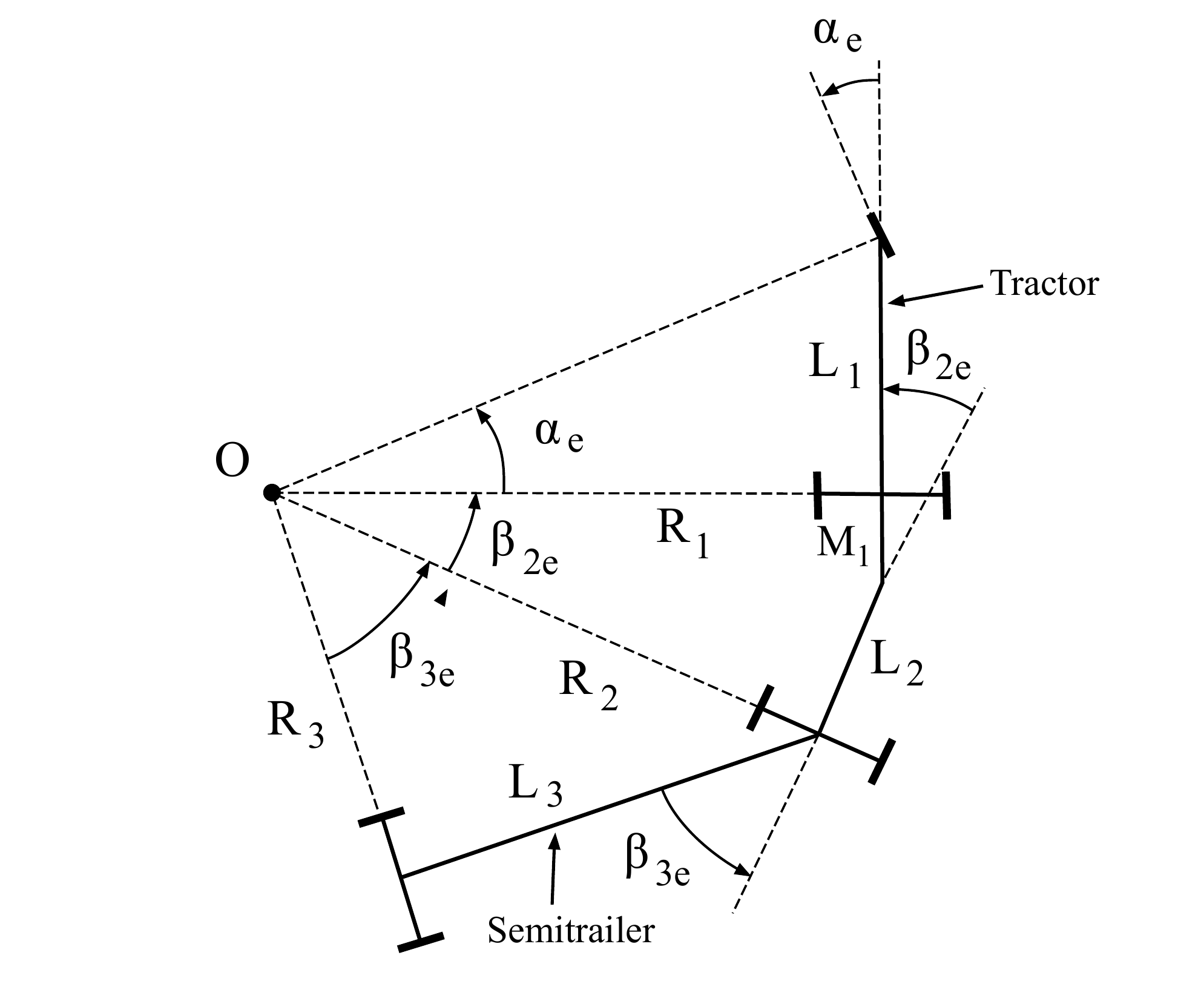}   
		\caption{Illustration of a circular equilibrium configuration for the G2T with a car-like tractor. Given a constant steering angle $\alpha_e$, there exists a unique pair of joint angles, $\beta_{2,e}$ and $\beta_{3,e}$, where $\dot\beta_2 = \dot\beta_3 =0$.}
		\label{j1:fig:eq_conf}
	\end{center}
\end{figure}

\subsubsection{Circular equilibrium configurations}
Given a constant steering angle $\alpha_e$ there exists a circular equilibrium configuration where $\dot \beta_2$ and $\dot\beta_3$ are equal to zero, as illustrated in Figure \ref{j1:fig:eq_conf}. 
In stationarity, the vehicle will travel along circles with radii determined by $\alpha_e$~\cite{hybridcontrol2001}. 
The equilibrium joint angles, $\beta_{2e}$ and $\beta_{3e}$, are related to $\alpha_e$ through the following equations
\begin{subequations} 
	\label{j1:eq:equ}
	\begin{align} 
	\beta_{3e} &=  \arctan \left( \frac{L_3}{R_3} \right)  \label{j1:eq:equ1},\\
	\beta_{2e} &=  \left( \arctan \left( \frac{M_1}{R_1} \right) + \arctan \left( \frac{L_2}{R_2} \right)\right),  \label{j1:eq:equ2}
	\end{align}  
\end{subequations}
where the absolute values of the signed radii are $|R_1| = L_1/ |\tan \alpha_e|$, $|R_2| = (R^2_1 + M_1^2 - L_2^2)^{1/2}$ and $|R_3| = (R_2^2 - L_3^2)^{1/2}$. 

\subsubsection{Symmetry}
A feasible path $(z(s),u_p(s))$, $s\in[0,s_G]$ to~\eqref{j1:driftless_system} that moves the system from an initial state $z(0)$ to a final state $z(s_G)$, is possible to reverse in distance and revisit the exact same points in $x$ and $\alpha$ by a simple transformation of the control signals. The result is formalized in Lemma~\ref{j1:L1} which is provided in Appendix A. Note that the actual state $x(\cdot)$ and steering angle $\alpha(\cdot)$ paths of the system~\eqref{j1:driftless_system} are fully distance-reversed and it is only the path of the steering angle rate $\omega(\cdot)$ that changes sign. Moreover, if $\omega(0)$ and $\omega(s_G)$ are equal to zero, the initial and final state constraints coincide. 
The practical interpretation of the result in Lemma~\ref{j1:L1} is that any path taken by the G2T with a car-like tractor~\eqref{j1:eq:time_scaling} with $|\alpha(\cdot)|\leq \alpha_{\text{max}}$ is feasible to follow in the reversed direction. 
Now, define the reverse optimal path planning problem to~\eqref{j1:eq:MotionPlanningOCP} as 
\begin{subequations}
	\label{j1:eq:revMotionPlanningOCP}
	\begin{align} 
	\minimize_{\bar u_{p}(\cdot), \hspace{0.5ex}\bar s_{G} }\hspace{3.7ex}
	& \bar J = \int_{0}^{\bar s_{G}}L(\bar x(\bar s),\bar \alpha(\bar s), \bar \omega(\bar s), \bar u_\omega(\bar s))\,\text d\bar s	\label{j1:eq:revMotionPlanningOCP_obj}\\
	\subjectto\hspace{3ex}
	& \frac{\text d\bar z}{\text d\bar s} = f_z(\bar z(\bar s),\bar u_p(\bar s)), \label{j1:eq:revMotionPlanningOCP_syseq} \\
	& \bar z(0) = z_G, \quad \bar z(\bar s_{G}) = z_I, \label{j1:eq:revMotionPlanningOCP_initfinal} \\ 
	& \bar z(\bar s) \in \mathbb Z_{\text{free}}, \quad
	\bar u_{p}(\bar s) \in {\mathbb  U}_p. \label{j1:eq:revMotionPlanningOCP_constraints} 
	\end{align}
\end{subequations}
Note that the only difference between the OCPs defined in~\eqref{j1:eq:MotionPlanningOCP} and~\eqref{j1:eq:revMotionPlanningOCP}, respectively, is that the initial and goal state constraints are switched. 
In other words,~\eqref{j1:eq:MotionPlanningOCP} defines a path planning problem from $z_I$ to $z_G$ and~\eqref{j1:eq:revMotionPlanningOCP} defines a path planning problem from $z_G$ to $z_I$. 
It is possible to show that also the optimal solutions to these OCPs are related through the result established in Lemma~\ref{j1:L1}.
\begin{assumption}\label{j1:A-optimal-symmetry}
	For all $z\in\mathbb Z_{\text{free}}$ and $u_p\in{\mathbb U}_p$, the cost function $L$ in~\eqref{j1:eq:MotionPlanningOCP} satisfies $L(x,\alpha, \omega,u_\omega)=L(x,\alpha, -\omega,u_\omega)$.
\end{assumption}
\begin{assumption}\label{j1:A-optimal-symmetry2}
	$z = \begin{bmatrix} x^T & \alpha & \omega\end{bmatrix}^T\in\mathbb Z_{\text{free}} \Leftrightarrow \bar z = \begin{bmatrix} x^T & \alpha & -\omega\end{bmatrix}^T\in\mathbb Z_{\text{free}}$.
\end{assumption}
\begin{theorem}
	\label{j1:T-optimal-symmetry}
	Under Assumption~\ref{j1:A-optimal-symmetry}--\ref{j1:A-optimal-symmetry2}, if $(z^*(s), u_p^*(s))$, $s\in [0, s_G^*]$ is an optimal solution to the optimal path planning problem~\eqref{j1:eq:MotionPlanningOCP} with optimal objective functional value $J^*$, then the distance-reversed path $(\bar z^*(\bar s),\bar u^*_p(\bar s))$, $\bar s\in [0, \bar s_G^*]$ given by~\eqref{j1:eq:reversed_controls}--\eqref{j1:eq:reversed_states} with $\bar s_G^*=s_G^*$, is an optimal solution to the reverse optimal path planning problem~\eqref{j1:eq:revMotionPlanningOCP} with optimal objective functional value $\bar J^* = J^*$.
\end{theorem}
\begin{proof} See Appendix A. \end{proof}
Theorem~\ref{j1:T-optimal-symmetry} shows that if an optimal solution to the optimal path planning problem in~\eqref{j1:eq:MotionPlanningOCP} or the reversed optimal path planning problem in~\eqref{j1:eq:revMotionPlanningOCP} is known, an optimal solution to the other one can immediately be derived using the invertible transformation defined in~\eqref{j1:eq:reversed_controls}--\eqref{j1:eq:reversed_states} and $\bar s_G=s_G$. 
\section{Lattice-based path planner}
\label{j1:sec:MotionPlanner}
As previously mentioned, the path planning problem defined in~\eqref{j1:pathplanningproblem} is hard to solve by directly invoking a numerical optimal control solver. 
Instead, it can be combined with classical search algorithms and a discretization of the state-space to build efficient algorithms to solve the path planning problem. 
By discretizing the state-space $\mathbb Z_d$ of the vehicle in a regular fashion and constraining the motion of the vehicle to a lattice graph $\pazocal{G} = \langle \pazocal{V},\pazocal{E}\rangle$, which is a directed graph embedded in an Euclidean space that forms a regular and repeated pattern, classical graph-search techniques can be used to traverse the graph and compute a path to the goal~\cite{pivtoraiko2009differentially,CirilloIROS2014}. 
Each vertex $\nu[k] \in \pazocal V$ represents a discrete augmented vehicle state $z[k]\in\mathbb Z_d$ and each edge $e_i \in \pazocal{E}$ represents a motion primitive $m_i$, which encodes a feasible path $(z^i(s), u_p^i(s))$, $s\in [0, s_f^i]$ that moves the vehicle from one discrete state $z[k] \in \mathbb Z_d$ to a neighboring state $z[k+1] \in \mathbb Z_d$, while respecting the vehicle model and its physically imposed constraints. 
For the remainder of this text, state and vertex will be used interchangeably. 

Each motion primitive $m_i$ is computed offline and stored in a library containing a set $\pazocal{P}$ of precomputed feasible motion segments that can be used to connect two vertices in the graph. 
In this work, an OCP solver is used to generate the motion primitives and the vehicle's nonholonomic constraints are in this way handled offline, and what remains during online planning is a search over the set of precomputed motions. 
Performing a search over a set of precomputed motion primitives is a well known technique and is known as lattice-based path planning~\cite{pivtoraiko2009differentially,CirilloIROS2014}. 

Let $z[k+1]=f_p(z[k],m_i)$ represent the state transition when $m_i$ is applied from $z[k]$, and let $J_p(m_i)$ denote the stage-cost associated with this transition. The complete set of motion primitives $\pazocal{P}$ is computed offline by solving a finite set of OCPs to connect a set of initial states with a set of neighboring states in an obstacle-free environment. The set $\pazocal{P}$ is constructed from the position of the semitrailer at the origin and since the G2T with a car-like tractor~\eqref{j1:eq:model_global_coord} is position-invariant, a motion primitive $m_i\in\pazocal P$ can be translated and reused from all other positions on the grid.
The cardinality of the complete set of motion primitives is $|\pazocal{P}|=M$, where $M$ is a positive integer-valued scalar.  
In general, all motion primitives in $\pazocal{P}$ cannot be used from each state $z[k]$ and the set of motion primitives that can be used from $z[k]$ is denoted $\pazocal P(z[k])\subseteq \pazocal{P}$. 
The cardinality of $\pazocal P(z[k])$ defines the number of motion primitives that can be used from a given state $z[k]$ and the average $|\pazocal P(z[k])|$ defines the branching factor of the search problem. 
Therefore, a trade off between planning time and maneuver resolution has to be made when designing the motion primitive set. 
Having a large library of diverse motions gives the lattice planner more flexibility, however, the planning time will increase exponentially with the size of $|\pazocal P(z[k])|$. 
As the branching factor increases, a well-informed heuristic function becomes more and more important in order to maintain real-time performance during online planning~\cite{knepper2006high,CirilloIROS2014}. 
The heuristic function estimates the true cost-to-go from a state $z[k]\in\mathbb Z_d$ to the goal state $z_G$, and is used as guidance for the online graph search to expand the most promising vertices~\cite{lavalle2006planning,CirilloIROS2014,knepper2006high}. It is desired that the heuristic function is admissible to maintain optimality guarantees, and close to the true cost-to-go for efficient online planning. For nonholonomic systems, the Euclidean distance to the goal is known to severely underestimate the true cost-to-go in many situations and precomputed free-space heuristic look-up tables (HLUTs) are often used to improve the online planning time~\cite{knepper2006high,CirilloIROS2014}.      

The nominal path taken by the vehicle when motion primitive $m_i\in\pazocal{P}$ is applied from $z[k]$, is declared collision-free if it does not collide with any obstacles $c(m_i,z[k])\in\mathbb Z_{\text{free}}$, otherwise it is declared as in collision $c(m_i,z[k])\notin\mathbb Z_{\text{free}}$. 
Define \mbox{$u_q:\mathbb Z_+\rightarrow \{1,\hdots,M\}$} as a discrete and integer-valued signal that is selected by the lattice planner, where $u_q[k]$ specifies which motion primitive that is applied a stage $k$. 
By specifying the set of allowed states $\mathbb Z_d$ and precomputing the set of motion primitives $\pazocal P$, the continuous-time optimal path planning problem~\eqref{j1:eq:MotionPlanningOCP} is approximated by the following  discrete-time OCP:
\begin{align}
\minimize_{\{u_q[k]\}^{N-1}_{k=0}, \hspace{0.5ex} N}\hspace{3.7ex}
& J_{\text{D}} = \sum_{k=0}^{N-1}J_p(m_{u_q[k]}) \label{j1:eq:OCP_discrete} \\
\subjectto\hspace{3ex}
& z[0] = z_I, \quad z[N] = z_G,  \nonumber\\ 
& z[k+1] = f_{p}(z[k],m_{u_q[k]}), \nonumber \\
& m_{u_q[k]} \in \pazocal P(z[k]), \nonumber \\
& c(m_{u_q[k]},z[k]) \in \mathbb Z_{\text{free}}. \nonumber
\end{align}
The decision variables to this problem are the integer-valued sequence $\{u_q[k]\}^{N-1}_{k=0}$ and its length $N$. 
A feasible solution is an ordered sequence of collision-free motion primitives $\{m_{u_q[k]}\}^{N-1}_{k=0}$, \textit{i.e.}, a nominal path $(z(s), u_p(s))$, $s\in [0, s_G]$, that connect the initial state $z(0)=z_I$ and the goal state $z(s_G)= z_G$. 
Given the set of all feasible solutions to~\eqref{j1:eq:OCP_discrete}, the optimal solution is the one that minimizes the cost function $J_{\text{D}}$. 

During online planning, the discrete-time OCP in~\eqref{j1:eq:OCP_discrete} is solved using the anytime repairing A$^*$ (ARA$^*$) search algorithm~\cite{arastar}. 
ARA$^*$ is based on standard A$^*$ but initially performs a greedy search with the heuristic function inflated by a factor $\gamma\geq1$. This provides a guarantee that the found solution has a cost $J_D$ that satisfies $J_D \leq \gamma J_D^*$, where $J_D^*$ denotes the optimal cost to~\eqref{j1:eq:OCP_discrete}.
When a solution with a guaranteed bound of $\gamma$-suboptimality has been found, $\gamma$ is gradually decreased until an optimal solution with $\gamma=1$ is found or if a maximum allowed planning time is reached. 
With this search algorithm, both real-time performance and suboptimality bounds for the produced solution can be guaranteed. 

In~\eqref{j1:eq:OCP_discrete}, it is assumed that $z_I\in \mathbb Z_d$ and $z_G\in \mathbb Z_d$ to make the problem well defined. 
If $z_I\notin \mathbb Z_d$ or $z_G\notin \mathbb Z_d$, they have to be projected to their closest neighboring state in $\mathbb Z_d$ using some distance metric. 
Thus, the discretization of the vehicle's state-space restricts the set of possible initial states the lattice planner can plan from and desired goal states that can be reached. 
Even though not considered in this work, these restrictions could be alleviated by the use of numerical optimal control~\cite{ipopt} as a post-processing step~\cite{lavalle2006planning,oliveira2018combining,andreasson2015fastsmoothing}. 

The main steps of the path planning framework used in this work are summarized in Workflow~\ref{j1:alg1} and each step is now explained more thoroughly.      

\begin{algorithm}[t]
	\caption{The lattice-based path planning framework for the G2T with a car-like tractor}
	\label{j1:alg1}
	\begin{description}
		\item \textbf{Step 1 -- State lattice construction:}
		\begin{enumerate}
			\item[a)] \textbf{\emph{State-space discretization:}} Specify the resolution of the discretized state-space $\mathbb Z_d$.
			\item[b)] \textbf{\emph{Motion primitive selection:}} Specify the connectivity in the state lattice by selecting pairs of discrete states $\{z_s^i,z_f^i\}$, $i=1,\hdots,M$, to connect.
			\item[c)] \textbf{\emph{Motion primitive generation:}} Design the cost functional $J_p$ and compute the set of motion primitives $\pazocal P$ that moves the vehicle between $\{z_s^i,z_f^i\}$, $i=1,\hdots,M$.
		\end{enumerate}
		\item\textbf{Step 2 -- Efficiency improvements:}
		\begin{enumerate}
			\item[a)]\textbf{\emph{Motion primitive reduction:}} Systematically remove redundant motion primitives from $\pazocal P$ to reduce the branching factor of the search problem and therefore enhance the online planning time.
			\item[b)] \textbf{\emph{Heuristic function:}} Precompute a HLUT by calculating the optimal cost-to-go in an obstacle-free environment. 
		\end{enumerate}
		\item\textbf{Step 3 -- Online path planning:}
		\begin{enumerate}
			\item[a)] \textbf{\emph{Initialization:}} Project the vehicle's initial state $z_I$ and desired goal state $z_G$ to $\mathbb Z_d$.
			\item[b)] \textbf{\emph{Graph search:}} Solve the discrete-time OCP in~\eqref{j1:eq:OCP_discrete} using ARA$^*$.
			\item[c)] \textbf{\emph{Return:}} Send the computed solution to the path-following controller or report failure.
		\end{enumerate}
	\end{description}
\end{algorithm}

\subsection{State lattice construction}
\label{j1:subsec:lattice_creation}
\begin{figure}[t!] 	
	\vspace{-20pt}
	\centering
	\subfloat[][]{%
		\includegraphics[width=0.42\linewidth]{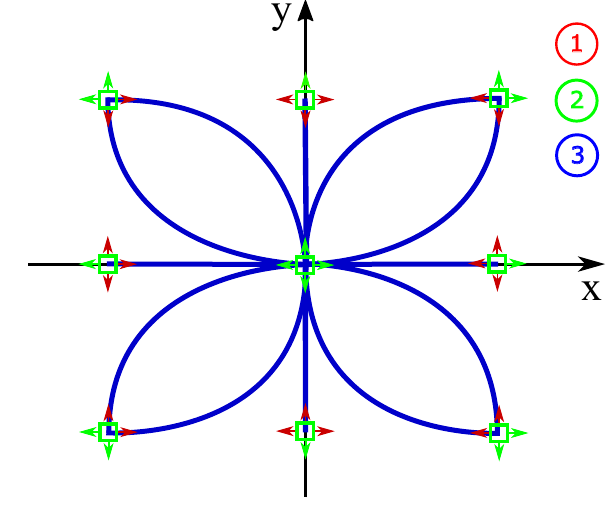}
		\label{j1:fig:state_lattice_construction}}
	\hfill
	\subfloat[][]{%
		\includegraphics[width=0.35\textwidth]{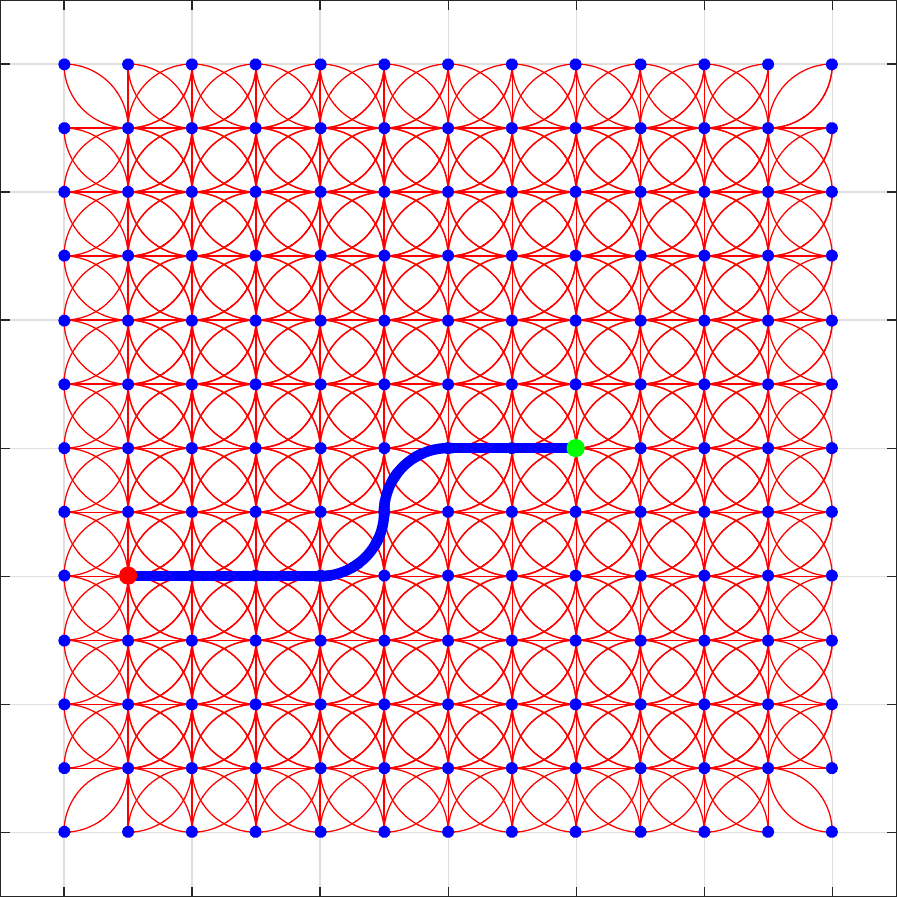}
		\label{j1:fig:state_lattice_planning}}
	\caption{In (a), an illustration of the three steps that are performed to generate the state lattice. (1) Discretize the state-space, (2) select which pair of states to connect, (3) compute optimal paths (motion primitives) between each pair of states. In (b), the resulting state lattice together with a solution (blue path) to a graph-search problem.}
	\label{j1:fig:state_lattice}
\end{figure}

The offline construction of the state lattice can be divided into three steps, as illustrated in Figure~\ref{j1:fig:state_lattice_construction}. 
First, the state-space of the vehicle is discretized with a certain resolution. 
Second, the connectivity in the state lattice is decided by specifying a finite amount of pairs of discrete vehicle states \mbox{$\{z_s^i,z_f^i\}$, $i=1,\hdots,M$}, to connect. 
Third, the motion primitives connecting each of these pairs of vehicle states are generated by the use of numerical optimal control~\cite{ipopt}. 
Together, these three steps define the resolution and the size of the lattice graph $\pazocal G$ and needs to be chosen carefully to maintain a reasonable search time during online planning, while at the same time allowing the vehicle to be flexible enough to maneuver in confined spaces.

To obtain a tractable search space, the augmented state-vector $z[k] = \begin{bmatrix} x[k]^T & \alpha[k] &\omega[k]\end{bmatrix}^T$ is discretized into circular equilibrium configurations~\eqref{j1:eq:equ} at each state in the state lattice. 
This implies that the joint angles, $\beta_{2}[k]$ and $\beta_{3}[k]$, are implicitly discretized since they are uniquely determined by the equilibrium steering angle $\alpha[k]$ through the relationships in~\eqref{j1:eq:equ}. 
However, in between two discrete states in the state lattice, the system is not restricted to circular equilibrium configurations.
The steering angle rate $\omega[k]$ is constrained to zero at each vertex in the state lattice to make sure that the steering angle is continuously differentiable, even when multiple motion primitives are combined during online planning. 
The position of the axle of the semitrailer $(x_{3}[k],y_{3}[k])$ is discretized to a uniform grid with resolution \mbox{$r=1$ m} and the orientation of the semitrailer $\theta_{3}[k]$ is discretized irregularly\footnote{$\Theta$ is the the set of unique angles $-\pi<\theta_{3}\leq \pi$ that can be generated by $\theta_{3} = \arctan2(i,j)$ for two integers $i,j\in\{-2,-1,0,1,2\}$.} into \mbox{$|\Theta|=16$} different orientations~\cite{pivtoraiko2009differentially}. 
This discretization of $\theta_{3}[k]$ is used to make it possible to construct short straight paths, compatible with the chosen discretization of the position from every orientation $\theta_{3}[k]\in\Theta$. 
Finally, the equilibrium steering angle $\alpha_{e}[k]$ is discretized into $|\Phi|=3$ different angles, where $\Phi = \{-0.1, 0, 0.1\}$. With the proposed state-space discretization, the actual dimension of the discretized state-space $\mathbb Z_d$ is four. Of course, the proposed discretization imposes restriction to the path planner, but is motivated to enable fast and deterministic online planning.  

\subsection{Motion primitive generation}
\label{j1:subsec:MPrimitiveGen}
The motion primitive set $\pazocal{P}$ is precomputed offline by solving a finite set of OCPs that connect a set of initial states $z_s^i \in \mathbb Z_d$ to a set of neighboring states $z_f^i\in \mathbb Z_d$ in a bounded neighborhood in an obstacle-free environment. 

Unlike our previous work in~\cite{LjungqvistIV2017}, the objective functional used during motion primitive generation coincides with the online planning stage-cost $J_p(m_i)$. 
This enables the resulting motion plan to be as close as possible to the optimal one and desirable behaviors can be favored in a systematic way. 
To promote and generate less complex paths that are easier for a path-following controller to execute, the cost function $L$ in~\eqref{j1:eq:MotionPlanningOCP} is chosen as
\begin{align}
L( z, u_\omega) = 1 + \left\lVert\begin{bmatrix} \beta_3 & \beta_2 \end{bmatrix}^T\right\rVert_\mathbf{Q_1}^2 + \left\lVert\begin{bmatrix} \alpha & \omega & u_\omega \end{bmatrix}^T\right\rVert_\mathbf{Q_2}^2, \label{j1:obj_rev}
\end{align}   
where the matrices $\mathbf Q_1 \succeq 0$ and $\mathbf Q_2 \succeq 0$ are design parameters that are used to trade off between simplicity of executing the maneuver and the path distance $s_f$. 
By tuning the weight matrix $\mathbf Q_1$, maneuvers in backward motion with large joint angles, $\beta_2$ and $\beta_3$, that have a higher risk to enter a jack knife state, can be penalized and therefore avoided during online planning if less complex motion primitives exist. 
In forward motion, the modes corresponding to the two joint angles $\beta_2$ and $\beta_3$ are stable and therefore not penalized.

To guarantee that the motion primitives in $\pazocal{P}$ move the vehicle between two discrete states in the state lattice, they are constructed by selecting initial states $z^i_{s}\in \mathbb Z_d$ and final states $z^i_{f}\in \mathbb Z_d$ that lie on the grid. 
A motion primitive in forward motion from $z^i_{s}=\begin{bmatrix} x_s^i & \alpha_s^i & 0\end{bmatrix}^T$ to $z^i_{f}=\begin{bmatrix} x_f^i & \alpha_f^i & 0\end{bmatrix}^T$ is computed by solving the following OCP:  
\begin{align} 
\minimize_{u^i_\omega(\cdot), \hspace{0.5ex} s^i_{f} }\hspace{3.7ex}
& J_p(m_i) = \int_{0}^{s^i_{f}}L(z^i(s), u^i_\omega(s))\,\text ds	\label{j1:OCP_mp_gen}\\
\subjectto\hspace{3ex}
& \frac{\text{d}z^i}{\text{d}s} = \begin{bmatrix}
f(x^i(s),\tan\alpha^i(s)/L_1) \\ \omega^i(s) \\ u_\omega^i(s) \end{bmatrix},  \nonumber \\
&  z^i(0) = z_s^i, \quad z^i(s_f) = z_f^i, \nonumber \\ 
&  z^i(s) \in \mathbb Z, \quad
|u^i_\omega(s)|\leq u_{\omega,\text{max}}. \nonumber 
\end{align}
Note the similarity of OCP in~\eqref{j1:OCP_mp_gen} with the optimal path planning problem~\eqref{j1:eq:MotionPlanningOCP}. 
Here, the obstacle imposed constraints are neglected and the vehicle is constrained to only move forwards. 
The established results in Lemma~\ref{j1:L1} and Theorem~\ref{j1:T-optimal-symmetry} are exploited to generate the motion primitives for backward motion. 
Here, each OCP is solved from the final state $z_f^i$ to the initial state $z_s^i$ in forward motion and the symmetry result in Lemma~\ref{j1:L1} is applied to recover the backward motion segment. Theorem~\ref{j1:T-optimal-symmetry} guarantees that the optimal solution $(z^i(s), u^i_p(s))$, $s\in [0, s^i_f]$ and the optimal objective functional value $J_p(m_i)$ remain unaffected.  
This technique is used to avoid the structurally unstable joint-angle kinematics in backward motion that can cause numerical problems for the OCP solver. 

\begin{figure}[t!] 	
	\vspace{-30pt}
	\centering
	\subfloat[][The set of motion primitives from $(\theta_{3,s},\alpha_{s})=(0,0.1)$ (green) and $(\theta_{3,s},\alpha_{s})=(0,-0.1)$ (blue) to different final states $z_f\in\mathbb Z_d$.]{%
		\includegraphics[width=0.47\textwidth]{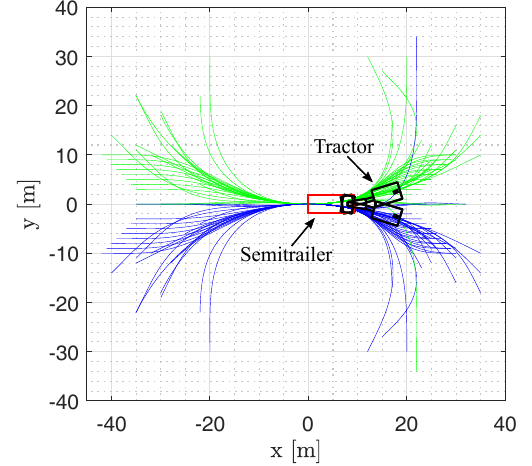}
		\label{j1:fig:all_prim_aID12_after_red}}
	\hfill
	\subfloat[][The set of motion primitives from $(\theta_{3,s},\alpha_{s})=(0,0)$ to different final states $z_f\in\mathbb Z_d$.]{%
		\includegraphics[width=0.47\textwidth]{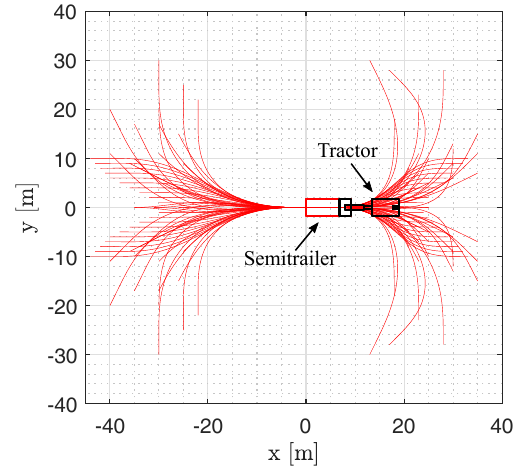}
		\label{j1:fig:all_prim_aID0_after_red}}
	\caption{The set of motion primitives from initial position of the semitrailer at the origin with orientation $\theta_{3,s}=0$ for different initial equilibrium configurations to different final states $z_f\in\mathbb Z_d$. The colored paths are the paths taken by the center of the axle of the semitrailer $(x_3,y_3)$ during the different motions.}
	\label{j1:fig:primitives}
\end{figure}

In this work, the OCP in~\eqref{j1:OCP_mp_gen} is solved by deploying the state-of-the-art numerical optimal control solver CasADi~\cite{casadi}, combined with the primal-dual interior-point solver IPOPT~\cite{ipopt}. 
Each generated motion primitive is represented as a distance sampled path in all vehicle states and control signals. 
Finally, since the system is orientation-invariant, rotational symmetries of the system are exploited\footnote{Essentially, it is only necessary to solve the OCPs from the initial orientations $\theta_{3,s}=0,\text{ }\arctan(1/2) \text{ and } \pi/4$. 
The motion primitives from the remaining initial orientations $\theta_{3,s}\in\Theta$ can be generated by mirroring the solutions.} to reduce the number of OCPs that need to be solved during the motion primitive generation~\cite{pivtoraiko2009differentially,CirilloIROS2014}. 

Even though the motion primitive generation is performed offline, it is not feasible to make an exhaustive generation of motion primitives to all grid points due to computation time and the high risk of creating redundant and undesirable segments. 
Instead, for each initial state $x_s^i\in\mathbb Z_d$ with position of the semitrailer at the origin, a careful selection of final states $x_f^i\in\mathbb Z_d$ is performed based on system knowledge and by visual inspection. 
The OCP solver is then only generating motion primitives from this specified set of OCPs. For our full-scale test vehicle, the set of motion primitives from all initial states with $\theta_{3,s}=0$, is illustrated in Figure~\ref{j1:fig:primitives}. 
The following can be noted regarding the manual specification of the motion primitive set:
\begin{itemize}
	\item[a)] A motion primitive $m_i\in \pazocal P$ is either a straight motion, a heading change maneuver or a parallel maneuver.
	\item[b)] The motion primitives in forward motion are more aggressive compared to the ones in backward motion, $i.e.$, a maneuver in forward motion has a shorter path distance compared to a similar maneuver in backward motion.
	\item[c)] The final position ($x^i_{3,f},y^i_{3,f}$) of motion primitive $m_i$ is selected such that the ratio between the stage-cost $J_p(m_i)$ and the path distance $s^i_f$ is sufficiently small, $i.e.$, such that the nominal path in all vehicle states and controls are sufficiently smooth to be executed by a path-following controller.
	\item[d)] While starting in a nonzero equilibrium configuration, the final position of the semitrailer ($x^i_{3,f},y^i_{3,f}$) is mainly restricted to the first and second quadrants for $\alpha^i_{s}=0.1$ and to the third and fourth quadrants for $\alpha^i_{s}=-0.1$.
\end{itemize}

\subsection{Efficiency improvements and online path planning}
\label{j1:subsec:Mreduction}
To improve the online planning time, the set of motion primitives $\pazocal P$ is reduced using the reduction technique presented in~\cite{CirilloIROS2014}. 
A motion primitive $m_i\in \pazocal P$ with stage-cost $J_p(m_i)$ is removed if its state transition \mbox{$z[k+1]=f_p(z[k],m_i)$} in free-space can be obtained by a combination of the other motion primitives in $\pazocal P$ with a combined total stage-cost $J_{\text{comb}}$ that satisfies $J_{\text{comb}}\leq \eta J_p(m_i)$, where $\eta\geq 1$ is a design parameter. 
This procedure can be used to reduce the size of the motion primitive set by choosing $\eta>1$, or by selecting $\eta = 1$ to verify that redundant motion primitives do not exist in $\pazocal P$. 

As previously mentioned, a heuristic function is used to guide the online search in the state lattice. The goal of the heuristic function is to perfectly estimate the cost-to-go at each vertex in the graph. 
In this work, we rely on a combination of two admissible heuristic functions: Euclidean distance and a free-space HLUT~\cite{knepper2006high}. The HLUT is generated offline using the techniques presented in~\cite{knepper2006high}. It is computed by solving several obstacle free path planning problems from all initial states $z_I\in\mathbb Z_d$ with position of the semitrailer at the origin, to all final states $z_G\in\mathbb Z_d$ with a specified maximum cut-off cost $J_{\text{cut}}$. 
As explained in~\cite{knepper2006high}, this computation step can be done efficiently by running a Dijkstra's algorithm from each initial state. 
During each Dijkstra's search, the optimal cost-to-come from explored vertices are simply recorded and stored in the HLUT. 
Moreover, in analogy to the motion primitive generation, the size of the HLUT is kept small by exploiting the position and orientation invariance properties of $\pazocal P$~\cite{knepper2006high,CirilloIROS2014}. 
The final heuristic function value used during the online graph search is the maximum of these two heuristics. As shown in~\cite{knepper2006high}, a HLUT significantly reduces the online planning time, since it takes the vehicle's nonholonomic constraints into account and enables perfect estimation of cost-to-go in free-space scenarios with no obstacles.  
\section{Path-following controller}
\label{j1:sec:Controller}
The motion plan received from the lattice planner is a feasible nominal path \mbox{$(x_r(s),u_r(s))$, $s \in[0,s_G]$} satisfying the time-scaled model of the G2T with a car-like tractor~\eqref{j1:eq:time_scaling}:
\begin{align}
\frac{\text dx_r}{\text ds} = v_{r}(s)f( x_r(s),\kappa_r(s)), \quad s \in[0, s_G], \label{j1:eq:tray:tractor}
\end{align}
where $x_r(s)$ is the nominal vehicle states for a specific $s$ and $u_r(s)=\begin{bmatrix}
v_r(s) & \kappa_r(s) \end{bmatrix}^T$ is the nominal velocity and curvature control signals.
The nominal path satisfies the system kinematics, its physically imposed constraints and moves the vehicle in free-space from the vehicle's initial state $x_r(0)=x_I$ to a desired goal state $x_r(s_G)=x_G$. 
Here, the nominal path is parametrized in $s$, which is the distance traveled by the rear axle of the car-like tractor. 
When backward motion tasks are considered and the axle of the semitrailer is to be controlled, it is more convenient to parameterize the nominal path in terms of distance traveled by the axle of the semitrailer $\tilde s$. 
Using the ratio $g_v>0$ defined in~\eqref{j1:relation_v_v3}, these different path parameterizations are related as  $\tilde s(s) = \int_0^{s}g_v(\beta_{2,r}(\tau),\beta_{3,r}(\tau),\kappa_r(\tau))\text d\tau$ and the nominal path~\eqref{j1:eq:tray:tractor} can equivalently be represented as 
\begin{align}\label{j1:eq:tray:semitrailer}
\frac{\text dx_r}{\text d\tilde s} = \frac{v_{r}(\tilde s)}{g_v(\beta_{2,r}(\tilde s),\beta_{3,r}(\tilde s),\kappa_{r}(\tilde s))}f(x_r(\tilde s),\kappa_r(\tilde s)),  \quad \tilde s \in[0,\tilde s_G],
\end{align}
where $\tilde s_G$ denotes the total distance of the nominal path taken by the axle of the semitrailer.
\begin{figure}[t!]	
	\vspace{-40pt}
	\begin{center}
		\includegraphics[width=0.7\linewidth]{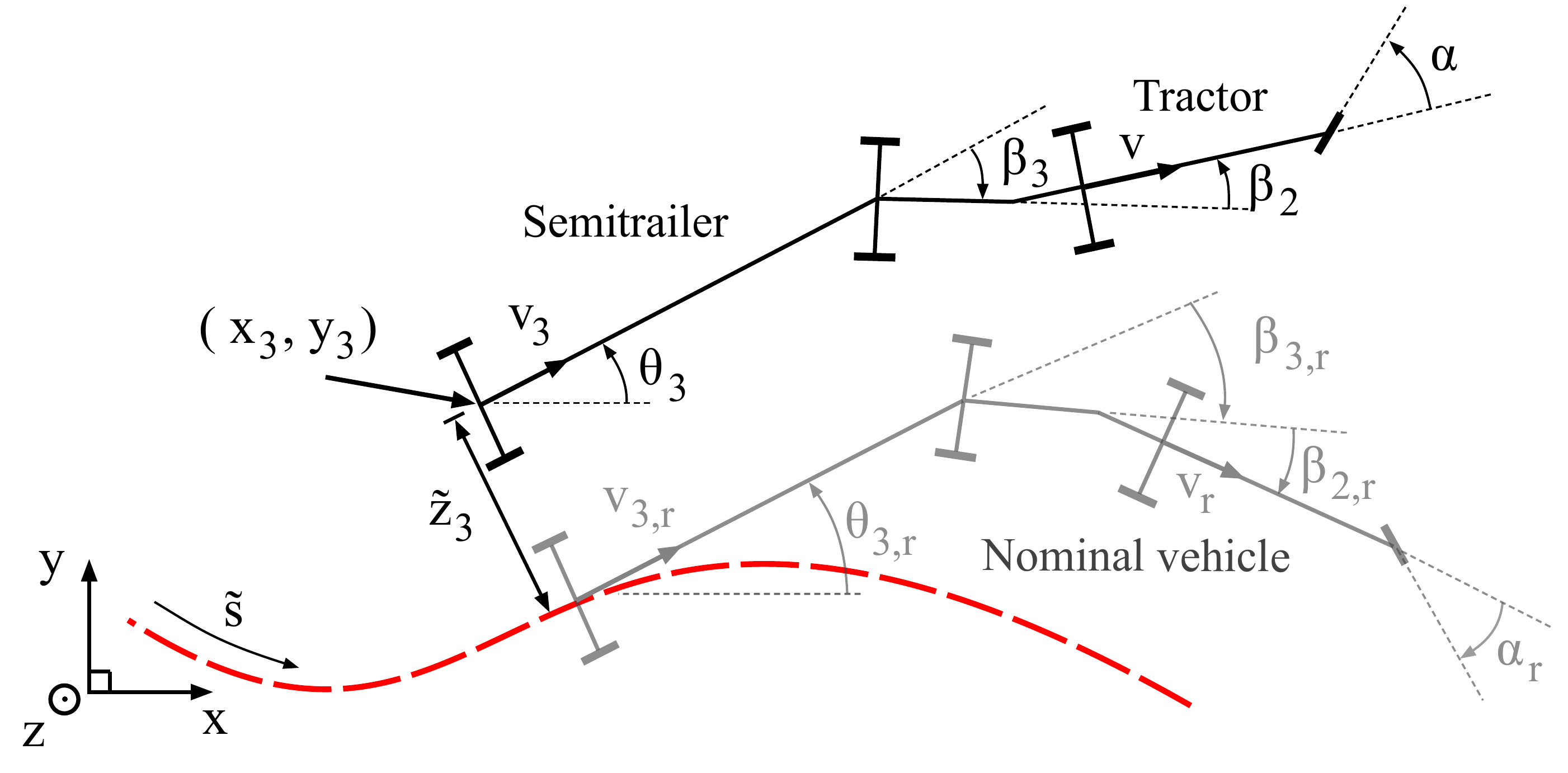}   
		\caption{An illustrative description of the Frenet frame with its moving coordinate system located at the orthogonal projection of the center of the axle of the semitrailer onto the reference path (dashed red curve) in the nominal position of the axle of the semitrailer $(x_{3,0}(\tilde s),y_{3,0}(\tilde s))$, $\tilde s \in [0,\tilde s_G]$. The black tractor-trailer system is the controlled vehicle and the gray tractor-trailer system is the nominal vehicle, or the desired vehicle configuration at this specific value of $\tilde s(t)$.} 
	\end{center} 
	\label{j1:fig:frenet_frame}
\end{figure}
According to the problem definition in Definition 2, the objective of the path-following controller is to stabilize the G2T with a car-like tractor~\eqref{j1:eq:model_global_coord} around this nominal path. 
It is done by first describing the controlled vehicle~\eqref{j1:eq:model_global_coord} in terms of deviation from the nominal path generated by the system in~\eqref{j1:eq:tray:semitrailer}, as depicted in Figure 7. 
During path execution, $\tilde s(t)$ is defined as the orthogonal projection of center of the axle of the semitrailer $(x_{3}(t),y_{3}(t))$ onto its nominal path $(x_{3,r}(\tilde s),y_{3,r}(\tilde s))$, $\tilde s\in[0,\tilde s_G]$ at time $t$:
\begin{align}\label{j1:eq:tildes_def}
\tilde s(t) = \argmin_{\tilde s\in[0,\tilde s_G]}  \left|\left|\begin{bmatrix} x_{3}(t)-x_{3,r}(\tilde s) \\ y_{3}(t)-y_{3,r}(\tilde s)\end{bmatrix}\right|\right|_2.
\end{align}
Using standard geometry, the curvature $\kappa_{3,r}(\tilde s)$ of the nominal path taken by the axle of the semitrailer is given by
\begin{align}
\kappa_{3,r}(\tilde s)=\frac{\text{d}\theta_{3,r}}{\text{d}\tilde s}=\frac{\tan\beta_{3,r}(\tilde s)}{L_3}, \quad  \tilde s \in[0,\tilde s_G].
\label{j1:eq:kappa3}
\end{align} 
Define $\tilde z_3(t)$ as the signed lateral distance between the center of the axle of the semitrailer $(x_3(t),y_3(t))$ and its projection to the nominal path in $(x_{3,r}(\tilde s),y_{3,r}(\tilde s))$, $\tilde s\in[0,\tilde s_G]$ at time $t$. 
Introduce the controlled curvature deviation as $\tilde \kappa(t)=\kappa(t)-\kappa_{r}(\tilde s(t))$, define the orientation error of the semitrailer as $\tilde\theta_3(t)=\theta_3(t)-\theta_{3,r}(\tilde s(t))$ and define the joint angular errors as $\tilde\beta_3(t)=\beta_3(t)-\beta_{3,r}(\tilde s(t))$ and 
$\tilde\beta_2(t)=\beta_2(t)-\beta_{2,r}(\tilde s(t))$, respectively. Define \mbox{$\Pi(a,b) = \{t\in\mathbb R_+ \mid a \leq \tilde s(t) \leq b \}$} as the time-interval when the distance traveled along the nominal path is between $a\in\mathbb R_+$ and $b\in\mathbb R_+$, where $0\leq a\leq b\leq \tilde s_G$. Then, using the Frenet-Serret formula, the distance traveled $\tilde s(t)$ along the nominal path and the signed lateral distance $\tilde z_3(t)$ to the nominal path can be modeled as:
\begin{subequations}
	\label{j1:eq:model_s_dot_sz}
	\begin{align}
	\dot {\tilde s} &= v_3  \frac{v_r\cos \tilde \theta_3}{1-\kappa_{3,r} \tilde z_3}, \quad t\in\Pi(0,\tilde s_G), \label{j1:eq:model_s1} \\
	\dot{\tilde z}_3 &= v_3\sin \tilde \theta_3 \label{j1:eq:model_s2}, \hspace{28pt}  t\in\Pi(0,\tilde s_G),
	\end{align}
\end{subequations}
where $v_3 = vg_v(\tilde \beta_{2}+\beta_{2,r},\tilde\beta_3 + \beta_{3,r}, \tilde \kappa+ \kappa_{r})$ and the dependencies of $\tilde s$ and $t$ are omitted for brevity. 
This transformation is valid in a tube around the nominal path in \mbox{$(x_{3,r}(\tilde s),y_{3,r}(\tilde s))$, $\tilde s\in[0,\tilde s_G]$} for which $\kappa_{3,r}\tilde z_3<1$. 
The width of this tube depends on the semitrailer's nominal curvature $\kappa_{3,r}$. When the nominal curvature tends to zero (a straight nominal path), $\tilde z_3$ can vary arbitrarily. 
Essentially, to avoid the singularities in the transformation, we must have that $|\tilde z_3| < |\kappa^{-1}_{3,r}|$, when $\tilde z_3$ and $\kappa_{3,r}$ have the same sign. 
Note that $v_r\in\{-1,1\}$ is included in~\eqref{j1:eq:model_s1} to make $\tilde s(t)$ a monotonically increasing function in time during tracking of nominal paths in both forward and backward motion. 
Here, it is assumed that the longitudinal velocity of the tractor $v(t)$ is chosen such that $\text{sign}(v(t))=v_r(\tilde s(t))$ and it is assumed that the orientation error of the semitrailer satisfies \mbox{$|\tilde \theta_3|<\pi/2$}. 
With the above assumptions, $\dot {\tilde s}(t)>0$ during path following of nominal paths in both forward and backward motion.

The models for the remaining path-following error states $\tilde\theta_3(t)$, $\tilde\beta_3(t)$ and $\tilde\beta_2(t)$ are derived by applying the chain rule, together with equations~\eqref{j1:eq:model_global_coord}--\eqref{j1:relation_v_v3},~\eqref{j1:eq:tray:semitrailer} and~\eqref{j1:eq:model_s1}:
\begin{subequations}
	\label{j1:eq:model_s}
	\begin{align} 
	\dot{\tilde\theta}_3 =& v_3 \left( \frac{\tan(\tilde{\beta}_3+\beta_{3,r})}{L_3} - \frac{\kappa_{3,r}\cos \tilde \theta_3}{1-\kappa_{3,r}\tilde z_3} \right), \hspace{61pt}  t\in\Pi(0,\tilde s_G), \label{j1:eq:model_s3} \\
	\dot{\tilde \beta}_3 =& v_3 \left(\frac{\sin(\tilde \beta_2+\beta_{2,r})-M_1\cos(\tilde \beta_2+\beta_{2,r}) (\tilde \kappa+ \kappa_r)}{L_2\cos(\tilde \beta_3+\beta_{3,r}) C_1(\tilde \beta_2+\beta_{2,r}, \tilde \kappa+ \kappa_r)} - \frac{\tan(\tilde \beta_3+\beta_{3,r})}{L_3} \nonumber \right. \\
	&\left. -\frac{\cos{\tilde{\theta}_3}}{1-\kappa_{3,r}\tilde z_3}\left(\frac{\sin\beta_{2,r} -M_1 \cos\beta_{2,r}\kappa_r}{L_2\cos\beta_{3,r} C_1(\beta_{2,r},\kappa_r)}-\kappa_{3,r}\right)\right), \quad  t\in\Pi(0,\tilde s_G),
	\label{j1:eq:model_s4} \\
	\dot{\tilde \beta}_2 =& v_3\left( \left( \frac{\tilde \kappa+ \kappa_r - \frac{\sin(\tilde \beta_2+\beta_{2,r})}{L_2} + \frac{M_1}{L_2}\cos(\tilde \beta_2+\beta_{2,r})(\tilde \kappa+ \kappa_r)}{\cos(\tilde \beta_3+\beta_{3,r}) C_1(\tilde \beta_2+\beta_{2,r}, \tilde \kappa+ \kappa_r)}\right) \nonumber \right. \\
	&\left. -\frac{\cos{\tilde{\theta}_3}}{1-\kappa_{3,r}\tilde z_3}\left( \frac{\kappa_r - \frac{\sin \beta_{2,r}}{L_2} + \frac{M_1}{L_2}\cos \beta_{2,r}\kappa_r}{\cos \beta_{3,r} C_1(\beta_{2,r}, \kappa_r)}\right)\right), \hspace{25pt}  t\in\Pi(0,\tilde s_G).\label{j1:eq:model_s5}
	\end{align}
\end{subequations}
A more detailed derivation of~\eqref{j1:eq:model_s} is provided in Appendix A. Together, the differential equations in~\eqref{j1:eq:model_s_dot_sz} and~\eqref{j1:eq:model_s} describe the model of the G2T with a car-like tractor~\eqref{j1:eq:model_global_coord} in terms of deviation from the nominal path generated by the system in~\eqref{j1:eq:tray:semitrailer}. 

When path-following control is considered, the speed at which the nominal path~\eqref{j1:eq:tray:tractor} is executed is not considered, but only that it is followed with a small path-following error. 
This means that the distance traveled $\tilde s(t)$ along the nominal path is not explicitly controlled by the path-following controller. 
However, the dependency of $\tilde s$ in~\eqref{j1:eq:model_s2} and \eqref{j1:eq:model_s} makes the nonlinear system distance-varying. 
Define the path-following error states as $\tilde x_e = \begin{bmatrix} \tilde z_3 & \tilde\theta_3 & \tilde\beta_3 & \tilde\beta_2\end{bmatrix}^T$, where its model is given by~\eqref{j1:eq:model_s2}--\eqref{j1:eq:model_s}. By replacing $v_3$ with $v$ using the relationship defined in~\eqref{j1:relation_v_v3}, the path-following error model~\eqref{j1:eq:model_s2}--\eqref{j1:eq:model_s} and the progression along the nominal path~\eqref{j1:eq:model_s1}, can compactly be expressed as (see Appendix A)
\begin{subequations}
	\label{j1:eq:error_model_and_progression}
	\begin{align}
	\dot{\tilde s} &= vf_{\tilde s}(\tilde s,\tilde x_e), \hspace{18pt}  t\in\Pi(0,\tilde s_G), \label{j1:eq:progression_compact} \\
	\dot{\tilde x}_e &= v \tilde f(\tilde s, \tilde x_e, \tilde \kappa),\quad  t\in\Pi(0,\tilde s_G),
	\label{j1:eq:error_model_compact}
	\end{align}
\end{subequations}
where $\tilde f(\tilde s, 0, 0)=0$, $\forall t\in\Pi(0,\tilde s_G)$, \textit{i.e.}, the origin $(\tilde x_e,\tilde \kappa)=(0,0)$ is an equilibrium point. 
Since $v$ enters linearly in~\eqref{j1:eq:error_model_and_progression}, in analogy to~\eqref{j1:eq:time_scaling}, time-scaling~\cite{sampei1986time} can be applied to eliminate the speed dependence $|v|$ from the model. 
Therefore, without loss of generality, it is hereafter assumed that the longitudinal velocity of the rear axle of the tractor is chosen as $v(t)=v_r(\tilde s(t))\in\{-1,1\}$ which implies that $\dot{\tilde s}(t) > 0$. 
Moreover, from the construction of the set of motion primitives $\pazocal P$, each motion primitive $m_i\in\pazocal P$ encodes a forward or backward motion segment (see Section~\ref{j1:subsec:MPrimitiveGen}). 

\subsection{Local behavior around a nominal path}
The path-following error model in~\eqref{j1:eq:model_s2} and \eqref{j1:eq:model_s} can be linearized around the nominal path \mbox{$(x_r(\tilde s),u_r(\tilde s))$, $\tilde s \in[0,\tilde s_G]$} by equivalently linearizing~\eqref{j1:eq:error_model_compact} around the origin $(\tilde x_e, \tilde \kappa) = (0,0)$. 
The origin is by construction an equilibrium point to~\eqref{j1:eq:error_model_compact} and hence a first-order Taylor series expansion yields
\begin{align}
\dot{\tilde x}_e = vA(\tilde s(t))\tilde x_e + vB(\tilde s(t))\tilde{\kappa},\quad  t\in\Pi(0,\tilde s_G).
\label{j1:eq:lin_sys} 
\end{align}
For the special case when the nominal path moves the system either straight forwards or backwards, the matrices $A$ and $B$ simplify to
\begin{align}
A = \begin{bmatrix}
0 & 1 & 0 & 0 \\
0 & 0 & \frac{1}{L_3} & 0 \\
0 & 0 & -\frac{1}{L_3} & \frac{1}{L_2} \\[2pt]
0 & 0 & 0 & -\frac{1}{L_2} \\
\end{bmatrix},
\quad B= \begin{bmatrix}
0 \\ 0 \\ -\frac{M_1}{L_2} \\[3pt] \frac{L_2 + M_1}{L_2}
\end{bmatrix},	
\label{j1:eq:lin_AB}
\end{align}
and the characteristic polynomial is
\begin{align}
\det{(\lambda I-vA)}=v^2\lambda^2\left(\lambda+\frac{v}{L_3}\right)\left(\lambda+\frac{v}{L_2}\right).
\end{align} 
Thus, around a straight nominal path, the linearized system in~\eqref{j1:eq:lin_sys} is marginally stable in forward motion \mbox{($v>0$)} because of the double integrator and unstable in backward motion \mbox{($v<0$)}, since the system has two poles in the right half plane. 
Due to the positive off-axle hitching $M_1>0$, the linearized system has a zero in some of the output channels~\cite{hybridcontrol2001,CascadeNtrailernonmin}. 
In forward motion, the system has non-minimum phase properties since the zero is located in the right half-plane (see~\cite{CascadeNtrailernonmin} for an extensive analysis).
In backward motion, this zero is located in the left half-plane and the system is instead minimum phase. 

In the sequel, we focus on stabilizing the path-following error model~\eqref{j1:eq:error_model_compact} in some neighborhood around the origin $(\tilde x_e,\tilde \kappa)=(0,0)$. 
This is done by utilizing the framework presented in~\cite{LjungqvistACC2018}, where the closed-loop system consisting of the controlled vehicle and the path-following controller, executing a nominal path computed by a lattice planner, is first modeled as a hybrid system. 
The framework is tailored for the lattice-based path planner considered in this work and is motivated because it is well-known from the theory of hybrid systems that switching between stable systems in an inappropriate way can lead to instability of the switched system~\cite{decarlo2000perspectives,pettersson96}.

\subsection{Connection to hybrid systems}\label{j1:sec:connection}
The nominal path~\eqref{j1:eq:tray:semitrailer} is computed online by the lattice planner and is thus a priori unknown. However, it is composed of a finite sequence of precomputed motion primitives $\{m_{u_q[k]}\}^{N-1}_{k=0}$ of length $N$. 
Each motion primitive $m_i$ is chosen from the set of $M$ possible motion primitives, \textit{i.e.}, $m_i\in\pazocal P$. Along motion primitive $m_i\in\pazocal P$, the nominal path is represented as $(x^i_r(\tilde s),u^i_r(\tilde s)),$ $\tilde s\in[0,\tilde s_f^i]$ and the path-following error model~\eqref{j1:eq:error_model_compact} becomes
\begin{align}\label{j1:eq:error_model_mi}
\dot{\tilde x}_e = v_r(\tilde s) \tilde f_i(\tilde s, \tilde x_e, \tilde \kappa),\quad t\in\Pi(0,\tilde s^i_f).
\end{align}
From the fact that the sequence of motion primitives is selected by the lattice planner, it follows that the system can be descried as a hybrid system. 
Define $q : \mathbb [0,\tilde s_G] \rightarrow \{1,\hdots,M\}$ as a piecewise integer-valued signal that is selected by the lattice planner. 
Then, the path-following error model can be written as a distance-switched continuous-time hybrid system
\begin{align}\label{j1:eq:error_model_hybrid}
\dot {\tilde x}_e = v_r(\tilde s)\tilde f_{q(\tilde s)}(\tilde s, \tilde x_e, \tilde \kappa), \quad t\in\Pi(0,\tilde s_G).
\end{align} 
This hybrid system is composed of $M$ different subsystems, where only one subsystem is active for each $\tilde s\in[0,\tilde s_G]$. 
Here, $q(\tilde s)$ is assumed to be right-continuous and from the construction of the motion primitives, it holds that there are finitely many switches in finite distance~\cite{decarlo2000perspectives,pettersson96}. 
We now turn to the problem of designing the hybrid path-following controller $\tilde\kappa = g_{q(\tilde s)}(\tilde x_e)$, such that the path-following error is upper bounded by an exponentially decaying function during the execution of each motion primitive $m_i\in\pazocal P$, individually.

\subsection{Design of the hybrid path-following controller}\label{j1:sec:feedback_design}
\label{j1:sec:lowlevelcontrol}
The synthesis of the path-following controller is performed separately for each motion primitive $m_i\in\pazocal P$. 
The class of hybrid path-following controllers is limited to piecewise linear state-feedback controllers with feedforward action. 
Denote the path-following controller dedicated for motion primitive $m_i\in\pazocal P$
as $\kappa (t) = \kappa_r(\tilde s(t))+K_i \tilde x_e(t)$. 
When applying this control law to the path-following error model in~\eqref{j1:eq:error_model_mi}, the nonlinear closed-loop system can, in a compact form, be written as
\begin{align}\label{j1:eq:error_model_mi_cl}
\dot{\tilde x}_e = v_r(\tilde s)\tilde f_i(\tilde s, \tilde x_e, K_i \tilde x_e) = v_r(\tilde s)\tilde f_{cl,i}(\tilde s, \tilde x_e), \quad  t\in\Pi(0,\tilde s^i_f),
\end{align}
where $\tilde x_e = 0$ is an equilibrium point, since $f_{cl,i}(\tilde s, 0)=\tilde f_{i}(\tilde s, 0, 0)=0$, $\forall \tilde s\in[0,\tilde s^i_f]$. 
The state-feedback controller $\tilde \kappa = K_i \tilde x_e$ is intended to be designed such that the path-following error is locally bounded and decays towards zero during the execution of $m_i\in\pazocal P$. This is guaranteed by Theorem~\ref{j1:T3}.
\begin{assumption}
	Assume $\tilde f_{cl,i}:[0,\tilde s_f^i] \times \tilde{\mathbb{X}}_e \rightarrow \mathbb R^4$ is continuously differentiable with respect to $\tilde x_e \in \tilde{\mathbb{X}}_e = \{ \tilde x_e \in \mathbb R^4 \mid \|\tilde x_e\|_2 < r \}$ and the Jacobian matrix  $[\partial f_{cl,i} / \partial \tilde{x}_e]$ is bounded and Lipschitz on $\tilde{\mathbb{X}}_e$, uniformly in $\tilde s\in [0,\tilde s_f^i]$.
	\label{j1:A2}
\end{assumption}
\vspace{6pt}
\begin{theorem}[\cite{LjungqvistACC2018}]
	Consider the closed-loop system in~\eqref{j1:eq:error_model_mi_cl}. Under Assumption~\ref{j1:A2}, let
	\begin{align}
	A_{cl,i}(\tilde s)=v_r(\tilde s)\frac{\partial \tilde f_{cl,i}}{\partial \tilde{x}_e}(\tilde s,0).
	\label{j1:Acli}
	\end{align} 
	If there exist a common matrix $ P_i\succ 0$ and a positive constant $\epsilon$ that satisfy
	\begin{align}
	A_{cl,i}(\tilde s)^{T}  P_i +P_iA_{cl,i}(\tilde s) \preceq -2\epsilon P_i \quad \forall \tilde s \in [0,\tilde s_f^i]. \label{j1:eq:lyap}
	\end{align} 
	Then, the following inequality holds 
	\begin{align}
	\label{j1:convergece_LTV}
	||\tilde x_e(t)|| \leq \rho_i||\tilde x_e(0)|| e^{-\epsilon t},\quad \forall t\in\Pi(0,\tilde s^i_f),
	\end{align}
	where $\rho_i=\text{Cond}(P_i)$ is the condition number of $ P_i$.
	\label{j1:T3}
\end{theorem}
\vspace{6pt}
\begin{proof} See, \textit{e.g.},~\cite{khalil}.
\end{proof}
Theorem~\ref{j1:T3} guarantees that if the feedback gain $K_i$ is designed such that there exists a quadratic Lyapunov function $V_i(\tilde x_e) = \tilde x_e^T P_i \tilde x_e$ for~\eqref{j1:eq:error_model_mi_cl} around the origin satisfying $\dot V_i \leq -2\epsilon V_i$, then a small disturbance in the initial path-following error $\tilde x_e(0)$ results in a path-following error state trajectory $\tilde x_e(t)$ whose norm is upper bounded by an exponentially decaying function. In analogy to~\cite{LjungqvistACC2018}, the condition in~\eqref{j1:eq:lyap} can be reformulated as a controller synthesis problem using linear matrix inequality (LMI) techniques. 
By using the chain rule, the matrix $A_{cl,i}(\tilde s)$ in~\eqref{j1:Acli} can be written as
\begin{align}\label{j1:eq:linearizaion_Acl}
A_{cl,i}(\tilde s) &= v_r(\tilde s)\frac{\partial \tilde f_i}{\partial \tilde x}(\tilde s,0,0) + v_r(\tilde s)\frac{\partial \tilde f_i}{\partial \tilde \kappa}(\tilde s,0,0)K_i \triangleq A_i(\tilde s)+B_i(\tilde s)K_i.
\end{align}      
Furthermore, assume the pairs $[A_i(\tilde s),B_i(\tilde s)]$ lie in the convex polytope $\mathbb S_i$, $\forall \tilde s\in[0,\tilde s^i_f]$, where $\mathbb S_i$  is represented by its $L_i$ vertices
\begin{align}
[A_i(\tilde s),B_i(\tilde s)] \in \mathbb S_i = \textbf{Co} \left\{[A_{i,1},B_{i,1}],\hdots,[A_{i,L_i},B_{i,L_i}] \right\},
\label{j1:def:polytope}
\end{align} 
where \textbf{Co} denotes the convex hull. Now, condition~\eqref{j1:eq:lyap} in Theorem~\ref{j1:T3} can be reformulated as~\cite{boyd1994linear}:
\begin{align}
\label{j1:matrixineq_nonconvex}
(A_{i,j}+B_{i,j}K_i)^TP_i + P_i(A_{i,j}+B_{i,j}K_i) \preceq -2\epsilon P_i, \quad j=1,\hdots,L_i.
\end{align}   
This matrix inequality is not jointly convex in $P_i$ and $K_i$. However, if $\epsilon>0$ is fixed, using the bijective transformation $Q_i=P_i^{-1}\succ 0$ and $Y_i=K_iP_i^{-1}\in \mathbb R ^{1\times 4}$, the matrix inequality in~\eqref{j1:matrixineq_nonconvex} can be rewritten as an LMI in $Q_i$ and $Y_i$~\cite{wolkowicz2012handbook}:
\begin{align}
\label{j1:eq:matrixineq_convex}
Q_iA_{i,j}^T + Y_i^T B_{i,j}^T + A_{i,j}Q_i+B_{i,j}Y_i + 2\epsilon Q_i \preceq 0, \quad j=1,\hdots,L_i.
\end{align} 
Hence, it is an LMI feasibility problem to find a linear state-feedback controller that satisfies condition~\eqref{j1:eq:lyap} in Theorem~\ref{j1:T3}. 
If $Q_i$ and $Y_i$ are feasible solutions to~\eqref{j1:eq:matrixineq_convex}, the quadratic Lyapunov function is $V_i(\tilde x) = \tilde x^T Q_i^{-1}\tilde x$ and the linear state-feedback controller is $\tilde \kappa = Y_iQ_i^{-1}\tilde x_e$. 
As in~\cite{LjungqvistACC2018}, the LMI feasibility problem in~\eqref{j1:eq:matrixineq_convex} is reformulated as a semidefinite programming (SDP) problem
\begin{align}
\minimize_{Y_{i}, Q_{i}} \hspace{3.7ex} & \|Y_{i}-K_{\text{nom}}^iQ_{i}\| \label{j1:eq:opt_LTV}\\
\subjectto\hspace{3ex}
& \eqref{j1:eq:matrixineq_convex} \text{ and } Q_i \succeq I,  \nonumber
\end{align}
where $K_{\text{nom}}^i$ is a nominal feedback gain that depends on $m_i\in\pazocal P$. 
Here, two nominal feedback gains are used; $K_{\text{nom}}^i=K_{\text{fwd}}$ for all forward motion primitives $m_i\in\pazocal P_{\text{fwd}}$ and $K_{\text{nom}}^i=K_{\text{rev}}$ for all backward motion primitives $m_i\in\pazocal P_{\text{rev}}$.
The motivation for this choice of objective function in~\eqref{j1:eq:opt_LTV} is that it is desired that the path-following controller inherits the nominal controller's properties. 
It is also used to reduce the number of different feedback gains, while not sacrificing desired convergence properties of the path-following error along the execution of each motion primitive. 
The nominal feedback gains $K_{\text{fwd}}$ and $K_{\text{rev}}$ are designed using infinite-horizon LQ-control~\cite{anderson2007optimal} where the path-following error model has been linearized around a straight nominal path in backward and forward motion, respectively. 
In these cases, the Jacobian linearization is given by the matrices $A$ and $B$ defined in~\eqref{j1:eq:lin_AB}. After these nominal feedback gains have been designed, the optimization problem in~\eqref{j1:eq:opt_LTV} can be solved separately for each motion primitive, e.g., using YALMIP~\cite{lofberg2004yalmip}. 
In this specific application, for all $m_i\in\pazocal P$, the optimal value of the objective function in~\eqref{j1:eq:opt_LTV} is zero, which implies that $K_i = K_{\text{nom}}^i$ since $Q_i\succ 0$. 
Thus, for this specific set of motion primitives $\pazocal P$ (see Figure~\ref{j1:fig:primitives}), the hybrid path-following controller $\tilde \kappa = K_{q(\tilde s)}\tilde x_e$ simplifies to
\begin{align}\label{j1:eq:hybrid_controller}
\kappa(t) = \kappa_r(\tilde s) + \begin{cases}
K_{\text{fwd}}\tilde x_e(t), \quad &m_i\in\pazocal P_{\text{fwd}}, \\
K_{\text{rev}}\tilde x_e(t), \quad &m_i\in\pazocal P_{\text{rev}},
\end{cases}
\end{align}
where $\kappa_r(\tilde s)$ is the feedforward computed by the lattice planner.
Note that if a common quadratic Lyapunov function exists that satisfies~\eqref{j1:eq:matrixineq_convex} $\forall m_i \in \pazocal P$ (\textit{i.e.}, $Q_i=Q$, but $Y_i$ can vary), then the path-following error is guaranteed to exponentially decay towards zero under an arbitrary sequence of motion primitives~\cite{boyd1994linear,decarlo2000perspectives}.
This is however not possible since the path-following error model~\eqref{j1:eq:error_model_hybrid} is underactuated\footnote{Here, a system is defined underactuated if the number of control signals is strictly less than the dimension of its configuration space~\cite{CascadeNtrailernonmin}.} and the Jacobian linearization takes on the form in~\eqref{j1:eq:lin_sys}. 
\begin{theorem}[\cite{LjungqvistACC2018}]
	\label{j1:P1}
	Consider the switched linear system
	\begin{align}
	\dot x = vAx+vBu, \quad v\in\{ -1, 1\},
	\label{j1.eq:switched_lin_v_sys}
	\end{align}
	where $A\in \mathbb R^{n\times n}$ and $B\in \mathbb R^{n\times m}$. When $\text{rank}(B)<n$, there exists no hybrid linear state-feedback control law in the form
	\begin{align}
	u=\begin{cases}
	K_{1}x, \quad v = 1 \\
	K_{2}x, \quad v = -1 \\
	\end{cases},
	\label{j1:hybrid_ctrl}
	\end{align} 
	where $K_1\in\mathbb R^{m \times n}$ and $K_2\in\mathbb R^{m \times n}$, such that the closed-loop system is quadratically stable with a quadratic Lyapunov function $V(x)=x^TPx$, $\dot V(x) < 0$ and $ P\succ 0$.
\end{theorem}
\begin{proof}
	See \cite{LjungqvistACC2018}.
\end{proof}
From Theorem~\ref{j1:P1}, it is clear that it is not possible to design a hybrid path-following controller $\tilde \kappa = K_{q(\tilde s)}\tilde x_e$ such that the closed-loop path-following error system is locally quadratically stable along nominal paths that are composed of forward and backward motion primitives.
In the next section, a systematic method is presented for analyzing the behavior of the distance-switched continuous-time hybrid system~\eqref{j1:eq:error_model_hybrid}, when the hybrid path-following controller already has been designed.  
\subsection{Convergence along a combination of motion primitives}\label{j1:sec:convergence}
Consider the path-following error model in~\eqref{j1:eq:error_model_hybrid} with the hybrid path-following controller $\tilde\kappa= K_{q(\tilde s)}\tilde x_e$ that has been designed following the steps presented in Section~\ref{j1:sec:lowlevelcontrol}. 
Assume motion primitive $m_i\in\pazocal P$ is switched in at distance $\tilde s_k$, $i.e.$, $q(\tilde s(t))=i$, for all $t\in\Pi(\tilde s_k, \tilde s_k + \tilde s_f^i)$. 
We are now interested in analyzing the evolution of the path-following error $\tilde x_e(t)$ during the execution of this motion primitive. 
Since the longitudinal velocity of the tractor is selected as $v(t)=v_r(\tilde s(t))$, then $\dot{\tilde s}(t)>0$ and it is possible to eliminate the time-dependency in the path-following error model~\eqref{j1:eq:error_model_compact}. 
By applying the chain rule, we get $\frac{\text d\tilde x_e}{\text d\tilde s}=\frac{\text d\tilde x_e}{\text dt}\frac{\text dt}{\text d\tilde s}=\frac{\text d\tilde x_e}{\text dt}\frac{1}{\dot{\tilde s}}$. 
Hence, using~\eqref{j1:eq:progression_compact}, the distance-based version of the path-following error model~\eqref{j1:eq:error_model_compact} can be represented as 
\begin{align}
\frac{\text d\tilde x_e}{\text d\tilde s}  = \frac{ f_{cl,i}(\tilde s, \tilde x_e(\tilde s))}{f_s(\tilde s,\tilde x_e(\tilde s))}, \quad \tilde s \in[\tilde s_k, \tilde s_k + \tilde s_f^i], \label{j1:eq:error_states_distance}
\end{align}
where $\tilde x_e(\tilde s_k)$ is given. The evolution of the path-following error $\tilde x_e(\tilde s)$ becomes
\begin{align}
\tilde x_e(\tilde s_k+\tilde s^i_f) = \tilde x_e(\tilde s_k) + \bigintsss_{\tilde s_k}^{\tilde s_k+\tilde s_f^i}\frac{ f_{cl,i}(\tilde s, \tilde x_e(\tilde s))}{f_s(\tilde s,\tilde x_e(\tilde s))}\text d\tilde s \triangleq  T_i(\tilde x_e(\tilde s_k)),
\label{j1:eq:error_states_lattice}
\end{align} 
where $\tilde x_e(s_k)$ denotes the path-following error when motion primitive $m_i\in\pazocal P$ is started and $\tilde x_e(\tilde s_k+\tilde s^i_f)$ denotes the path-following error when the execution of $m_i$ is finished. 
The solution to the integral in~\eqref{j1:eq:error_states_lattice} has no analytical expression. 
However, numerical integration can be used to compute a local approximation of the evolution of $\tilde x_e(\tilde s)$ between the two switching points $\tilde s_k$ and $\tilde s_k + \tilde s_f^i$. 
A first-order Taylor series expansion of~\eqref{j1:eq:error_states_lattice} around the origin $\tilde x_e(\tilde s_k) = 0$ yields
\begin{align}
\tilde x_e(\tilde s_k+\tilde s^i_f) =  T_i(0) + \underbrace{\left.\frac{\text dT_i(\tilde x_e(\tilde s_k))}{\text d\tilde x_e(\tilde s_k)}\right|_{(0)}}_{=F_i}\tilde x_e(\tilde s_k).
\label{j1:eq:lin_disc}
\end{align}
The term $ T_i(0)=0$, since $\tilde f_{cl,i}(\tilde s,0) = 0$, $\forall\tilde s\in[\tilde s_k, \tilde s_k + \tilde s_f^i]$.
Denote $\tilde x_e[k] = \tilde x_e (\tilde s_k)$, $\tilde x_e[k+1] = \tilde x_e(\tilde s_k+ \tilde s_f^i)$ and $u_q[k] = q(\tilde s_k) = i$. 
By, \textit{e.g.}, the use of finite differences, the evolution of the path-following error~\eqref{j1:eq:error_states_lattice} after motion primitive $m_i\in\pazocal P$ has be executed can be approximated as a linear discrete-time system
\begin{align}
\tilde x_e[k+1] = F_i \tilde x_e[k].
\label{j1:eq:lin_disc_transition}
\end{align}
Repeating this procedure for all $M$ motion primitives, a set of $M$ transition matrices $\mathbb F=\{F_1,\hdots,F_M\}$ can be computed. Then, the local evolution of the path-following error~\eqref{j1:eq:error_states_lattice} between each switching point can be described as a linear discrete-time switched system
\begin{align}
\tilde x_e[k+1]=F_{u_q[k]}\tilde x_e[k], \quad u_q[k]\in\{1,\hdots,M\},
\label{j1:eq:swithing_system}
\end{align} 
where the motion primitive sequence $\{u_q[k]\}_{k=0}^{N-1}$ and its length $N$ are unknown at the time of the analysis. Exponential decay of the solution $\tilde x[k]$ to~\eqref{j1:eq:swithing_system} is guaranteed by Theorem~\ref{j1:T4}.
\begin{theorem}[\cite{LjungqvistACC2018}]
	\label{j1:T4} 
	Consider the linear discrete-time switched system in \eqref{j1:eq:swithing_system}. If there exist a matrix $ S\succ 0$ and a $\eta\geq 1$ that satisfy
	\begin{subequations}
		\label{j1:eq:dlmitotal}
		\begin{align}
		I \preceq  S &\preceq \eta I \label{j1:cond_S},\\
		\label{j1:dlmi}
		F_j^T  S F_j - S &\preceq - \mu  S, \quad \forall j \in \{ 1,\hdots,M\},
		\end{align}
	\end{subequations}
	where $0<\mu<1$ is a constant. Then, under arbitrary switching for $k \geq 0$ the following inequality holds
	\begin{align}
	\label{ineq:discretetime}
	\|\tilde x_e[k]\| \leq \|\tilde x_e[0]\|\eta^{1/2}\lambda^{k},
	\end{align}
	where $\lambda = \sqrt{1 - \mu}$ and $\eta=\text{Cond}(S)$ denotes the condition number of $S$.  
\end{theorem}
\begin{proof}
	See~\cite{LjungqvistACC2018}.
\end{proof}
Note that for a fixed $\mu$,~\eqref{j1:eq:dlmitotal} is a set of LMIs in the variables $S$ and $\eta$. 
The result in Theorem~\ref{j1:T4} establishes that the upper bound on the path-following error at the switching points exponentially decays towards zero. Thus, the norm of the initial path-following error $\Vert \tilde x_e(\tilde s_k)\Vert$, when starting the execution of a new motion primitive, will decrease as $k$ grows. 
Moreover, combining Theorem~\ref{j1:T3} and Theorem~\ref{j1:T4}, this implies that the upper bound on the continuous-time path-following error $\Vert \tilde x_e(t)\Vert$ will exponentially decay towards zero. This result is formalized in Corollary~\ref{j1:C1}.
\begin{corollary}[\cite{LjungqvistACC2018}]
	\label{j1:C1}
	Consider the hybrid system in~\eqref{j1:eq:error_model_hybrid} with the path-following controller $\tilde \kappa = K_{q(\tilde s)}\tilde x_e$. Assume the conditions in Theorem~\ref{j1:T3} are satisfied for each mode $i\in\{1,\hdots,M\}$ of \eqref{j1:eq:error_model_hybrid} and assume the conditions in Theorem~\ref{j1:T4} are satisfied for the resulting discrete-time switched system~\eqref{j1:eq:swithing_system}. Then, $\forall k\in\mathbb Z_{+}$ and $t\in\Pi(\tilde s_k,\tilde s_k+\tilde s_f^i)$ with $q(\tilde s(t))=i$, the continuous-time path-following error $\tilde x_e(t)$ satisfies 
	\begin{align}
	\lVert\tilde x_e(t)\rVert\leq \lVert\tilde x_e(t_0)\rVert\eta^{1/2}\rho_i^{1/2}\lambda^{k},
	\end{align} 
	where $0<\lambda<1$, $\eta=\text{Cond}(S)$ and $\rho_i=\text{Cond}(P_i)$.  
\end{corollary}
\begin{proof}
	See \cite{LjungqvistACC2018}.
\end{proof}
The practical interpretation of Corollary~\ref{j1:C1} is that the upper bound on the continuous-time path-following error is guaranteed to exponentially decay towards zero as a function of the number of executed motion primitives. The analysis method presented in this section will be later used for this specific application in Section~\ref{j1:sec:StabilityAnalysis}. 

In this application, none of the vehicle states are directly observed from the vehicle's onboard sensors and we instead need to rely on dynamic output feedback~\cite{rugh1996linear}, \textit{i.e.}, the hybrid state-feedback controller $\tilde\kappa = K_{q(\tilde s)}\tilde x_e$ is operating in series with a nonlinear observer. 
Naturally, the observer is operating in a discrete-time fashion and we make the assumption that the observer is operating sufficiently fast and estimates the state $\hat x(t_k)$ with good accuracy. 
This means that it is further assumed that the separation principle of estimation and control holds. 
That is, the current state estimate from the observer $\hat x(t_k)$ is interpreted as the true vehicle state $x(t_k)$, which is then used to construct the path-following error $\tilde x_e(t_k)$ used by the hybrid path-following controller.

\section{State observer}
\label{j1:sec:stateEstimation}
The state-vector $x=\begin{bmatrix} x_3 & y_3 & \theta_3 & \beta_3 & \beta_2 \end{bmatrix}^T$ for the G2T with a car-like tractor is not directly observed from the sensors on the car-like tractor and therefore needs to be inferred using the available measurements, the vehicle model~\eqref{j1:eq:model_global_coord} and the geometry of the vehicle. 

High accuracy measurements of the position of the rear axle of the car-like tractor $(x_1,y_1)$ and its orientation $\theta_1$ are obtained from the localization system that was briefly described in Section~\ref{j1:sec:loc}. 
To obtain information about the joint angles $\beta_2$ and $\beta_3$, a LIDAR sensor is mounted in the rear of the tractor as illustrated in Figure~\ref{j1:fig:ransac_meas}. 
This sensor provides a point-cloud from which the $y$-coordinate $L_y$, given in the tractor's local coordinate system, of the midpoint of the semitrailer's front and the relative orientation $\phi$ between the tractor and semitrailer can be extracted\footnote{Other features could be extracted from the point cloud, but using $L_y$ and $\phi$ have shown to yield good performance in practice.}. 
To estimate $L_y$ and $\phi$, an iterative RANSAC algorithm~\cite{fischler1981random} is first used to find the visible edges of the semitrailer's body. 
Logical reasoning and the known width $b$ of the semitrailer's front are used to classify an edge to the front, the left or the right side of the semitrailer's body. 
Once the front edge and its corresponding corners are found, $L_y$ and $\phi$ can easily be calculated~\cite{Patrik2016,Daniel2018}.  

The measurements $y_{k}^{\text{loc}}=\begin{bmatrix} x_{1,k} & y_{1,k} & \theta_{1,k}\end{bmatrix}^T$ from the localization system and the constructed measurements \mbox{$y_{k}^{\text{ran}}=\begin{bmatrix} L_{y,k} & \phi_{k}\end{bmatrix}^T$} from the iterative RANSAC algorithm are treated as synchronous observations with different sampling rates. 
These observations are fed to an EKF to estimate the full state vector $\hat x$ of the G2T with car-like tractor~\eqref{j1:eq:model_global_coord}.

\subsection{Extended Kalman filter}
The EKF algorithm performs two steps, a time update where the next state $\hat x_{k\mid k-1}$ is predicted using a prediction model of the vehicle and a measurement update that corrects $\hat x_{k\mid k-1}$ to give a filtered estimate $\hat x_{k\mid k}$ using the available measurements~\cite{gustafsson2010statistical}.

\begin{figure}[t!]
	\vspace{-30pt}
	\begin{center}
		\includegraphics[width=0.60\linewidth]{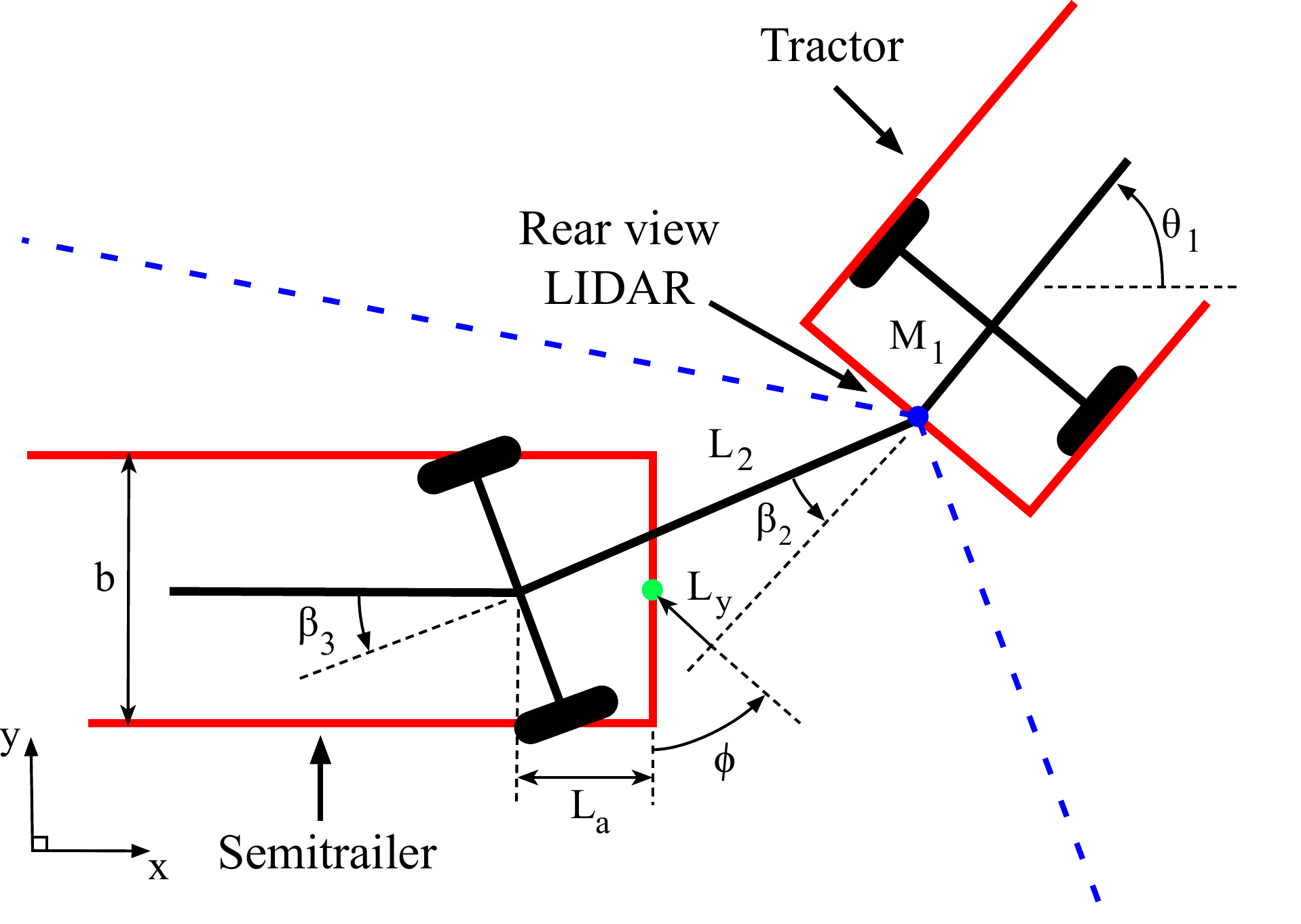} 
		\caption{A bird's-eye view of the connection between the car-like tractor and the semitrailer, as well as the geometric properties of the semitrailer that are used by the nonlinear observer. The green dot represents the midpoint of the front of the semitrailer's body, where $L_y$ is the $y$-coordinate in the tractor's local coordinate system. The LIDAR sensor is mounted at the blue dot and the dashed blue lines illustrate the LIDAR's field of view.} 
		\label{j1:fig:ransac_meas}
	\end{center} 
\end{figure}

To construct the prediction model, the continuous-time model of the G2T with a car-like tractor~\eqref{j1:eq:model_global_coord} is discretized using Euler forward with a sampling time of $T_s$ seconds. 
The control signals to the prediction model are the longitudinal velocity $v$ of the car-like tractor and its curvature $\kappa$. 
Given the control signals $u_k = \begin{bmatrix} v_k & \kappa_k\end{bmatrix}^T$, the vehicle state $x_k$ and a process noise model $w_k$ with covariance $\Sigma^w$, the prediction model for the G2T with a car-like tractor can be written as
\begin{align}\label{j1:eq:discrete_time_model}
x_{k+1} = \hat f(x_k, u_k,w_k), \quad w_k \thicksim \pazocal{N}(0,\Sigma^w).
\end{align} 
Since the observations $y_{k}^{\text{loc}}$ and $y_{k}^{\text{ran}}$ are updated at different sampling rates, independent measurement equations for each observation are derived. 
Assuming measurements with normally distributed zero mean noise, the measurement equation for the observation from the iterative RANSAC algorithm \mbox{$y_{k}^{\text{ran}}=\begin{bmatrix} L_{y,k} & \phi_k\end{bmatrix}^T$} can be written as
\begin{align}\label{j1:eq:meas_eq_ransac}
y_{k}^{\text{ran}} = h^{\text{ran}}(x_{k}) + e_{k}^{\text{ran}}, \quad e_{k}^{\text{ran}} \thicksim \pazocal{N}(0,  \Sigma^{e}_\text{ran}),
\end{align}
where $e_{k}^{\text{ran}}$ is the measurement noise with covariance matrix $\Sigma^{e}_\text{ran}$ and $h^{\text{ran}}(x_k)$ defines the relationship between the states and the measurements. From Figure~\ref{j1:fig:ransac_meas}, the two components of $h^{\text{ran}}(x_k)$ can be derived as 
\begin{subequations}
	\label{j1:eq:meas_eq_hx_ransac}
	\begin{align}
	%
	L_{y,k}=h^{\text{ran}}_1(x_k) &= L_2\sin{\beta_{2,k}} - L_a \sin{(\beta_{2,k} +\beta_{3,k})}, \label{j1:eq:meas_eq_hx_ransac_ly}\\
	\phi_k=h^{\text{ran}}_2(x_k) &=\beta_{2,k} + \beta_{3,k}. \label{j1:eq:meas_eq_hx_ransac_phi}
	\end{align}
\end{subequations} 
The second measurement equation, corresponding to the observation $y_{k}^{\text{loc}}=\begin{bmatrix} x_{1,k} & y_{1,k} & \theta_{1,k}\end{bmatrix}^T$ from the localization system is given by 
\begin{align}\label{j1:eq:meas_eq_localization}
y_{k}^{\text{loc}} = h^{\text{loc}}(x_k) + e_k^{\text{loc}}, \quad e_k^{\text{loc}} \thicksim \pazocal{N}(0,  \Sigma^{e}_\text{loc}),
\end{align}
where the components of $h^{\text{loc}}(x_k)$ can be derived from Figure~\ref{j1:fig:schematic_model_description} as  
\begin{subequations}
	\label{j1:eq:meas_eq_hx_localization}
	\begin{align}
	x_{1,k}=h^{\text{loc}}_1(x_k) &= x_{3,k} + L_3\cos{\theta_{3,k}} + L_2\cos{(\theta_{3,k}+\beta_{3,k})} + M_1 \cos{(\theta_{3,k}+\beta_{3,k}+\beta_{2,k})}, \\
	y_{1,k}=h^{\text{loc}}_2(x_k) &= y_{3,k} + L_3\sin{\theta_{3,k}} + L_2\sin{(\theta_{3,k}+\beta_{3,k})} + M_1 \sin{(\theta_{3,k}+\beta_{3,k}+\beta_{2,k})},\\
	\theta_{1,k}=h^{\text{loc}}_3(x_k) &=\theta_{3,k}+\beta_{3,k}+\beta_{2,k},
	\end{align}
\end{subequations}
and $e_{k}^{\text{loc}}$ is the the measurement noise with covariance matrix $\Sigma^{e}_\text{loc}$. 
The standard EKF framework is now applied using the prediction model in~\eqref{j1:eq:discrete_time_model} and the measurement equations in~\eqref{j1:eq:meas_eq_ransac} and \eqref{j1:eq:meas_eq_localization} \cite{gustafsson2010statistical}.

The process noise $w_k$ is assumed to enter additively with $u_k$ into the prediction model~\eqref{j1:eq:discrete_time_model} and the time update of the EKF is performed as follows
\begin{subequations} \label{j1:eq:time_update_ekf}
	\begin{align}
	\hat x_{k+1\mid k} &= \hat f(\hat x_{k\mid k},u_k,0), \\
	\Sigma^x_{k+1\mid k} &= F_k\Sigma^x_{k\mid k}F_k^T + G_{w,k}\Sigma^wG_{w,k}^T,
	\end{align}
\end{subequations}
where $F_k = \hat f'_x(\hat x_{k\mid k},u_k,0)$ and $G_{w,k} = \hat f'_u(\hat x_{k\mid k},u_k,0)$ are the linearization of the prediction model around the current state estimate $\hat x_{k \mid k}$ with respect to $x$ and $u$, respectively. 

Since the observations $y_{k}^{\text{loc}}$ and $y_{k}^{\text{ran}}$ are updated at different sampling rates, the measurement update of the state estimate $\hat x_{k\mid k}$ and the covariance matrix $P_{k\mid k}$ is performed sequentially for $y_{k}^{\text{loc}}$ and $y_{k}^{\text{ran}}$. Let $H_k$ be defined as the block matrix
\begin{align}
H_k = 
\begin{bmatrix} H_{1,k} \\[1ex] H_{2,k}\end{bmatrix}
=
\begin{bmatrix}
\left(\dfrac{\partial h^{\text{ran}}(x_{k\mid k-1})}{\partial x}\right)^T  & \left(\dfrac{\partial h^{\text{loc}}(x_{k\mid k-1})}{\partial x}\right)^T
\end{bmatrix}^T.
\end{align}
Each time an observation from the localization system $y_{k}^{\text{loc}}$ is available, the following measurement update is performed
\begin{subequations} \label{j1:eq:meas_update_ekf}
	\begin{align}
	K_k &= \Sigma^x_{k\mid k-1}H_{2,k}^T\left(\Sigma^e_{\text{loc}} + H_{2,k} \Sigma^x_{k\mid k-1}  H_{2,k}^T\right)^{-1}, \\
	\hat x_{k\mid k} &= \hat x_{k\mid k-1} + K_k\left(y_{k}^{\text{loc}} - h^{\text{loc}}(\hat x_{k\mid k-1}) \right), \\
	\Sigma^x_{k\mid k} &= \Sigma^x_{k\mid k-1} - K_k H_{2,k} \Sigma^x_{k\mid k-1},
	\end{align}
\end{subequations} 
where $K_k$ is the Kalman gain~\cite{gustafsson2010statistical}. Similarly, when the observation $y_{k}^{\text{ran}}$ is updated, the same measurement update~\eqref{j1:eq:meas_update_ekf} is performed with $\Sigma^e_{\text{loc}}$, $H_{2,k}$, $y_{k}^{\text{loc}}$ and $h^{\text{loc}}$ replaced with $\Sigma^e_{\text{ran}}$, $H_{1,k}$, $y_{k}^{\text{ran}}$ and $h^{\text{ran}}$, respectively. 

To decrease the convergence time of the estimation error, the EKF is initialized as follows. 
Define the combined measurement equation of $y_{k}^{\text{ran}}$ and $y_{k}^{\text{loc}}$ as $y_k=h(x_k)$. 
Assuming noise-free observations and that \mbox{$|\beta_{2,k}|\leq \pi/2$}, this system of equations has a unique solution \mbox{$x_k = h^{-1}(y_k)$} given by
\begin{subequations} \label{j1:eq:invers_h}
	\begin{align}
	\beta_{2,k} &= \arcsin\left(\frac{L_{y,k}+L_a\sin{\phi_k}}{L_2}\right)= h_{\beta_{2,k}}^{-1}( y_k), \label{j1:eq:invers_h_beta2}\\
	\beta_{3,k} &= \phi_k -  h_{\beta_{2,k}}^{-1}(y_k)= h_{\beta_{3,k}}^{-1}( y_k), \label{j1:eq:invers_h_beta3}\\
	\theta_{3,k} &= \theta_{1,k}-\phi_k= h_{\theta_{3,k}}^{-1}( y_k), \\
	x_{3,k} &= x_{1,k} - L_3\cos{( h_{\theta_{3,k}}^{-1}( y_k))} - L_2\cos{( h_{\theta_{3,k}}^{-1}( y_k)+ h_{\beta_{3,k}}^{-1}( y_k))} \nonumber \\ &+M_1 \cos{( h_{\theta_{3,k}}^{-1}( y_k)+ h_{\beta_{3,k}}^{-1}( y_k)+ h_{\beta_{2,k}}^{-1}( y_k))}, \\
	y_{3,k} &= y_{1,k} - L_3\sin{( h_{\theta_{3,k}}^{-1}( y_k))} - L_2\sin{(h_{\theta_{3,k}}^{-1}( y_k)+h_{\beta_{3,k}}^{-1}( y_k))} \nonumber \\ &+M_1 \sin{( h_{\theta_{3,k}}^{-1}( y_k)+ h_{\beta_{3,k}}^{-1}( y_k)+ h_{\beta_{2,k}}^{-1}( y_k))}.
	\end{align}      
\end{subequations}
This relationship is used to initialize the EKF with the initial state estimate $\hat x_{1\mid 0}= h^{-1}(y_0)$, the first time both measurements are obtained.
The state covariance matrix is at the same time initialized to $\Sigma^x_{1\mid 0} = \Sigma^x_0$, where $\Sigma^x_0\succeq 0$ is a design parameter. 
Since no ground truth is available for all vehicle states, the filter cannot be individually evaluated but will be seen as part of the full system and thus be evaluated through the overall system performance.        

\section{Implementation details: Application to full-scale tractor-trailer system}
\label{j1:sec:implementation}
The path planning and path-following control framework has been deployed on a modified version of a Scania G580 6x4 tractor that is shown in Figure~\ref{j1:fig:truck_scania}. 
The car-like tractor is equipped with a sensor platform as described in Section~\ref{j1:sec:systemArchitecture}, including a real time kinematic GPS (RTK-GPS), IMUs and a rear view LIDAR sensor with 120 degrees field of view in the horizontal scan field. The tractor is also equipped with a servo motor for automated control of the steering column and additional computation power compared to the commercially available version. 
The triple axle semitrailer and the double axle dolly are both commercially available and are not equipped with any sensors that are used by the system. 
The vehicle lengths and the physical parameters for the car-like tractor are summarized in Table~\ref{j1:tab:vehicle_parameters}, where it is assumed that the rotational centers are located at the longitudinal center for each axle pair and triple, respectively. 
The total distance from the front axle of the car-like tractor to the center of the axle of the semitrailer is approximately \mbox{$18.4$ m}. 
In the remainder of this section, implementation details for each module within the path planning and path-following control framework are presented. 

\begin{table}[b!]
	\centering
	\caption{Vehicle parameters for the research platform used for the real-world experiments.}
	\begin{tabular}{l l}
		\hline \noalign{\smallskip} Vehicle Parameters  & Value   \\  \hline \noalign{\smallskip}	
		The tractor's wheelbase $L_1$            &   4.62 m  \\ 
		Maximum steering angle $\alpha_{\text{max}}$ & $42\pi/180$ rad \\
		Maximum steering angle rate $\omega_{\text{max}}$ & $0.6$ rad/s \\
		Maximum steering angle acceleration $u_{\omega,\text{max}}$ & $40$ rad/s$^2$ \\
		Length of the off-hitch $M_1$      &   1.66 m  \\  
		Length of the dolly $L_2$          &   3.87 m  \\    
		Length of the semitrailer $L_3$        &   8.00 m  \\
		Length of the semitrailer's overhang $L_a$       &   1.73 m  \\
		Width of the semitrailer's front $b$   &   2.45 m  \\
		\hline \noalign{\smallskip}
	\end{tabular}
	\label{j1:tab:vehicle_parameters}
\end{table}

\subsection{Lattice planner}\label{j1:sec:implementation_lattice_planner}
The lattice planner is implemented in C++ and the motion primitive set is calculated offline using the numerical optimal control solver CasADi~\cite{casadi}, together with the primal-dual interior-point solver IPOPT~\cite{ipopt}. The resulting paths are represented as discretely sampled points containing full state information including the control signals. For generation of the set of backward motion primitives $\pazocal P_{\text{rev}}$, the weight matrices $\mathbf{Q}_1\succeq 0$ and $\mathbf{Q}_2 \succeq 0$ in the cost function~\eqref{j1:obj_rev} are chosen as
\begin{align*}
\mathbf{Q}_1 = \begin{bmatrix}
11 & -10 \\ -10 & 11 
\end{bmatrix}, \quad
\mathbf{Q}_2 = \text{diag}\left(\begin{bmatrix}1& 10& 1\end{bmatrix}\right),
\end{align*}
giving the integrand $||\begin{bmatrix} \beta_3 & \beta_2\end{bmatrix}^T||_\mathbf{Q_1}^2=\beta_3^2 + \beta_2^2 + 10(\beta_3-\beta_2)^2$. 
This means that large joint angles with opposite signs are highly penalized during backward motion, which is directly related to motion plans that have an increased risk of leading to a jack-knife state during path execution. 
For the set of forward motion primitives $\pazocal P_{\text{fwd}}$, the weight $\mathbf{Q}_1$ is chosen as $\mathbf{Q}_1=0_{2\times 2}$. During motion primitive generation, the physical limitation on steering angle $\alpha_{\text{max}}$ is additionally 20 \% tightened to enable the path-following controller to reject
disturbances during plan execution. 
The complete set of motion primitives from the initial orientation $\theta_{3,i}=0$ is presented in Figure~\ref{j1:fig:primitives}. 
The generated motion primitive set $\pazocal P$ was then reduced using the reduction technique described in Section~\ref{j1:subsec:Mreduction}, with $\eta=1.2$, yielding a reduction factor of about 7 \%. 
The size of the reduced motion primitive set was $|\pazocal{P}'|=3888$, with between 66--111 different state transitions from each discrete state $z[k]\in\mathbb Z_d$. 
For the reduced motion primitive set $\pazocal{P}'$, a free-space HLUT~\cite{CirilloIROS2014,knepper2006high} was precomputed using a Dijkstra's search with cut-off cost $J_\text{cut}=170$.  
The surrounding environment is represented by an occupancy gridmap~\cite{occupancyGridMap} and efficient collision checking is performed using grid inflation and circle approximations for the semitrailer's and tractor's bounding boxes~\cite{lavalle2006planning}.

In the experiments, the lattice planner is given a desired goal state $z_G$ that can be specified by an operator or selected by an algorithm. At the goal state, the system is constrained to end up in a straight vehicle configuration where all vehicle segments are lined up, $i.e.$, the steering angle and the joint angles are constrained to zero at the goal state.
When a desired goal state $z_G$ has been specified, the vehicle's initial state $z(0)$ is first projected down to its closest neighboring state in $\mathbb Z_d$. 
The ARA$^*$ search algorithm is initialized with heuristic inflation factor $\gamma=2$ and then iteratively decreased by 0.1 in every subsequent iteration. 
If $\gamma$ reaches 1 or if a specified maximum allowed planning time is reached and a motion plan with a proven $\gamma$-suboptimality cost has been found, portions of the resulting motion plan are iteratively sent to the path-following controller for path execution.

\subsection{Path-following controller}\label{j1:implementation:details:feedback}
The framework presented in Section~\ref{j1:sec:lowlevelcontrol} is deployed to synthesize the hybrid path-following controller for this specific application.
First, a feedback gain $K_i$ and a corresponding Lyapunov function $V_i(\tilde x_e) = \tilde x_e^TP_i\tilde x_e$ is computed for each motion primitive $m_i\in \pazocal P$, separately. 
The convex polytope $\mathbb S_i$ in~\eqref{j1:def:polytope} is estimated by evaluating the Jacobian linearization~\eqref{j1:eq:linearizaion_Acl} of the path-following error model~\eqref{j1:eq:error_model_mi_cl} at each sampled point of the nominal path. 
Each resulting pair $[A_{i,j},B_{i,j}]$ of the linearization is assumed to be a vertex of the convex polytope $\mathbb S_i$ in~\eqref{j1:def:polytope}. The nominal feedback gains are designed using infinite-horizon LQ-control~\cite{anderson2007optimal} where the path-following error model has been linearized around a straight nominal path in backward and forward motion, respectively. 
In these cases, the Jacobian linearization is given by the matrices $A$ and $B$ defined in~\eqref{j1:eq:lin_AB}.  
The weight matrices $\tilde Q_{\text{fwd}}$ and $\tilde Q_{\text{rev}}$ that are used in the LQ-design are listed in Table~\ref{j1:tab:design_parameters}. By choosing the penalty on the curvature deviation as $\tilde R_{\text{rev}}=\tilde R_{\text{fwd}}=1$, the nominal feedback gains are   
\begin{subequations}
	\begin{align}
	K_{\text{rev}} &= 
	\begin{bmatrix}
	-0.12 & 1.67 & -1.58 & 0.64
	\end{bmatrix}, \\
	K_{\text{fwd}} &= -\begin{bmatrix}
	0.20 & 2.95 & 1.65 & 1.22
	\end{bmatrix}, 
	\end{align}
\end{subequations}
where positive feedback is assumed. Here, $K_{\text{fwd}}$ and $K_{\text{rev}}$ are dedicated for the set of forward $\pazocal P_{\text{fwd}}$ and backward $\pazocal P_{\text{rev}}$ motion primitives, respectively. 
Using these nominal feedback gains, the SDP problem in~\eqref{j1:eq:opt_LTV} with decay rate $\epsilon=0.01$ is solved separately for each motion primitive $m_i\in\pazocal P$ using YALMIP~\cite{lofberg2004yalmip}. 
In this specific application, $\forall m_i\in\pazocal P$, the optimal value of the objective function in~\eqref{j1:eq:opt_LTV} is zero, which implies that $K_i = K_{\text{nom}}^i$. 
Thus, for this specific set of motion primitives $m_i\in\pazocal P$ (see Figure~\ref{j1:fig:primitives}), the hybrid path-following controller can be written as in~\eqref{j1:eq:hybrid_controller}. However, the continuous-time quadratic Lyapunov functions are not equal $\forall m_i\in\pazocal P$.  

\begin{table}[t!]
	\centering
	\caption{Design parameters for the EKF and the path-following controller during the real-world experiments.}
	\begin{tabular}{l l}
		\hline \noalign{\smallskip} EKF parameters  & Value  \\  \hline \noalign{\smallskip}
		Process noise $\Sigma^w$                  &  
		$10^{-3}\times\text{diag}\left(\begin{bmatrix} 1 & 1 \end{bmatrix}\right)$ \\[2pt] 
		Measurement noise $\Sigma^e_{\text{loc}}$ &  $10^{-3}\times\text{diag}\left(\begin{bmatrix}1&1&0.5\end{bmatrix}\right)$ \\[2pt]  
		Measurement noise $\Sigma^e_{\text{ran}}$ &  $10^{-3}\times\text{diag}\left(\begin{bmatrix}0.5&0.1\end{bmatrix}\right)$ \\[2pt]   
		Initial state covariance $\Sigma^x_0$     &  $0.5\times\text{diag}\left(\begin{bmatrix}1&1&0.1&0.1&0.1\end{bmatrix}\right)$ \\[2pt]
		EKF frequency					   & 100 Hz	\\[1pt]
		\hline \noalign{\smallskip} \hline \noalign{\smallskip}
		Controller parameters  & Value   \\  \hline \noalign{\smallskip}
		Nominal LQ weight $\tilde Q_\text{fwd}$ & 
		$0.05\times\text{diag}\left(\begin{bmatrix}0.8& 6& 8 & 8\end{bmatrix}\right)$ \\[2pt]
		Nominal LQ weight $\tilde Q_\text{rev}$ & 
		$0.05\times\text{diag}\left(\begin{bmatrix}0.3 & 6 & 7 & 5\end{bmatrix}\right)$ \\[2pt]
		Controller frequency				& 50 Hz	\\[1pt]
		\hline 
	\end{tabular}
	\label{j1:tab:design_parameters}
\end{table}

In this setup, it is possible to find a common quadratic Lyapunov function $V_{\text{fwd}}(\tilde x_e)$ with decay-rate $\epsilon=0.01$ and path-following controller $\tilde \kappa = K_{\text{fwd}}\tilde x_e$, for all forward motion primitives $\pazocal P_{\text{fwd}}$.
It is also possible to find a common quadratic Lyapunov function $V_{\text{rev}}(\tilde x_e)$ with decay-rate $\epsilon=0.01$ and path-following controller $\tilde \kappa = K_{\text{rev}}\tilde x_e$, for all backward motion primitives $\pazocal P_{\text{rev}}$. 
It is however not possible to find a common quadratic Lyapunov function $V(\tilde x_e)$ with a decay-rate $\epsilon>0$ and hybrid path-following controller $\tilde \kappa = K_{i}\tilde x_e$, $i\in\{1,\hdots,M\}$, for the complete set of forward and backward motion primitives $\pazocal P$. This follows directly from Theorem~\ref{j1:P1}. 
Thus, if the lattice planner is constrained to only compute nominal paths using either $\pazocal P_{\text{fwd}}$ or $\pazocal P_{\text{rev}}$, it is possible to guarantee that the continuous-time path-following error exponentially decays towards zero. To guarantee similar properties for the path-following error when the motion plan is composed by a combination of forward and backward motion primitives, the analysis method presented in Section~\ref{j1:sec:convergence} needs to be applied. This analysis is presented in Section~\ref{j1:sec:StabilityAnalysis}.  

The hybrid path-following controller~\eqref{j1:eq:hybrid_controller} was implemented in Matlab/Simulink and C-code was then auto-generated where the path-following controller was specified to operate at 50 Hz. 
During the real-world experiments, the tractor's cruise controller was used for longitudinal control with \mbox{$v=1$ m/s} along forward motion primitives and \mbox{$v=-0.8$ m/s} along backward motion primitives.  

\subsection{State observer}
The design parameters for the EKF are summarized in Table~\ref{j1:tab:design_parameters}, which were tuned using collected data from manual tests with the vehicle. 
This data was then used offline to tune the covariance matrices in the EKF and to calibrate the position and orientation of the rear view LIDAR sensor. 
The pitch angle of the rear view LIDAR sensor was adjusted such that that body of the semitrailer was visible in the LIDAR's point-cloud for all vehicle configurations that are of relevance for this application. 

The EKF and the iterative RANSAC algorithm~\cite{Patrik2016,Daniel2018} was implemented in Matlab/Simulink and C-code was then auto-generated. 
The EKF was specified to operate at 100 Hz and the measurements from the localization system is updated at the same sampling rate. 
The observation from the iterative RANSAC algorithm is received at a sampling rate of 20 Hz. 
The iterative RANSAC algorithm is specified to extract at most two edges of the semitrailer's body and 500 random selections of data pairs are performed for each edge extraction with an inlier threshold of 5 centimeters. 

\section{Results}
\label{j1:sec:Results}
In this section, we first analyze the behavior of the closed-loop system consisting of the controlled G2T with a car-like tractor and the path-following controller when executing a nominal path computed by the lattice planner. 
Then, the planning capabilities of the lattice planner and the ideal tracking performance of the path-following controller are evaluated in simulations. 
Finally, the complete framework is evaluated in three different real-world experiments on the full-scale test vehicle that can be seen in Figure~\ref{j1:fig:truck_scania}.    

\subsection{Analysis of the closed-loop hybrid system}
\label{j1:sec:StabilityAnalysis}
To verify that the path-following error $\tilde x_e(t)$ is bounded and decays toward zero when the nominal path is constructed by any sequence of motion primitives, backward as well as forward ones, the method presented in Section~\ref{j1:sec:convergence} is applied. 
The closed-loop system in~\eqref{j1:eq:error_states_distance} is implemented in MATLAB/Simulink. 
Central differences with step size $\delta=0.01$ is used to compute the state-transition matrix $F_i$ for the linear discrete-time system~\eqref{j1:eq:lin_disc_transition} that describes the evolution of the path-following error~\eqref{j1:eq:error_states_lattice} when motion primitive $m_i\in\pazocal P$ is executed.
Since there are four path-following error states, eight simulations are performed in order to generate each transition matrix $F_i$. 
This numerical differentiation is performed for all $m_i\in\pazocal P$ and $M$ state-transition matrices are produced, $i.e.$, $\mathbb F = \{F_1,\hdots,F_M\}$.

\tikzexternaldisable
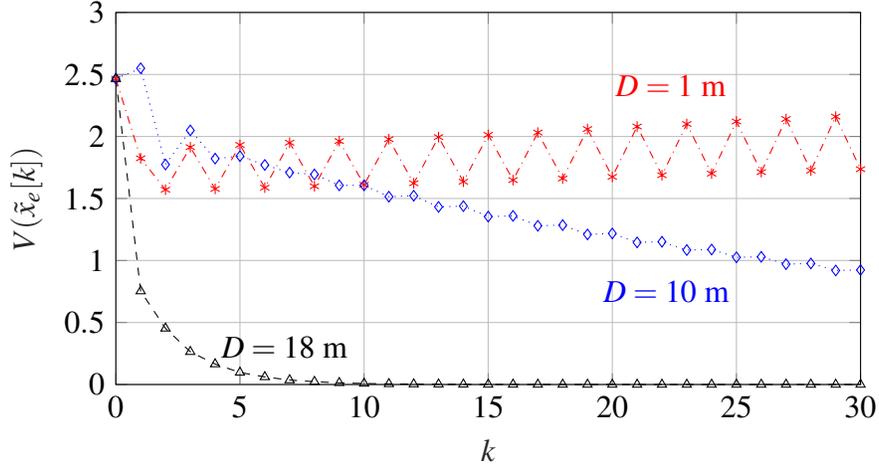
\begin{figure*}[t!]
	\centering
	\setlength\figureheight{0.30\textwidth}
	\setlength\figurewidth{0.6\textwidth}
	\begin{tikzpicture}
	\node[anchor=south west] (myplot) at (0,0) {
%
%
\begin{tikzpicture}

\begin{axis}[%
width=\figurewidth,
height=\figureheight,
at={(0\figurewidth,0\figureheight)},
scale only axis,
xmin=0,
xmax=30,
xlabel style={font=\color{white!15!black}},
xlabel={$k$},
ymin=0,
ymax=3,
ylabel style={font=\color{white!15!black}},
ylabel={$V(\tilde x_e[k])$},
axis background/.style={fill=white},
xmajorgrids,
ymajorgrids,
xtick={0,5,10,15,20,25,30},
ytick={0.0,0.5,1.0,1.5,2.0,2.5,3.0},
]
\addplot [color=red, dashdotted, mark=asterisk, mark options={solid, red}, forget plot]
  table[row sep=crcr]{%
0	2.465212\\
1	1.82379223696891\\
2	1.5705613054153\\
3	1.91357476858931\\
4	1.57727183121781\\
5	1.93177535365028\\
6	1.58721872778968\\
7	1.94984518792937\\
8	1.59831652422298\\
9	1.96293205167303\\
10	1.61020345121844\\
11	1.97823020175583\\
12	1.62258128687864\\
13	1.99503590116928\\
14	1.6352549994426\\
15	2.01292165540863\\
16	1.64806416960931\\
17	2.03154031278751\\
18	1.66087612212293\\
19	2.05979082256671\\
20	1.67434690286342\\
21	2.08020933793103\\
22	1.68747835960543\\
23	2.10069172548125\\
24	1.70028800928014\\
25	2.12095433987836\\
26	1.71272451602308\\
27	2.14084265023421\\
28	1.72474301730048\\
29	2.1602405180299\\
30  1.73631058974478\\
};
\addplot [color=blue, dotted, mark=diamond, mark options={solid, blue}, forget plot]
  table[row sep=crcr]{%
0	2.465212\\
1	2.55002713163937\\
2	1.77395842749014\\
3	2.0491170950099\\
4	1.82127052964684\\
5	1.8414175650963\\
6	1.77046237790235\\
7	1.70883936798654\\
8	1.69385290151998\\
9	1.60596210531345\\
10	1.60545486994787\\
11	1.51440967123855\\
12	1.52220738835996\\
13	1.43199450614004\\
14	1.43856470593648\\
15	1.35393068599508\\
16	1.35971230324364\\
17	1.28001421641959\\
18	1.28525138494961\\
19	1.21009219614682\\
20	1.21788542203593\\
21	1.14573879801393\\
22	1.15117118172691\\
23	1.08422418365252\\
24	1.08857916180272\\
25	1.02575919143816\\
26	1.02955482897438\\
27	0.970356251525441\\
28	0.976190179041762\\
29	0.919325759341287\\
30	0.923287167659439\\
};
\addplot [color=black, dashed, mark=triangle, mark options={solid, black}, forget plot]
  table[row sep=crcr]{%
0	2.465212\\
1	0.753048783266017\\
2	0.452620930787525\\
3	0.263880343551136\\
4	0.164660442123175\\
5	0.0970993122484494\\
6	0.0610029987604115\\
7	0.0360196377163824\\
8	0.0226914643561713\\
9	0.0134184805906692\\
10	0.00845604444480941\\
11	0.00500131980728824\\
12	0.00315168986712422\\
13	0.00186406925996607\\
14	0.00117466893560092\\
15	0.000694756979743971\\
16	0.000437808446260846\\
17	0.000258941139438917\\
18	0.000163174177224334\\
19	9.65090748969999e-05\\
20	6.08160508548394e-05\\
21	3.59695382279313e-05\\
22	2.26665179283714e-05\\
23	1.34060682945858e-05\\
24	8.44794989873776e-06\\
25	4.9965235504601e-06\\
26	3.14860241701354e-06\\
27	1.86223476461578e-06\\
28	1.17350328565635e-06\\
29	6.94066226714085e-07\\
30	4.37371813433174e-07\\
};
\end{axis}
\end{tikzpicture}%
	};
	\begin{scope}[x={(myplot.south east)}, y={(myplot.north west)}]
	%
	\node[text=red]   at (0.75,0.79)  {$D=1$ m};
	\node[text=blue]  at (0.745,0.38) {$D=10$ m};
	\node[text=black] at (0.325,0.27) {$D=18$ m};
	\end{scope}
	\end{tikzpicture}
	\caption{Simulation results of the closed-loop path-following error system when switching thirty times between a straight forward and backward motion primitive of different path lengths $D$ m. The path-following error at each switching instance $\tilde x_e[k]$ is evaluated using the discrete-time Lyapunov function $V_d(\tilde x_e[k])= \tilde x_e^T[k]S\tilde x_e[k]$, where $S$ is given in~\eqref{j1:eq:lyapunov_matrix_discrete} for $D=1$ m (red stars), $D=10$ m (blue diamonds) and $D=18$ m (black triangles).} 
	\label{j1:fig:sim_fwd_backward}
\end{figure*}
\tikzexternalenable 

The matrix inequalities in \eqref{j1:eq:dlmitotal} are solved to show that the norm of the path-following error for the discrete-time switched system in~\eqref{j1:eq:swithing_system}, exponentially decays towards zero at the switching instants $\tilde x_e[k]$. 
By selecting $0 < \mu < 1$, the following SDP problem can be solved 
\begin{align}
\minimize_{\eta,S} \hspace{3.7ex} & \eta \label{j1:eq:opt_DISC}\\
\subjectto\hspace{3ex}
&F_j^T S F_j - S \preceq - \mu S, \quad j=1,\hdots,M \nonumber\\
& I \preceq S \preceq \eta I,  \nonumber
\end{align} 
where the condition number of $S$ is minimized to compute a guaranteed upper bound~\eqref{ineq:discretetime} of the path-following error that is as tight as possible.
It turns out that it is not possible to select $0<\mu<1$ such that a feasible solution to \eqref{j1:eq:opt_DISC} exists for the original motion primitive set $\pazocal P$. 
This is because there exist short motion primitives of about 1 m that moves the vehicle either straight forwards or backwards. 
If these motion primitives are switched between, it is not possible to guarantee that the norm of the path-following error at the switching points will exponentially decay towards zero. 
In order to resolve this, the short motion primitives were extended to about 18 m (as the length of the tractor-trailer system) and their corresponding state-transition matrices $F_i$ were again computed. 
With this adjusted motion primitive set $\pazocal P_{\text{adj}}$ and $\mu = 0.3$, the optimization problem in~\eqref{j1:eq:opt_DISC} is feasible to solve using YALMIP~\cite{lofberg2004yalmip} and the optimal solutions are $\eta = 51.58$ and 
\begin{align}\label{j1:eq:lyapunov_matrix_discrete}
S = \begin{bmatrix}
1.04 &   1.29 &    0.29 & 0.34 \\
1.29 &  50.54 &   -0.22 & 6.62 \\
0.29 &  -0.23 &   51.09 & 2.58 \\
0.34 &   6.62 &    2.58 & 5.16
\end{bmatrix}.
\end{align}
Extending the short motion primitives manually is equivalent to adding constraints on the switching sequence $\{u_q[k]\}_{k=0}^{N-1}$ in the lattice planner. For this case, when a short motion primitive $p_i\in \pazocal P$ is activated, $u_q[k]$ needs to remain constant for a certain amount of switching instances. This constraint can easily be added within the lattice planner.
To illustrate the phenomenon, Figure~\ref{j1:fig:sim_fwd_backward} shows the behavior of the closed-loop path-following error system when switching thirty times between a straight forward and backward motion primitive of three different path lengths $D=1$ m, 10 m and 18 m, with the initial path-following error state \mbox{$\tilde{x}_e[0]=\begin{bmatrix} 1\text{ m} & 0.1 \text{ rad} & -0.1\text{ rad} & 0.1\text{ rad} \end{bmatrix}^T$}. 
In the figure, the path-following error at each switching instance is evaluated using $V_d(\tilde x_e[k])=\tilde x^T_e[k]S\tilde x_e[k]$, where $S$ is given in~\eqref{j1:eq:lyapunov_matrix_discrete}. As can be seen, $V_d(\tilde x_e[k])$ is a valid discrete-time Lyapunov function for \mbox{$D=18$ m}, since $V_d(\tilde x_e[k])$ (black triangles) is monotonically decreasing towards zero as a function of the number of executed motion primitives. 
When $D=10$ m,  $V_d(\tilde x_e[k])$ (blue diamonds) decays towards zero, but not monotonously. 
When $D=1$ m, $V_d(\tilde x_e[k])$ (red star) does not converge towards zero. 
From our practical experience, allowing the short motion primitives have not caused any problems, since repeated switching between short straight forward and backward motion primitives is of limited practical relevance for the planning problems typically considered in this work.

\subsection{Simulation results}
Results from a quantitative analysis of the lattice planner is first presented, where its performance has been statistically evaluated in Monte Carlo simulations in two practically relevant scenarios. 
Then, simulation results for the path-following controller during ideal conditions where perfect state information is available is given to demonstrate its performance. The simulations have been performed on a standard laptop computer with an Intel Core i7-6820HQ@2.7GHz CPU.

\begin{figure}[b!]
	\begin{center}
		\includegraphics[width=0.85\linewidth]{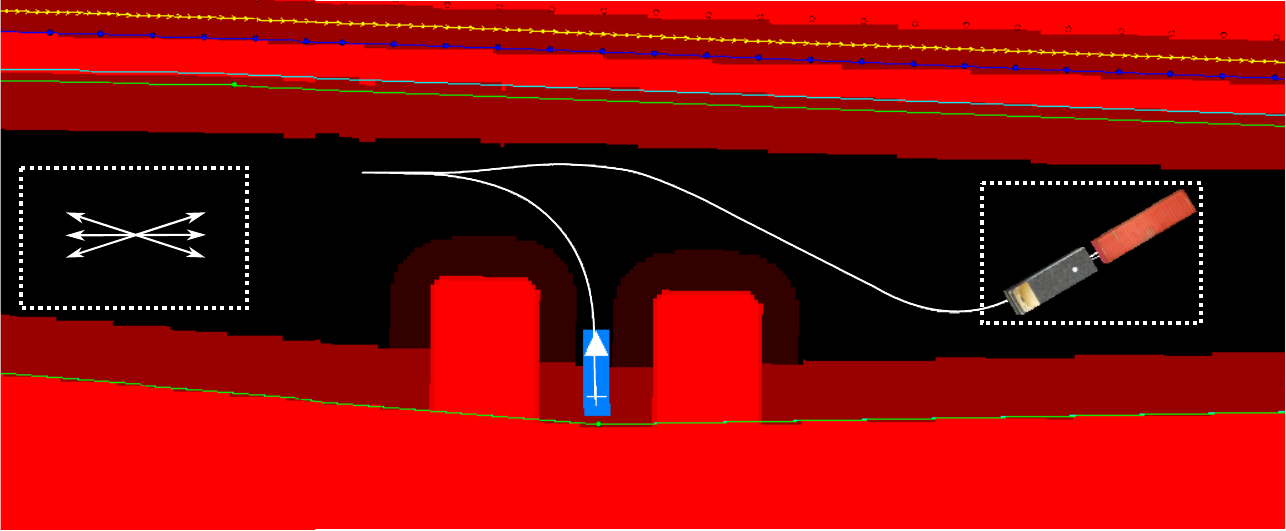}   
		\caption{An overview of the parking planning problem. The goal position of the axle of the semitrailer $(x_{3,G},y_{3,G}$) is marked by the white cross inside the blue rectangle, where the white arrow specifies its goal orientation $\theta_{3,G}$. The initial position $(x_{3,I},y_{3,I}$) is uniformly sampled within the two white-dotted rectangles, and the initial orientation $\theta_{3,I}\in\Theta$ is sampled from six different initial orientations. The white path illustrates the planned path for the axle of the semitrailer for one out of 1000 Monte Carlo simulations. The area occupied by obstacles is colored in red and the black area is free-space.} 
		\label{j1:fig:parking_example}
	\end{center} 
\end{figure}

\subsubsection{Simulation results for the lattice planner}
Two different path planning scenarios are used to evaluate the performance of the lattice planner. One thousand Monte Carlo simulations are performed for each scenario, where the goal state $z_G\in\mathbb Z_d$ and/or the initial state $z_I\in\mathbb Z_d$ are randomly selected from specified regions that are compliant with the specified state-space discretization $\mathbb Z_d$. 
For simplicity, it is assumed that the vehicle starts and ends in a straight vehicle configuration where all vehicle segments are lined up, $i.e.$, the steering angle and the joint angles are constrained to zero. A goal state and an initial state, respectively, is thus specified by a position $(x_{3},y_{3})$ and an orientation $\theta_{3}$ of the semitrailer. 
As explained in Section~\ref{j1:sec:implementation_lattice_planner}, the ARA$^*$ search is initialized with heuristic inflation factor $\gamma=2$. This factor is then iteratively decreased by 0.1 every time a path to the goal for a specific $\gamma$ has been found. 
To evaluate the computation time and the quality of the produced solution, the lattice planner was allowed to plan until an optimal solution with $\gamma =1$ was found. 
In the Monte Carlo simulations, each time a solution for a specific $\gamma$ is found, the accumulated planning time and the value of the cost function $J_D$ are stored. 
During the simulations, a planning problem is marked unsolved if the planning time exceeds 60 s and a solution with $\gamma = 1$ has not yet been found. 

The first planning scenario is illustrated in Figure~\ref{j1:fig:parking_example}, and the objective is to plan a parking maneuver from a randomly selected initial state \mbox{$z_I\in\mathbb Z_d$} to a fixed goal state $z_G\in\mathbb Z_d$. 
In Figure~\ref{j1:fig:parking_example}, the goal position of the axle of the semitrailer $(x_{3,G},y_{3,G})$ is illustrated by the white cross inside the blue rectangle, and the white arrow specifies its goal orientation $\theta_{3,G}$. 
The initial position of the axle of the semitrailer $(x_{3,I},y_{3,I})$ is uniformly sampled from two different 20 m $\times$ 15 m rectangles on each side of the goal location and the initial orientation of the semitrailer $\theta_{3,I}$ is randomly selected from six different initial orientations, as depicted in Figure~\ref{j1:fig:parking_example}. 

\tikzexternaldisable
\begin{figure}[t!] 	
	\centering
	\setlength\figureheight{0.34\columnwidth}
	\setlength\figurewidth{0.36\columnwidth} 
	\subfloat[][]{
%
%
\begin{tikzpicture}

\begin{axis}[%
width=\figurewidth,
height=\figureheight,
at={(0\figurewidth,0\figureheight)},
scale only axis,
xmin=0,
xmax=12,
xtick={1,3,5,7,9,11},
xticklabels={{1.0},{1.2},{1.4},{1.6},{1.8},{2.0}},
xlabel style={font=\color{white!15!black}},
xlabel={$\gamma$},
ymode=log,
ymin=0.005,
ymax=50,
ylabel style={font=\color{white!15!black}},
ylabel={Planning Time [s]},
axis background/.style={fill=white},
ymajorgrids,
yminorgrids,
minor y grid style={dotted,black},
xmajorgrids,
xminorgrids,
]
\addplot [color=black, dashed, forget plot]
  table[row sep=crcr]{%
1	17.841\\
1	30.388\\
};
\addplot [color=black, dashed, forget plot]
  table[row sep=crcr]{%
2	13.392\\
2	23.524\\
};
\addplot [color=black, dashed, forget plot]
  table[row sep=crcr]{%
3	9.7\\
3	17.223\\
};
\addplot [color=black, dashed, forget plot]
  table[row sep=crcr]{%
4	6.536\\
4	11.986\\
};
\addplot [color=black, dashed, forget plot]
  table[row sep=crcr]{%
5	3.632\\
5	6.897\\
};
\addplot [color=black, dashed, forget plot]
  table[row sep=crcr]{%
6	1.585\\
6	3.228\\
};
\addplot [color=black, dashed, forget plot]
  table[row sep=crcr]{%
7	0.999\\
7	2.099\\
};
\addplot [color=black, dashed, forget plot]
  table[row sep=crcr]{%
8	0.898\\
8	1.895\\
};
\addplot [color=black, dashed, forget plot]
  table[row sep=crcr]{%
9	0.837\\
9	1.745\\
};
\addplot [color=black, dashed, forget plot]
  table[row sep=crcr]{%
10	0.811\\
10	1.688\\
};
\addplot [color=black, dashed, forget plot]
  table[row sep=crcr]{%
11	0.757\\
11	1.561\\
};
\addplot [color=black, dashed, forget plot]
  table[row sep=crcr]{%
1	0.287\\
1	9.138\\
};
\addplot [color=black, dashed, forget plot]
  table[row sep=crcr]{%
2	0.188\\
2	6.61\\
};
\addplot [color=black, dashed, forget plot]
  table[row sep=crcr]{%
3	0.119\\
3	4.55\\
};
\addplot [color=black, dashed, forget plot]
  table[row sep=crcr]{%
4	0.064\\
4	2.72\\
};
\addplot [color=black, dashed, forget plot]
  table[row sep=crcr]{%
5	0.05\\
5	1.453\\
};
\addplot [color=black, dashed, forget plot]
  table[row sep=crcr]{%
6	0.015\\
6	0.373\\
};
\addplot [color=black, dashed, forget plot]
  table[row sep=crcr]{%
7	0.011\\
7	0.259\\
};
\addplot [color=black, dashed, forget plot]
  table[row sep=crcr]{%
8	0.011\\
8	0.228\\
};
\addplot [color=black, dashed, forget plot]
  table[row sep=crcr]{%
9	0.011\\
9	0.221\\
};
\addplot [color=black, dashed, forget plot]
  table[row sep=crcr]{%
10	0.011\\
10	0.218\\
};
\addplot [color=black, dashed, forget plot]
  table[row sep=crcr]{%
11	0.01\\
11	0.217\\
};
\addplot [color=black, forget plot]
  table[row sep=crcr]{%
0.875	30.388\\
1.125	30.388\\
};
\addplot [color=black, forget plot]
  table[row sep=crcr]{%
1.875	23.524\\
2.125	23.524\\
};
\addplot [color=black, forget plot]
  table[row sep=crcr]{%
2.875	17.223\\
3.125	17.223\\
};
\addplot [color=black, forget plot]
  table[row sep=crcr]{%
3.875	11.986\\
4.125	11.986\\
};
\addplot [color=black, forget plot]
  table[row sep=crcr]{%
4.875	6.897\\
5.125	6.897\\
};
\addplot [color=black, forget plot]
  table[row sep=crcr]{%
5.875	3.228\\
6.125	3.228\\
};
\addplot [color=black, forget plot]
  table[row sep=crcr]{%
6.875	2.099\\
7.125	2.099\\
};
\addplot [color=black, forget plot]
  table[row sep=crcr]{%
7.875	1.895\\
8.125	1.895\\
};
\addplot [color=black, forget plot]
  table[row sep=crcr]{%
8.875	1.745\\
9.125	1.745\\
};
\addplot [color=black, forget plot]
  table[row sep=crcr]{%
9.875	1.688\\
10.125	1.688\\
};
\addplot [color=black, forget plot]
  table[row sep=crcr]{%
10.875	1.561\\
11.125	1.561\\
};
\addplot [color=black, forget plot]
  table[row sep=crcr]{%
0.875	0.287\\
1.125	0.287\\
};
\addplot [color=black, forget plot]
  table[row sep=crcr]{%
1.875	0.188\\
2.125	0.188\\
};
\addplot [color=black, forget plot]
  table[row sep=crcr]{%
2.875	0.119\\
3.125	0.119\\
};
\addplot [color=black, forget plot]
  table[row sep=crcr]{%
3.875	0.064\\
4.125	0.064\\
};
\addplot [color=black, forget plot]
  table[row sep=crcr]{%
4.875	0.05\\
5.125	0.05\\
};
\addplot [color=black, forget plot]
  table[row sep=crcr]{%
5.875	0.015\\
6.125	0.015\\
};
\addplot [color=black, forget plot]
  table[row sep=crcr]{%
6.875	0.011\\
7.125	0.011\\
};
\addplot [color=black, forget plot]
  table[row sep=crcr]{%
7.875	0.011\\
8.125	0.011\\
};
\addplot [color=black, forget plot]
  table[row sep=crcr]{%
8.875	0.011\\
9.125	0.011\\
};
\addplot [color=black, forget plot]
  table[row sep=crcr]{%
9.875	0.011\\
10.125	0.011\\
};
\addplot [color=black, forget plot]
  table[row sep=crcr]{%
10.875	0.01\\
11.125	0.01\\
};
\addplot [color=blue, forget plot]
  table[row sep=crcr]{%
0.75	9.138\\
0.75	17.841\\
1.25	17.841\\
1.25	9.138\\
0.75	9.138\\
};
\addplot [color=blue, forget plot]
  table[row sep=crcr]{%
1.75	6.61\\
1.75	13.392\\
2.25	13.392\\
2.25	6.61\\
1.75	6.61\\
};
\addplot [color=blue, forget plot]
  table[row sep=crcr]{%
2.75	4.55\\
2.75	9.7\\
3.25	9.7\\
3.25	4.55\\
2.75	4.55\\
};
\addplot [color=blue, forget plot]
  table[row sep=crcr]{%
3.75	2.72\\
3.75	6.536\\
4.25	6.536\\
4.25	2.72\\
3.75	2.72\\
};
\addplot [color=blue, forget plot]
  table[row sep=crcr]{%
4.75	1.453\\
4.75	3.632\\
5.25	3.632\\
5.25	1.453\\
4.75	1.453\\
};
\addplot [color=blue, forget plot]
  table[row sep=crcr]{%
5.75	0.373\\
5.75	1.585\\
6.25	1.585\\
6.25	0.373\\
5.75	0.373\\
};
\addplot [color=blue, forget plot]
  table[row sep=crcr]{%
6.75	0.259\\
6.75	0.999\\
7.25	0.999\\
7.25	0.259\\
6.75	0.259\\
};
\addplot [color=blue, forget plot]
  table[row sep=crcr]{%
7.75	0.228\\
7.75	0.898\\
8.25	0.898\\
8.25	0.228\\
7.75	0.228\\
};
\addplot [color=blue, forget plot]
  table[row sep=crcr]{%
8.75	0.221\\
8.75	0.837\\
9.25	0.837\\
9.25	0.221\\
8.75	0.221\\
};
\addplot [color=blue, forget plot]
  table[row sep=crcr]{%
9.75	0.218\\
9.75	0.811\\
10.25	0.811\\
10.25	0.218\\
9.75	0.218\\
};
\addplot [color=blue, forget plot]
  table[row sep=crcr]{%
10.75	0.217\\
10.75	0.757\\
11.25	0.757\\
11.25	0.217\\
10.75	0.217\\
};
\addplot [color=red, forget plot]
  table[row sep=crcr]{%
0.75	12.983\\
1.25	12.983\\
};
\addplot [color=red, forget plot]
  table[row sep=crcr]{%
1.75	9.703\\
2.25	9.703\\
};
\addplot [color=red, forget plot]
  table[row sep=crcr]{%
2.75	7.016\\
3.25	7.016\\
};
\addplot [color=red, forget plot]
  table[row sep=crcr]{%
3.75	4.4865\\
4.25	4.4865\\
};
\addplot [color=red, forget plot]
  table[row sep=crcr]{%
4.75	2.4665\\
5.25	2.4665\\
};
\addplot [color=red, forget plot]
  table[row sep=crcr]{%
5.75	0.921\\
6.25	0.921\\
};
\addplot [color=red, forget plot]
  table[row sep=crcr]{%
6.75	0.5565\\
7.25	0.5565\\
};
\addplot [color=red, forget plot]
  table[row sep=crcr]{%
7.75	0.5035\\
8.25	0.5035\\
};
\addplot [color=red, forget plot]
  table[row sep=crcr]{%
8.75	0.467\\
9.25	0.467\\
};
\addplot [color=red, forget plot]
  table[row sep=crcr]{%
9.75	0.4575\\
10.25	0.4575\\
};
\addplot [color=red, forget plot]
  table[row sep=crcr]{%
10.75	0.4465\\
11.25	0.4465\\
};
\end{axis}
\end{tikzpicture}%
		\label{j1:fig:eval_planner_planning_time_tt}
	} 
	~
	\subfloat[][]{
%
%
\begin{tikzpicture}

\begin{axis}[%
width=\figurewidth,
height=\figureheight,
at={(0\figurewidth,0\figureheight)},
scale only axis,
xmin=0,
xmax=12,
xtick={1,3,5,7,9,11},
xticklabels={{1.0},{1.2},{1.4},{1.6},{1.8},{2.0}},
xlabel style={font=\color{white!15!black}},
xlabel={$\gamma$},
ymin=-0.1,
ymax=10,
ytick={0,2,4,6,8,10},
yticklabels={{0\%},{2\%},{4\%},{6\%},{8\%},{10\%}},
ylabel style={font=\color{white!15!black}},
ylabel style={font=\color{white!15!black}},
ylabel={$\Delta J_D$},
axis background/.style={fill=white},
xmajorgrids,
ymajorgrids
]
\addplot [color=black, dashed, forget plot]
  table[row sep=crcr]{%
1	0\\
1	0\\
};
\addplot [color=black, dashed, forget plot]
  table[row sep=crcr]{%
2	0\\
2	0\\
};
\addplot [color=black, dashed, forget plot]
  table[row sep=crcr]{%
3	0\\
3	0\\
};
\addplot [color=black, dashed, forget plot]
  table[row sep=crcr]{%
4	0\\
4	0\\
};
\addplot [color=black, dashed, forget plot]
  table[row sep=crcr]{%
5	0\\
5	0\\
};
\addplot [color=black, dashed, forget plot]
  table[row sep=crcr]{%
6	1.08406425544496\\
6	2.70852570021545\\
};
\addplot [color=black, dashed, forget plot]
  table[row sep=crcr]{%
7	1.5400570391496\\
7	3.8249483115093\\
};
\addplot [color=black, dashed, forget plot]
  table[row sep=crcr]{%
8	1.88570530616521\\
8	4.69584874640362\\
};
\addplot [color=black, dashed, forget plot]
  table[row sep=crcr]{%
9	1.92606067091113\\
9	4.80142887161168\\
};
\addplot [color=black, dashed, forget plot]
  table[row sep=crcr]{%
10	1.92606067091113\\
10	4.80142887161168\\
};
\addplot [color=black, dashed, forget plot]
  table[row sep=crcr]{%
11	1.92636555678503\\
11	4.80142887161168\\
};
\addplot [color=black, dashed, forget plot]
  table[row sep=crcr]{%
1	0\\
1	0\\
};
\addplot [color=black, dashed, forget plot]
  table[row sep=crcr]{%
2	0\\
2	0\\
};
\addplot [color=black, dashed, forget plot]
  table[row sep=crcr]{%
3	0\\
3	0\\
};
\addplot [color=black, dashed, forget plot]
  table[row sep=crcr]{%
4	0\\
4	0\\
};
\addplot [color=black, dashed, forget plot]
  table[row sep=crcr]{%
5	0\\
5	0\\
};
\addplot [color=black, dashed, forget plot]
  table[row sep=crcr]{%
6	0\\
6	0\\
};
\addplot [color=black, dashed, forget plot]
  table[row sep=crcr]{%
7	0\\
7	0\\
};
\addplot [color=black, dashed, forget plot]
  table[row sep=crcr]{%
8	0\\
8	0\\
};
\addplot [color=black, dashed, forget plot]
  table[row sep=crcr]{%
9	0\\
9	0\\
};
\addplot [color=black, dashed, forget plot]
  table[row sep=crcr]{%
10	0\\
10	0\\
};
\addplot [color=black, dashed, forget plot]
  table[row sep=crcr]{%
11	0\\
11	0\\
};
\addplot [color=black, forget plot]
  table[row sep=crcr]{%
0.875	0\\
1.125	0\\
};
\addplot [color=black, forget plot]
  table[row sep=crcr]{%
1.875	0\\
2.125	0\\
};
\addplot [color=black, forget plot]
  table[row sep=crcr]{%
2.875	0\\
3.125	0\\
};
\addplot [color=black, forget plot]
  table[row sep=crcr]{%
3.875	0\\
4.125	0\\
};
\addplot [color=black, forget plot]
  table[row sep=crcr]{%
4.875	0\\
5.125	0\\
};
\addplot [color=black, forget plot]
  table[row sep=crcr]{%
5.875	2.70852570021545\\
6.125	2.70852570021545\\
};
\addplot [color=black, forget plot]
  table[row sep=crcr]{%
6.875	3.8249483115093\\
7.125	3.8249483115093\\
};
\addplot [color=black, forget plot]
  table[row sep=crcr]{%
7.875	4.69584874640362\\
8.125	4.69584874640362\\
};
\addplot [color=black, forget plot]
  table[row sep=crcr]{%
8.875	4.80142887161168\\
9.125	4.80142887161168\\
};
\addplot [color=black, forget plot]
  table[row sep=crcr]{%
9.875	4.80142887161168\\
10.125	4.80142887161168\\
};
\addplot [color=black, forget plot]
  table[row sep=crcr]{%
10.875	4.80142887161168\\
11.125	4.80142887161168\\
};
\addplot [color=black, forget plot]
  table[row sep=crcr]{%
0.875	0\\
1.125	0\\
};
\addplot [color=black, forget plot]
  table[row sep=crcr]{%
1.875	0\\
2.125	0\\
};
\addplot [color=black, forget plot]
  table[row sep=crcr]{%
2.875	0\\
3.125	0\\
};
\addplot [color=black, forget plot]
  table[row sep=crcr]{%
3.875	0\\
4.125	0\\
};
\addplot [color=black, forget plot]
  table[row sep=crcr]{%
4.875	0\\
5.125	0\\
};
\addplot [color=black, forget plot]
  table[row sep=crcr]{%
5.875	0\\
6.125	0\\
};
\addplot [color=black, forget plot]
  table[row sep=crcr]{%
6.875	0\\
7.125	0\\
};
\addplot [color=black, forget plot]
  table[row sep=crcr]{%
7.875	0\\
8.125	0\\
};
\addplot [color=black, forget plot]
  table[row sep=crcr]{%
8.875	0\\
9.125	0\\
};
\addplot [color=black, forget plot]
  table[row sep=crcr]{%
9.875	0\\
10.125	0\\
};
\addplot [color=black, forget plot]
  table[row sep=crcr]{%
10.875	0\\
11.125	0\\
};
\addplot [color=blue, forget plot]
  table[row sep=crcr]{%
0.75	0\\
0.75	0\\
1.25	0\\
1.25	0\\
0.75	0\\
};
\addplot [color=blue, forget plot]
  table[row sep=crcr]{%
1.75	0\\
1.75	0\\
2.25	0\\
2.25	0\\
1.75	0\\
};
\addplot [color=blue, forget plot]
  table[row sep=crcr]{%
2.75	0\\
2.75	0\\
3.25	0\\
3.25	0\\
2.75	0\\
};
\addplot [color=blue, forget plot]
  table[row sep=crcr]{%
3.75	0\\
3.75	0\\
4.25	0\\
4.25	0\\
3.75	0\\
};
\addplot [color=blue, forget plot]
  table[row sep=crcr]{%
4.75	0\\
4.75	0\\
5.25	0\\
5.25	0\\
4.75	0\\
};
\addplot [color=blue, forget plot]
  table[row sep=crcr]{%
5.75	0\\
5.75	1.08406425544496\\
6.25	1.08406425544496\\
6.25	0\\
5.75	0\\
};
\addplot [color=blue, forget plot]
  table[row sep=crcr]{%
6.75	0\\
6.75	1.5400570391496\\
7.25	1.5400570391496\\
7.25	0\\
6.75	0\\
};
\addplot [color=blue, forget plot]
  table[row sep=crcr]{%
7.75	0\\
7.75	1.88570530616521\\
8.25	1.88570530616521\\
8.25	0\\
7.75	0\\
};
\addplot [color=blue, forget plot]
  table[row sep=crcr]{%
8.75	0\\
8.75	1.92606067091113\\
9.25	1.92606067091113\\
9.25	0\\
8.75	0\\
};
\addplot [color=blue, forget plot]
  table[row sep=crcr]{%
9.75	0\\
9.75	1.92606067091113\\
10.25	1.92606067091113\\
10.25	0\\
9.75	0\\
};
\addplot [color=blue, forget plot]
  table[row sep=crcr]{%
10.75	0\\
10.75	1.92636555678503\\
11.25	1.92636555678503\\
11.25	0\\
10.75	0\\
};
\addplot [color=red, forget plot]
  table[row sep=crcr]{%
0.75	0\\
1.25	0\\
};
\addplot [color=red, forget plot]
  table[row sep=crcr]{%
1.75	0\\
2.25	0\\
};
\addplot [color=red, forget plot]
  table[row sep=crcr]{%
2.75	0\\
3.25	0\\
};
\addplot [color=red, forget plot]
  table[row sep=crcr]{%
3.75	0\\
4.25	0\\
};
\addplot [color=red, forget plot]
  table[row sep=crcr]{%
4.75	0\\
5.25	0\\
};
\addplot [color=red, forget plot]
  table[row sep=crcr]{%
5.75	0\\
6.25	0\\
};
\addplot [color=red, forget plot]
  table[row sep=crcr]{%
6.75	0\\
7.25	0\\
};
\addplot [color=red, forget plot]
  table[row sep=crcr]{%
7.75	0\\
8.25	0\\
};
\addplot [color=red, forget plot]
  table[row sep=crcr]{%
8.75	0\\
9.25	0\\
};
\addplot [color=red, forget plot]
  table[row sep=crcr]{%
9.75	0\\
10.25	0\\
};
\addplot [color=red, forget plot]
  table[row sep=crcr]{%
10.75	0\\
11.25	0\\
};
\end{axis}
\end{tikzpicture}%
		\label{j1:fig:eval_planner_obj_tt}
	}
	\caption{A statistical evaluation of the parking problem scenario (see Figure~\ref{j1:fig:parking_example}) over 1000 Monte Carlo simulations. In (a), a box plot of the planning time as a function of the heuristic inflation factor. In (b), a box plot of the level of suboptimality $\Delta J_D$ as a function of the heuristic inflation factor.}
	\label{j1:fig:eval_planner_tt}
\end{figure}
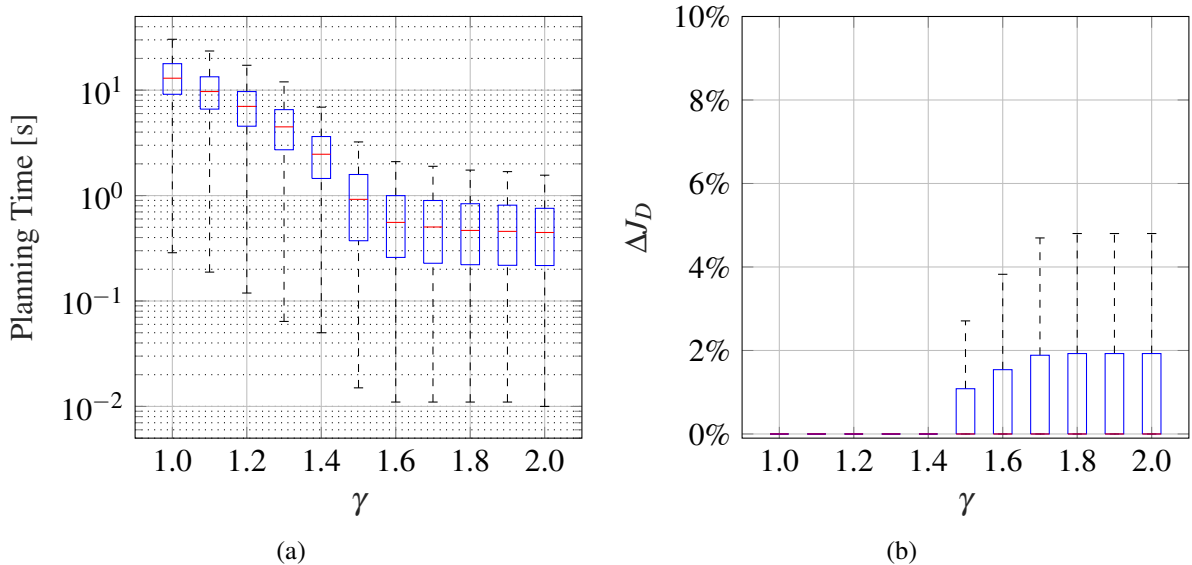
\tikzexternalenable

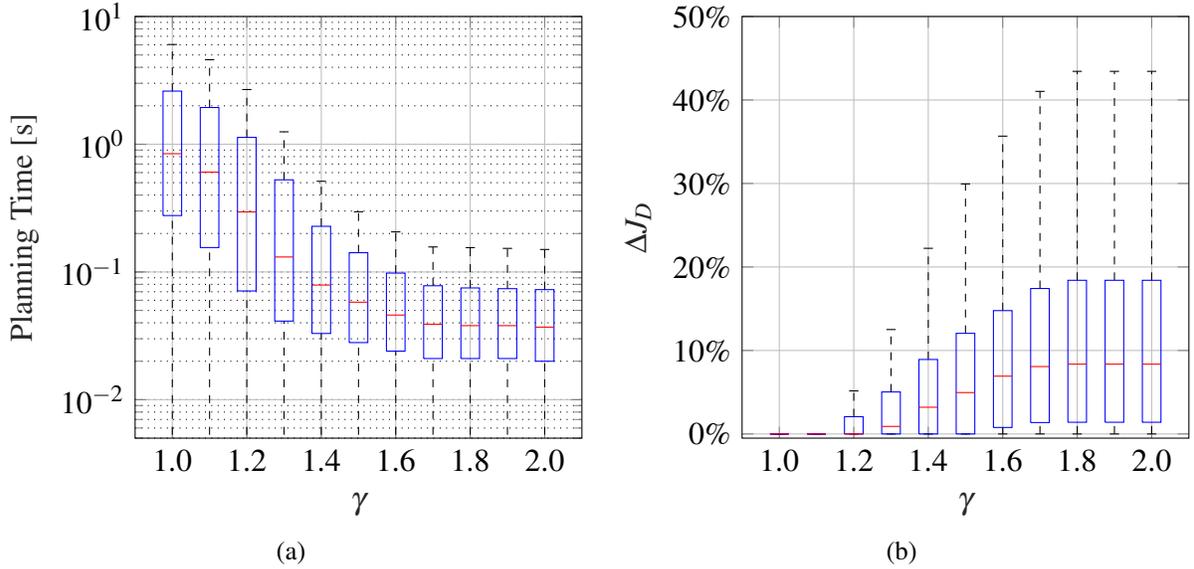
\begin{figure}[t!] 	
	\centering
	\setlength\figureheight{0.34\columnwidth}
	\setlength\figurewidth{0.36\columnwidth} 
	\subfloat[][]{
%
%
\begin{tikzpicture}

\begin{axis}[%
width=\figurewidth,
height=\figureheight,
at={(0\figurewidth,0\figureheight)},
scale only axis,
xmin=0,
xmax=12,
xtick={1,3,5,7,9,11},
xticklabels={{1.0},{1.2},{1.4},{1.6},{1.8},{2.0}},
xlabel style={font=\color{white!15!black}},
xlabel={$\gamma$},
ymode=log,
ymin=0.005,
ymax=10,
ylabel style={font=\color{white!15!black}},
ylabel={Planning Time [s]},
axis background/.style={fill=white},
ymajorgrids,
yminorgrids,
minor y grid style={dotted,black},
xmajorgrids,
xminorgrids,
]
\addplot [color=black, dashed, forget plot]
  table[row sep=crcr]{%
1	2.60575\\
1	6.037\\
};
\addplot [color=black, dashed, forget plot]
  table[row sep=crcr]{%
2	1.939\\
2	4.589\\
};
\addplot [color=black, dashed, forget plot]
  table[row sep=crcr]{%
3	1.1315\\
3	2.681\\
};
\addplot [color=black, dashed, forget plot]
  table[row sep=crcr]{%
4	0.5265\\
4	1.251\\
};
\addplot [color=black, dashed, forget plot]
  table[row sep=crcr]{%
5	0.22775\\
5	0.513\\
};
\addplot [color=black, dashed, forget plot]
  table[row sep=crcr]{%
6	0.1415\\
6	0.296\\
};
\addplot [color=black, dashed, forget plot]
  table[row sep=crcr]{%
7	0.098\\
7	0.206\\
};
\addplot [color=black, dashed, forget plot]
  table[row sep=crcr]{%
8	0.078\\
8	0.157\\
};
\addplot [color=black, dashed, forget plot]
  table[row sep=crcr]{%
9	0.075\\
9	0.155\\
};
\addplot [color=black, dashed, forget plot]
  table[row sep=crcr]{%
10	0.074\\
10	0.153\\
};
\addplot [color=black, dashed, forget plot]
  table[row sep=crcr]{%
11	0.07275\\
11	0.15\\
};
\addplot [color=black, dashed, forget plot]
  table[row sep=crcr]{%
1	0.003\\
1	0.2765\\
};
\addplot [color=black, dashed, forget plot]
  table[row sep=crcr]{%
2	0.003\\
2	0.15525\\
};
\addplot [color=black, dashed, forget plot]
  table[row sep=crcr]{%
3	0.003\\
3	0.071\\
};
\addplot [color=black, dashed, forget plot]
  table[row sep=crcr]{%
4	0.003\\
4	0.04125\\
};
\addplot [color=black, dashed, forget plot]
  table[row sep=crcr]{%
5	0.003\\
5	0.033\\
};
\addplot [color=black, dashed, forget plot]
  table[row sep=crcr]{%
6	0.003\\
6	0.028\\
};
\addplot [color=black, dashed, forget plot]
  table[row sep=crcr]{%
7	0.003\\
7	0.024\\
};
\addplot [color=black, dashed, forget plot]
  table[row sep=crcr]{%
8	0.003\\
8	0.021\\
};
\addplot [color=black, dashed, forget plot]
  table[row sep=crcr]{%
9	0.003\\
9	0.021\\
};
\addplot [color=black, dashed, forget plot]
  table[row sep=crcr]{%
10	0.003\\
10	0.021\\
};
\addplot [color=black, dashed, forget plot]
  table[row sep=crcr]{%
11	0.003\\
11	0.02\\
};
\addplot [color=black, forget plot]
  table[row sep=crcr]{%
0.875	6.037\\
1.125	6.037\\
};
\addplot [color=black, forget plot]
  table[row sep=crcr]{%
1.875	4.589\\
2.125	4.589\\
};
\addplot [color=black, forget plot]
  table[row sep=crcr]{%
2.875	2.681\\
3.125	2.681\\
};
\addplot [color=black, forget plot]
  table[row sep=crcr]{%
3.875	1.251\\
4.125	1.251\\
};
\addplot [color=black, forget plot]
  table[row sep=crcr]{%
4.875	0.513\\
5.125	0.513\\
};
\addplot [color=black, forget plot]
  table[row sep=crcr]{%
5.875	0.296\\
6.125	0.296\\
};
\addplot [color=black, forget plot]
  table[row sep=crcr]{%
6.875	0.206\\
7.125	0.206\\
};
\addplot [color=black, forget plot]
  table[row sep=crcr]{%
7.875	0.157\\
8.125	0.157\\
};
\addplot [color=black, forget plot]
  table[row sep=crcr]{%
8.875	0.155\\
9.125	0.155\\
};
\addplot [color=black, forget plot]
  table[row sep=crcr]{%
9.875	0.153\\
10.125	0.153\\
};
\addplot [color=black, forget plot]
  table[row sep=crcr]{%
10.875	0.15\\
11.125	0.15\\
};
\addplot [color=black, forget plot]
  table[row sep=crcr]{%
0.875	0.003\\
1.125	0.003\\
};
\addplot [color=black, forget plot]
  table[row sep=crcr]{%
1.875	0.003\\
2.125	0.003\\
};
\addplot [color=black, forget plot]
  table[row sep=crcr]{%
2.875	0.003\\
3.125	0.003\\
};
\addplot [color=black, forget plot]
  table[row sep=crcr]{%
3.875	0.003\\
4.125	0.003\\
};
\addplot [color=black, forget plot]
  table[row sep=crcr]{%
4.875	0.003\\
5.125	0.003\\
};
\addplot [color=black, forget plot]
  table[row sep=crcr]{%
5.875	0.003\\
6.125	0.003\\
};
\addplot [color=black, forget plot]
  table[row sep=crcr]{%
6.875	0.003\\
7.125	0.003\\
};
\addplot [color=black, forget plot]
  table[row sep=crcr]{%
7.875	0.003\\
8.125	0.003\\
};
\addplot [color=black, forget plot]
  table[row sep=crcr]{%
8.875	0.003\\
9.125	0.003\\
};
\addplot [color=black, forget plot]
  table[row sep=crcr]{%
9.875	0.003\\
10.125	0.003\\
};
\addplot [color=black, forget plot]
  table[row sep=crcr]{%
10.875	0.003\\
11.125	0.003\\
};
\addplot [color=blue, forget plot]
  table[row sep=crcr]{%
0.75	0.2765\\
0.75	2.60575\\
1.25	2.60575\\
1.25	0.2765\\
0.75	0.2765\\
};
\addplot [color=blue, forget plot]
  table[row sep=crcr]{%
1.75	0.15525\\
1.75	1.939\\
2.25	1.939\\
2.25	0.15525\\
1.75	0.15525\\
};
\addplot [color=blue, forget plot]
  table[row sep=crcr]{%
2.75	0.071\\
2.75	1.1315\\
3.25	1.1315\\
3.25	0.071\\
2.75	0.071\\
};
\addplot [color=blue, forget plot]
  table[row sep=crcr]{%
3.75	0.04125\\
3.75	0.5265\\
4.25	0.5265\\
4.25	0.04125\\
3.75	0.04125\\
};
\addplot [color=blue, forget plot]
  table[row sep=crcr]{%
4.75	0.033\\
4.75	0.22775\\
5.25	0.22775\\
5.25	0.033\\
4.75	0.033\\
};
\addplot [color=blue, forget plot]
  table[row sep=crcr]{%
5.75	0.028\\
5.75	0.1415\\
6.25	0.1415\\
6.25	0.028\\
5.75	0.028\\
};
\addplot [color=blue, forget plot]
  table[row sep=crcr]{%
6.75	0.024\\
6.75	0.098\\
7.25	0.098\\
7.25	0.024\\
6.75	0.024\\
};
\addplot [color=blue, forget plot]
  table[row sep=crcr]{%
7.75	0.021\\
7.75	0.078\\
8.25	0.078\\
8.25	0.021\\
7.75	0.021\\
};
\addplot [color=blue, forget plot]
  table[row sep=crcr]{%
8.75	0.021\\
8.75	0.075\\
9.25	0.075\\
9.25	0.021\\
8.75	0.021\\
};
\addplot [color=blue, forget plot]
  table[row sep=crcr]{%
9.75	0.021\\
9.75	0.074\\
10.25	0.074\\
10.25	0.021\\
9.75	0.021\\
};
\addplot [color=blue, forget plot]
  table[row sep=crcr]{%
10.75	0.02\\
10.75	0.07275\\
11.25	0.07275\\
11.25	0.02\\
10.75	0.02\\
};
\addplot [color=red, forget plot]
  table[row sep=crcr]{%
0.75	0.843\\
1.25	0.843\\
};
\addplot [color=red, forget plot]
  table[row sep=crcr]{%
1.75	0.605\\
2.25	0.605\\
};
\addplot [color=red, forget plot]
  table[row sep=crcr]{%
2.75	0.295\\
3.25	0.295\\
};
\addplot [color=red, forget plot]
  table[row sep=crcr]{%
3.75	0.131\\
4.25	0.131\\
};
\addplot [color=red, forget plot]
  table[row sep=crcr]{%
4.75	0.079\\
5.25	0.079\\
};
\addplot [color=red, forget plot]
  table[row sep=crcr]{%
5.75	0.058\\
6.25	0.058\\
};
\addplot [color=red, forget plot]
  table[row sep=crcr]{%
6.75	0.046\\
7.25	0.046\\
};
\addplot [color=red, forget plot]
  table[row sep=crcr]{%
7.75	0.039\\
8.25	0.039\\
};
\addplot [color=red, forget plot]
  table[row sep=crcr]{%
8.75	0.038\\
9.25	0.038\\
};
\addplot [color=red, forget plot]
  table[row sep=crcr]{%
9.75	0.038\\
10.25	0.038\\
};
\addplot [color=red, forget plot]
  table[row sep=crcr]{%
10.75	0.037\\
11.25	0.037\\
};
\end{axis}
\end{tikzpicture}%
		\label{j1:fig:eval_planner_planning_time_cs}
	} 
	~
	\subfloat[][]{
%
%
\begin{tikzpicture}

\begin{axis}[%
width=\figurewidth,
height=\figureheight,
at={(0\figurewidth,0\figureheight)},
scale only axis,
xmin=0,
xmax=12,
xtick={1,3,5,7,9,11},
xticklabels={{1.0},{1.2},{1.4},{1.6},{1.8},{2.0}},
xlabel style={font=\color{white!15!black}},
xlabel={$\gamma$},
ymin=-0.5,
ymax=50,
ytick={0,10,20,30,40,50},
yticklabels={{0\%},{10\%},{20\%},{30\%},{40\%},{50\%}},
ylabel style={font=\color{white!15!black}},
ylabel style={font=\color{white!15!black}},
ylabel={$\Delta J_D$},
axis background/.style={fill=white},
xmajorgrids,
ymajorgrids
]
\addplot [color=black, dashed, forget plot]
  table[row sep=crcr]{%
1	0\\
1	0\\
};
\addplot [color=black, dashed, forget plot]
  table[row sep=crcr]{%
2	0\\
2	0\\
};
\addplot [color=black, dashed, forget plot]
  table[row sep=crcr]{%
3	2.06852643249851\\
3	5.15895953757225\\
};
\addplot [color=black, dashed, forget plot]
  table[row sep=crcr]{%
4	5.03305801975678\\
4	12.5071022727273\\
};
\addplot [color=black, dashed, forget plot]
  table[row sep=crcr]{%
5	8.92139257877975\\
5	22.2364217252396\\
};
\addplot [color=black, dashed, forget plot]
  table[row sep=crcr]{%
6	12.0536068396939\\
6	29.9473234624146\\
};
\addplot [color=black, dashed, forget plot]
  table[row sep=crcr]{%
7	14.7740413169889\\
7	35.6696471493126\\
};
\addplot [color=black, dashed, forget plot]
  table[row sep=crcr]{%
8	17.4222701851573\\
8	41.0321355073611\\
};
\addplot [color=black, dashed, forget plot]
  table[row sep=crcr]{%
9	18.4058411254415\\
9	43.4308705193855\\
};
\addplot [color=black, dashed, forget plot]
  table[row sep=crcr]{%
10	18.4058411254415\\
10	43.4308705193855\\
};
\addplot [color=black, dashed, forget plot]
  table[row sep=crcr]{%
11	18.4058411254415\\
11	43.4308705193855\\
};
\addplot [color=black, dashed, forget plot]
  table[row sep=crcr]{%
1	0\\
1	0\\
};
\addplot [color=black, dashed, forget plot]
  table[row sep=crcr]{%
2	0\\
2	0\\
};
\addplot [color=black, dashed, forget plot]
  table[row sep=crcr]{%
3	0\\
3	0\\
};
\addplot [color=black, dashed, forget plot]
  table[row sep=crcr]{%
4	0\\
4	0\\
};
\addplot [color=black, dashed, forget plot]
  table[row sep=crcr]{%
5	0\\
5	0\\
};
\addplot [color=black, dashed, forget plot]
  table[row sep=crcr]{%
6	0\\
6	0\\
};
\addplot [color=black, dashed, forget plot]
  table[row sep=crcr]{%
7	0\\
7	0.776607926565172\\
};
\addplot [color=black, dashed, forget plot]
  table[row sep=crcr]{%
8	0\\
8	1.36727120689108\\
};
\addplot [color=black, dashed, forget plot]
  table[row sep=crcr]{%
9	0\\
9	1.4021080992544\\
};
\addplot [color=black, dashed, forget plot]
  table[row sep=crcr]{%
10	0\\
10	1.4021080992544\\
};
\addplot [color=black, dashed, forget plot]
  table[row sep=crcr]{%
11	0\\
11	1.4021080992544\\
};
\addplot [color=black, forget plot]
  table[row sep=crcr]{%
0.875	0\\
1.125	0\\
};
\addplot [color=black, forget plot]
  table[row sep=crcr]{%
1.875	0\\
2.125	0\\
};
\addplot [color=black, forget plot]
  table[row sep=crcr]{%
2.875	5.15895953757225\\
3.125	5.15895953757225\\
};
\addplot [color=black, forget plot]
  table[row sep=crcr]{%
3.875	12.5071022727273\\
4.125	12.5071022727273\\
};
\addplot [color=black, forget plot]
  table[row sep=crcr]{%
4.875	22.2364217252396\\
5.125	22.2364217252396\\
};
\addplot [color=black, forget plot]
  table[row sep=crcr]{%
5.875	29.9473234624146\\
6.125	29.9473234624146\\
};
\addplot [color=black, forget plot]
  table[row sep=crcr]{%
6.875	35.6696471493126\\
7.125	35.6696471493126\\
};
\addplot [color=black, forget plot]
  table[row sep=crcr]{%
7.875	41.0321355073611\\
8.125	41.0321355073611\\
};
\addplot [color=black, forget plot]
  table[row sep=crcr]{%
8.875	43.4308705193855\\
9.125	43.4308705193855\\
};
\addplot [color=black, forget plot]
  table[row sep=crcr]{%
9.875	43.4308705193855\\
10.125	43.4308705193855\\
};
\addplot [color=black, forget plot]
  table[row sep=crcr]{%
10.875	43.4308705193855\\
11.125	43.4308705193855\\
};
\addplot [color=black, forget plot]
  table[row sep=crcr]{%
0.875	0\\
1.125	0\\
};
\addplot [color=black, forget plot]
  table[row sep=crcr]{%
1.875	0\\
2.125	0\\
};
\addplot [color=black, forget plot]
  table[row sep=crcr]{%
2.875	0\\
3.125	0\\
};
\addplot [color=black, forget plot]
  table[row sep=crcr]{%
3.875	0\\
4.125	0\\
};
\addplot [color=black, forget plot]
  table[row sep=crcr]{%
4.875	0\\
5.125	0\\
};
\addplot [color=black, forget plot]
  table[row sep=crcr]{%
5.875	0\\
6.125	0\\
};
\addplot [color=black, forget plot]
  table[row sep=crcr]{%
6.875	0\\
7.125	0\\
};
\addplot [color=black, forget plot]
  table[row sep=crcr]{%
7.875	0\\
8.125	0\\
};
\addplot [color=black, forget plot]
  table[row sep=crcr]{%
8.875	0\\
9.125	0\\
};
\addplot [color=black, forget plot]
  table[row sep=crcr]{%
9.875	0\\
10.125	0\\
};
\addplot [color=black, forget plot]
  table[row sep=crcr]{%
10.875	0\\
11.125	0\\
};
\addplot [color=blue, forget plot]
  table[row sep=crcr]{%
0.75	0\\
0.75	0\\
1.25	0\\
1.25	0\\
0.75	0\\
};
\addplot [color=blue, forget plot]
  table[row sep=crcr]{%
1.75	0\\
1.75	0\\
2.25	0\\
2.25	0\\
1.75	0\\
};
\addplot [color=blue, forget plot]
  table[row sep=crcr]{%
2.75	0\\
2.75	2.06852643249851\\
3.25	2.06852643249851\\
3.25	0\\
2.75	0\\
};
\addplot [color=blue, forget plot]
  table[row sep=crcr]{%
3.75	0\\
3.75	5.03305801975678\\
4.25	5.03305801975678\\
4.25	0\\
3.75	0\\
};
\addplot [color=blue, forget plot]
  table[row sep=crcr]{%
4.75	0\\
4.75	8.92139257877975\\
5.25	8.92139257877975\\
5.25	0\\
4.75	0\\
};
\addplot [color=blue, forget plot]
  table[row sep=crcr]{%
5.75	0\\
5.75	12.0536068396939\\
6.25	12.0536068396939\\
6.25	0\\
5.75	0\\
};
\addplot [color=blue, forget plot]
  table[row sep=crcr]{%
6.75	0.776607926565172\\
6.75	14.7740413169889\\
7.25	14.7740413169889\\
7.25	0.776607926565172\\
6.75	0.776607926565172\\
};
\addplot [color=blue, forget plot]
  table[row sep=crcr]{%
7.75	1.36727120689108\\
7.75	17.4222701851573\\
8.25	17.4222701851573\\
8.25	1.36727120689108\\
7.75	1.36727120689108\\
};
\addplot [color=blue, forget plot]
  table[row sep=crcr]{%
8.75	1.4021080992544\\
8.75	18.4058411254415\\
9.25	18.4058411254415\\
9.25	1.4021080992544\\
8.75	1.4021080992544\\
};
\addplot [color=blue, forget plot]
  table[row sep=crcr]{%
9.75	1.4021080992544\\
9.75	18.4058411254415\\
10.25	18.4058411254415\\
10.25	1.4021080992544\\
9.75	1.4021080992544\\
};
\addplot [color=blue, forget plot]
  table[row sep=crcr]{%
10.75	1.4021080992544\\
10.75	18.4058411254415\\
11.25	18.4058411254415\\
11.25	1.4021080992544\\
10.75	1.4021080992544\\
};
\addplot [color=red, forget plot]
  table[row sep=crcr]{%
0.75	0\\
1.25	0\\
};
\addplot [color=red, forget plot]
  table[row sep=crcr]{%
1.75	0\\
2.25	0\\
};
\addplot [color=red, forget plot]
  table[row sep=crcr]{%
2.75	0\\
3.25	0\\
};
\addplot [color=red, forget plot]
  table[row sep=crcr]{%
3.75	0.904977375565611\\
4.25	0.904977375565611\\
};
\addplot [color=red, forget plot]
  table[row sep=crcr]{%
4.75	3.19862520095349\\
5.25	3.19862520095349\\
};
\addplot [color=red, forget plot]
  table[row sep=crcr]{%
5.75	4.95672698662471\\
6.25	4.95672698662471\\
};
\addplot [color=red, forget plot]
  table[row sep=crcr]{%
6.75	6.92218946288937\\
7.25	6.92218946288937\\
};
\addplot [color=red, forget plot]
  table[row sep=crcr]{%
7.75	8.07236881649336\\
8.25	8.07236881649336\\
};
\addplot [color=red, forget plot]
  table[row sep=crcr]{%
8.75	8.3637311099531\\
9.25	8.3637311099531\\
};
\addplot [color=red, forget plot]
  table[row sep=crcr]{%
9.75	8.3637311099531\\
10.25	8.3637311099531\\
};
\addplot [color=red, forget plot]
  table[row sep=crcr]{%
10.75	8.3637311099531\\
11.25	8.3637311099531\\
};
\end{axis}
\end{tikzpicture}%
		\label{j1:fig:eval_planner_obj_cs}
	}
	\caption{A statistical evaluation of the loading/offloading scenario (see Figure~\ref{j1:fig:construction_site_example}) over 1000 Monte Carlo simulations. In (a), a box plot of the planning time as a function of the heuristic inflation factor and. In (b), a box plot of the level of suboptimality $\Delta J_D$ as a function of the heuristic inflation factor.}
	\label{j1:fig:eval_planner_cs}
\end{figure}

In all simulations, the lattice planner was able to find an optimal path ($\gamma=1$) to the goal within the allowed planning time of 60 s (max: 40 s). 
A statistical evaluation of the simulation results from one thousand Monte Carlo simulations are provided in Figure~\ref{j1:fig:eval_planner_tt}, where the planning time (Figure~\ref{j1:fig:eval_planner_planning_time_tt}), and the level of suboptimality $\Delta J_D$ (Figure~\ref{j1:fig:eval_planner_obj_cs}) between the cost $J_D$ for a specific $\gamma$ and the optimal cost $J_D^*$ for each planning experiment are plotted.
In the box plots, the red central mark of each bar is the median, the bottom and top edges of the boxes indicate the 25th and 75th percentiles, respectively, and the whiskers extends to the most extreme data points.    

As can be seen in Figure~\ref{j1:fig:eval_planner_planning_time_tt}, the planning time is drastically increasing with decreasing $\gamma$. 
For most of the problems, a feasible solution to the goal with $\gamma=2$ was found within $0.7$ s, while a median planning time of \mbox{13 s} was needed to find an optimal solution with $\gamma = 1$. 
In Figure~\ref{j1:fig:eval_planner_obj_tt}, the quality of the produced solution in terms of level of suboptimality $\Delta J_D$ as a function of $\gamma$ is displayed. 
For $\gamma\geq 1$, the provided theoretical guarantee is that the cost for a feasible solution $J_D$ satisfies $J_D \leq \gamma J_D^*$, where $J_D^*$ denotes the optimal cost. 
For all iterations of the ARA$^*$, the median level of suboptimality is $0$ \% and the extreme values for large $\gamma$ are about $5$ \%. For this scenario, it is clear that the guaranteed upper bound of $\gamma$-suboptimality is a conservative bound.

\tikzexternaldisable
\tikzexternalenable 

\begin{figure}[t!]
	\begin{center}
		\includegraphics[width=0.75\linewidth]{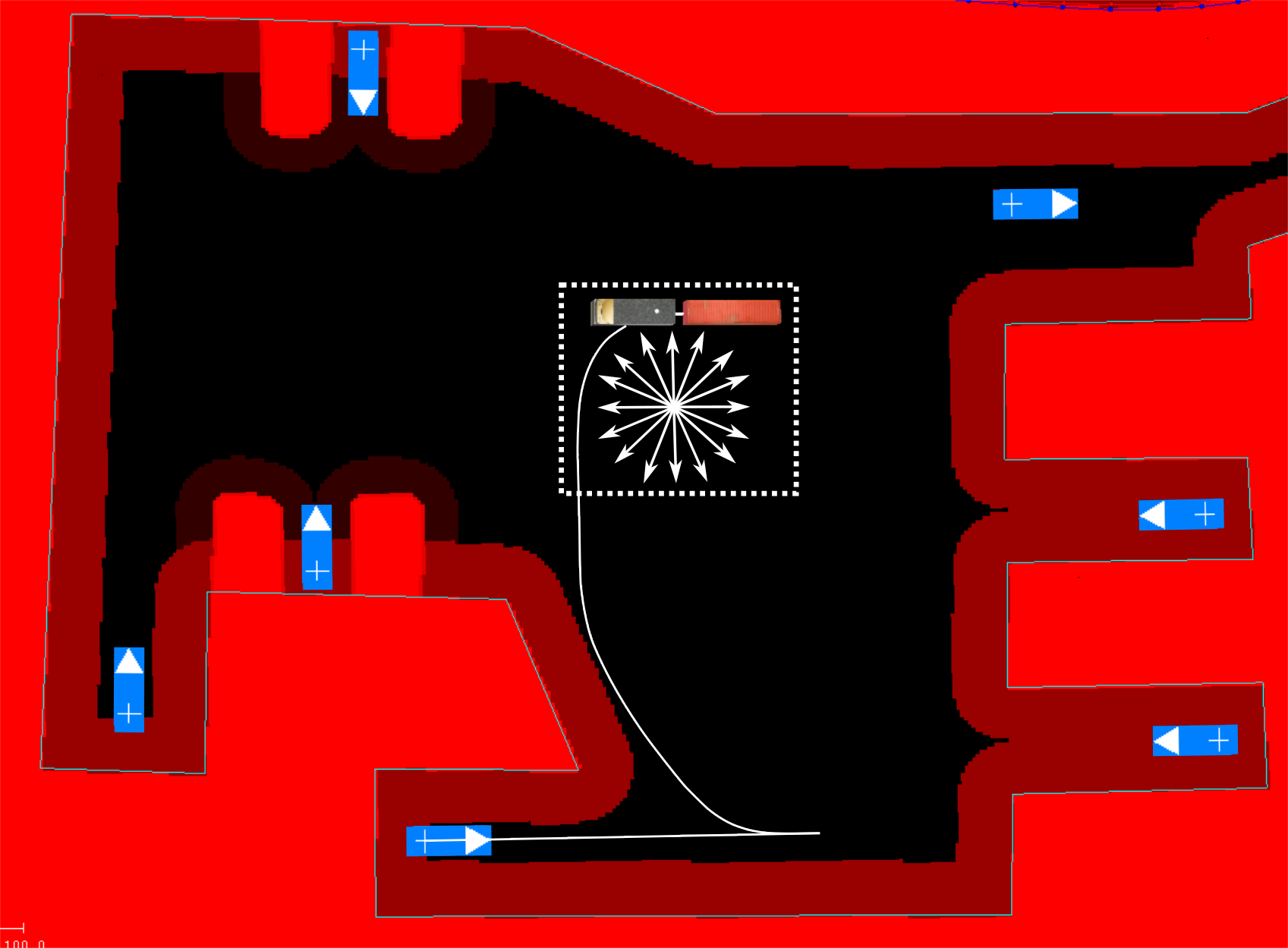}   
		\caption{An overview of the loading/offloading site planning problem. The goal positions of the axle of the semitrailer $(x_{3,G},y_{3,G})$ are illustrated by the white crosses inside the blue rectangles, where the white arrow specifies its goal orientation $\theta_{3,G}$. The initial position $(x_{3,I},y_{3,I})$ is uniformly sampled within the two white-dotted rectangles, and the initial orientation $\theta_{3,I}$ are sampled from sixteen different initial orientations. The white path illustrates the planned path for the axle of the semitrailer for one case out of 1000 Monte Carlo simulations. The area occupied by obstacles is colored in red and the black area is free-space.} 
		\label{j1:fig:construction_site_example}
	\end{center} 
\end{figure}

A loading/offloading site is used as the second planning scenario and the setup is illustrated in Figure~\ref{j1:fig:construction_site_example}. 
In this scenario, the lattice planner has to plan a path from a randomly selected initial state $z_I\in\mathbb Z_d$ to one of the six loading bays, or
plan how to exit the site. 
In the Monte Carlo simulations, the initial position of the semitrailer ($x_{3,I},y_{3,I}$) is uniformly sampled from a \mbox{25 m $\times$ 25 m} square (see,~Figure~\ref{j1:fig:construction_site_example}), and the initial orientation of the semitrailer $\theta_{3,I}$ is randomly selected from one of its sixteen discretized orientations, $i.e.$, $\theta_{3,I}\in\Theta$. 

Also in this scenario, the lattice planner was always able to find an optimal path to the goal within the allowed planning time of 60 s (max: 27 s). 
A statistical evaluation of the simulation results from one thousand Monte Carlo simulations are presented in Figure~\ref{j1:fig:eval_planner_cs}.  
From the box plots in~\ref{j1:fig:eval_planner_cs}, it can be seen that the planning time is also in this scenario increasing with decreasing heuristic inflation factor $\gamma$. However, the median planning time to find an optimal solution with $\gamma=1$ was only 0.84 s and most problems where solved within $3$ s. 
The main reason for this improvement in terms of planning time compared to the parking scenario is because the precomputed HLUT yields a better estimation of the true cost-to-go in this less constrained environment. 
However, as can be seen in Figure~\ref{j1:fig:eval_planner_obj_cs}, the extreme values for the level of suboptimality $\Delta J_D$ is about $43$ \% for large $\gamma$.   
Compared to the parking scenario, a heuristic inflation factor of $\gamma = 1.2$ is needed in this scenario to obtain a median level of suboptimality of \mbox{$0$ \%}. One reason for this greedy behavior in this scenario compared to parking scenario is that the there exist more alternative paths to the goal. This implies that the probability of finding a suboptimal path to the goal increases~\cite{arastar}.

\tikzexternaldisable
\tikzexternalenable

\tikzexternaldisable
\begin{figure*}[b!]
	\centering
	\setlength\figureheight{0.36\textwidth}
	\setlength\figurewidth{0.74\textwidth}
	\begin{tikzpicture}
	\node[anchor=south west] (myplot) at (0,0) {
		\input{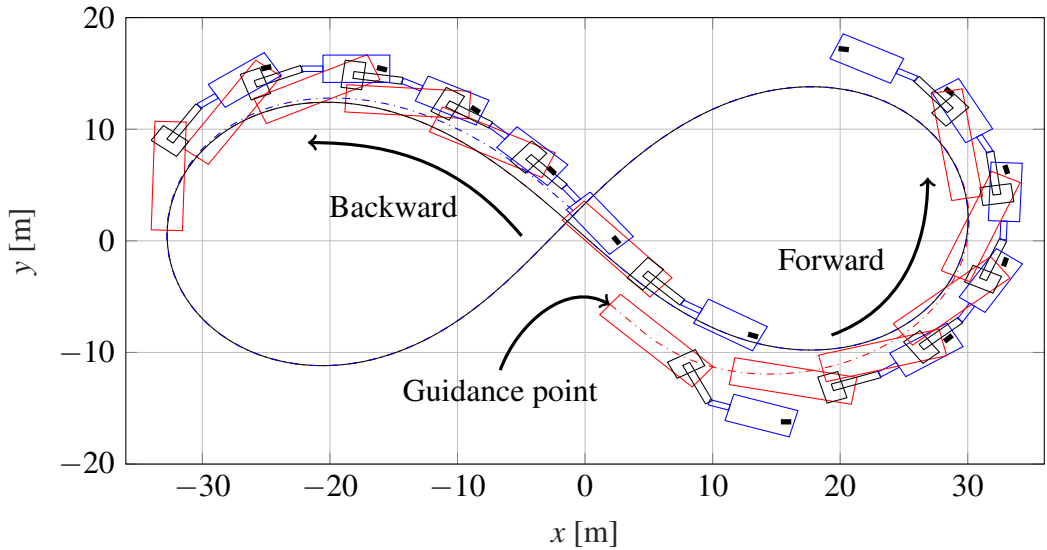}
	};
	\begin{scope}[x={(myplot.south east)}, y={(myplot.north west)}]
	\draw[<-,very thick] (0.3,0.73) to [out=0,in=130] (0.5,0.57);  
	\node at (0.38,0.62) {Backward};
	\draw[->,very thick] (0.79,0.4) to [out=20,in=270] (0.88,0.67);  
	\node at (0.79,0.53) {Forward};
	\draw[->,very thick] (0.48,0.34) to [out=70,in=150] (0.5825,0.452);  
	\node at (0.48,0.3) {Guidance point};
	\end{scope}
	\end{tikzpicture}
	\caption{Simulation results of backward and forward tracking of the figure-eight nominal path, where the nominal path of the axle of the semitrailer $(x_{3,r}(\cdot),y_{3,r}(\cdot))$ is the black solid line. The dashed-dotted blue (red) path is the path taken by the axle of the semitrailer $(x_3(\cdot),y_3(\cdot))$ during the backward (forward) path execution.} 
	\label{j1:fig:eight_rev_sim}
\end{figure*}
\tikzexternalenable

\tikzexternaldisable
\begin{figure}[t!]
	\vspace{-40pt}
	\centering
	\setlength\figureheight{0.14\columnwidth}
	\setlength\figurewidth{0.36\columnwidth}
	\subfloat[][Lateral error for the axle of the semitrailer.]{
		\input{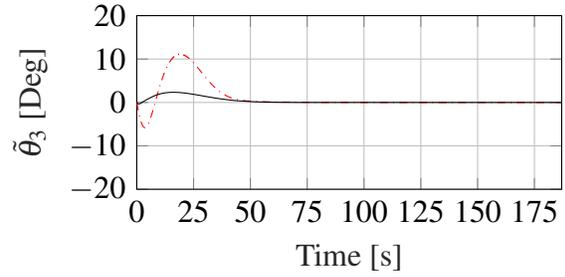}
		\label{j1:fig:sim_eight_rev_z3}
	}
	~
	\subfloat[][Orientation error of the semitrailer.]{
		\input{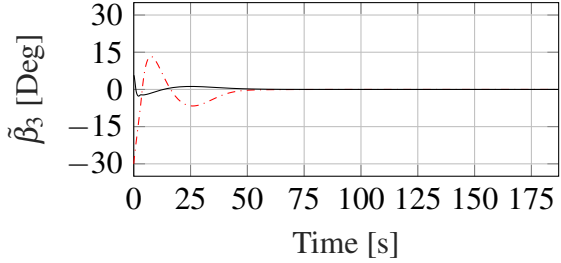}
		\label{j1:fig:sim_eight_rev_theta3}
	}
	\quad
	\subfloat[][The joint angle error between the semitrailer and the dolly.]{
%
%
\begin{tikzpicture}

\begin{axis}[%
width=0.951\figurewidth,
height=\figureheight,
at={(0\figurewidth,0\figureheight)},
scale only axis,
xmin=0,
xmax=187,
xlabel style={font=\color{white!15!black}},
xlabel={Time [s]},
ymin=-35,
ymax=35,
ylabel style={font=\color{white!15!black}},
ylabel={$\tilde\beta_3$ [Deg]},
axis background/.style={fill=white},
xmajorgrids,
ymajorgrids,
xtick={0,25,50,75,100,125,150,175},
ytick={-30.0,-15.0,0.0,15.0,30.0},
]
\addplot [color=red, dashdotted, forget plot]
  table[row sep=crcr]{%
0	-29.9999985154569\\
0.240000000000009	-27.7112380732116\\
0.680000000000007	-24.1064323614202\\
3	-6.60033919757242\\
3.96000000000001	1.87066271880784\\
4.31999999999999	4.52142297702088\\
4.63999999999999	6.48739185661165\\
4.88	7.75056359050313\\
5.19999999999999	9.17714918925915\\
5.52000000000001	10.3335141186668\\
5.72	10.9496910347893\\
6.03999999999999	11.7560851453821\\
6.36000000000001	12.3715743789923\\
6.68000000000001	12.8157208012659\\
7.12	13.1758205988257\\
7.44	13.2853296061503\\
7.84	13.2622072635901\\
8.31999999999999	13.0353365066148\\
8.88	12.5463426159982\\
9.59999999999999	11.6431921470838\\
10.72	9.82765027451879\\
12.12	7.22530709584382\\
13.8	4.05433519028466\\
14.96	2.01144770496475\\
15.24	1.53641069781159\\
16.4	-0.265064305796443\\
18.16	-2.5599539305048\\
19.48	-3.92872569938382\\
20.6	-4.85692959396616\\
22	-5.72593815492317\\
23	-6.16561701070717\\
23.96	-6.4522276377846\\
24.88	-6.61478296866517\\
25.76	-6.67382501333796\\
26.64	-6.64892657294237\\
27.52	-6.54558290348285\\
28.48	-6.35737858219613\\
29.48	-6.08821102241279\\
30.16	-5.87030209419669\\
30.96	-5.58382043024696\\
32.4	-5.0085831053982\\
36.92	-3.07723611428347\\
38.24	-2.56780699337\\
39.52	-2.12242024473855\\
40.72	-1.75210212571133\\
41.92	-1.43188806899863\\
43.16	-1.14931838602186\\
44.36	-0.923546977966737\\
45.72	-0.717032293632144\\
47.12	-0.551675504581084\\
48.68	-0.414254514004455\\
50.44	-0.304513447133132\\
52.64	-0.215664423900762\\
55.44	-0.151667738020052\\
59.72	-0.103116846829039\\
70.48	-0.0416013452617108\\
82.88	-0.0100990827639578\\
98.84	-0.00363247359027241\\
187	-0.0023544291628923\\
};
\addplot [color=black, forget plot]
  table[row sep=crcr]{%
0	5.72957795130824\\
0.0800000000000125	5.59446853948532\\
0.199999999999989	5.17386105617931\\
0.360000000000014	4.30503076976026\\
0.560000000000002	2.84771530593972\\
0.919999999999987	0.0558066385106031\\
1.12	-0.990879619129942\\
1.28	-1.5637431456139\\
1.44	-1.97328766383669\\
1.63999999999999	-2.33041532007934\\
1.80000000000001	-2.52359848804849\\
1.96000000000001	-2.64398223030486\\
2.12	-2.69504561447872\\
2.24000000000001	-2.68618132307734\\
2.36000000000001	-2.63553969509724\\
2.56	-2.47291055113649\\
2.84	-2.2584494596249\\
3.03999999999999	-2.1811622201549\\
3.28	-2.16168146809511\\
3.63999999999999	-2.1971676247438\\
4.16	-2.22397016042942\\
4.72	-2.17238344176528\\
5.80000000000001	-1.96729987222352\\
6.59999999999999	-1.77147254595064\\
11.92	-0.261454943068031\\
14.56	0.321388466824345\\
17.24	0.745816261143347\\
20.12	1.03315808958564\\
23.24	1.18047876547979\\
26.64	1.19018728727619\\
30.44	1.07475998172146\\
36.4	0.770853270733767\\
42.96	0.447830009374798\\
47.32	0.288588650530443\\
51.84	0.176215093252694\\
59.2	0.0699593283315494\\
66.24	0.0245894958114548\\
77.32	0.00294846213952837\\
120.72	-0.00230055458646916\\
187.04	0.00246715564495048\\
};
\end{axis}
\end{tikzpicture}%
		\label{j1:fig:sim_eight_rev_b3}
	}
	~
	\subfloat[][The joint angle error between the dolly and the tractor.]{
%
%
\begin{tikzpicture}

\begin{axis}[%
width=0.951\figurewidth,
height=\figureheight,
at={(0\figurewidth,0\figureheight)},
scale only axis,
xmin=0,
xmax=187,
xlabel style={font=\color{white!15!black}},
xlabel={Time [s]},
ymin=-35,
ymax=35,
ylabel style={font=\color{white!15!black}},
ylabel={$\tilde\beta_2$ [Deg]},
axis background/.style={fill=white},
xmajorgrids,
ymajorgrids,
xtick={0,25,50,75,100,125,150,175},
ytick={-30.0,-15.0,0.0,15.0,30.0},
]
\addplot [color=red, dashdotted, forget plot]
  table[row sep=crcr]{%
0	30.0001189182506\\
0.0800000000000125	29.5817348709179\\
0.159999999999997	29.3553117217293\\
0.240000000000009	29.2702700349465\\
0.319999999999993	29.2869113536026\\
0.400000000000006	29.3746762310556\\
0.560000000000002	29.6787338147336\\
1.44	31.9105024797492\\
1.84	32.8245409729728\\
2.44	34.0219703271827\\
2.72	34.494717426355\\
2.80000000000001	34.5293434724455\\
2.84	34.5040005780101\\
2.91999999999999	34.3462416982499\\
3	34.0477875241145\\
3.12	33.3509233611403\\
3.28	32.0429159919255\\
3.52000000000001	29.5329509781735\\
4.52000000000001	18.4474383460874\\
4.88	15.2232909463866\\
5.28	12.0874380491771\\
5.68000000000001	9.31916698450468\\
6.16	6.41240220982203\\
6.64000000000001	3.89564332109484\\
7.12	1.72145958739915\\
7.47999999999999	0.289927037129644\\
7.96000000000001	-1.37546782674423\\
8.44	-2.79720382985394\\
8.91999999999999	-4.00548006648799\\
9.40000000000001	-5.02200292036116\\
9.88	-5.86642654426737\\
10.36	-6.55865426832534\\
10.84	-7.12017201348314\\
11.36	-7.60221927673484\\
11.88	-7.96763883909313\\
12.2	-8.14253067330176\\
12.52	-8.2832169474415\\
13.12	-8.4619722885777\\
13.76	-8.54954559298048\\
14.12	-8.56209792352504\\
14.84	-8.50176660490897\\
15.68	-8.33316628778618\\
16.56	-8.06229477006409\\
17.2	-7.81796111921028\\
17.8	-7.55864468075148\\
19.08	-6.929545947283\\
20.84	-5.93246091789828\\
25.68	-2.99892395646592\\
27.08	-2.21439893616298\\
28.4	-1.55046133099634\\
29.52	-1.04793391206232\\
30.52	-0.658010869351244\\
31.64	-0.291484181447373\\
32.6	-0.0302988831504933\\
33.56	0.177654014507425\\
34.56	0.339252106256623\\
35.72	0.47083624414131\\
36.84	0.542559667132537\\
38.24	0.571255659136085\\
39.68	0.551392901827086\\
42.08	0.446776363311159\\
49.64	0.0879264331248635\\
52.96	0.0200088910962677\\
57.2	-0.00997097497952382\\
102.96	-0.00409089065914259\\
187	-0.00105357783422733\\
};
\addplot [color=black, forget plot]
  table[row sep=crcr]{%
0	5.72957795130824\\
0.039999999999992	5.84844658063088\\
0.120000000000005	6.38873336682769\\
0.240000000000009	7.75786394376993\\
0.400000000000006	10.2951905079114\\
0.919999999999987	19.2752651349907\\
1.03999999999999	20.092894094445\\
1.12	20.3234401783132\\
1.16	20.3522048163202\\
1.19999999999999	20.3294387706547\\
1.28	20.1564611985214\\
1.40000000000001	19.6513457295326\\
1.56	18.6725971445642\\
1.80000000000001	16.8260538948337\\
2.16	13.5698153624986\\
2.88	6.80111212404853\\
3.08000000000001	5.63660454593014\\
3.28	4.77357237493374\\
3.52000000000001	3.97392970925418\\
3.91999999999999	2.87677293958203\\
4.44	1.60031959941796\\
4.84	0.74573579577023\\
5.19999999999999	0.0952715754053202\\
5.56	-0.448573323231727\\
5.96000000000001	-0.946424390743118\\
6.31999999999999	-1.31492147008163\\
6.75999999999999	-1.67585590189913\\
7.16	-1.93546724383958\\
7.64000000000001	-2.169275488838\\
8.03999999999999	-2.30818763196285\\
8.47999999999999	-2.4111368367173\\
8.91999999999999	-2.46974568623273\\
9.40000000000001	-2.49086350125302\\
9.96000000000001	-2.47259016370131\\
10.56	-2.41275845765961\\
11.24	-2.30612621956649\\
12.8	-1.95821679492039\\
14.04	-1.63864799286824\\
18.8	-0.472358178635744\\
21.4	0.00569120638755294\\
23.96	0.354866572233988\\
26.52	0.591674460322082\\
29.24	0.737321574717811\\
32.16	0.792024435081203\\
33.8	0.78669912481277\\
37.4	0.712393222913079\\
41.44	0.57614627960146\\
49.64	0.296134678399284\\
56.96	0.140897508886042\\
63.48	0.0641179814693942\\
71.8	0.0188138757813476\\
97.28	-0.00353788951451861\\
103.48	-0.00796660209769584\\
108.8	0.00118057531321369\\
178.4	0.000910105450117271\\
187.04	0.00198056827289861\\
};
\end{axis}
\end{tikzpicture}%
		\label{j1:fig:sim_eight_rev_b2}
	}
	\quad
	\subfloat[][Forward tracking: The controlled curvature $\kappa(t)$ (black line) and the nominal feedforward $\kappa_r(\tilde s(t))$ (red dashed line).]{
		\input{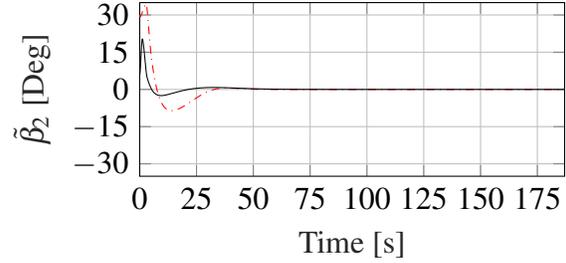}
		\label{j1:fig:sim_kappa_state_eight_fwd}
	}
	~
	\subfloat[][Backward tracking: The controlled curvature $\kappa(t)$ (black line) and the nominal feedforward $\kappa_r(\tilde s(t))$ (red dashed line).]{
		\input{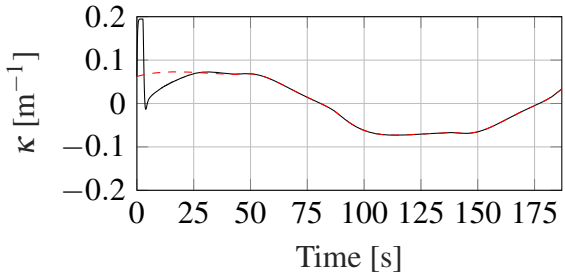}
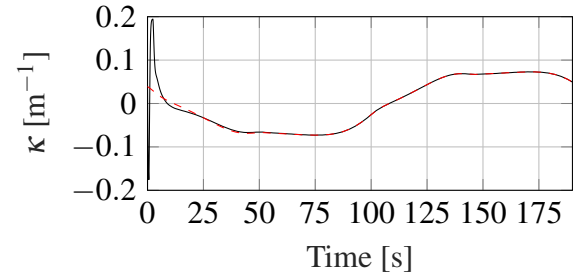
		\label{j1:fig:sim_kappa_state_eight_rev}
	}
	\caption{Simulation results of backward and forward tracking of the figure-eight nominal path in Figure~\ref{j1:fig:eight_rev_sim}. In (a)--(d), the forward (dashed-dotted red line) and backward (black line) path-following error states are plotted, and in~(e)--(f), the controlled curvature of the tractor are displayed.}
	\label{j1:fig:sim_eight_rev_states}
\end{figure}
\tikzexternalenable

\subsubsection{Path following of a figure-eight nominal path}
Nominal paths of the shape of a figure-eight are used to evaluate the performance of the proposed path-following controller in backward and forward motion. 
These nominal paths are used as a benchmark since they expose the closed-loop system for a wide range of practically relevant maneuvers, \textit{e.g.}, enter, exit and keep a narrow turn. 
To generate the figure-eight nominal path in forward motion, a list of waypoints of the same shape was first constructed manually. 
The nominal path was generated by simulating the model of the G2T with a car-like tractor~\eqref{j1:eq:model_global_coord} in forward motion with $v(t)=1$ m/s, together with the pure pursuit controller in~\cite{evestedtLjungqvist2016planning}. 
The path taken by the vehicle $(x_r(\tilde s),u_r(\tilde s))$, $\tilde s\in[0,\tilde s_G]$ was then stored and used as the nominal path in forward motion. 
The established symmetry result in Lemma~\ref{j1:L1} was then used to construct the figure-eight nominal path in backward motion. 

In analogy to the design of the hybrid path-following controller~\eqref{j1:eq:hybrid_controller}, the OCP in~\eqref{j1:eq:opt_LTV} is solved with decay-rate $\epsilon=0.01$.
In both cases, the optimal objective function to~\eqref{j1:eq:opt_LTV} is zero, which implies that the proposed hybrid path-following controller~\eqref{j1:eq:opt_LTV} is able the locally stabilize the path-following error model~\eqref{j1:eq:error_model_mi_cl} around the origin while tracking the figure-eight nominal path in forward and backward motion, respectively.
To confirm the theoretical analysis and to illustrate how the proposed path-following controller handles disturbance rejection, the closed-loop system is simulated with a perturbation in the initial path-following error states. 
For backward tracking, the initial path-following error is chosen as $\tilde x_e(0)=\begin{bmatrix} 1\text{ m} & 0 & 0.1\text{ rad} & 0.1 \text{ rad} \end{bmatrix}^T$ and for forward tracking it is chosen as $\tilde x_e(0)=\begin{bmatrix} -3\text{ m} & 0 & -\pi/6 \text{ rad}& \pi/6 \text{ rad}  \end{bmatrix}^T$. 
To perform realistic simulations, the steering angle of the car-like tractor is constrained according to the values in Table~\ref{j1:tab:vehicle_parameters}. 
The velocity of the car-like tractor is set to $v=v_r(s)$, $i.e.$, 1 m/s for forward tracking and $v=-1$ m/s for backward tracking. 

The simulation results are provided in Figure~\ref{j1:fig:eight_rev_sim}--\ref{j1:fig:sim_eight_rev_states}. In Figure~\ref{j1:fig:eight_rev_sim}, the resulting paths taken by the axle of the semitrailer $(x_3(\cdot),y_3(\cdot))$ is plotted together with its nominal path $(x_{3,r}(\tilde s),y_{3,r}(\tilde s))$, $\tilde s\in[0,\tilde s_G]$. 
The resulting trajectories for the path-following error states are presented in Figure~\ref{j1:fig:sim_eight_rev_z3}--\ref{j1:fig:sim_eight_rev_b2}. As theoretically verified, the path-following error states are converging towards the origin. 
The controlled curvature of the car-like tractor is plotted in Figure~\ref{j1:fig:sim_kappa_state_eight_fwd} and Figure~\ref{j1:fig:sim_kappa_state_eight_rev} for the forward and backward tracking simulation, respectively. 
From these plots, it is clear that the feedback part in the path-following controller $\tilde\kappa(t)$ is responsible for disturbance rejection and the feedforward part $\kappa_r(\tilde s)$ takes care of path-following.

\subsection{Results from real-world experiments}
The path planning and path-following control framework is finally evaluated in three different real-world experiments. First, the performance of the path-following controller and the nonlinear observer are evaluated by path-tracking of a precomputed figure-eight nominal path in backward motion. 
Then, two real-world experiments with the complete path planning and path-following control framework are presented. 
To validate the performance of the path-following controller and the nonlinear observer, a high-precision RTK-GPS\footnote{The RTK-GPS is a Trimble SPS356 with a horizontal accuracy of about 0.1 m.} was mounted above the midpoint of the axle of the semitrailer.  
The authors recommend the supplemental video material\footnote{\url{youtu.be/IBA-8wom5zQ}} for real-world demonstration of the proposed framework.

\subsubsection{Path following of a figure-eight nominal path}
\begin{figure*}[t!]
	\centering
	\setlength\figureheight{0.36\textwidth}
	\setlength\figurewidth{0.74\textwidth}
	\begin{tikzpicture}
	\node[anchor=south west] (myplot) at (0,0) {
		\input{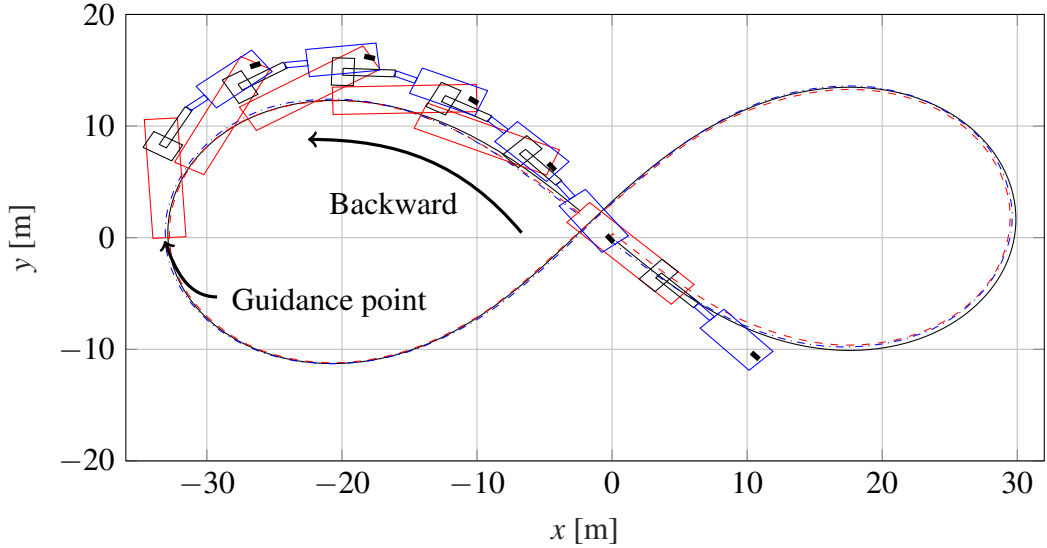}
	};
	\begin{scope}[x={(myplot.south east)}, y={(myplot.north west)}]
	\draw[<-,very thick] (0.3,0.73) to [out=0,in=130] (0.5,0.57);  
	\node at (0.38,0.62) {Backward};
	\draw[->,very thick] (0.215,0.46) to [out=180,in=290] (0.166,0.556);  
	\node at (0.32,0.45) {Guidance point};
	\end{scope}
	\end{tikzpicture}
	\caption{Results from real-world experiments of backward tracking of the figure-eight nominal path, where the nominal path of the axle of the semitrailer $(x_{3,r}(\cdot),y_{3,r}(\cdot))$ is the black solid line. The dashed red line is the estimated path taken by the axle of the semitrailer $(\hat x_3(\cdot),\hat y_3(\cdot))$ and the dashed-dotted blue line is the ground truth path $(x_{3,GT}(\cdot),y_{3,GT}(\cdot))$ measured by the external RTK-GPS.} 
	\label{j1:fig:eight_rev}
\end{figure*}

\begin{figure}[t!]
	\centering
	\setlength\figureheight{0.17\columnwidth}
	\setlength\figurewidth{0.36\columnwidth}
	\subfloat[][The norm of the position estimation error $\lVert e(t)\rVert$ for the axle of the semitrailer.]{
		\input{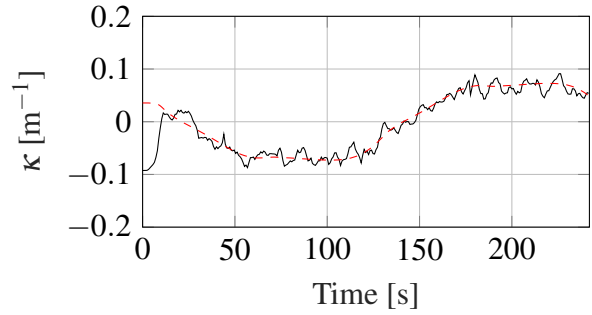}
		\label{j1:fig:eight_est_poes_error} 
	}
	~
	\subfloat[][The controlled curvature $\kappa(t)$ (black line) and the nominal feed-forward $\kappa_r(\tilde s(t))$ (red dashed line).]{
		\input{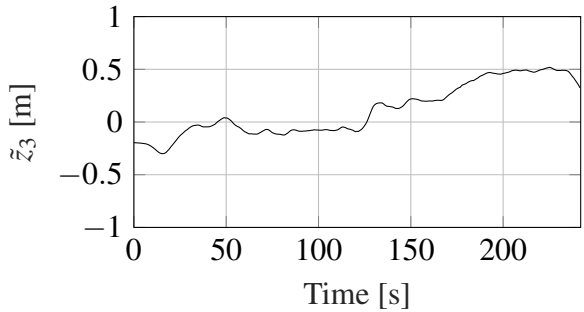}
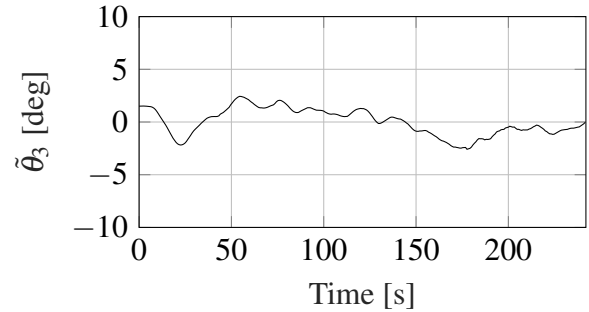
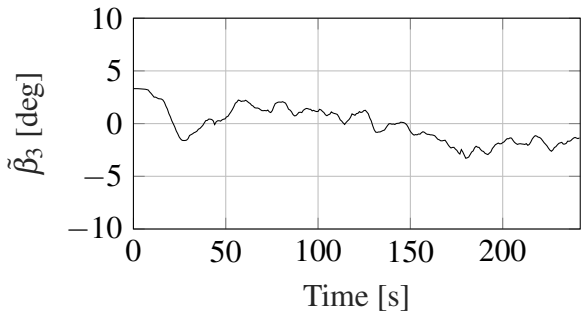
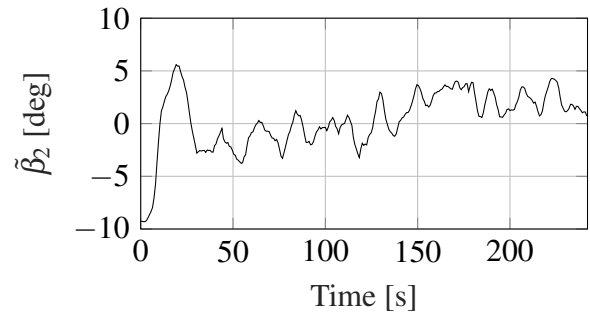
		\label{j1:fig:kappa_state_eight_rev}
	}
	\quad
	\subfloat[][Estimated lateral error for the axle of the semitrailer.]{
%
%
%
\begin{tikzpicture}

\begin{axis}[%
width=\figurewidth,
height=\figureheight,
at={(0\figurewidth,0\figureheight)},
scale only axis,
xmin=0,
xmax=242,
xlabel style={font=\color{white!15!black}},
xlabel={Time [s]},
ymin=-1,
ymax=1,
ylabel style={font=\color{white!15!black}},
ylabel={$\tilde z_3$ [m]},
axis background/.style={fill=white},
xmajorgrids,
ymajorgrids
]
\addplot [color=black, forget plot]
  table[row sep=crcr]{%
0	-0.196134763981206\\
2.40000000000001	-0.198447722974578\\
4	-0.201118930317335\\
6.40000000000001	-0.206734603035358\\
7.19999999999999	-0.21073343140776\\
8	-0.216295983735989\\
8.80000000000001	-0.224051900331176\\
9.59999999999999	-0.232990898726172\\
11.2	-0.253876901832399\\
12.8	-0.278646908696004\\
13.6	-0.289225014416814\\
14.4	-0.296487784010679\\
15.2	-0.300130289086042\\
16	-0.299830146126453\\
16.8	-0.295570008170671\\
17.6	-0.285477548969396\\
18.4	-0.270767405006097\\
19.2	-0.255162986960869\\
20	-0.236499549766052\\
21.6	-0.20248397919093\\
22.4	-0.185385292022346\\
24	-0.149325466401052\\
24.8	-0.132887789747741\\
25.6	-0.119780711753464\\
26.4	-0.108003719714276\\
28	-0.0820927356144239\\
28.8	-0.0731409676976398\\
30.4	-0.0506591138415047\\
31.2	-0.0423351021790381\\
32	-0.0358791214139558\\
33.6	-0.0309405599599017\\
34.4	-0.0291606647009246\\
35.2	-0.0296183238388039\\
36	-0.0346276664367338\\
36.8	-0.0421810663887072\\
37.6	-0.0461496919971012\\
40	-0.0460281197506163\\
40.8	-0.0451189824169944\\
41.6	-0.0404922026160648\\
42.4	-0.0349348067279891\\
44	-0.0203999819879357\\
44.8	-0.0064748538777053\\
45.6	0.00577849175689948\\
46.4	0.0150136160176544\\
47.2	0.025788442291514\\
48.8	0.0410773968661431\\
49.6	0.0380334390877124\\
50.4	0.0389584113244155\\
51.2	0.0311648259412607\\
52	0.020860674075351\\
52.8	0.00967167599924323\\
53.6	-0.00725654887531846\\
54.4	-0.0219948241803536\\
55.2	-0.0329640671192806\\
56	-0.0415059101311783\\
56.8	-0.0470695005095081\\
57.6	-0.0567549897915001\\
58.4	-0.067710729670182\\
59.2	-0.0809642354314235\\
60	-0.0923248308080531\\
60.8	-0.0980012081818984\\
61.6	-0.106417756445808\\
62.4	-0.112801594835446\\
63.2	-0.111331855907707\\
64.8	-0.114493675714471\\
65.6	-0.114333846953315\\
66.4	-0.115113613199128\\
67.2	-0.112554861878124\\
68	-0.104033867929729\\
68.8	-0.0973597232738825\\
69.6	-0.087472826130039\\
70.4	-0.0804682181347118\\
71.2	-0.0722262852534925\\
72	-0.0703336860057391\\
72.8	-0.0727185510838524\\
73.6	-0.0815731494781744\\
74.4	-0.0934560480885409\\
75.2	-0.102691673209023\\
76	-0.109051928286249\\
76.8	-0.112922192719083\\
77.6	-0.11549303481479\\
78.4	-0.115527011025762\\
79.2	-0.117096871847195\\
80.8	-0.123237462659858\\
81.6	-0.122724219058142\\
82.4	-0.115287204489562\\
83.2	-0.106184951889304\\
84.8	-0.0823791845409971\\
85.6	-0.0762453323469003\\
86.4	-0.0736286959900383\\
87.2	-0.0747433361144374\\
89.6	-0.0877530961697346\\
90.4	-0.0874029251249056\\
91.2	-0.0884610014516625\\
92	-0.0854530965967513\\
92.8	-0.086308545950061\\
93.6	-0.0835489411927881\\
94.4	-0.0820438230385889\\
95.2	-0.0814800826222495\\
96	-0.0786453740541333\\
96.8	-0.0775581272272348\\
97.6	-0.0735635239252019\\
98.4	-0.0757185067693342\\
99.2	-0.0754529579306791\\
100.8	-0.0779095195636899\\
101.6	-0.0782621579897977\\
104	-0.0715520269914691\\
104.8	-0.0734364082865397\\
106.4	-0.0799493776076474\\
107.2	-0.0798631295184862\\
108	-0.0816881460002321\\
108.8	-0.08094428410962\\
109.6	-0.0779714278664301\\
110.4	-0.0742346993530134\\
111.2	-0.0651091444331655\\
112	-0.0550906008711536\\
112.8	-0.0477849179748659\\
113.6	-0.0481325029743971\\
114.4	-0.0519499921874171\\
115.2	-0.0588442308952892\\
116	-0.0648905796022916\\
116.8	-0.0674684367290865\\
117.6	-0.0724267303064039\\
118.4	-0.0764435373920662\\
119.2	-0.0844425666544169\\
120	-0.0913080968723818\\
120.8	-0.0883677523142126\\
121.6	-0.0898057556355241\\
122.4	-0.0797739312329497\\
123.2	-0.0718782370830979\\
124	-0.0566335596881231\\
124.8	-0.0385543794186276\\
125.6	-0.0178644747239787\\
126.4	0.00905289521017494\\
127.2	0.0441947106245379\\
128	0.0804992376096152\\
128.8	0.115529766401067\\
129.6	0.146541279164779\\
130.4	0.163216776172249\\
131.2	0.17343291888767\\
132	0.178697824866731\\
132.8	0.181015689338352\\
133.6	0.181505633034647\\
134.4	0.181114785717114\\
135.2	0.172446206313992\\
136	0.16056818820644\\
136.8	0.151252147061967\\
137.6	0.148255742156294\\
138.4	0.147550916301498\\
139.2	0.144694182449513\\
140	0.14273079452019\\
140.8	0.137208325367595\\
141.6	0.129823009785468\\
142.4	0.128172805027305\\
143.2	0.128051560484408\\
144	0.132399386182044\\
144.8	0.140181100270667\\
145.6	0.149192831264088\\
146.4	0.163738030163643\\
148	0.195766825295323\\
148.8	0.208952733202267\\
149.6	0.216618043170655\\
150.4	0.22024750827012\\
151.2	0.220495029212486\\
152.8	0.21703134989292\\
153.6	0.213802673990045\\
154.4	0.209353597276134\\
155.2	0.203228440350529\\
156	0.199316257599349\\
156.8	0.199189232294771\\
157.6	0.197548510411394\\
158.4	0.196951688732184\\
159.2	0.197898057194948\\
160	0.199710894078919\\
160.8	0.199568210937258\\
161.6	0.198470104726709\\
162.4	0.201642185981711\\
163.2	0.203235993288843\\
164	0.206343248004742\\
164.8	0.206364038863313\\
166.4	0.204433852596452\\
167.2	0.207667788424232\\
168	0.216139050536754\\
169.6	0.241798258918919\\
170.4	0.254392589705162\\
171.2	0.265392076534368\\
172	0.274021453338349\\
172.8	0.284589421351285\\
174.4	0.308867397141313\\
175.2	0.317855008711092\\
176	0.324580210523266\\
176.8	0.330077022512086\\
177.6	0.347765184130139\\
178.4	0.354049775813678\\
179.2	0.358888664001682\\
180	0.365379184679114\\
180.8	0.375823623549763\\
182.4	0.389357095078196\\
183.2	0.392990925264741\\
184	0.392986803688729\\
185.6	0.411065069310268\\
186.4	0.418780054743337\\
187.2	0.430160489718702\\
188	0.437989521850284\\
188.8	0.443515628445226\\
191.2	0.464947912091503\\
192	0.468205554515322\\
192.8	0.469748354626347\\
193.6	0.46715751999065\\
194.4	0.463680598626723\\
195.2	0.461412750341083\\
196	0.460070108325311\\
196.8	0.455660302892255\\
197.6	0.455498346499752\\
198.4	0.45354899144823\\
199.2	0.456134901973428\\
200	0.461020439307077\\
200.8	0.463289704430252\\
203.2	0.473934664327743\\
204	0.479680023376176\\
204.8	0.486860811271725\\
206.4	0.491500503027822\\
207.2	0.49173000846119\\
208.8	0.486707927278786\\
209.6	0.485898523754486\\
210.4	0.489125653812408\\
211.2	0.491530718778137\\
212.8	0.493446178196052\\
213.6	0.486323294344487\\
216	0.474223666039961\\
216.8	0.472012047778122\\
217.6	0.475916590433371\\
218.4	0.482120107032785\\
219.2	0.487012751859425\\
220	0.49444245930485\\
220.8	0.496485365066576\\
221.6	0.500024967179911\\
222.4	0.505567197913877\\
224	0.514000782878753\\
224.8	0.517585251218151\\
225.6	0.51764033726576\\
227.2	0.503325169606029\\
228	0.497440688782376\\
228.8	0.490194628019935\\
229.6	0.48856627356497\\
230.4	0.489127903677939\\
231.2	0.48874506168093\\
232	0.491867711924584\\
232.8	0.489586740112145\\
233.6	0.490507024284256\\
234.4	0.487139874645351\\
235.2	0.481548293867007\\
236	0.467907708159515\\
237.6	0.430595959985396\\
238.4	0.414129925294048\\
239.2	0.392827679974374\\
240.8	0.351768552514187\\
241.6	0.324286500884767\\
};
\end{axis}
\end{tikzpicture}%
		\label{j1:fig:eight_rev_z3}
	}
	~
	\subfloat[][Estimated orientation error of the semitrailer.]{
%
%
%
\begin{tikzpicture}

\begin{axis}[%
width=\figurewidth,
height=\figureheight,
at={(0\figurewidth,0\figureheight)},
scale only axis,
xmin=0,
xmax=242,
xlabel style={font=\color{white!15!black}},
xlabel={Time [s]},
ymin=-10,
ymax=10,
ylabel style={font=\color{white!15!black}},
ylabel={$\tilde\theta_3$ [deg]},
axis background/.style={fill=white},
xmajorgrids,
ymajorgrids
]
\addplot [color=black, forget plot]
  table[row sep=crcr]{%
0	1.5029790714365\\
2.40000000000001	1.5150349956518\\
4	1.49933772683522\\
6.40000000000001	1.45283683847043\\
7.19999999999999	1.40212567209085\\
8	1.31711895194985\\
8.80000000000001	1.18758430761886\\
9.59999999999999	1.00489617067149\\
10.4	0.79314248387621\\
11.2	0.566650262708265\\
14.4	-0.288444118013217\\
15.2	-0.528072980750068\\
16.8	-1.04071741793874\\
18.4	-1.53989240191899\\
20	-1.92749667269393\\
20.8	-2.0638624666903\\
21.6	-2.14504353384413\\
22.4	-2.18023864486355\\
23.2	-2.16593921694931\\
24	-2.10694708653133\\
24.8	-2.00827561246038\\
25.6	-1.85081285822173\\
26.4	-1.64318236452934\\
27.2	-1.44520805077605\\
28	-1.23689400455814\\
28.8	-1.01639596807794\\
29.6	-0.832664039487184\\
30.4	-0.68536055130869\\
32.8	-0.277190527581979\\
33.6	-0.125525923597792\\
35.2	0.127435655161889\\
36	0.250102157416279\\
36.8	0.361661264636098\\
37.6	0.426174135574598\\
38.4	0.474837579224499\\
39.2	0.509126462879266\\
40.8	0.520691360526286\\
41.6	0.515403033200556\\
42.4	0.528996977278723\\
43.2	0.585617591311689\\
44	0.749467586280502\\
45.6	0.90301002537899\\
47.2	1.14523596082603\\
48.8	1.40437792696966\\
49.6	1.63171810545089\\
50.4	1.80042292417656\\
51.2	1.98790179849647\\
52	2.14352274953674\\
52.8	2.26490460259799\\
53.6	2.36824002865498\\
54.4	2.43417461947746\\
55.2	2.42552717310141\\
56.8	2.34575696689799\\
58.4	2.19699181823714\\
59.2	2.12614141220294\\
60	2.00203986319158\\
60.8	1.88969072186427\\
61.6	1.79142265584119\\
62.4	1.667627184682\\
64	1.46411390767693\\
64.8	1.37752781301211\\
65.6	1.34735908089155\\
66.4	1.34467887197258\\
67.2	1.32460708043601\\
68	1.32892354654805\\
68.8	1.35694653205215\\
69.6	1.40383887563374\\
71.2	1.53280362605588\\
72.8	1.62969060648445\\
73.6	1.75973538056908\\
74.4	1.92481838476564\\
75.2	2.02823470587563\\
76	2.059456140902\\
76.8	2.04449866694989\\
77.6	1.97143165108662\\
78.4	1.88066057145127\\
79.2	1.76796415769337\\
80	1.63958424221892\\
80.8	1.50013632535959\\
81.6	1.33238837344794\\
82.4	1.18374029849832\\
83.2	1.05550637831738\\
84	0.965278191999346\\
84.8	0.907515388910241\\
85.6	0.893434953936264\\
86.4	0.915534244439272\\
87.2	1.00350942882656\\
88	1.0804638895446\\
90.4	1.28344505212465\\
91.2	1.35361841565737\\
92	1.35726253502295\\
92.8	1.35211225222366\\
93.6	1.29704022350242\\
94.4	1.23239675187651\\
96	1.12306196512898\\
98.4	1.10446207935857\\
99.2	1.07948599855453\\
100	1.03703658630633\\
100.8	0.941192063499813\\
102.4	0.797844535511246\\
103.2	0.758562249964655\\
104	0.738492112614523\\
104.8	0.759095378245888\\
105.6	0.761264123389907\\
106.4	0.741314047958809\\
107.2	0.712122246332996\\
108	0.664694533314332\\
108.8	0.602863338718294\\
109.6	0.561447846355691\\
110.4	0.529775992370077\\
111.2	0.51177177380049\\
112.8	0.534732895438509\\
113.6	0.639319420052374\\
115.2	0.881494822771742\\
116	1.00323120533957\\
116.8	1.0997793443523\\
117.6	1.18408586338165\\
118.4	1.22977289201816\\
119.2	1.26328105303429\\
120	1.28802682839725\\
120.8	1.26995105292391\\
121.6	1.26901713127663\\
122.4	1.21094013948573\\
123.2	1.1439732296032\\
124	1.02675192224186\\
124.8	0.87269076229245\\
126.4	0.504063371713585\\
127.2	0.296607928299096\\
128	0.11666555407848\\
128.8	-0.0262840286795551\\
129.6	-0.137330077968954\\
130.4	-0.138640940681057\\
131.2	-0.0866212728071218\\
132	-0.0125357758537348\\
136	0.435046898151967\\
136.8	0.467121173009673\\
137.6	0.456335290454007\\
140	0.318174517407698\\
140.8	0.290723346803162\\
141.6	0.27787117194103\\
143.2	0.116289106601556\\
144	0.02254364133276\\
144.8	-0.0912228326159266\\
145.6	-0.185930035544118\\
148	-0.64296636754321\\
148.8	-0.764014181665345\\
149.6	-0.843334235713769\\
150.4	-0.883373178154329\\
151.2	-0.884836973070833\\
152	-0.851427156106581\\
152.8	-0.85150927845919\\
153.6	-0.834170042838309\\
154.4	-0.806726236031494\\
155.2	-0.806436860126269\\
156	-0.861347045380683\\
156.8	-0.964607552800203\\
157.6	-1.02917098314683\\
158.4	-1.11169909167103\\
160	-1.304770503539\\
160.8	-1.38150180917029\\
161.6	-1.48627242872664\\
162.4	-1.61018326604977\\
163.2	-1.70357265601476\\
164	-1.77675823969315\\
164.8	-1.81139669029523\\
166.4	-1.93887640818141\\
168	-2.1129807048894\\
168.8	-2.23242598587751\\
169.6	-2.31811796609131\\
170.4	-2.38212601668104\\
171.2	-2.37482902676993\\
172	-2.4102243477094\\
172.8	-2.43586530474045\\
173.6	-2.47166766226579\\
174.4	-2.47981543761537\\
175.2	-2.43409799329271\\
176.8	-2.41340721586263\\
177.6	-2.58751727554272\\
179.2	-2.49875704628769\\
180	-2.37674766959648\\
180.8	-2.17818083693268\\
181.6	-2.01670427933959\\
182.4	-1.87958225533606\\
183.2	-1.69642219500571\\
184	-1.57433341736538\\
184.8	-1.58918075451709\\
185.6	-1.58594922126605\\
186.4	-1.60987334621618\\
187.2	-1.69322491942219\\
188	-1.64156873233793\\
188.8	-1.65492794202424\\
189.6	-1.63511569767613\\
190.4	-1.5827976225726\\
191.2	-1.39972261944177\\
192	-1.25298059415738\\
192.8	-1.13453401489195\\
193.6	-0.945858470085113\\
194.4	-0.876187925869203\\
195.2	-0.842159336454813\\
196	-0.737646129652802\\
196.8	-0.709014083373802\\
197.6	-0.697833223785068\\
198.4	-0.568128678498454\\
200	-0.487059669135704\\
200.8	-0.40182504479202\\
202.4	-0.490461263424919\\
203.2	-0.474387160265252\\
204	-0.59575732729968\\
204.8	-0.69756583255122\\
205.6	-0.691996453613314\\
206.4	-0.755734124761432\\
207.2	-0.795916928061416\\
208	-0.741776313027401\\
208.8	-0.756217360645223\\
210.4	-0.747760892761647\\
211.2	-0.670008527210314\\
212	-0.62938939221624\\
212.8	-0.54532737261485\\
213.6	-0.449501526106843\\
214.4	-0.387005588685923\\
215.2	-0.297185079283963\\
216	-0.318860844697554\\
216.8	-0.387495681439304\\
217.6	-0.492564001193927\\
218.4	-0.583892221494324\\
219.2	-0.69190243179051\\
220	-0.843512514323265\\
220.8	-0.910066571286961\\
221.6	-1.03032902037626\\
222.4	-1.12054461353301\\
223.2	-1.12983580523834\\
224	-1.17302543923788\\
224.8	-1.15566555296692\\
228	-0.830870901386618\\
228.8	-0.771354340506434\\
230.4	-0.726543443713524\\
232	-0.696532286191598\\
232.8	-0.640971525418735\\
233.6	-0.600171283689946\\
234.4	-0.583001255333613\\
235.2	-0.584170178177033\\
236	-0.558610116198963\\
237.6	-0.487768994953171\\
238.4	-0.464370581125706\\
240.8	-0.173196508538382\\
241.6	-0.0375924640624135\\
};
\end{axis}
\end{tikzpicture}%
		\label{j1:fig:eight_rev_theta3}
	}
	\quad
	\subfloat[][Estimated joint angle error between the semitrailer and the dolly.]{
%
%
%
\begin{tikzpicture}

\begin{axis}[%
width=\figurewidth,
height=\figureheight,
at={(0\figurewidth,0\figureheight)},
scale only axis,
xmin=0,
xmax=242,
xlabel style={font=\color{white!15!black}},
xlabel={Time [s]},
ymin=-10,
ymax=10,
ylabel style={font=\color{white!15!black}},
ylabel={$\tilde\beta_3$ [deg]},
axis background/.style={fill=white},
xmajorgrids,
ymajorgrids
]
\addplot [color=black, forget plot]
  table[row sep=crcr]{%
0	3.31661350245855\\
2.40000000000001	3.3141665871884\\
4	3.29310313756173\\
6.40000000000001	3.2591876197946\\
7.19999999999999	3.22868399496133\\
8	3.18849302745085\\
8.80000000000001	3.00419595406711\\
9.59999999999999	2.85771422302312\\
11.2	2.50722925562286\\
12	2.51068148936147\\
13.6	2.36158315553516\\
14.4	2.34886619707424\\
15.2	2.32819014723955\\
16	2.22944012599316\\
16.8	2.01334532037183\\
17.6	1.65721530834423\\
18.4	1.37968209736499\\
19.2	0.947042120523946\\
20	0.651992195727104\\
20.8	0.265214217759194\\
21.6	-0.0193927013123414\\
22.4	-0.319733854931286\\
23.2	-0.695816503710915\\
24	-0.960646363246724\\
24.8	-1.27245879956638\\
25.6	-1.46800604828678\\
26.4	-1.5906466838677\\
27.2	-1.60446372967183\\
28	-1.60301434218562\\
28.8	-1.56685468038933\\
29.6	-1.43006392056054\\
30.4	-1.12752442653454\\
32	-0.973239844694064\\
32.8	-0.867150091493386\\
33.6	-0.808446750011257\\
34.4	-0.717967714323748\\
35.2	-0.530240925481309\\
36	-0.463446637957901\\
36.8	-0.318102480256272\\
37.6	-0.163900447655237\\
38.4	0.070413994298832\\
39.2	0.283472087142997\\
40	0.302591280733736\\
40.8	0.429532187407858\\
41.6	0.419418785000744\\
42.4	0.389792279186878\\
43.2	0.268477287274294\\
44	-0.0922943585626115\\
44.8	0.163035548271154\\
45.6	0.268672721398019\\
46.4	0.272776357748313\\
47.2	0.234488878582141\\
48	0.353953182714321\\
48.8	0.417834154432882\\
49.6	0.443391388285761\\
50.4	0.631082958965294\\
51.2	0.689370459320855\\
52	0.908531166105888\\
52.8	1.1094320024751\\
53.6	1.33920959799468\\
54.4	1.61113490350021\\
55.2	1.93170236167538\\
56	2.03152088064004\\
56.8	2.23064714848556\\
57.6	2.19035586257164\\
58.4	2.09855363781503\\
59.2	2.14094584722895\\
60	2.13674954975761\\
60.8	2.21407881710638\\
61.6	2.11657445160986\\
62.4	1.99401289681293\\
63.2	1.90834588947828\\
64	1.74463980872972\\
64.8	1.66863623841016\\
65.6	1.51475122263611\\
66.4	1.47538777523229\\
67.2	1.48070689177013\\
68	1.43119911934494\\
68.8	1.32033383661482\\
69.6	1.23249280141169\\
70.4	1.2251549370873\\
71.2	1.24362739714098\\
72	1.2128869568426\\
72.8	1.28386895074001\\
73.6	1.10920372474644\\
74.4	1.04925675522682\\
75.2	1.22848529463789\\
76	1.59579429862438\\
76.8	1.89397026226609\\
77.6	1.96920842451181\\
78.4	2.03427835607403\\
79.2	2.04326688502178\\
80	2.02787889570857\\
80.8	2.07616854658394\\
81.6	2.01233189105372\\
82.4	1.93903371597574\\
83.2	1.67841098616438\\
84	1.36953777234365\\
84.8	1.28521161652043\\
85.6	1.12964146501213\\
86.4	0.868312771117019\\
87.2	0.734399068393543\\
88	0.754260458087145\\
88.8	0.892324514587614\\
89.6	1.06171470010781\\
90.4	1.07889913589446\\
91.2	1.04626342482308\\
92	1.2592755739729\\
92.8	1.36291515689138\\
93.6	1.43702825891472\\
94.4	1.41961586556454\\
95.2	1.36502675256199\\
96	1.35166679582372\\
96.8	1.20649699470891\\
97.6	1.25826321187185\\
98.4	1.18107310962742\\
99.2	1.14241541286393\\
100	1.1511737430923\\
100.8	1.35409108609227\\
101.6	1.30086608993298\\
102.4	1.15802292872593\\
103.2	1.04144109782564\\
104	0.833529499160193\\
104.8	0.750627651222345\\
105.6	0.840303339269553\\
106.4	0.897098389930619\\
107.2	1.11087818831015\\
108	1.01165699073272\\
108.8	0.959739335469294\\
109.6	0.92656830832172\\
110.4	0.871513515155982\\
111.2	0.657259552237662\\
112	0.356411623928722\\
112.8	0.239528971220238\\
113.6	0.0623570991517965\\
114.4	-0.0788482307170568\\
115.2	0.108699785927996\\
116	0.256655830646594\\
116.8	0.387930652431635\\
117.6	0.601161821202169\\
118.4	0.922363793279033\\
120	0.784209610298035\\
120.8	0.950928071274859\\
121.6	0.973805256291286\\
122.4	1.13856425338909\\
123.2	1.05435901325853\\
124	1.09734000632324\\
124.8	1.23310026136102\\
125.6	1.26261329532122\\
126.4	1.1059298229782\\
127.2	0.934542447455982\\
128	0.537553069874633\\
128.8	0.261758061997881\\
129.6	-0.199432038872772\\
130.4	-0.578414890027659\\
131.2	-0.807859407103479\\
132	-0.834238184647461\\
132.8	-0.793451496575301\\
133.6	-0.792613304298101\\
135.2	-0.657951429056254\\
136	-0.608873211496814\\
136.8	-0.393997914807443\\
137.6	-0.190949210727894\\
138.4	-0.0155222135360873\\
139.2	-0.00992662941681033\\
140	-0.0581774369263712\\
140.8	-0.0371846937648286\\
141.6	-0.0806969708262955\\
142.4	0.00873673492444027\\
144	0.0428067932253953\\
144.8	0.133836929796388\\
146.4	0.0931077513834282\\
147.2	0.0171680206542533\\
148	-0.109020124468117\\
148.8	-0.34135599268771\\
149.6	-0.634627080092599\\
150.4	-0.833784937562996\\
151.2	-0.993552400398841\\
152	-1.09333644825256\\
152.8	-1.0421107765294\\
153.6	-1.06740424768643\\
154.4	-0.985507611495336\\
155.2	-1.00508560169925\\
156	-0.863725818841687\\
156.8	-0.771954543379735\\
157.6	-0.824955904204558\\
158.4	-0.947413512060052\\
159.2	-1.01865270526841\\
160	-1.02151049910583\\
160.8	-1.06663659845674\\
161.6	-1.06950730562056\\
162.4	-1.10002149590272\\
163.2	-1.15373228926669\\
164	-1.3121967334358\\
164.8	-1.47846253429753\\
165.6	-1.53303617265277\\
166.4	-1.62808529858609\\
167.2	-1.63584268593593\\
168	-1.60253228024061\\
168.8	-1.57707392404814\\
169.6	-1.77243259551443\\
170.4	-1.99720088760802\\
171.2	-2.18744919019298\\
172	-2.31153601160895\\
172.8	-2.25749747588651\\
173.6	-2.25536474931818\\
174.4	-2.42047390033267\\
175.2	-2.54102439130585\\
176	-2.76423625404234\\
176.8	-2.8877896803728\\
177.6	-2.46909289439211\\
178.4	-2.69380462398163\\
180	-3.29099623764617\\
180.8	-3.24647685119263\\
181.6	-3.17346020797447\\
183.2	-2.69302342109532\\
184	-2.59240000702565\\
184.8	-2.2902239004078\\
185.6	-2.15096616162819\\
186.4	-2.23953429181279\\
187.2	-2.25696322702737\\
188	-2.45605598425726\\
188.8	-2.59520059802875\\
189.6	-2.64064910365283\\
190.4	-2.81315748397256\\
191.2	-2.88251890876768\\
192	-2.93569235834306\\
192.8	-2.80159440059455\\
193.6	-2.67812016176975\\
194.4	-2.36322368159642\\
195.2	-2.19467446311779\\
196	-2.00653156634451\\
197.6	-1.84611112675182\\
198.4	-1.91790986186359\\
199.2	-1.90931385534972\\
200	-1.80639443050185\\
200.8	-1.72576906327001\\
201.6	-1.55900401826864\\
202.4	-1.42893288392267\\
203.2	-1.43958104211703\\
204	-1.35302312198249\\
204.8	-1.38044302312721\\
205.6	-1.59491416471681\\
206.4	-1.77873803405711\\
207.2	-1.8897007537046\\
208	-1.97418592486943\\
208.8	-1.92045645520693\\
209.6	-1.90871045082523\\
210.4	-1.93558852586969\\
211.2	-1.95093361305717\\
212	-1.97674297693996\\
212.8	-1.90017484879684\\
213.6	-2.0625702802767\\
214.4	-1.97560489666949\\
215.2	-1.75100650553151\\
216	-1.46644983237309\\
216.8	-1.27591210052205\\
217.6	-1.14525730516718\\
218.4	-1.18508008279269\\
219.2	-1.28493200565129\\
220	-1.31380769758459\\
220.8	-1.42493807703187\\
221.6	-1.6094433644636\\
222.4	-1.80592784320416\\
223.2	-1.95393910973542\\
224	-2.11413265898159\\
224.8	-2.33599748034771\\
225.6	-2.5220563156505\\
226.4	-2.6255300530382\\
227.2	-2.49530075712426\\
228	-2.27036939426424\\
228.8	-2.19089800883702\\
229.6	-2.01811321929932\\
230.4	-1.87752466084859\\
231.2	-1.82282743875808\\
232	-1.77910280884748\\
232.8	-1.94418431894016\\
233.6	-1.9423832834334\\
234.4	-1.87530079394352\\
235.2	-1.66501079534305\\
236	-1.67803191066545\\
236.8	-1.63060873331929\\
237.6	-1.49050458469111\\
238.4	-1.37169440457407\\
239.2	-1.31293716863053\\
240	-1.37538174623054\\
240.8	-1.42312999789252\\
241.6	-1.35283465145008\\
};
\end{axis}
\end{tikzpicture}%
		\label{j1:fig:eight_rev_b3}
	}
	~
	\subfloat[][Estimated joint angle error between the dolly and the tractor.]{
%
%
%
\begin{tikzpicture}

\begin{axis}[%
width=\figurewidth,
height=\figureheight,
at={(0\figurewidth,0\figureheight)},
scale only axis,
xmin=0,
xmax=242,
xlabel style={font=\color{white!15!black}},
xlabel={Time [s]},
ymin=-10,
ymax=10,
ylabel style={font=\color{white!15!black}},
ylabel={$\tilde\beta_2$ [deg]},
axis background/.style={fill=white},
xmajorgrids,
ymajorgrids
]
\addplot [color=black, forget plot]
  table[row sep=crcr]{%
0	-9.2615853966019\\
1.59999999999999	-9.29681807366177\\
2.40000000000001	-9.30534641861416\\
3.19999999999999	-9.17173425807039\\
4	-8.91073309309033\\
5.59999999999999	-8.34205666543696\\
6.40000000000001	-7.95470895364272\\
7.19999999999999	-7.05794054722637\\
8	-5.79232100817498\\
8.80000000000001	-3.74315668169172\\
9.59999999999999	-1.78879452711462\\
10.4	-0.0502769991350931\\
11.2	1.19260257566813\\
12	1.59192691745028\\
12.8	2.11437705168782\\
13.6	2.56381096959407\\
14.4	2.78640598087114\\
15.2	3.12295095906225\\
16	3.5841678070604\\
16.8	4.18311606000211\\
17.6	4.88586995483649\\
18.4	5.20928111958389\\
19.2	5.59294223677361\\
20	5.46114946786997\\
20.8	5.45906199130491\\
21.6	4.9259123657325\\
22.4	4.46583253078774\\
23.2	4.12420909350533\\
24	3.45015202502535\\
24.8	2.96196553475914\\
25.6	2.03558377597915\\
26.4	0.998524351227786\\
27.2	0.0150561948649965\\
28	-0.748366058194279\\
29.6	-2.06725322872637\\
30.4	-2.80008919765621\\
31.2	-2.66923814108441\\
32	-2.55609114828468\\
32.8	-2.60154776475841\\
33.6	-2.55919482893219\\
34.4	-2.5391443817368\\
35.2	-2.7324745937413\\
36	-2.5067165645483\\
36.8	-2.55550460578968\\
37.6	-2.5348341054837\\
38.4	-2.69947410787958\\
39.2	-2.70364201737715\\
40	-2.09552871778635\\
40.8	-1.92334669244292\\
41.6	-1.44187171680991\\
42.4	-1.14674436541554\\
43.2	-0.830567011329407\\
44	-0.46544815018899\\
44.8	-1.41932622285506\\
45.6	-1.73172786971233\\
46.4	-1.86115120330419\\
47.2	-1.79334236588477\\
48	-2.16381993731591\\
48.8	-2.2128995422575\\
49.6	-2.50642009164963\\
50.4	-2.91500158091296\\
51.2	-2.99965686647681\\
52	-3.36534647529976\\
52.8	-3.47070534489839\\
54.4	-3.77076302565644\\
55.2	-3.70760768149592\\
56	-3.15456769128681\\
56.8	-2.96675733934873\\
57.6	-2.07414254172141\\
58.4	-1.35276720333204\\
59.2	-1.15795462683423\\
60	-0.849470040105103\\
60.8	-0.786518641124331\\
61.6	-0.390048731445859\\
62.4	-0.102171745917389\\
63.2	0.0513867914833952\\
64	0.29330032915226\\
64.8	0.103205252256373\\
65.6	0.118272769627339\\
66.4	-0.21359058079517\\
67.2	-0.60953259155599\\
68	-0.72945267741315\\
68.8	-0.72112123400791\\
69.6	-0.761286641669841\\
70.4	-1.12138163187146\\
71.2	-1.37153346824178\\
72	-1.39877557102466\\
72.8	-1.62516774059344\\
73.6	-1.50233404544673\\
74.4	-1.82692000273138\\
75.2	-2.44027782038771\\
76	-3.09860282877503\\
76.8	-3.29460556573571\\
78.4	-2.21470249875497\\
79.2	-1.61897489345034\\
80	-1.06002412156653\\
80.8	-0.748062328219618\\
81.6	-0.222523494393982\\
82.4	0.141298644660367\\
83.2	0.80513026628239\\
84	1.22325475711585\\
84.8	0.900836851284168\\
85.6	0.736564131638744\\
86.4	0.762729198966895\\
87.2	0.297672957561957\\
88	-0.36230711771907\\
88.8	-1.11660028600903\\
89.6	-1.78672700750849\\
90.4	-1.78325492353261\\
91.2	-1.69123423944427\\
92	-2.01580141370869\\
92.8	-1.95906324571681\\
93.6	-1.69932171703007\\
94.4	-1.1956351796596\\
95.2	-0.783220726972274\\
96	-0.613817254599695\\
96.8	-0.303189455894653\\
97.6	-0.507271159385112\\
98.4	-0.381573613944227\\
100	-0.365917859871189\\
100.8	-0.6679259598437\\
101.6	-0.336870025826471\\
102.4	0.170434057204858\\
103.2	0.3272657393168\\
104	0.567088995640745\\
104.8	0.31214836956903\\
105.6	-0.264377119379475\\
106.4	-0.492708022135901\\
107.2	-0.963913434825798\\
108	-0.376155379622105\\
108.8	-0.122960640615702\\
109.6	-0.0459439844885594\\
110.4	0.0115639413204747\\
111.2	0.411904296253283\\
112	0.798500980723503\\
112.8	0.573769987866058\\
113.6	0.204445384383348\\
114.4	-0.191563674289142\\
115.2	-1.35127811537126\\
116	-1.99860790657706\\
116.8	-2.41337263802995\\
117.6	-2.87244459318532\\
118.4	-3.2179678550404\\
119.2	-2.40718448664194\\
120	-1.86251254538314\\
120.8	-2.06570811187009\\
121.6	-1.94097220903956\\
122.4	-2.01973112035896\\
123.2	-1.32887829654481\\
124	-1.03527663865628\\
124.8	-0.884413981589972\\
125.6	-0.402551766247541\\
126.4	0.478708642297448\\
127.2	1.14720572334963\\
128	2.09546532509313\\
128.8	2.35085826443435\\
129.6	2.98432565311109\\
130.4	2.84368736234546\\
131.2	2.22709118124891\\
132	1.22337577953019\\
132.8	0.393524684083587\\
133.6	-0.0680171715293056\\
134.4	-0.55298465500465\\
135.2	-0.771301353179524\\
136	-0.907623071111374\\
136.8	-1.14023989089156\\
137.6	-1.20069944476404\\
138.4	-0.980695956668484\\
139.2	-0.391946803663615\\
140	0.175137309459075\\
140.8	0.311814650795043\\
141.6	0.543093633951997\\
142.4	0.621328488335166\\
143.2	0.869900896820752\\
144	1.04178943747331\\
144.8	1.04816435184645\\
145.6	1.27131869158353\\
146.4	1.66114095956971\\
147.2	2.12491155126551\\
148	2.63949586890016\\
148.8	3.27644772127064\\
149.6	3.68918068488887\\
150.4	3.61304328555028\\
151.2	3.33474018418553\\
152	2.94964088528869\\
152.8	2.38680832480048\\
153.6	2.11627178885229\\
154.4	1.70314150032343\\
155.2	1.75096957982132\\
156	1.56927512867645\\
156.8	1.75312546499981\\
157.6	2.16351065781257\\
158.4	2.6539282720598\\
159.2	2.89611738851835\\
160	2.90202080639537\\
160.8	3.04349515500408\\
161.6	3.05798945675306\\
162.4	3.20705769505835\\
163.2	3.26617874529694\\
164	3.49153955630104\\
164.8	3.67299783231101\\
165.6	3.52670243736532\\
166.4	3.52868959552697\\
168	3.29791970158703\\
168.8	3.26145951146728\\
169.6	3.68453572936644\\
170.4	3.99796585330233\\
171.2	4.03554939609717\\
172	3.92228095399625\\
172.8	3.40400000991676\\
173.6	3.21569703058478\\
174.4	3.42332368229577\\
175.2	3.50719008101046\\
176	3.7825647387686\\
176.8	3.78277860906698\\
177.6	3.02869392162924\\
178.4	3.56178826535802\\
179.2	3.92132558128608\\
180	3.88766555528264\\
180.8	2.92580886117062\\
181.6	2.11441760593604\\
182.4	1.22352860501289\\
183.2	0.692546710728323\\
184	0.657190663376582\\
184.8	0.586552953702011\\
185.6	1.04897821157473\\
186.4	1.79652950290622\\
187.2	2.38178493410936\\
188	3.00991602405358\\
188.8	3.30168398959336\\
189.6	3.1436024629046\\
190.4	3.2705268010339\\
191.2	2.91522420239124\\
192	2.5075639348922\\
192.8	1.78235301839706\\
193.6	1.31706399027138\\
194.4	0.645569576751484\\
195.2	0.631496185410072\\
196	0.59727725783344\\
196.8	0.810122174070301\\
197.6	0.985925426973466\\
198.4	1.33574103583507\\
199.2	1.39480285552693\\
200	1.24045722887075\\
200.8	1.30454564090493\\
201.6	1.27545106275309\\
202.4	1.33265901969381\\
203.2	1.7970663130863\\
204	2.09166475170684\\
204.8	2.50032628224952\\
205.6	3.17875467179832\\
206.4	3.48109491639963\\
207.2	3.47234437685577\\
208	3.3296559621221\\
208.8	2.86748544896108\\
209.6	2.58485648997419\\
210.4	2.49486490567784\\
211.2	2.38453131048593\\
212	2.20957494767899\\
212.8	1.94027873139538\\
213.6	2.04415929192552\\
214.4	1.63145076757789\\
215.2	1.12142869614371\\
216	0.727970473002387\\
216.8	0.848568401773804\\
217.6	1.21456771698695\\
218.4	2.04483484032312\\
219.2	2.65992325716039\\
220.8	3.52956321941051\\
221.6	4.01318735675454\\
222.4	4.27493642744869\\
223.2	4.27208823856481\\
224	4.20861315617375\\
224.8	4.10120032062568\\
225.6	3.92485380087922\\
226.4	3.3852867856948\\
227.2	2.50550676729156\\
228	1.75288715543078\\
229.6	1.16632651776595\\
230.4	1.11971346488838\\
231.2	1.23237967640358\\
232	1.43359865017774\\
232.8	1.82729886478691\\
233.6	1.82494821249537\\
234.4	1.66716678724325\\
235.2	1.35659242943237\\
236	1.62966104457195\\
236.8	1.64315440344242\\
237.6	1.42015352951827\\
238.4	1.2559750984999\\
239.2	1.03031194490333\\
240	1.11025109156699\\
240.8	1.10342730569931\\
241.6	0.706705746238612\\
};
\end{axis}
\end{tikzpicture}%
		\label{j1:fig:eight_rev_b2}
	}
	\caption{Results from real-world experiments of backward tracking of the figure-eight nominal path in Figure~\ref{j1:fig:eight_rev}. In (a), the norm of the position estimation error is plotted and in (b), the controlled curvature of the tractor is plotted. In (c)--(f), the estimated path-following error states are plotted.}
	\label{j1:fig:eight_rev_states}
\end{figure}

\begin{figure}[t!]
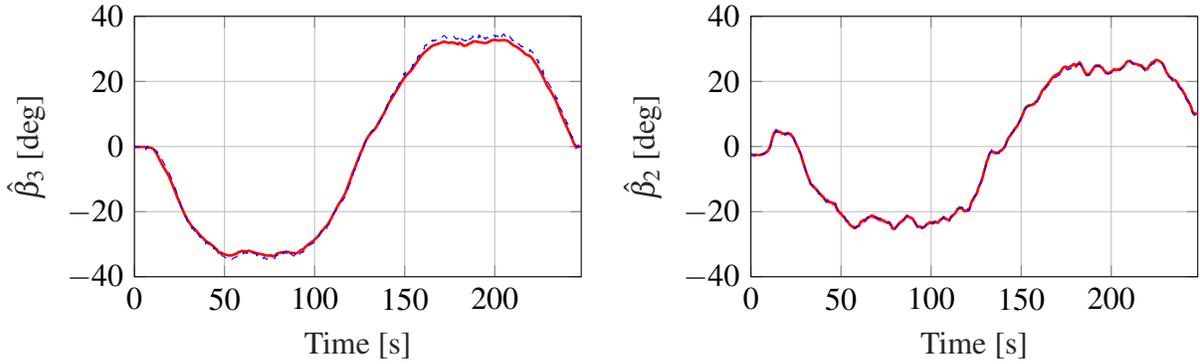
 	
	\centering
	\setlength\figureheight{0.21\columnwidth}
	\setlength\figurewidth{0.36\columnwidth} 
	\subfloat[][Estimated joint angle between the semitrailer and the dolly.]{
		\input{beta3_eight.tex}
		\label{j1:fig:eight_est_beta3}
	}
	~
	\subfloat[][Estimated joint angle between the dolly and the tractor.]{
		\input{beta2_eight.tex}
		\label{j1:fig:eight_est_beta2}
	}
	\caption{Results from real-world experiments of backward tracking of the figure-eight nominal path in Figure~\ref{j1:fig:eight_rev}. The red solid lines are the estimated joint angles, $\hat\beta_2$ and $\hat\beta_3$, and the dashed blue lines are the computed joint angles, $\beta_2$ and $\beta_3$, from the RANSAC measurements.}
	\label{j1:fig:estimation_eight_rev}
\end{figure}

The figure-eight nominal path in backward motion that was used in the simulations, is also used here as the nominal path to evaluate the joint performance of the path-following controller and the nonlinear observer. The real-world path-following experiment is performed on an open gravel surface at Scania's test facility in S\"odert\"alje, Sweden.
During the experiment, the longitudinal velocity of the rear axle of the car-like tractor was set to \mbox{$v=-0.8$ m/s} and results from one lap around the figure-eight nominal path are provided in Figure~\ref{j1:fig:eight_rev}--\ref{j1:fig:estimation_eight_rev}.

\tikzexternaldisable
\tikzexternalenable 

\tikzexternaldisable
\tikzexternalenable

\tikzexternaldisable
\tikzexternalenable

Figure~\ref{j1:fig:eight_rev} shows the nominal path for the position of the axle of the semitrailer $(x_{3,r}(\cdot),y_{3,r}(\cdot))$, compared to its ground truth path $(x_{3,GT}(\cdot),y_{3,GT}(\cdot))$ and its estimated path $(\hat x_3(\cdot),\hat y_3(\cdot))$ around one lap of the figure-eight nominal path. 
A more detailed plot is provided in Figure~\ref{j1:fig:eight_rev_states}, where all four estimated error states $\tilde x_e(t)$ are plotted. 
From these plots, we conclude that the path-following controller is able to keep its estimated lateral path-following error $\tilde z_3(\cdot)$ within \mbox{$\pm0.5$ m} (avg. 0.21 m), while at the same time keep the orientation and joint angle errors within acceptable error tolerances. As in the simulation trails, it can be seen from Figure~\ref{j1:fig:kappa_state_eight_rev} that the feedforward part $\kappa_r(s)$ of the path-following controller takes care of path-following and the feedback part $\tilde\kappa=K_{\text{rev}}\tilde x_e$ is responsible for disturbance rejection.

\tikzexternaldisable
\begin{figure}[b!] 	
	\centering
	\setlength\figureheight{0.21\linewidth}
	\setlength\figurewidth{0.36\linewidth} 
	\subfloat[][The norm of the position estimation error of the axle of the semitrailer.]{
		\input{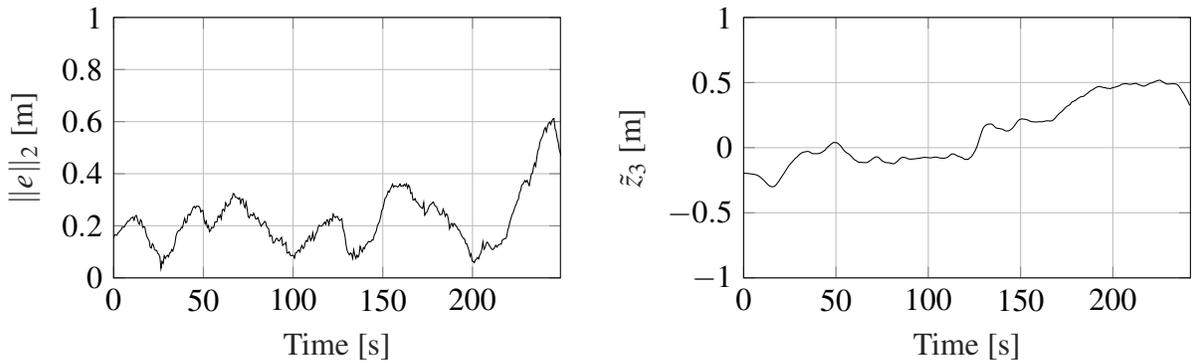}
		\label{j1:fig:eight_many_laps_poes_error}
	} 
	~
	\setlength\figureheight{0.21\linewidth}
	\setlength\figurewidth{0.36\linewidth}
	\subfloat[][Estimated lateral path-following error for the axle of the semitrailer.]{
%
%
%
\begin{tikzpicture}

\begin{axis}[%
width=\figurewidth,
height=\figureheight,
at={(0\figurewidth,0\figureheight)},
scale only axis,
xmin=0,
xmax=242,
xlabel style={font=\color{white!15!black}},
xlabel={Time [s]},
ymin=-1,
ymax=1,
ylabel style={font=\color{white!15!black}},
ylabel={$\tilde z_3$ [m]},
axis background/.style={fill=white},
xmajorgrids,
ymajorgrids
]
\addplot [color=black, forget plot]
  table[row sep=crcr]{%
0	-0.196134763981206\\
2.40000000000001	-0.198447722974578\\
4	-0.201118930317335\\
6.40000000000001	-0.206734603035358\\
7.19999999999999	-0.21073343140776\\
8	-0.216295983735989\\
8.80000000000001	-0.224051900331176\\
9.59999999999999	-0.232990898726172\\
11.2	-0.253876901832399\\
12.8	-0.278646908696004\\
13.6	-0.289225014416814\\
14.4	-0.296487784010679\\
15.2	-0.300130289086042\\
16	-0.299830146126453\\
16.8	-0.295570008170671\\
17.6	-0.285477548969396\\
18.4	-0.270767405006097\\
19.2	-0.255162986960869\\
20	-0.236499549766052\\
21.6	-0.20248397919093\\
22.4	-0.185385292022346\\
24	-0.149325466401052\\
24.8	-0.132887789747741\\
25.6	-0.119780711753464\\
26.4	-0.108003719714276\\
28	-0.0820927356144239\\
28.8	-0.0731409676976398\\
30.4	-0.0506591138415047\\
31.2	-0.0423351021790381\\
32	-0.0358791214139558\\
33.6	-0.0309405599599017\\
34.4	-0.0291606647009246\\
35.2	-0.0296183238388039\\
36	-0.0346276664367338\\
36.8	-0.0421810663887072\\
37.6	-0.0461496919971012\\
40	-0.0460281197506163\\
40.8	-0.0451189824169944\\
41.6	-0.0404922026160648\\
42.4	-0.0349348067279891\\
44	-0.0203999819879357\\
44.8	-0.0064748538777053\\
45.6	0.00577849175689948\\
46.4	0.0150136160176544\\
47.2	0.025788442291514\\
48.8	0.0410773968661431\\
49.6	0.0380334390877124\\
50.4	0.0389584113244155\\
51.2	0.0311648259412607\\
52	0.020860674075351\\
52.8	0.00967167599924323\\
53.6	-0.00725654887531846\\
54.4	-0.0219948241803536\\
55.2	-0.0329640671192806\\
56	-0.0415059101311783\\
56.8	-0.0470695005095081\\
57.6	-0.0567549897915001\\
58.4	-0.067710729670182\\
59.2	-0.0809642354314235\\
60	-0.0923248308080531\\
60.8	-0.0980012081818984\\
61.6	-0.106417756445808\\
62.4	-0.112801594835446\\
63.2	-0.111331855907707\\
64.8	-0.114493675714471\\
65.6	-0.114333846953315\\
66.4	-0.115113613199128\\
67.2	-0.112554861878124\\
68	-0.104033867929729\\
68.8	-0.0973597232738825\\
69.6	-0.087472826130039\\
70.4	-0.0804682181347118\\
71.2	-0.0722262852534925\\
72	-0.0703336860057391\\
72.8	-0.0727185510838524\\
73.6	-0.0815731494781744\\
74.4	-0.0934560480885409\\
75.2	-0.102691673209023\\
76	-0.109051928286249\\
76.8	-0.112922192719083\\
77.6	-0.11549303481479\\
78.4	-0.115527011025762\\
79.2	-0.117096871847195\\
80.8	-0.123237462659858\\
81.6	-0.122724219058142\\
82.4	-0.115287204489562\\
83.2	-0.106184951889304\\
84.8	-0.0823791845409971\\
85.6	-0.0762453323469003\\
86.4	-0.0736286959900383\\
87.2	-0.0747433361144374\\
89.6	-0.0877530961697346\\
90.4	-0.0874029251249056\\
91.2	-0.0884610014516625\\
92	-0.0854530965967513\\
92.8	-0.086308545950061\\
93.6	-0.0835489411927881\\
94.4	-0.0820438230385889\\
95.2	-0.0814800826222495\\
96	-0.0786453740541333\\
96.8	-0.0775581272272348\\
97.6	-0.0735635239252019\\
98.4	-0.0757185067693342\\
99.2	-0.0754529579306791\\
100.8	-0.0779095195636899\\
101.6	-0.0782621579897977\\
104	-0.0715520269914691\\
104.8	-0.0734364082865397\\
106.4	-0.0799493776076474\\
107.2	-0.0798631295184862\\
108	-0.0816881460002321\\
108.8	-0.08094428410962\\
109.6	-0.0779714278664301\\
110.4	-0.0742346993530134\\
111.2	-0.0651091444331655\\
112	-0.0550906008711536\\
112.8	-0.0477849179748659\\
113.6	-0.0481325029743971\\
114.4	-0.0519499921874171\\
115.2	-0.0588442308952892\\
116	-0.0648905796022916\\
116.8	-0.0674684367290865\\
117.6	-0.0724267303064039\\
118.4	-0.0764435373920662\\
119.2	-0.0844425666544169\\
120	-0.0913080968723818\\
120.8	-0.0883677523142126\\
121.6	-0.0898057556355241\\
122.4	-0.0797739312329497\\
123.2	-0.0718782370830979\\
124	-0.0566335596881231\\
124.8	-0.0385543794186276\\
125.6	-0.0178644747239787\\
126.4	0.00905289521017494\\
127.2	0.0441947106245379\\
128	0.0804992376096152\\
128.8	0.115529766401067\\
129.6	0.146541279164779\\
130.4	0.163216776172249\\
131.2	0.17343291888767\\
132	0.178697824866731\\
132.8	0.181015689338352\\
133.6	0.181505633034647\\
134.4	0.181114785717114\\
135.2	0.172446206313992\\
136	0.16056818820644\\
136.8	0.151252147061967\\
137.6	0.148255742156294\\
138.4	0.147550916301498\\
139.2	0.144694182449513\\
140	0.14273079452019\\
140.8	0.137208325367595\\
141.6	0.129823009785468\\
142.4	0.128172805027305\\
143.2	0.128051560484408\\
144	0.132399386182044\\
144.8	0.140181100270667\\
145.6	0.149192831264088\\
146.4	0.163738030163643\\
148	0.195766825295323\\
148.8	0.208952733202267\\
149.6	0.216618043170655\\
150.4	0.22024750827012\\
151.2	0.220495029212486\\
152.8	0.21703134989292\\
153.6	0.213802673990045\\
154.4	0.209353597276134\\
155.2	0.203228440350529\\
156	0.199316257599349\\
156.8	0.199189232294771\\
157.6	0.197548510411394\\
158.4	0.196951688732184\\
159.2	0.197898057194948\\
160	0.199710894078919\\
160.8	0.199568210937258\\
161.6	0.198470104726709\\
162.4	0.201642185981711\\
163.2	0.203235993288843\\
164	0.206343248004742\\
164.8	0.206364038863313\\
166.4	0.204433852596452\\
167.2	0.207667788424232\\
168	0.216139050536754\\
169.6	0.241798258918919\\
170.4	0.254392589705162\\
171.2	0.265392076534368\\
172	0.274021453338349\\
172.8	0.284589421351285\\
174.4	0.308867397141313\\
175.2	0.317855008711092\\
176	0.324580210523266\\
176.8	0.330077022512086\\
177.6	0.347765184130139\\
178.4	0.354049775813678\\
179.2	0.358888664001682\\
180	0.365379184679114\\
180.8	0.375823623549763\\
182.4	0.389357095078196\\
183.2	0.392990925264741\\
184	0.392986803688729\\
185.6	0.411065069310268\\
186.4	0.418780054743337\\
187.2	0.430160489718702\\
188	0.437989521850284\\
188.8	0.443515628445226\\
191.2	0.464947912091503\\
192	0.468205554515322\\
192.8	0.469748354626347\\
193.6	0.46715751999065\\
194.4	0.463680598626723\\
195.2	0.461412750341083\\
196	0.460070108325311\\
196.8	0.455660302892255\\
197.6	0.455498346499752\\
198.4	0.45354899144823\\
199.2	0.456134901973428\\
200	0.461020439307077\\
200.8	0.463289704430252\\
203.2	0.473934664327743\\
204	0.479680023376176\\
204.8	0.486860811271725\\
206.4	0.491500503027822\\
207.2	0.49173000846119\\
208.8	0.486707927278786\\
209.6	0.485898523754486\\
210.4	0.489125653812408\\
211.2	0.491530718778137\\
212.8	0.493446178196052\\
213.6	0.486323294344487\\
216	0.474223666039961\\
216.8	0.472012047778122\\
217.6	0.475916590433371\\
218.4	0.482120107032785\\
219.2	0.487012751859425\\
220	0.49444245930485\\
220.8	0.496485365066576\\
221.6	0.500024967179911\\
222.4	0.505567197913877\\
224	0.514000782878753\\
224.8	0.517585251218151\\
225.6	0.51764033726576\\
227.2	0.503325169606029\\
228	0.497440688782376\\
228.8	0.490194628019935\\
229.6	0.48856627356497\\
230.4	0.489127903677939\\
231.2	0.48874506168093\\
232	0.491867711924584\\
232.8	0.489586740112145\\
233.6	0.490507024284256\\
234.4	0.487139874645351\\
235.2	0.481548293867007\\
236	0.467907708159515\\
237.6	0.430595959985396\\
238.4	0.414129925294048\\
239.2	0.392827679974374\\
240.8	0.351768552514187\\
241.6	0.324286500884767\\
};
\end{axis}
\end{tikzpicture}%
		\label{j1:fig:eight_many_laps_position_error.tex}
	}
	\caption{Results from real-world experiments of backward tracking of the figure-eight nominal path in Figure~\ref{j1:fig:eight_rev} over four consecutive laps. In (a), the norm of the position estimation error and in (b), the estimated lateral path-following error $z_3$. This experiment was performed under rougher ground surface conditions compared to the first experiment.}
	\label{j1:fig:eight_many_laps}
\end{figure}
\tikzexternalenable

The performance of the nonlinear observer are presented in Figure~\ref{j1:fig:eight_est_poes_error} and Figure~\ref{j1:fig:estimation_eight_rev}. 
In Figure~\ref{j1:fig:eight_est_poes_error}, the Euclidean norm of the difference between the estimated position of the axle of the semitrailer $(\hat x_3(\cdot),\hat y_3(\cdot))$ and its ground truth $(x_{3,GT}(\cdot),y_{3,GT}(\cdot))$ measured by the external RTK-GPS is presented. The maximum estimation error is $0.6$ m and the average error is \mbox{$0.23$ m}. It can be seen from Figure~\ref{j1:fig:eight_rev_z3} that the estimated lateral path-following error for the axle of the semitrailer $\tilde z_3$ is increasing at the end of the maneuver. The reason for this is because the nonlinear observer is not able to track the absolute position of the axle of the semitrailer with high precision in this part of the maneuver. 
Likely causes to this estimation error are probably because of known asymmetries in the tractor's steering column~\cite{truls2018} and the unavoidable lateral slip-effects of the wheels of the dolly and the semitrailer which are not captured by the kinematic model of the vehicle~\cite{winkler1998simplified}. 
Note that the absolute position of the axle of the semitrailer ($\hat x_3(\cdot),\hat y_3(\cdot)$) is estimated from GPS measurements of the car-like tractor's position and its orientation, propagated about 14 m through two hitch connections whose angles are estimated using only a LIDAR sensor on the car-like tractor.

In Figure~\ref{j1:fig:eight_est_beta2} and~\ref{j1:fig:eight_est_beta3}, the estimated trajectories of the joint-angles, $\hat\beta_2$ and $\hat\beta_3$, are compared with their derived angles based on the outputs from the RANSAC algorithm, respectively. 
The maximum (avg.) errors in the residuals $\hat\beta_2-\beta_2$ and $\hat\beta_3-\beta_3$, are $0.83\degree$ (avg. $0.27\degree$) and $2.18\degree$ (avg. $0.8\degree$), respectively.

To illustrate the repeatability of the system, the same figure-eight nominal path was executed multiple times. This experiment was performed at another occasion on rougher ground surface conditions compared to the first experiment. The resulting estimated lateral control error $\tilde z_3$ and the Euclidean norm of the position error $||e(t)||_2$ over four consecutive laps are presented in Figure~\ref{j1:fig:eight_many_laps}. As can be seen, both errors are bounded and have a periodic behavior of approximately 250 seconds, \textit{i.e.}, one lap time around the figure-eight nominal path.

\begin{figure}[t!]
	\begin{center}
		\includegraphics[width=0.75\linewidth]{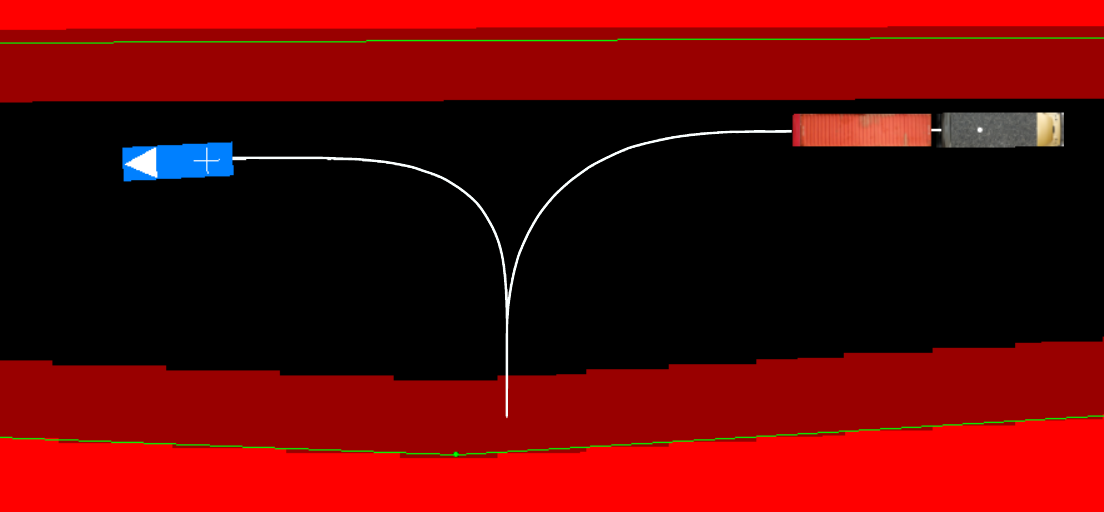}   
		\caption{Illustration of the planned two-point turn maneuver. The goal position of the semitrailer is illustrated by the white cross inside the blue rectangle, where the white arrow specifies its goal orientation. The white path is the planned path for axle of the semitrailer $(x_{3,r}(\cdot),y_{3,r}(\cdot))$.} 
		\label{j1:fig:two_point_turn_gui}
	\end{center} 
	\vspace{-15pt}
\end{figure}

\tikzexternaldisable
\begin{figure*}[b!]
	\setlength\figureheight{0.25\textwidth}
	\setlength\figurewidth{0.8\textwidth}
	\begin{tikzpicture}
	\node[anchor=south west] (myplot) at (0,0) {
		\input{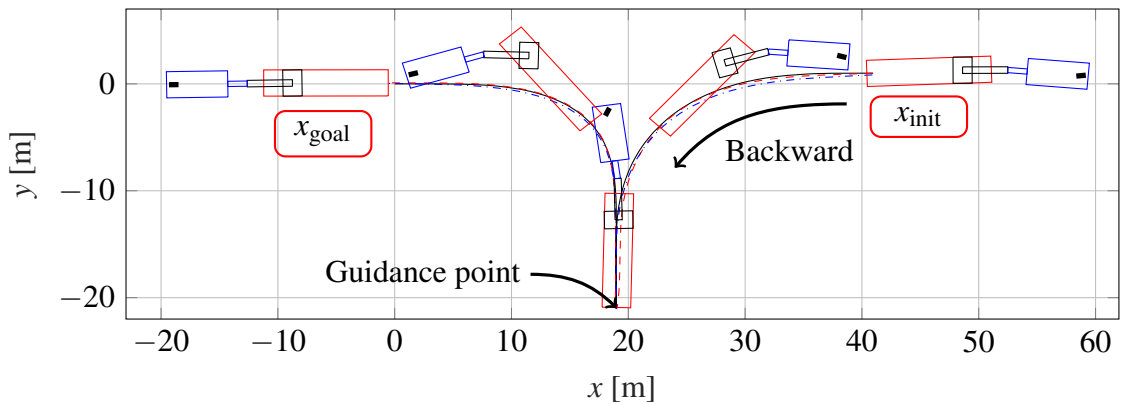}
	};
	\begin{scope}[x={(myplot.south east)}, y={(myplot.north west)}]
	
	%
	\draw[thick,red] (0.77,0.72) 
	node[right,draw=red,rounded corners,
	text width=1cm,align=center,text=black] 
	{$x_{\text{init}}$};
	
	\draw[thick,red] (0.25,0.68) 
	node[right,draw=red,rounded corners,
	text width=1cm,align=center,text=black] 
	{$x_{\text{goal}}$};
	\draw[->,very thick] (0.474,0.35) to [out=0,in=135] (0.549,0.271);  
	\node at (0.38,0.35) {Guidance point};
	\draw[<-,very thick] (0.6,0.6) to [out=50,in=180] (0.75,0.75);  
	\node at (0.7,0.64) {Backward};
	
	\end{scope}
	\end{tikzpicture}
	\caption{Results from real-world experiments while executing the planned two-point turn maneuver. The black line is the planned path for the axle of the semitrailer $(x_{3,r}(\cdot),y_{3,r}(\cdot))$. The dashed red line is the estimated path taken by the axle of the semitrailer $(\hat x_3(\cdot),\hat y_3(\cdot))$ and the dashed-dotted blue line is the ground truth path $(x_{3,GT}(\cdot),y_{3,GT}(\cdot))$ measured by the external RTK-GPS.}
	\label{j1:fig:two_point_turn}
\end{figure*}
\tikzexternalenable

\tikzexternaldisable
\begin{figure}[t!]
	\centering
	\setlength\figureheight{0.17\columnwidth}
	\setlength\figurewidth{0.36\columnwidth} 
	\subfloat[][The norm of the position estimation error of the axle of the semitrailer.]{
		\begin{tikzpicture}
		\node[anchor=south west] (myplot) at (0,0) {
			\input{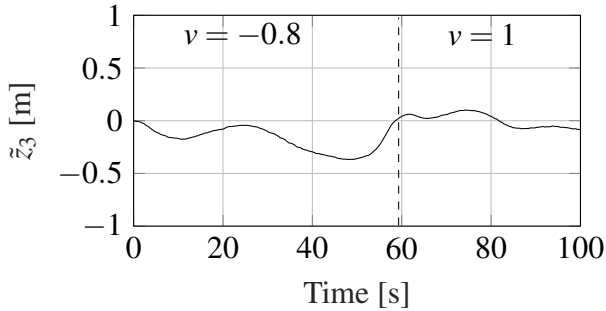}
		};
		\begin{scope}[x={(myplot.south east)}, y={(myplot.north west)}]
		
		%
		\node at (0.775,0.85) {$v=1$};
		\node at (0.38,0.85) {$v=-0.8$};
		\draw [dashed] (0.636,0.3) -- (0.636,0.9); 
		\end{scope}
		\end{tikzpicture}
		\label{j1:fig:two_point_turn_poes_error}
	} 
	~
	\subfloat[][The controlled curvature $\kappa(t)$ (black line) and the nominal feed-forward $\kappa_r(\tilde s(t))$ (red dashed line).]{
		\begin{tikzpicture}
		\node[anchor=south west] (myplot) at (0,0) {
			\input{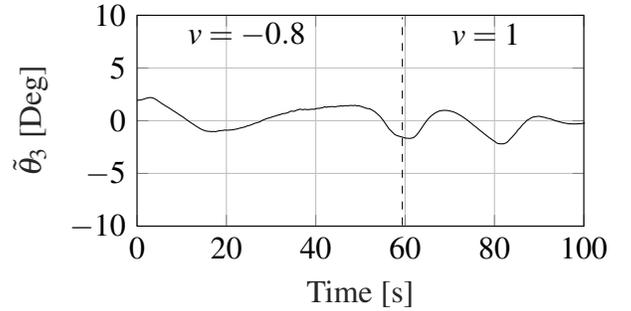}
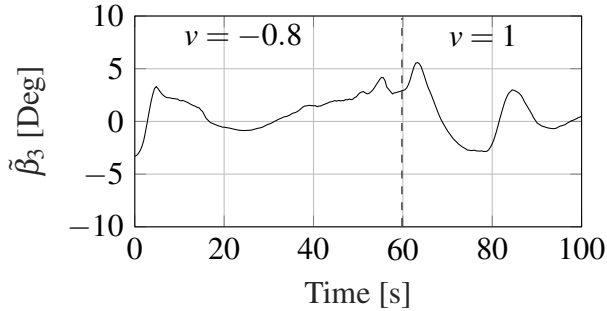
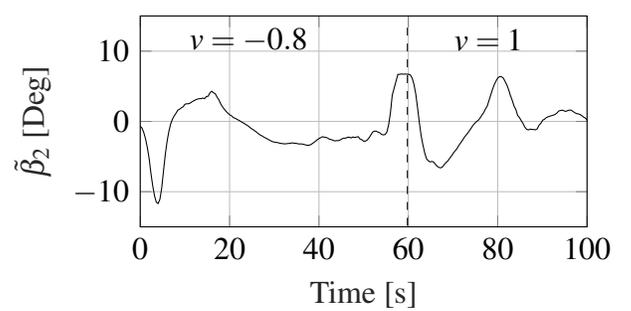
		};
		\begin{scope}[x={(myplot.south east)}, y={(myplot.north west)}]
		\node at (0.775,0.35) {$v=1$};
		\node at (0.40,0.85) {$v=-0.8$};
		\draw [dashed] (0.65,0.3) -- (0.65,0.9); 
		\end{scope}
		\end{tikzpicture}
		\label{j1:fig:two_point_turn_kappa}
		
	}
	\quad
	\subfloat[][Estimated lateral error for the axle of the semitrailer.]{
		\begin{tikzpicture}
		\node[anchor=south west] (myplot) at (0,0) {
%
%
%
\begin{tikzpicture}

\begin{axis}[%
width=\figurewidth,
height=\figureheight,
at={(0.0\figurewidth,0\figureheight)},
scale only axis,
xmin=0,
xmax=100,
xlabel style={font=\color{white!15!black}},
xlabel={Time [s]},
ymin=-1,
ymax=1,
ylabel style={font=\color{white!15!black}},
ylabel={$\tilde z_3$ [m]},
axis background/.style={fill=white},
xmajorgrids,
xminorgrids,
ymajorgrids,
yminorgrids
]
\addplot [color=black, forget plot]
  table[row sep=crcr]{%
0	-0.00310230292798508\\
0.400000000000006	-0.00426178416134348\\
0.799999999999997	-0.00639075886230955\\
1.2	-0.00967139253961591\\
1.60000000000001	-0.0147521293675936\\
2	-0.0220325602820708\\
2.40000000000001	-0.0315088690445009\\
3.2	-0.054940122409505\\
4.80000000000001	-0.10440456062959\\
5.60000000000001	-0.117329981820191\\
6	-0.128550690724069\\
6.40000000000001	-0.135473148760099\\
6.80000000000001	-0.137727174536792\\
7.2	-0.144307173144242\\
7.60000000000001	-0.154332911240601\\
8	-0.159255985378437\\
8.40000000000001	-0.160091961629988\\
9.2	-0.165704457419906\\
9.60000000000001	-0.165894452326285\\
10	-0.168616390736801\\
10.4	-0.173018260350304\\
10.8	-0.176455040089763\\
11.2	-0.174345541555113\\
11.6	-0.174293486201435\\
12	-0.169911283609252\\
12.4	-0.164363679185186\\
12.8	-0.161026272751471\\
13.2	-0.155130153809154\\
13.6	-0.153142178404366\\
14	-0.145537703141031\\
14.4	-0.14278810676592\\
14.8	-0.133776608846887\\
15.6	-0.124484098246143\\
16	-0.115337769934143\\
16.4	-0.116347341576414\\
16.8	-0.10681261958581\\
17.2	-0.10584787296672\\
17.6	-0.0973129541422537\\
18	-0.0932378841119146\\
18.4	-0.0925291056014146\\
19.2	-0.0832366636710873\\
19.6	-0.0822486906809132\\
20	-0.0799217534139274\\
20.8	-0.0635626913982463\\
21.2	-0.0568424676356329\\
21.6	-0.0548744598678468\\
22	-0.0559603313410832\\
22.4	-0.0510436690693297\\
22.8	-0.0517681633747742\\
23.2	-0.0481128072225374\\
23.6	-0.0454000937195929\\
24	-0.0438834868821942\\
25.2	-0.0435101391458659\\
26	-0.0475017173872914\\
26.4	-0.0507752986420655\\
26.8	-0.0555242284305137\\
27.6	-0.0588013155782647\\
28	-0.0667571460179346\\
28.4	-0.0729596837928739\\
28.8	-0.0783702410596874\\
29.2	-0.0849051176352589\\
29.6	-0.0928601320345166\\
30	-0.102776011681371\\
30.4	-0.111559213916721\\
30.8	-0.116594592132586\\
31.2	-0.120449672979092\\
31.6	-0.127316796481409\\
32.4	-0.147853077347122\\
32.8	-0.161545445293356\\
33.2	-0.166523419737274\\
33.6	-0.173883041994884\\
34	-0.183982383458996\\
34.4	-0.199840902153369\\
34.8	-0.211954241326083\\
35.2	-0.212895824037474\\
35.6	-0.220522027645558\\
36	-0.229843071443909\\
36.4	-0.244671569915695\\
36.8	-0.252365175978184\\
37.2	-0.253620834391938\\
37.6	-0.262472014979082\\
38	-0.27222063054613\\
38.4	-0.276950108559404\\
38.8	-0.280190200189722\\
39.2	-0.287294799286968\\
39.6	-0.291910505747751\\
40	-0.294877181014272\\
40.4	-0.301531307928613\\
40.8	-0.310084659825407\\
41.6	-0.313934853979575\\
42	-0.31958001760367\\
42.4	-0.321826188198784\\
42.8	-0.325177710088639\\
43.6	-0.335108863023237\\
44	-0.335657716208402\\
44.4	-0.340300376366457\\
44.8	-0.344087873248242\\
45.2	-0.346617880855803\\
45.6	-0.355948420117457\\
46	-0.360019677933437\\
46.4	-0.35973240912638\\
46.8	-0.365175959416106\\
47.2	-0.364192170809375\\
47.6	-0.365391629558829\\
48	-0.367712093889793\\
48.4	-0.366205858467495\\
48.8	-0.366713875184303\\
49.2	-0.366169929551759\\
49.6	-0.363872316961206\\
50.4	-0.354490197867392\\
51.2	-0.346901621033183\\
51.6	-0.339741128749353\\
52.8	-0.324131482641604\\
53.2	-0.310755534554971\\
53.6	-0.293717426931224\\
54	-0.280058512355723\\
54.4	-0.263925440347336\\
54.8	-0.243117763954714\\
55.2	-0.220980642370364\\
55.6	-0.194136032295276\\
56.4	-0.135292492911248\\
57.66	-0.0412166286379261\\
58.06	-0.0200707156278668\\
58.46	-0.00464092596816101\\
59.26	0.0209811986990616\\
59.66	0.0327473537957417\\
60.06	0.0424486093406529\\
60.46	0.049829860873345\\
60.86	0.055551457364615\\
61.26	0.0599166760621301\\
61.66	0.0619545083111603\\
62.06	0.0609404173673198\\
62.46	0.0570495699098359\\
62.86	0.0513924903844156\\
63.66	0.0387366389092989\\
64.06	0.0330112921380561\\
64.46	0.0282908734110521\\
64.86	0.024732997205021\\
65.26	0.0225567433922862\\
65.66	0.0218272632557159\\
66.06	0.0226237619723406\\
66.86	0.0271091133352144\\
68.06	0.0353532271234513\\
68.46	0.0423764436891503\\
68.86	0.0507329758956416\\
70.06	0.0581540673862264\\
70.46	0.0638359217225286\\
70.86	0.0724460352277561\\
71.66	0.0796791741438341\\
72.46	0.0890305602325441\\
72.86	0.0945416560958137\\
73.26	0.0981217191296366\\
73.66	0.0958317076637343\\
74.06	0.0989796397402074\\
74.46	0.100669161787678\\
75.66	0.0954824095747853\\
76.06	0.0954970756850173\\
76.46	0.0943321063717519\\
76.86	0.0888633026137029\\
77.26	0.086214778483793\\
77.66	0.0815844580890683\\
78.06	0.0747144187046445\\
78.46	0.0709653378090849\\
78.86	0.0644488922789037\\
79.26	0.0568198398611202\\
79.66	0.0457858527372679\\
80.06	0.0409102861322737\\
80.86	0.0196888004205675\\
81.26	0.0157133404090644\\
82.06	-0.0103470128163963\\
82.46	-0.0156632963133063\\
82.86	-0.0298144339716515\\
83.66	-0.0449969382072766\\
84.06	-0.0534308574212332\\
84.46	-0.0573068978545308\\
84.86	-0.0664237665539815\\
85.26	-0.06684291992282\\
85.66	-0.0706133826544715\\
86.46	-0.0713542543022641\\
86.86	-0.0774382596133449\\
87.26	-0.0738057704711963\\
87.66	-0.0749535251832043\\
88.06	-0.0737164190280595\\
88.46	-0.0733705791402883\\
88.86	-0.0678560135047803\\
89.66	-0.0688723661466497\\
90.06	-0.0646592692841779\\
90.46	-0.0620113137085241\\
90.86	-0.0614859488242274\\
91.26	-0.0572667615584095\\
91.66	-0.0545369034749257\\
92.06	-0.0611437940863624\\
92.46	-0.0563793824536702\\
92.86	-0.0579040039184662\\
93.26	-0.0571566579688465\\
93.66	-0.0506668407495567\\
94.06	-0.0515446087190838\\
94.46	-0.0554978913095709\\
94.86	-0.0546516428283326\\
95.66	-0.0637608459583277\\
96.06	-0.0612539557979517\\
96.46	-0.0655678261977073\\
96.86	-0.0646638329361338\\
97.26	-0.0687806620915552\\
97.66	-0.0707511402552967\\
98.06	-0.07542300689623\\
98.46	-0.0778291931216302\\
98.86	-0.0761276742830432\\
99.26	-0.0809053246282616\\
100.06	-0.0847324346471225\\
100.46	-0.090925311744968\\
100.86	-0.0854387215208732\\
101.26	-0.088573471425363\\
102.06	-0.0922522159592631\\
102.86	-0.0939835374248048\\
104.06	-0.0948620194641734\\
};
\end{axis}
\end{tikzpicture}%
		};
		\begin{scope}[x={(myplot.south east)}, y={(myplot.north west)}]
		
		%
		\node at (0.775,0.85) {$v=1$};
		\node at (0.40,0.85) {$v=-0.8$};
		\draw [dashed] (0.644,0.3) -- (0.644,0.90); 
		\end{scope}
		\end{tikzpicture}
		\label{j1:fig:two_point_turn_z3}
	}
	~
	\subfloat[][Estimated orientation error of the semitrailer.]{
		\begin{tikzpicture}
		\node[anchor=south west] (myplot) at (0,0) {
			\begin{tikzpicture}

\begin{axis}[%
width=\figurewidth,
height=\figureheight,
at={(0\figurewidth,0\figureheight)},
scale only axis,
xmin=0,
xmax=100,
xlabel style={font=\color{white!15!black}},
xlabel={Time [s]},
ymin=-10,
ymax=10,
ylabel style={font=\color{white!15!black}},
ylabel={$\tilde\theta_3$ [Deg]},
axis background/.style={fill=white},
xmajorgrids,
ymajorgrids
]
\addplot [color=black, forget plot]
  table[row sep=crcr]{%
0	1.93521542678629\\
0.400000000000006	1.94871913766222\\
0.799999999999997	1.9734079236343\\
1.2	2.00864743088333\\
1.60000000000001	2.05732298744684\\
2	2.11525558476983\\
2.40000000000001	2.16500390791984\\
2.8	2.194153874637\\
3.2	2.19040282012219\\
3.60000000000001	2.14782085951539\\
4	2.07313679636611\\
4.40000000000001	1.96536516244319\\
5.60000000000001	1.61615146400219\\
6.40000000000001	1.40042356760533\\
7.2	1.19965559975981\\
8.40000000000001	0.866082287466185\\
8.80000000000001	0.762874287125896\\
9.60000000000001	0.529257645045092\\
10	0.4299415451908\\
10.4	0.315780776227427\\
10.8	0.181510102738329\\
12	-0.175881693356914\\
12.4	-0.281309318414486\\
13.2	-0.513980326140782\\
13.6	-0.611334480836405\\
14	-0.71999387926887\\
14.4	-0.815831998773021\\
14.8	-0.891717161360006\\
15.6	-0.991492884920675\\
16	-1.01490389190943\\
16.4	-1.01887607379861\\
16.8	-1.0397036684555\\
17.2	-1.01481915487503\\
17.6	-1.04464755147583\\
18	-0.984441034767755\\
18.8	-0.931895108822374\\
19.2	-0.927207784782752\\
19.6	-0.887315317883349\\
20.4	-0.906452333998971\\
20.8	-0.873885244984407\\
21.2	-0.874652520688116\\
21.6	-0.848858311287572\\
22	-0.807275639372847\\
22.4	-0.783208880862929\\
22.8	-0.725494956315089\\
23.2	-0.688118346907771\\
23.6	-0.639445405026066\\
24	-0.610584600770736\\
24.8	-0.472625771872174\\
25.2	-0.446818168897536\\
26.4	-0.238945443613446\\
26.8	-0.185293320507213\\
27.2	-0.150211267794859\\
27.6	-0.0779652420038417\\
28	0.0306445008199603\\
28.4	0.0907142103063876\\
28.8	0.169972499167585\\
30	0.320749205240091\\
30.4	0.386111074028491\\
30.8	0.427542894812092\\
31.6	0.541104287995367\\
32	0.561842770027937\\
32.4	0.65290007727819\\
32.8	0.67966945626533\\
33.2	0.695137326775324\\
33.6	0.741305190057219\\
34	0.813185887439033\\
34.4	0.907529689618826\\
35.2	0.893129783218583\\
35.6	0.980376858768182\\
36	1.02592173409259\\
36.4	1.09401242353698\\
36.8	1.06348521320673\\
37.2	1.04707098797071\\
37.6	1.08711542657915\\
38	1.05264794414974\\
38.4	1.05314964288861\\
39.2	1.0963128099051\\
39.6	1.07531296641507\\
40	1.13474756502524\\
40.4	1.17196409187386\\
40.8	1.17236867386049\\
41.6	1.24523078404866\\
42	1.32804533363118\\
42.4	1.26703445542663\\
42.8	1.33149895475196\\
43.2	1.32277808384234\\
43.6	1.3451298779012\\
44	1.32707886186363\\
44.4	1.32248545292508\\
44.8	1.36202893984658\\
45.2	1.36948302410859\\
45.6	1.41964713502433\\
46	1.37621236126664\\
46.4	1.42333191535985\\
46.8	1.4252745705702\\
47.2	1.41517859048156\\
47.6	1.4429163088661\\
48	1.44468554130454\\
48.4	1.43235491541856\\
48.8	1.4586805989139\\
49.2	1.4113655306557\\
49.6	1.42482473858033\\
50	1.37402307219423\\
50.8	1.25777399957256\\
51.2	1.18979248643319\\
51.6	1.15305887530322\\
52	1.13928921393646\\
52.4	1.06253133460045\\
52.8	0.965353964410454\\
53.2	0.855841752827075\\
53.6	0.715935466241419\\
54	0.563283086169164\\
54.4	0.401071562868779\\
54.8	0.191959918642482\\
55.2	-0.0285235670001072\\
56.4	-0.775110571023973\\
57.66	-1.29939567827074\\
58.06	-1.3914666282825\\
58.46	-1.4499603683345\\
60.06	-1.64513985405678\\
60.46	-1.66839664818067\\
60.86	-1.67804770547791\\
61.26	-1.66978317658219\\
61.66	-1.63125407947415\\
62.06	-1.55212633793954\\
62.46	-1.40203976111769\\
62.86	-1.20120487876264\\
63.26	-0.967250409122187\\
64.46	-0.212275449214829\\
64.86	0.0135647556916467\\
65.26	0.213771260643057\\
65.66	0.387488317641612\\
66.06	0.532628076596225\\
66.46	0.656547665099239\\
66.86	0.765514208600948\\
67.26	0.851520324155729\\
67.66	0.91914389269391\\
68.06	0.955794794501898\\
68.46	0.974456045253135\\
68.86	0.981025833461047\\
69.26	0.974992108381272\\
69.66	0.956520608822913\\
70.46	0.879078323187727\\
70.86	0.828789561319624\\
71.26	0.757502556787088\\
71.66	0.673021921294762\\
72.46	0.479766147680039\\
72.86	0.37572683861255\\
74.06	0.0128563295047996\\
74.46	-0.135339998300609\\
74.86	-0.260927553683487\\
75.26	-0.40819851842214\\
76.06	-0.667745013191535\\
76.46	-0.818884132836672\\
78.86	-1.62395954416228\\
79.26	-1.73903233014283\\
79.66	-1.8647490054495\\
80.06	-1.96095819632465\\
80.46	-2.10730246981157\\
80.86	-2.17028876679015\\
81.26	-2.1956898134266\\
81.66	-2.19730387250483\\
82.06	-2.19092596992392\\
82.46	-2.1282246198998\\
82.86	-2.0367807666312\\
83.66	-1.67339293155854\\
84.06	-1.4999576732241\\
84.86	-1.11277047087536\\
85.26	-0.867896556464473\\
85.66	-0.711363294922336\\
86.06	-0.49064994740074\\
86.46	-0.303809695768223\\
86.86	-0.160368012554713\\
87.26	0.0350443632749631\\
87.66	0.118635143179318\\
88.06	0.236849962920431\\
88.46	0.316034613804632\\
88.86	0.372187573322861\\
89.26	0.386955456164642\\
89.66	0.412012298619516\\
90.06	0.424260401946512\\
90.46	0.374167260965109\\
91.26	0.313713976992844\\
91.66	0.252376983935847\\
92.46	0.181690502441739\\
93.26	0.064636174529312\\
93.66	-0.0115071628717089\\
94.06	-0.0458213779238719\\
95.26	-0.189172000841921\\
96.46	-0.279259621333239\\
96.86	-0.298306991215995\\
97.26	-0.297157518403452\\
97.66	-0.31059500846095\\
98.46	-0.302602966760531\\
99.26	-0.276364554940088\\
99.66	-0.260489274318275\\
101.66	-0.129515434423951\\
102.46	-0.0828915092725993\\
103.26	-0.0344924553085662\\
104.06	0.0161602275762505\\
};
\end{axis}
\end{tikzpicture}%
		};
		\begin{scope}[x={(myplot.south east)}, y={(myplot.north west)}]
		\node at (0.775,0.85) {$v=1$};
		\node at (0.40,0.85) {$v=-0.8$};
		\draw [dashed] (0.644,0.3) -- (0.644,0.9); 
		\end{scope}
		\end{tikzpicture}
		\label{j1:fig:two_point_turn_theta3}
	}
	\quad
	\subfloat[][Estimated joint angle error between the semitrailer and the dolly.]{
		\begin{tikzpicture}
		\node[anchor=south west] (myplot) at (0,0) {
%
%
\definecolor{mycolor1}{rgb}{0.00000,0.44700,0.74100}%
\begin{tikzpicture}

\begin{axis}[%
width=\figurewidth,
height=\figureheight,
at={(0\figurewidth,0\figureheight)},
scale only axis,
xmin=0,
xmax=100,
xlabel style={font=\color{white!15!black}},
xlabel={Time [s]},
ymin=-10,
ymax=10,
ylabel style={font=\color{white!15!black}},
ylabel={$\tilde\beta_3$ [Deg]},
axis background/.style={fill=white},
xmajorgrids,
ymajorgrids
]
\addplot [color=black, forget plot]
  table[row sep=crcr]{%
0	-3.30355567486936\\
0.400000000000006	-3.22537049944023\\
0.799999999999997	-3.0426815087192\\
1.2	-2.76850981802865\\
1.60000000000001	-2.3677751489046\\
2	-1.8051899693304\\
2.40000000000001	-0.958570199182446\\
2.8	-0.0504955985390581\\
3.2	0.894348278294345\\
3.60000000000001	1.80248372031519\\
4	2.52881777604199\\
4.40000000000001	3.10616085997847\\
4.80000000000001	3.30059862984557\\
5.2	3.12266217012639\\
5.60000000000001	2.8569731474982\\
6	2.64000642554062\\
6.40000000000001	2.46736101455068\\
6.80000000000001	2.33972889806864\\
7.2	2.2545308259092\\
7.60000000000001	2.24953817106933\\
8	2.22385772700714\\
8.40000000000001	2.17107679986563\\
8.80000000000001	2.16386732175327\\
9.2	2.13391075238383\\
9.60000000000001	2.11867900364788\\
10	1.98259965070983\\
10.4	1.95404156257295\\
10.8	1.98463718151662\\
11.2	1.92065965333475\\
11.6	1.83682591146201\\
12	1.76364523698209\\
12.4	1.65332071313516\\
12.8	1.56775028894424\\
13.6	1.42671445982459\\
14	1.33146523069294\\
14.4	1.28883556548367\\
14.8	1.0802587704261\\
15.2	0.92406973974596\\
15.6	0.582765933972354\\
16	0.277773068012991\\
16.4	0.112535810930936\\
16.8	-0.061615029780981\\
17.2	-0.159116820616347\\
18.8	-0.431678681076235\\
19.2	-0.476695657482992\\
19.6	-0.539961031722186\\
20	-0.54960525883024\\
21.2	-0.633485851531816\\
21.6	-0.671223393748079\\
22	-0.718791339088128\\
22.8	-0.786676535888034\\
23.2	-0.826714251029003\\
24	-0.857841237345752\\
24.8	-0.865201726671785\\
25.2	-0.841330589048567\\
25.6	-0.843715285564258\\
26	-0.83234675585588\\
26.8	-0.736681258710831\\
27.6	-0.641757210616703\\
28	-0.620980842378572\\
28.4	-0.570620523606749\\
28.8	-0.498327987162341\\
29.6	-0.317093639971532\\
30	-0.254413741010268\\
30.4	-0.199764503666714\\
32	0.161982685282766\\
32.4	0.242789319331862\\
32.8	0.311877269755925\\
33.2	0.393278050221525\\
33.6	0.45865840854546\\
34	0.473493259257609\\
34.4	0.545106938931553\\
34.8	0.631380712524418\\
35.2	0.754016545628588\\
36	0.854534678735519\\
36.4	0.943253926855519\\
36.8	1.06125686399001\\
37.6	1.36067432123509\\
38	1.45391013866546\\
38.4	1.47383241309223\\
38.8	1.51062973708055\\
39.2	1.52139891768925\\
39.6	1.49619776080772\\
40	1.48366201580322\\
40.4	1.49739283046907\\
40.8	1.40278916093568\\
41.2	1.4605819528241\\
41.6	1.46294402433816\\
42	1.54474076901265\\
43.2	1.74952670235915\\
43.6	1.83331775554957\\
44	1.86906867321271\\
44.4	1.9171771106317\\
44.8	1.97513702871136\\
45.2	1.91886203292258\\
45.6	1.99710099609142\\
46	1.94162897441693\\
46.4	1.98323677929335\\
46.8	1.97420897675789\\
47.2	2.07771942641237\\
47.6	2.04673619423332\\
48	2.12778983567044\\
48.4	2.06327336287438\\
48.8	2.10833745691424\\
49.2	2.26560170878014\\
49.6	2.33796927058373\\
50	2.55760467551606\\
50.4	2.71360214073434\\
50.8	2.78937199051822\\
51.2	2.82905133417118\\
51.6	2.69986150767156\\
52	2.62237215776412\\
52.4	2.65900779836818\\
52.8	2.78794490786851\\
53.2	2.94435858773933\\
53.6	3.17905093125356\\
54	3.46360685736003\\
54.4	3.70147331784761\\
54.8	3.9782855169826\\
55.2	4.17234058281277\\
55.6	4.1524365577977\\
56	3.87375929013248\\
56.4	3.26147701382703\\
57.66	2.61540130938589\\
58.06	2.70528894674078\\
58.46	2.72748428677897\\
58.86	2.79096000950025\\
59.26	2.8309475654395\\
59.66	2.89187840236471\\
60.06	2.93766436202249\\
60.46	3.00604095891666\\
60.86	3.17854122838673\\
61.26	3.51546015943489\\
61.66	3.98527151984305\\
62.06	4.60035524092525\\
62.46	5.16106766800641\\
62.86	5.52069804526191\\
63.26	5.5838730262163\\
63.66	5.48557550500242\\
64.06	5.21300143824742\\
64.46	4.75069538852006\\
64.86	4.24072443678293\\
65.26	3.68569045822301\\
65.66	3.204883987355\\
66.06	2.82311525945219\\
66.46	2.41484198247515\\
67.66	1.0681602642502\\
68.06	0.630117394332231\\
68.46	0.256325180121664\\
68.86	-0.0740806386144612\\
69.66	-0.657930671684468\\
70.06	-0.93773011808598\\
71.26	-1.63221425466107\\
72.06	-1.97688854534665\\
72.46	-2.15068228189079\\
72.86	-2.27189508777008\\
73.26	-2.40185096267379\\
74.06	-2.60546829334199\\
74.46	-2.7182366375296\\
74.86	-2.72905096141984\\
75.26	-2.76656513510591\\
76.06	-2.73175528944435\\
76.46	-2.7978296857014\\
77.26	-2.81387998992069\\
77.66	-2.79673194623574\\
78.06	-2.84664269982196\\
78.46	-2.85404641143884\\
78.86	-2.81745922248919\\
79.26	-2.65809319334964\\
79.66	-2.40004331042127\\
80.06	-2.05232307553098\\
80.46	-1.62085010054263\\
80.86	-1.06680724646429\\
81.26	-0.467423622323849\\
81.66	0.195708171712511\\
82.46	1.38116947777849\\
82.86	1.92293581387565\\
83.26	2.32692805634699\\
83.66	2.66607298873433\\
84.06	2.81946430989741\\
84.46	2.99216884471237\\
84.86	2.94520851940418\\
85.26	2.87713351805054\\
85.66	2.78162805010331\\
86.06	2.65924962975349\\
86.46	2.46193827123651\\
86.86	2.18608101792486\\
87.26	1.82906660312148\\
87.66	1.56030164748923\\
88.06	1.26723989471026\\
88.46	0.988021974883551\\
88.86	0.68896070133151\\
89.26	0.352232668734572\\
90.06	-0.157426815704227\\
90.46	-0.256734057656942\\
90.86	-0.380343059315706\\
91.26	-0.461746027576282\\
91.66	-0.504625154772825\\
92.06	-0.567635090609215\\
92.46	-0.615092525622941\\
93.26	-0.673753231783621\\
93.66	-0.68435469923169\\
94.06	-0.673234564293523\\
94.46	-0.616630598846669\\
94.86	-0.518162589707785\\
95.26	-0.430628352728363\\
95.66	-0.404744870620917\\
96.06	-0.346196942622825\\
96.46	-0.262500444863718\\
96.86	-0.135484739614071\\
97.26	-0.051885421973239\\
97.66	0.052094863320832\\
98.06	0.10772955222032\\
98.46	0.177779658852955\\
99.26	0.353704787944125\\
100.06	0.494408045548127\\
100.46	0.494755584783007\\
101.66	0.421228278202705\\
102.46	0.426353508081576\\
103.26	0.376858675663215\\
103.66	0.433485319472553\\
104.06	0.535487898146059\\
};
\end{axis}
\end{tikzpicture}%
		};
		\begin{scope}[x={(myplot.south east)}, y={(myplot.north west)}]
		\node at (0.775,0.85) {$v=1$};
		\node at (0.40,0.85) {$v=-0.8$};
		\draw [dashed] (0.648,0.3) -- (0.648,0.9); 
		\end{scope}
		\end{tikzpicture}
		\label{j1:fig:two_point_turn_b3}
	}
	~
	\subfloat[][Estimated joint angle error between the dolly and the tractor.]{
		\begin{tikzpicture}
		\node[anchor=south west] (myplot) at (0,0) {
%
%
\definecolor{mycolor1}{rgb}{0.00000,0.44700,0.74100}%
\begin{tikzpicture}

\begin{axis}[%
width=\figurewidth,
height=\figureheight,
at={(0\figurewidth,0\figureheight)},
scale only axis,
xmin=0,
xmax=100,
xlabel style={font=\color{white!15!black}},
xlabel={Time [s]},
ymin=-15,
ymax=15,
ylabel style={font=\color{white!15!black}},
ylabel={$\tilde\beta_2$ [Deg]},
axis background/.style={fill=white},
xmajorgrids,
ymajorgrids
]
\addplot [color=black, forget plot]
  table[row sep=crcr]{%
0	-0.670664332092343\\
0.400000000000006	-0.957673203245363\\
0.799999999999997	-1.59862808007799\\
1.2	-2.51386734572247\\
1.60000000000001	-3.75560342079361\\
2	-5.33008795609437\\
2.40000000000001	-7.41694681594645\\
2.8	-9.19944262798583\\
3.2	-10.6112965171672\\
3.60000000000001	-11.4966307600261\\
4	-11.6823384809169\\
4.40000000000001	-11.0799115272535\\
4.80000000000001	-9.5019406072683\\
5.2	-7.08313967639548\\
5.60000000000001	-4.73507559966188\\
6	-2.876894674873\\
6.40000000000001	-1.42614680707155\\
6.80000000000001	-0.379650755610299\\
7.2	0.320818654390123\\
7.60000000000001	0.782357561815914\\
8	1.16804356329028\\
8.40000000000001	1.59737153871413\\
8.80000000000001	1.73343701783041\\
9.2	1.94010050614672\\
9.60000000000001	2.08731114644118\\
10	2.43648386374717\\
10.4	2.51712193086199\\
10.8	2.51000764851118\\
11.2	2.62593559552016\\
11.6	2.80617959196174\\
12	2.93649917443877\\
12.4	3.07954105146199\\
12.8	3.18003682637156\\
13.2	3.23637329386288\\
13.6	3.24013267267074\\
14	3.33020515202004\\
14.4	3.23444120410142\\
14.8	3.54308336542913\\
15.2	3.58946205744081\\
15.6	4.06648118841403\\
16	4.29771169323611\\
16.4	4.07705645568332\\
16.8	3.93022702532191\\
17.2	3.44034771535861\\
17.6	3.07963667411038\\
18	2.50285345224242\\
18.4	2.08161946442846\\
18.8	1.74761435611325\\
19.2	1.4407068493913\\
19.6	1.18632958052399\\
20	0.954801322142785\\
20.4	0.813169932544014\\
20.8	0.565381661791648\\
21.2	0.461712196795546\\
22.4	0.0393382310657984\\
22.8	-0.136164957086294\\
23.2	-0.253434622720576\\
23.6	-0.432768202970564\\
24	-0.571515651913998\\
24.8	-0.992863298344616\\
25.2	-1.14652377164968\\
25.6	-1.33851580018539\\
26	-1.50542022373453\\
26.4	-1.732319961736\\
26.8	-1.91243888033841\\
27.2	-2.03866995426694\\
27.6	-2.20329565595424\\
28	-2.34825340103964\\
28.4	-2.46287806257762\\
28.8	-2.62047400550509\\
29.6	-2.87959580124632\\
30	-2.93948631498358\\
30.4	-2.98466225337111\\
30.8	-3.01669434647458\\
31.2	-3.09569466229249\\
31.6	-3.14129727405972\\
32.4	-3.19455055147101\\
33.6	-3.10822656531558\\
34	-3.05018767317362\\
34.4	-3.09602400547652\\
34.8	-3.10494828795738\\
35.2	-3.16681678178959\\
35.6	-3.21103542647626\\
36	-3.19268206784619\\
36.4	-3.24104087438694\\
37.2	-3.37243578477141\\
37.6	-3.39854653075132\\
38	-3.35452468089011\\
38.4	-3.20214871003583\\
39.2	-2.83751223100553\\
39.6	-2.62030730620066\\
40	-2.43726037087843\\
40.8	-2.17699558566828\\
41.2	-2.23958338379745\\
41.6	-2.27992781712084\\
42	-2.38769475528871\\
42.4	-2.54816087177146\\
42.8	-2.58641419999176\\
43.2	-2.70450432875988\\
43.6	-2.77446809086243\\
44	-2.72797883516357\\
44.4	-2.7386196295602\\
44.8	-2.61322226559724\\
45.2	-2.45937204109929\\
45.6	-2.383967606436\\
46	-2.37674340072824\\
46.4	-2.25490648631354\\
46.8	-2.24566054461147\\
47.2	-2.35023543655942\\
47.6	-2.21179646670655\\
48	-2.29158695003404\\
48.4	-2.21805505442039\\
48.8	-2.18057417790654\\
49.2	-2.51670934474743\\
49.6	-2.60386511987774\\
50	-2.82079747744604\\
50.4	-2.79704400050343\\
50.8	-2.64656788945548\\
51.2	-2.33953381079048\\
52	-1.53013206159505\\
52.4	-1.37873825059907\\
52.8	-1.39373297042648\\
53.2	-1.51519102278564\\
53.6	-1.61083522895039\\
54	-1.85171721795652\\
54.4	-1.91299532731234\\
54.8	-1.87261119351442\\
55.2	-1.51379062056238\\
55.6	-0.452086859429414\\
56	1.25911719896152\\
56.4	3.73007003924198\\
57.66	6.69287846203341\\
58.06	6.70341232128085\\
58.46	6.75802907520522\\
58.86	6.71448405792545\\
59.26	6.74774366408882\\
59.66	6.72519806341225\\
60.06	6.71458181500793\\
60.46	6.5936187379202\\
60.86	6.1684450116156\\
61.26	5.26076196290556\\
61.66	3.91901783112027\\
62.06	1.9694893160414\\
62.46	-0.132244815368892\\
62.86	-2.00729398621061\\
63.26	-3.32448209382581\\
63.66	-4.46547693406409\\
64.06	-5.30081547285899\\
64.46	-5.60594179431143\\
64.86	-5.76057343513912\\
65.26	-5.69709280321938\\
65.66	-5.77239350438465\\
66.06	-6.10138621434442\\
66.46	-6.38609368527311\\
66.86	-6.55329799681549\\
67.26	-6.60462686842814\\
67.66	-6.46886106386168\\
68.06	-6.12284837626389\\
68.46	-5.8349588909556\\
68.86	-5.61398604960641\\
69.26	-5.23683879845076\\
69.66	-4.91882177560039\\
70.06	-4.52691790074293\\
70.46	-4.19051955817558\\
70.86	-3.90072018122063\\
71.26	-3.44989610637488\\
72.06	-2.73653023200441\\
72.46	-2.32697821554871\\
72.86	-1.99289488022316\\
73.26	-1.68373714633482\\
73.66	-1.32966910438465\\
74.06	-0.955534286818093\\
74.46	-0.560934034703209\\
74.86	-0.27016375702992\\
75.26	-0.0194866547665384\\
75.66	0.166074862389877\\
76.06	0.476193905383269\\
76.46	0.894310445449648\\
76.86	1.26054588243682\\
77.26	1.76591606143771\\
77.66	2.18830296383923\\
78.46	3.81079903325012\\
78.86	4.59861311580008\\
79.26	5.28336860839784\\
79.66	5.79481744231661\\
80.06	6.22304127497245\\
80.46	6.37981076501035\\
80.86	6.35436532214429\\
81.26	6.16988233954906\\
81.66	5.68228964885384\\
82.06	5.17325400614618\\
82.46	4.49141927319226\\
82.86	3.66455486472321\\
83.26	2.98694460377095\\
83.66	2.1776365562332\\
84.06	1.70320916188487\\
84.46	0.857942142718571\\
84.86	0.559556470924974\\
85.26	0.11437978858406\\
85.66	-0.283958284554103\\
86.06	-0.795078842505916\\
86.46	-1.12014515418828\\
86.86	-1.20973557159256\\
87.26	-1.09432639879539\\
87.66	-1.13674326092334\\
88.06	-1.15772015924662\\
88.46	-1.19335624435435\\
88.86	-1.04931139674018\\
89.26	-0.525010868206934\\
89.66	-0.268946462573354\\
90.06	0.175927863834872\\
90.46	0.30345618011782\\
90.86	0.538673861228745\\
91.26	0.652816140656597\\
91.66	0.745891715445737\\
92.46	0.892392122788223\\
92.86	1.0025392217993\\
93.26	1.19385271372964\\
93.66	1.47958087105243\\
94.06	1.55374093062632\\
94.46	1.60766623461892\\
94.86	1.50028306770504\\
95.26	1.42351910523179\\
95.66	1.54485250663861\\
96.06	1.61350956817557\\
96.46	1.59831314432424\\
97.26	1.23953035085604\\
97.66	1.10694517328699\\
98.06	1.07137440186321\\
98.46	0.977005413360843\\
98.86	0.804078470511044\\
99.26	0.596744663160734\\
99.66	0.462499642901662\\
100.06	0.164625474903659\\
100.46	0.0552598755986651\\
100.86	0.0568518732992942\\
101.26	0.0835197732844364\\
101.66	0.181051386166459\\
102.46	0.266571637815446\\
102.86	0.442903536688192\\
103.26	0.732521620145448\\
103.66	0.765082946187889\\
104.06	0.599629457589714\\
};
\end{axis}
\end{tikzpicture}%
		};
		\begin{scope}[x={(myplot.south east)}, y={(myplot.north west)}]
		\node at (0.775,0.89) {$v=1$};
		\node at (0.40,0.89) {$v=-0.8$};
		\draw [dashed] (0.648,0.32) -- (0.648,0.94); 
		\end{scope}
		\end{tikzpicture}
		\label{j1:fig:two_point_turn_b2}
	}
	\caption{The absolute position estimation error (a), the controlled curvature of the tractor (b) and the estimated path-following error states (c)--(f) during the execution of the planned two-point turn maneuver in Figure~\ref{j1:fig:two_point_turn}.}
	\label{j1:fig:two_point_error_states}
\end{figure}
\tikzexternalenable

\subsubsection{Two-point turn}
In this section, the complete path planning and path-following control framework is evaluated in a real-world experiment. 
The G2T with a car-like tractor is operating on dry asphalt on a relatively narrow road at Scania's test facility. 
The scenario setup is shown in Figure~\ref{j1:fig:two_point_turn_gui} and the objective is to change the orientation of the semitrailer with $180\degree$ while at the same time move the vehicle about 40 m longitudinally. 
Similarly to the parking planning problem in Figure~\ref{j1:fig:parking_example}, the precomputed HLUT may underestimate the cost-to-go due to the confined environment. 
Despite this, the lattice planner found an optimal solution ($\gamma=1$) in 0.6 s and the ARA$^*$ search expanded 720 vertices. 
As a comparison, a planning time of only 29 ms was needed for this example to find a motion plan with $\gamma=1.3$, \textit{i.e.}, a solution that is guaranteed to be less than 30 \% worse than the optimal one.

In Figure~\ref{j1:fig:two_point_turn_gui}, the white path illustrates the planned path for the axle of the semitrailer. 
As can be seen, the solution is mainly composed of a $90\degree$-turn in backward motion followed by a $90\degree$-turn in forward motion.
The execution of the planned two-point turn maneuver is visualized in Figure~\ref{j1:fig:two_point_turn},  where the estimated path taken by the axle of the semitrailer $(\hat x_3(\cdot),\hat y_3(\cdot))$ is plotted together with its ground truth path $(x_{3,GT}(\cdot),\hat y_{3,GT}(\cdot))$ measured by the external RTK-GPS.

More detailed plots are provided in Figure~\ref{j1:fig:two_point_error_states}, where the vehicle is changing from backward to forward motion at $t=60$ s. 
In Figure~\ref{j1:fig:two_point_turn_poes_error}, the norm of the position estimation error for the axle of the semitrailer is plotted.  
In this scenario, the maximum position estimation error was \mbox{0.35 m} and the mean absolute error was \mbox{0.21 m}. 
The path-following error states are plotted in Figure~\ref{j1:fig:two_point_turn_z3}--\ref{j1:fig:two_point_turn_b2}. 
From these plots, it can be seen that the estimated lateral control error $\tilde z_3(t)$, which is plotted in Figure~\ref{j1:fig:two_point_turn_z3}, has a maximum absolute error of 0.37 m and a mean absolute error of 0.12 m. 
Except from initial transients, the joint angle errors, $\tilde\beta_3$ and $\tilde\beta_2$, attain their peak values when the vehicle is changing from backward to forward motion at \mbox{$t=60$ s.} 
There are multiple possible explanations to this phenomenon. 

Except from possible estimation errors in the joint angles, one possibility is that lateral dynamical effects arise when the vehicle is exiting the tight $90\degree$-turn in backward motion. 
However, the path-following controller is still able to compensate for these disturbances, as can be seen for \mbox{$t\in[50,\hspace{1pt}80]$ s} in Figure~\ref{j1:fig:two_point_turn_kappa}.

\subsubsection{T-turn}
\begin{figure}[t!]
	\centering
	\setlength\figureheight{0.50\textwidth}
	\setlength\figurewidth{0.50\textwidth}
	\begin{tikzpicture}
	\node[anchor=south west] (myplot) at (0,0) {
		\input{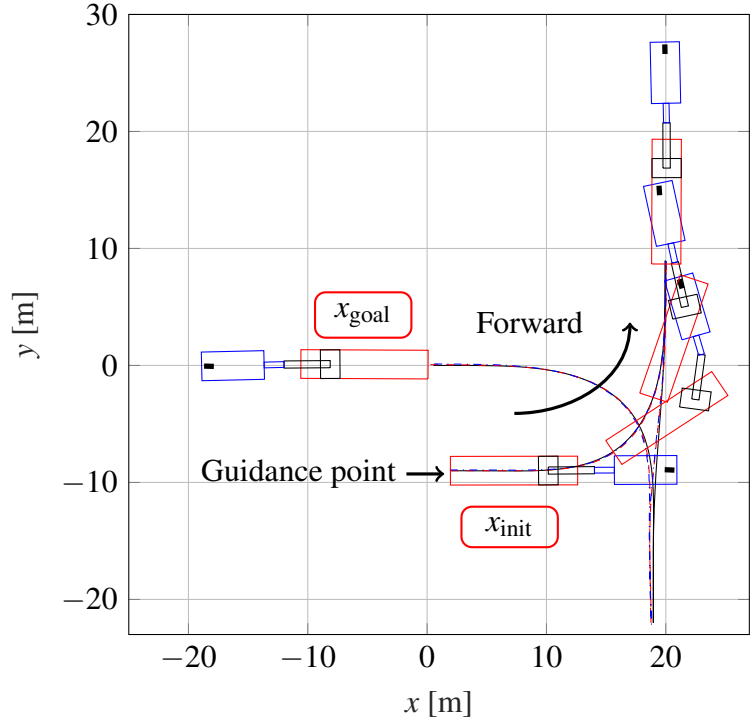}
	};
	\begin{scope}[x={(myplot.south east)}, y={(myplot.north west)}]
	
	%

	\draw[thick,red] (0.61,0.28) 
	node[right,draw=red,rounded corners,
	text width=1cm,align=center,text=black] 
	{$x_{\text{init}}$};
	
	\draw[thick,red] (0.42,0.56) 
	node[right,draw=red,rounded corners,
	text width=1cm,align=center,text=black] 
	{$x_{\text{goal}}$};
	
	\draw[->,very thick] (0.68,0.43) to [out=0,in=-90] (0.83,0.55);  
	\node at (0.70,0.55) {Forward};
	\draw[->,very thick] (0.54,0.35) to [out=0,in=180] (0.589,0.35);  
	\node at (0.4,0.35) {Guidance point};
	\end{scope}
	\end{tikzpicture}
	\caption{Results from real-world experiments while executing the planned T-turn maneuver $(x_{3,r}(\cdot),y_{3,r}(\cdot))$ (black line). The dashed red line is the estimated path taken by the axle of the semitrailer $(\hat x_3(\cdot),\hat y_3(\cdot))$ and the dashed-dotted blue line is the ground truth path $(x_{3,GT}(\cdot),y_{3,GT}(\cdot))$ measured by the external RTK-GPS.}
	\label{j1:fig:T-turn}
\end{figure}

The final real-world experiment is an open area planning problem on the same gravel surface as the execution of the figure-eight nominal path was performed. 
The open area planning problem is shown in Figure~\ref{j1:fig:T-turn}, where the G2T with a car-like tractor is intended to change the orientation of the semitrailer with $180\degree$ together with a small lateral and longitudinal movement. 
In this scenario, the planning time for finding an optimal solution ($\gamma=1$) was only 38 ms and the ARA$^*$ search expanded only 22 vertices. 
The reason why such a small amount of vertex expansions was needed is because the precomputed HLUT perfectly estimates the cost-to-go in free-space scenarios.

\tikzexternaldisable
\tikzexternalenable

\tikzexternaldisable
\begin{figure}[t!]	
	\vspace{-40pt}
	\centering
	\setlength\figureheight{0.19\columnwidth}
	\setlength\figurewidth{0.36\columnwidth} 
	\subfloat[][The norm of the position estimation error of the axle of the semitrailer.]{
		\begin{tikzpicture}
		\node[anchor=south west] (myplot) at (0,0) {
%
%
\begin{tikzpicture}

\begin{axis}[%
width=\figurewidth,
height=\figureheight,
at={(0\figurewidth,0\figureheight)},
scale only axis,
xmin=0,
xmax=145,
xlabel style={font=\color{white!15!black}},
xlabel={Time [s]},
ymin=0,
ymax=1,
ylabel style={font=\color{white!15!black}},
ylabel={$||e(t)||_2$ [m]},
axis background/.style={fill=white},
xmajorgrids,
ymajorgrids
]
\addplot [color=black, forget plot]
  table[row sep=crcr]{%
0	0.0561981201171875\\
0.600006103515625	0.05999755859375\\
1.19999694824219	0.0654296875\\
1.80000305175781	0.069244384765625\\
2.39999389648438	0.0710906982421875\\
3	0.073394775390625\\
3.60000610351563	0.0696563720703125\\
4.80000305175781	0.0699920654296875\\
5.39999389648438	0.0748748779296875\\
6	0.0758209228515625\\
6.60000610351563	0.0799560546875\\
7.19999694824219	0.08880615234375\\
7.80000305175781	0.0943603515625\\
8.39999389648438	0.10052490234375\\
9	0.124160766601563\\
9.60000610351563	0.13873291015625\\
10.1999969482422	0.16552734375\\
10.8000030517578	0.1771240234375\\
11.3999938964844	0.180526733398438\\
12	0.183258056640625\\
13.1999969482422	0.184890747070313\\
13.8000030517578	0.17645263671875\\
14.3999938964844	0.174728393554688\\
15	0.178985595703125\\
15.6000061035156	0.171905517578125\\
16.1999969482422	0.164108276367188\\
16.8000030517578	0.1671142578125\\
17.3999938964844	0.157150268554688\\
18	0.15911865234375\\
19.1999969482422	0.154800415039063\\
19.8000030517578	0.155731201171875\\
20.3999938964844	0.158172607421875\\
21	0.162261962890625\\
21.6000061035156	0.151107788085938\\
22.1999969482422	0.151046752929688\\
22.8000030517578	0.15728759765625\\
24	0.1436767578125\\
24.6000061035156	0.14208984375\\
25.1999969482422	0.1480712890625\\
25.8000030517578	0.15155029296875\\
26.3999938964844	0.158676147460938\\
27	0.156234741210938\\
27.6000061035156	0.160400390625\\
28.1999969482422	0.160964965820313\\
28.8000030517578	0.15460205078125\\
29.3999938964844	0.164215087890625\\
30	0.163192749023438\\
30.6000061035156	0.153732299804688\\
31.1999969482422	0.164520263671875\\
31.8000030517578	0.19903564453125\\
32.4000091552734	0.165939331054688\\
33	0.156936645507813\\
33.6000061035156	0.13909912109375\\
34.1999969482422	0.134719848632813\\
34.8000030517578	0.126846313476563\\
35.4000091552734	0.117691040039063\\
36	0.116424560546875\\
36.6000061035156	0.109085083007813\\
37.1999969482422	0.102493286132813\\
37.8000030517578	0.107711791992188\\
38.4000091552734	0.105026245117188\\
39	0.114547729492188\\
39.6000061035156	0.0991973876953125\\
40.1999969482422	0.112045288085938\\
40.8000030517578	0.101974487304688\\
41.4000091552734	0.100616455078125\\
42	0.107437133789063\\
42.6000061035156	0.11004638671875\\
43.1999969482422	0.103927612304688\\
43.8000030517578	0.098419189453125\\
44.4000091552734	0.099456787109375\\
45	0.105621337890625\\
45.6000061035156	0.0911407470703125\\
46.1999969482422	0.080291748046875\\
46.8000030517578	0.061920166015625\\
47.4000091552734	0.05340576171875\\
48	0.05731201171875\\
48.6000061035156	0.0640411376953125\\
49.1999969482422	0.0594940185546875\\
49.8000030517578	0.0606536865234375\\
50.4000091552734	0.0731658935546875\\
51	0.0771331787109375\\
52.1999969482422	0.08660888671875\\
52.8000030517578	0.09649658203125\\
53.4000091552734	0.0977935791015625\\
54	0.10809326171875\\
54.6000061035156	0.102432250976563\\
55.1999969482422	0.0908203125\\
55.8000030517578	0.0894317626953125\\
56.4000091552734	0.0836944580078125\\
57	0.075714111328125\\
57.6000061035156	0.0707855224609375\\
58.1999969482422	0.0866851806640625\\
58.8000030517578	0.0937957763671875\\
59.4000091552734	0.106643676757813\\
60	0.11077880859375\\
60.6000061035156	0.107818603515625\\
61.1999969482422	0.105422973632813\\
61.8000030517578	0.10552978515625\\
62.4000091552734	0.108154296875\\
63	0.09039306640625\\
63.6000061035156	0.0910491943359375\\
64.1999969482422	0.104248046875\\
64.8000030517578	0.086334228515625\\
65.4000091552734	0.085205078125\\
66	0.08612060546875\\
66.6000061035156	0.08099365234375\\
67.1999969482422	0.0778961181640625\\
67.8000030517578	0.068389892578125\\
68.4000091552734	0.0771942138671875\\
69	0.08184814453125\\
69.6000061035156	0.07220458984375\\
70.1999969482422	0.0697021484375\\
70.8000030517578	0.0765838623046875\\
71.4000091552734	0.0694427490234375\\
72	0.090118408203125\\
72.6000061035156	0.082611083984375\\
73.1999969482422	0.0819854736328125\\
73.8000030517578	0.0952911376953125\\
74.4000091552734	0.092132568359375\\
75	0.09405517578125\\
75.6000061035156	0.089447021484375\\
76.1999969482422	0.0996246337890625\\
76.8000030517578	0.0501251220703125\\
77.4000091552734	0.0826263427734375\\
78	0.0818939208984375\\
78.6000061035156	0.0801849365234375\\
79.1999969482422	0.081146240234375\\
79.8000030517578	0.0812835693359375\\
80.4000091552734	0.0829010009765625\\
81	0.08880615234375\\
81.6000061035156	0.0851898193359375\\
82.1999969482422	0.086029052734375\\
83.4000091552734	0.077484130859375\\
84	0.03802490234375\\
84.6000061035156	0.076324462890625\\
85.1999969482422	0.0706024169921875\\
85.8000030517578	0.071533203125\\
86.4000091552734	0.075775146484375\\
87	0.0875701904296875\\
87.6000061035156	0.0887298583984375\\
88.1999969482422	0.10137939453125\\
88.8000030517578	0.0969696044921875\\
89.4000091552734	0.110244750976563\\
90	0.119720458984375\\
90.6000061035156	0.111709594726563\\
91.1999969482422	0.114852905273438\\
91.8000030517578	0.079254150390625\\
92.4000091552734	0.0982818603515625\\
93	0.108688354492188\\
93.6000061035156	0.0983123779296875\\
94.1999969482422	0.0968780517578125\\
94.8000030517578	0.0963897705078125\\
95.4000091552734	0.098602294921875\\
96	0.11083984375\\
96.6000061035156	0.121826171875\\
97.1999969482422	0.131515502929688\\
97.8000030517578	0.132049560546875\\
98.4000091552734	0.135650634765625\\
99	0.13031005859375\\
99.6000061035156	0.127914428710938\\
100.199996948242	0.130538940429688\\
100.800003051758	0.12677001953125\\
101.400009155273	0.116012573242188\\
102	0.10992431640625\\
102.600006103516	0.119720458984375\\
103.199996948242	0.115325927734375\\
103.800003051758	0.120574951171875\\
104.400009155273	0.101181030273438\\
105	0.0968017578125\\
105.600006103516	0.1002197265625\\
106.199996948242	0.106369018554688\\
106.800003051758	0.1138916015625\\
107.400009155273	0.122390747070313\\
108	0.168075561523438\\
108.600006103516	0.131561279296875\\
109.199996948242	0.134307861328125\\
109.800003051758	0.160781860351563\\
110.400009155273	0.174026489257813\\
111	0.173263549804688\\
111.600006103516	0.179595947265625\\
112.199996948242	0.171310424804688\\
112.800003051758	0.160690307617188\\
114	0.146881103515625\\
114.600006103516	0.149246215820313\\
115.199996948242	0.148513793945313\\
115.800003051758	0.142501831054688\\
116.400009155273	0.152786254882813\\
117	0.147216796875\\
117.600006103516	0.14520263671875\\
118.199996948242	0.14605712890625\\
118.800003051758	0.122207641601563\\
119.400009155273	0.117813110351563\\
120	0.11932373046875\\
120.600006103516	0.128158569335938\\
121.199996948242	0.113006591796875\\
121.800003051758	0.1212158203125\\
122.400009155273	0.121429443359375\\
123	0.139907836914063\\
123.600006103516	0.142608642578125\\
124.199996948242	0.141708374023438\\
124.800003051758	0.136566162109375\\
125.400009155273	0.121810913085938\\
126	0.1214599609375\\
126.600006103516	0.119110107421875\\
127.199996948242	0.1065673828125\\
127.800003051758	0.110870361328125\\
128.399993896484	0.107040405273438\\
129	0.0895538330078125\\
129.600006103516	0.0875244140625\\
130.199996948242	0.060089111328125\\
130.800003051758	0.0566253662109375\\
131.399993896484	0.06024169921875\\
132	0.0496673583984375\\
133.199996948242	0.0532073974609375\\
133.800003051758	0.0571746826171875\\
134.399993896484	0.0643310546875\\
135	0.07757568359375\\
135.600006103516	0.0864410400390625\\
136.199996948242	0.09930419921875\\
136.800003051758	0.1026611328125\\
137.399993896484	0.107635498046875\\
138	0.111587524414063\\
138.600006103516	0.126861572265625\\
139.199996948242	0.130935668945313\\
139.800003051758	0.137542724609375\\
140.399993896484	0.150772094726563\\
141	0.158905029296875\\
141.600006103516	0.161041259765625\\
142.199996948242	0.172760009765625\\
142.800003051758	0.1737060546875\\
143.399993896484	0.175765991210938\\
144	0.171005249023438\\
144.600006103516	0.169509887695313\\
145.199996948242	0.161880493164063\\
145.800003051758	0.165664672851563\\
146.399993896484	0.173782348632813\\
147	0.1746826171875\\
148.199996948242	0.17413330078125\\
148.800003051758	0.166824340820313\\
149.399993896484	0.157989501953125\\
150	0.1514892578125\\
150.600006103516	0.149246215820313\\
151.199996948242	0.144271850585938\\
151.800003051758	0.14166259765625\\
};
\end{axis}
\end{tikzpicture}%
		};
		\begin{scope}[x={(myplot.south east)}, y={(myplot.north west)}]
		
		%
		\node at (0.32,0.85) {$v=1$};
		\node at (0.59,0.85) {$v=-0.8$};
		\node at (0.85,0.85) {$v=1$};
		\draw [dashed] (0.475,0.28) -- (0.475,0.92);
		\draw [dashed] (0.738,0.28) -- (0.738,0.92); 
		\end{scope}
		\end{tikzpicture}
		\label{j1:fig:T_turn_poes_error}
	} 
	~
	\subfloat[][The controlled curvature $\kappa(t)$ (black line) and the nominal feed-forward $\kappa_r(\tilde s(t))$ (red dashed line).]{
		\begin{tikzpicture}
		\node[anchor=south west] (myplot) at (0,0) {
			\input{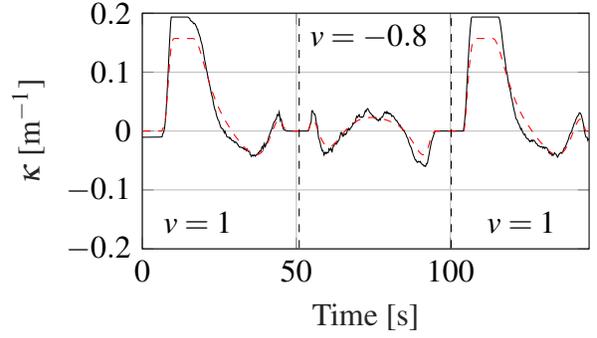}
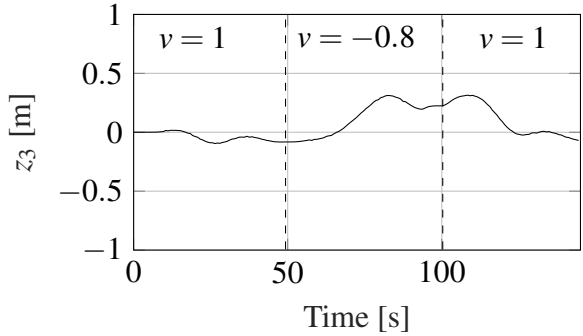
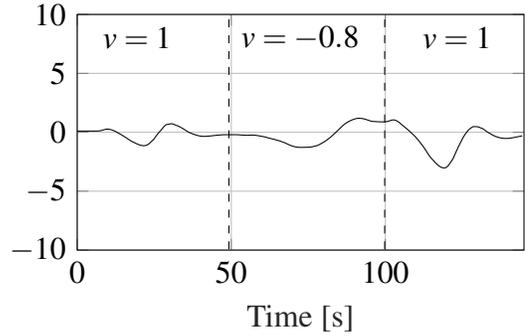
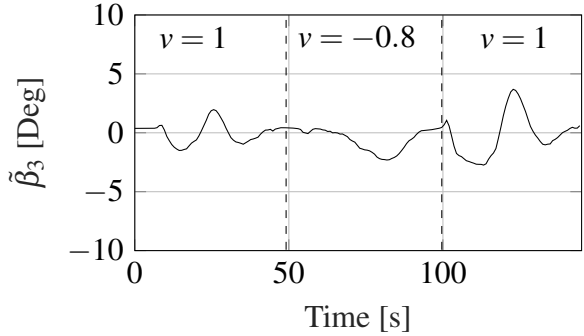
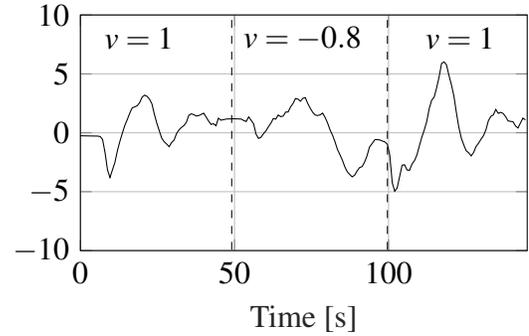
		};
		\begin{scope}[x={(myplot.south east)}, y={(myplot.north west)}]
		
		%
		\node at (0.34,0.35) {$v=1$};
		\node at (0.62,0.85) {$v=-0.8$};
		\node at (0.87,0.35) {$v=1$};
		\draw [dashed] (0.506,0.28) -- (0.506,0.92);
		\draw [dashed] (0.755,0.28) -- (0.755,0.92); 
		\end{scope}
		\end{tikzpicture}
		\label{j1:fig:T_turn_kapp}
	}
	\quad
	\subfloat[][Estimated lateral error for the axle of the semitrailer.]{
		\begin{tikzpicture}
		\node[anchor=south west] (myplot) at (0,0) {
%
%
\begin{tikzpicture}

\begin{axis}[%
width=\figurewidth,
height=\figureheight,
at={(0\figurewidth,0\figureheight)},
scale only axis,
xmin=0,
xmax=145,
xlabel style={font=\color{white!15!black}},
xlabel={Time [s]},
ymin=-1,
ymax=1,
ylabel style={font=\color{white!15!black}},
ylabel={$z_3$ [m]},
axis background/.style={fill=white},
xmajorgrids,
ymajorgrids
]
\addplot [color=black, forget plot]
  table[row sep=crcr]{%
0	0.00014137896766897\\
7.20000000000002	0.000900291220460758\\
8	0.00214886272598847\\
9.59999999999999	0.00555536569476089\\
10.4	0.0147790836122681\\
11.2	0.0122026704734708\\
12	0.0165719180221515\\
13.6	0.0161913035364591\\
15.2	0.0138435945794697\\
16.8	0.00107791535430124\\
17.6	-0.00748001191436742\\
18.4	-0.0113699728615018\\
19.2	-0.0282486947171208\\
20	-0.0361072531924265\\
20.8	-0.0427217924147953\\
21.6	-0.0581666421300895\\
22.4	-0.0650808022919875\\
23.2	-0.0756674731585463\\
24	-0.0839085254248744\\
24.8	-0.0888242851203529\\
25.6	-0.0895146862107481\\
26.4	-0.0957004674706354\\
27.2	-0.0915959610031791\\
28	-0.0936146688303552\\
28.8	-0.0867533025209184\\
29.6	-0.082274475465681\\
30.4	-0.0764399674259266\\
31.2	-0.0653749711378566\\
32	-0.0579133741150883\\
32.8	-0.0548176533414733\\
34.4	-0.0410982955324641\\
36	-0.0416223647051766\\
36.8	-0.0342780119144379\\
37.6	-0.0421235709443408\\
38.4	-0.0382625226937989\\
39.2	-0.0435186186996077\\
40.8	-0.0517760500219993\\
41.6	-0.0593821928339651\\
42.4	-0.0628908792195091\\
43.2	-0.0691430366404404\\
44	-0.071930251587645\\
44.8	-0.0767277824974997\\
45.6	-0.0803188407043649\\
47.4	-0.0827624007243912\\
52.2	-0.0816700938873112\\
53.8	-0.0782852407993744\\
55.4	-0.0727085063599304\\
56.2	-0.069308571822404\\
57	-0.0650690500740723\\
57.8	-0.0691029260691209\\
58.6	-0.0608129293274828\\
59.4	-0.0600407284696018\\
61	-0.0512908252123623\\
61.8	-0.0455438410036209\\
62.6	-0.034639756100006\\
63.4	-0.0302012533084337\\
64.2	-0.0267346916459132\\
65	-0.0129803877176755\\
65.8	-0.00114470788230392\\
66.6	0.00733303945102648\\
67.4	0.0207242734746842\\
68.2	0.0354694814167829\\
69.8	0.0722642553071751\\
72.2	0.122273571840338\\
73.8	0.16304844106125\\
74.6	0.179734105901019\\
75.4	0.200575240230677\\
76.2	0.215974294202198\\
77	0.23352459868417\\
77.8	0.254044588500506\\
78.6	0.269856392592061\\
79.4	0.275806769324106\\
80.2	0.293443218923699\\
81	0.299722051877353\\
81.8	0.309611844458942\\
82.6	0.312415154188784\\
83.4	0.310300538701199\\
84.2	0.307046859559819\\
85	0.297947165861927\\
85.8	0.292342757295273\\
86.6	0.283740958454047\\
87.4	0.26250779250708\\
88.2	0.259384140380433\\
89.8	0.231426422656597\\
90.6	0.218858633538645\\
91.4	0.208229911474632\\
92.2	0.20167607689288\\
93	0.196330184716061\\
93.8	0.196211434291087\\
95.6	0.216655110614397\\
96.4	0.220510074146745\\
97.2	0.223160448729089\\
98	0.224275649424271\\
99.6	0.224146710266041\\
100.4	0.22549685629707\\
101.2	0.233811451732265\\
102	0.246541937390589\\
103.6	0.276173614843231\\
104.4	0.287549396762131\\
105.2	0.297279948126288\\
106	0.304118648214541\\
107.6	0.311208980239314\\
108.4	0.314860259489365\\
109.2	0.30986650001077\\
110	0.313597279736427\\
110.8	0.306134091097675\\
111.6	0.299888868622361\\
113.2	0.276095868624196\\
114	0.257661835969344\\
114.8	0.235618713234288\\
115.6	0.21815893195577\\
116.4	0.187936232584178\\
118	0.143310341223213\\
118.8	0.113594165396563\\
119.6	0.0915414192454023\\
120.4	0.0653109655789308\\
121.2	0.0423879323578547\\
122	0.0235919441053341\\
122.8	0.0106309256512134\\
123.6	-0.00522779605969959\\
124.4	-0.0146617665749602\\
125.2	-0.0186733612249554\\
126	-0.0235376387607289\\
126.8	-0.0226474948256055\\
127.6	-0.019063026498344\\
128.4	-0.0173624181805963\\
129.2	-0.00617822194638507\\
130	-0.00458787387350412\\
130.8	-0.00674259956230117\\
131.6	0.000805498868317045\\
132.4	0.00649702250575501\\
133.2	0.00819811056672393\\
134	0.00274898684756408\\
134.8	0.00597047923741911\\
135.6	-0.00167539090671198\\
137.2	-0.0112455628947998\\
138	-0.0201997263996816\\
138.8	-0.0231906773857418\\
139.6	-0.0336949982626322\\
140.4	-0.0422752382440876\\
141.2	-0.04465717248479\\
142	-0.05313236504918\\
142.8	-0.0593753638243015\\
144.4	-0.0692705363018433\\
};
\end{axis}
\end{tikzpicture}%
		};
		\begin{scope}[x={(myplot.south east)}, y={(myplot.north west)}]
		
		\node at (0.34,0.85) {$v=1$};
		\node at (0.61,0.85) {$v=-0.8$};
		\node at (0.87,0.85) {$v=1$};	
		\draw [dashed] (0.493,0.28) -- (0.493,0.92);
		\draw [dashed] (0.753,0.28) -- (0.753,0.92); 
		\end{scope}
		\end{tikzpicture}
		\label{j1:fig:T_turn_z3}
	}
	~
	\subfloat[][Estimated orientation error of the semitrailer.]{
		\begin{tikzpicture}
		\node[anchor=south west] (myplot) at (0,0) {
%
%
\begin{tikzpicture}

\begin{axis}[%
width=\figurewidth,
height=\figureheight,
at={(0\figurewidth,0\figureheight)},
scale only axis,
xmin=0,
xmax=145,
xlabel style={font=\color{white!15!black}},
xlabel={Time [s]},
ymin=-10,
ymax=10,
ylabel style={font=\color{white!15!black}},
ylabel={$\tilde\theta_3$ [Deg]},
axis background/.style={fill=white},
xmajorgrids,
ymajorgrids
]
\addplot [color=black, forget plot]
  table[row sep=crcr]{%
0	0.102043406396064\\
6.40000000000001	0.106247988139813\\
7.20000000000002	0.116391386016346\\
8	0.155100562662682\\
8.80000000000001	0.221456517460012\\
9.59999999999999	0.269503763320927\\
10.4	0.256438385171066\\
11.2	0.200034936469166\\
12	0.121219942893362\\
12.8	0.0202847346901081\\
14.4	-0.223581560065526\\
16	-0.480432037763251\\
16.8	-0.631015510847249\\
17.6	-0.758138884957816\\
19.2	-0.962453908777547\\
20	-1.05562131759453\\
20.8	-1.11617585378514\\
21.6	-1.14483639475958\\
22.4	-1.11468932714533\\
23.2	-1.006639256292\\
24	-0.800770951405468\\
24.8	-0.614659715745148\\
25.6	-0.32249439861431\\
26.4	-0.05336278722163\\
27.2	0.178151290512915\\
28	0.435251723785797\\
28.8	0.642759647599092\\
29.6	0.672229299185915\\
30.4	0.731145390388548\\
31.2	0.693974433708263\\
32	0.594827834603194\\
32.8	0.532159301503583\\
33.6	0.411873800525854\\
34.4	0.299338163017694\\
35.2	0.157674196618217\\
36	0.0280951127019478\\
37.6	-0.155479798244158\\
39.2	-0.284056361801731\\
40	-0.325057026893944\\
40.8	-0.349167888148799\\
41.6	-0.347557857814962\\
43.2	-0.315246020440185\\
45.6	-0.242844574495024\\
47.4	-0.20964582638436\\
51.4	-0.205356467585602\\
52.2	-0.211906618999649\\
54.6	-0.255167116211794\\
55.4	-0.254642152141059\\
57	-0.238121178518838\\
57.8	-0.256340921573269\\
58.6	-0.283960365627365\\
61	-0.398976295619178\\
61.8	-0.464346408659338\\
64.2	-0.621849290259377\\
65	-0.68724104112539\\
65.8	-0.74017431350191\\
67.4	-0.896456458870347\\
69	-1.09284531496255\\
69.8	-1.16862679741237\\
70.6	-1.22801358469835\\
71.4	-1.26678835517311\\
72.2	-1.28496455577155\\
73.8	-1.28236001254197\\
75.4	-1.26553971121544\\
76.2	-1.23735664315268\\
77.8	-1.13642416302307\\
78.6	-1.04520538583196\\
80.2	-0.806606120751638\\
81	-0.6616095289653\\
81.8	-0.507636787949139\\
82.6	-0.329032597197312\\
84.2	0.0545004639506317\\
85.8	0.467691760609625\\
86.6	0.65204320870248\\
87.4	0.801641960607213\\
88.2	0.92524378523126\\
89	1.0245391543942\\
89.8	1.10684175995408\\
90.6	1.16669129255632\\
91.4	1.18998522583382\\
92.2	1.17289449170488\\
93	1.14078502536469\\
93.8	1.07845271899916\\
95.6	0.924233877729449\\
98	0.885902856107322\\
99.6	0.880192817961216\\
100.4	0.896592102158394\\
101.2	0.951595615544505\\
102	1.03246505531746\\
102.8	1.05292830207082\\
103.6	0.98400486315623\\
104.4	0.853459457549775\\
107.6	0.14861491117955\\
108.4	-0.050344090402433\\
109.2	-0.235476401206171\\
110	-0.461198118338785\\
110.8	-0.705849091106302\\
111.6	-0.972833982046978\\
112.4	-1.25184624914664\\
113.2	-1.55709135185762\\
114	-1.89390473513876\\
114.8	-2.18023843142015\\
115.6	-2.39855774427679\\
116.4	-2.62910821838963\\
117.2	-2.82057635007166\\
118	-2.96096363138528\\
118.8	-3.02506901217066\\
119.6	-3.01158280365448\\
120.4	-2.84226604266786\\
121.2	-2.55182249349681\\
122	-2.22490061151603\\
123.6	-1.28384685934691\\
124.4	-0.874519438720995\\
125.2	-0.456249486202552\\
126	-0.128857455087172\\
126.8	0.141406513624474\\
127.6	0.332360820313795\\
128.4	0.471967459030054\\
129.2	0.479304202776547\\
130	0.465982772574108\\
131.6	0.305880123540817\\
133.2	0.00940716401521513\\
134.8	-0.232374908353592\\
135.6	-0.338131636471388\\
136.4	-0.410760693134534\\
137.2	-0.469715364233792\\
138	-0.505054869721164\\
138.8	-0.525054837549703\\
139.6	-0.52084866832061\\
140.4	-0.502589971843577\\
141.2	-0.471276702789396\\
144.4	-0.305422394144699\\
};
\end{axis}
\end{tikzpicture}%
		};
		\begin{scope}[x={(myplot.south east)}, y={(myplot.north west)}]
		\node at (0.34,0.85) {$v=1$};
		\node at (0.61,0.85) {$v=-0.8$};
		\node at (0.87,0.85) {$v=1$};
		\draw [dashed] (0.492,0.28) -- (0.492,0.92);
		\draw [dashed] (0.75,0.28) -- (0.75,0.92); 
		\end{scope}
		\end{tikzpicture}
		\label{j1:fig:T_turn_theta3error}
	}
	\quad
	\subfloat[][Estimated joint angle error between the semitrailer and the dolly.]{
		\begin{tikzpicture}
		\node[anchor=south west] (myplot) at (0,0) {
%
%
\begin{tikzpicture}

\begin{axis}[%
width=\figurewidth,
height=\figureheight,
at={(0\figurewidth,0\figureheight)},
scale only axis,
xmin=0,
xmax=145,
xlabel style={font=\color{white!15!black}},
xlabel={Time [s]},
ymin=-10,
ymax=10,
ylabel style={font=\color{white!15!black}},
ylabel={$\tilde\beta_3$ [Deg]},
axis background/.style={fill=white},
xmajorgrids,
ymajorgrids
]
\addplot [color=black, forget plot]
  table[row sep=crcr]{%
0	0.373573034507331\\
6.40000000000001	0.383798201220799\\
7.20000000000002	0.43129512326567\\
8	0.61637537388728\\
8.80000000000001	0.632251561643585\\
9.59999999999999	0.240688395831853\\
10.4	-0.310896897486174\\
11.2	-0.668268002910736\\
12	-0.997264181325932\\
12.8	-1.22508727552582\\
13.6	-1.37839396637622\\
14.4	-1.49993889020388\\
15.2	-1.49023340483672\\
16	-1.41702946502494\\
16.8	-1.36367661626213\\
17.6	-1.11642366158597\\
18.4	-0.878579932821737\\
19.2	-0.734147845498143\\
20	-0.605807523776548\\
20.8	-0.267939009670783\\
21.6	0.152354119585027\\
23.2	1.05708191085984\\
24	1.57622047056699\\
24.8	1.87154098824885\\
25.6	1.96358909981558\\
26.4	1.87639554520143\\
27.2	1.60201222536833\\
28	1.24446206760192\\
28.8	0.870274636234399\\
29.6	0.397605001741169\\
30.4	-0.0121455887909576\\
31.2	-0.412048530638629\\
32	-0.625723181457545\\
32.8	-0.775656452549043\\
33.6	-0.823509238121829\\
34.4	-0.890767391539413\\
35.2	-0.970999542646297\\
36.8	-0.718434141538893\\
37.6	-0.576766519920994\\
38.4	-0.527661995461102\\
39.2	-0.457827046457425\\
40	-0.371774373558509\\
40.8	-0.116437649875792\\
41.6	0.0808220353252693\\
42.4	0.226157008290073\\
43.2	0.238106664366853\\
44	0.291884265518661\\
44.8	0.146858138500278\\
45.6	0.313325003055695\\
47.4	0.429367142289863\\
49	0.424716984084284\\
51.4	0.396170795417589\\
53	0.348660266355239\\
53.8	0.281622146414549\\
54.6	0.0382320072245079\\
56.2	-0.0671380181273946\\
57	0.121184097740411\\
57.8	0.296734393520467\\
58.6	0.351406109123303\\
60.2	0.316989426225831\\
61	0.244285530966494\\
61.8	0.220358084303257\\
62.6	0.110958951037276\\
63.4	0.129834599025315\\
64.2	-0.00519778021160278\\
65	0.0127445018324011\\
65.8	-0.0136672327083431\\
66.6	-0.11026149800955\\
67.4	-0.100659966253914\\
69	-0.329492674470913\\
69.8	-0.568381769023716\\
70.6	-0.695499860409853\\
71.4	-0.793296643380216\\
73	-1.3112474438629\\
73.8	-1.37381154977336\\
74.6	-1.44867330456276\\
75.4	-1.47313349264107\\
76.2	-1.5339689203613\\
77	-1.65769737028927\\
77.8	-1.81166776342792\\
78.6	-2.06119890635739\\
79.4	-2.19106428125463\\
80.2	-2.25720633907315\\
81	-2.27318663385344\\
81.8	-2.31017509641683\\
82.6	-2.28775564079649\\
83.4	-2.18039744676176\\
84.2	-2.09631461642772\\
85	-1.96759735364131\\
85.8	-1.76395569042421\\
86.6	-1.50581701503663\\
87.4	-1.23073125293607\\
88.2	-0.916738812332426\\
89	-0.674723438810872\\
89.8	-0.548014198279589\\
90.6	-0.389863542605895\\
91.4	-0.139420756185217\\
92.2	0.0517800342887824\\
93	0.174168863339986\\
93.8	0.193059312294594\\
96.4	0.266472359948636\\
98	0.345522061235897\\
98.8	0.385318291837109\\
99.6	0.450621570409027\\
100.4	0.680805828988497\\
101.2	1.05712118444757\\
102	0.582569726116276\\
102.8	-0.258688505615766\\
103.6	-1.13483432977202\\
104.4	-1.75379023450512\\
105.2	-1.93402291844589\\
106	-2.02053345414302\\
107.6	-2.37074243921214\\
108.4	-2.43347879395577\\
109.2	-2.51147357786181\\
110	-2.5810467385212\\
110.8	-2.66631524700645\\
111.6	-2.67782326198429\\
112.4	-2.68143024219629\\
113.2	-2.74034958841094\\
114	-2.63252672806922\\
114.8	-2.2449465763998\\
115.6	-1.94470362676049\\
116.4	-1.719064378826\\
117.2	-1.33436219139639\\
118	-0.463807997653674\\
118.8	0.459309944559266\\
119.6	1.443468060085\\
120.4	2.26360985486343\\
121.2	2.92558410525825\\
122	3.46414814985175\\
122.8	3.68829617539137\\
123.6	3.59574497770356\\
124.4	3.29301840052733\\
125.2	2.87303625832217\\
126.8	1.80660979490412\\
127.6	1.16379283774901\\
128.4	0.542248024516226\\
129.2	0.141893004520284\\
130	-0.195376147145623\\
130.8	-0.453462182051823\\
131.6	-0.744297186439667\\
132.4	-0.973583702030083\\
133.2	-1.02391013666033\\
134	-0.967886150323523\\
134.8	-0.948503460841209\\
135.6	-0.667705419438704\\
136.4	-0.560233940313282\\
137.2	-0.409223767169152\\
138.8	0.00819414509697936\\
139.6	0.115211324094872\\
140.4	0.21075485167043\\
141.2	0.36472508472653\\
142	0.46516795235047\\
142.8	0.388891894835609\\
143.6	0.395873975673396\\
144.4	0.603295294822459\\
};
\end{axis}
\end{tikzpicture}%
		};
		\begin{scope}[x={(myplot.south east)}, y={(myplot.north west)}]
		\node at (0.34,0.85) {$v=1$};
		\node at (0.61,0.85) {$v=-0.8$};
		\node at (0.87,0.85) {$v=1$};
		\draw [dashed] (0.493,0.28) -- (0.493,0.92);
		\draw [dashed] (0.75,0.28) -- (0.75,0.92); 
		\end{scope}
		\end{tikzpicture}
		\label{j1:fig:T_turn_b3error}
	}
	~
	\subfloat[][Estimated joint angle error between the dolly and the tractor.]{
		\begin{tikzpicture}
		\node[anchor=south west] (myplot) at (0,0) {
%
%
\begin{tikzpicture}

\begin{axis}[%
width=\figurewidth,
height=\figureheight,
at={(0\figurewidth,0\figureheight)},
scale only axis,
xmin=0,
xmax=145,
xlabel style={font=\color{white!15!black}},
xlabel={Time [s]},
ymin=-10,
ymax=10,
ylabel style={font=\color{white!15!black}},
ylabel={$\tilde\beta_2$ [Deg]},
axis background/.style={fill=white},
xmajorgrids,
ymajorgrids
]
\addplot [color=black, forget plot]
  table[row sep=crcr]{%
0	-0.255051429882741\\
5.59999999999999	-0.280870692203735\\
6.40000000000001	-0.31755689221589\\
7.20000000000002	-0.567603180819191\\
8	-1.6901422629507\\
8.80000000000001	-3.1557303175496\\
9.59999999999999	-3.79669981525592\\
10.4	-3.14623635460327\\
11.2	-2.51715373393031\\
12	-1.52810584328907\\
12.8	-0.740165455600334\\
13.6	-0.0424321166617858\\
14.4	0.616872750397079\\
15.2	1.1133285186749\\
16	1.65225893888436\\
16.8	2.1091229286688\\
18.4	2.28816651936376\\
19.2	2.54854293547817\\
20	3.01224767988148\\
20.8	3.19513624019234\\
21.6	3.10269347175762\\
22.4	2.86483042636465\\
23.2	2.466650674877\\
24	1.57475891680633\\
24.8	0.810812610145859\\
25.6	0.165994754409468\\
26.4	-0.462379100734978\\
27.2	-0.766260993737149\\
28	-0.957465629527604\\
28.8	-1.18129348994987\\
29.6	-0.828362727857353\\
30.4	-0.621835629881048\\
31.2	-0.0670020146536103\\
32	0.249183657055454\\
32.8	0.455480076064731\\
33.6	0.615603615871919\\
34.4	0.897242037346615\\
35.2	1.49150979643787\\
36	1.60947932967844\\
36.8	1.53986038540725\\
37.6	1.42431460882668\\
38.4	1.46000885726406\\
40	1.66362159889889\\
40.8	1.33622320450181\\
41.6	0.994722924067844\\
42.4	0.709397107575853\\
43.2	0.768616555227169\\
44	0.599152625008287\\
44.8	1.24187801493983\\
45.6	1.06863912454403\\
47.4	1.16681765105596\\
49.8	1.18971255187554\\
52.2	1.16509847109754\\
53	1.04046598159991\\
53.8	0.954796839831147\\
54.6	1.3477970667721\\
55.4	1.06265502505747\\
56.2	0.763019321924048\\
57	-0.115091902506094\\
57.8	-0.481897860392166\\
58.6	-0.381817846567884\\
59.4	-0.0823223490754685\\
60.2	0.159163578063698\\
61	0.501608985924037\\
61.8	0.808405341989413\\
62.6	1.20795161155871\\
63.4	1.16149159760724\\
64.2	1.51046794697459\\
65	1.36913308363532\\
65.8	1.41503590356092\\
66.6	1.75591100824241\\
67.4	1.76454468751842\\
68.2	2.19384331447316\\
69	2.48558929534155\\
69.8	2.90713555086666\\
70.6	2.8585349129406\\
71.4	2.67907788235743\\
72.2	2.93126063278876\\
73	2.97819491800075\\
73.8	2.47546631466136\\
74.6	2.08287771333963\\
75.4	1.73402548131608\\
76.2	1.50576973732097\\
77	1.46777179414261\\
77.8	1.52001089518384\\
78.6	1.67247715260123\\
79.4	1.30926796966406\\
80.2	0.840918056294868\\
81	0.254065748208149\\
81.8	-0.194335317029555\\
82.6	-0.74605195808698\\
83.4	-1.40465689750857\\
84.2	-1.85100986649061\\
85	-2.31341217919658\\
85.8	-2.884101378221\\
86.6	-3.30645018097357\\
87.4	-3.53904651018624\\
88.2	-3.74236096465341\\
89	-3.5991195180404\\
89.8	-3.23799525034735\\
90.6	-2.95553192344784\\
91.4	-2.87490431506018\\
92.2	-2.50218195926965\\
93	-1.97055695998606\\
93.8	-1.22514159687407\\
95.6	-0.57616738426475\\
96.4	-0.582774204904553\\
97.2	-0.613276442355669\\
98	-0.700631573632364\\
98.8	-0.774748037389429\\
99.6	-0.967649228136679\\
100.4	-1.90215624398635\\
101.2	-4.2366241888468\\
102	-4.9587971652652\\
102.8	-4.68588799235403\\
103.6	-3.71048062981555\\
104.4	-2.75367699502891\\
105.2	-2.71220686115294\\
106	-3.15909503948947\\
106.8	-3.18981488249867\\
107.6	-2.6554507639145\\
108.4	-2.37045813258794\\
109.2	-1.88911250366235\\
110	-1.11472016971828\\
110.8	-0.369867443439176\\
111.6	0.242242157146308\\
112.4	0.963272037374054\\
113.2	2.05307930553815\\
114	2.76583473175154\\
114.8	3.02367565358\\
115.6	3.81114267714275\\
116.4	4.73881726266657\\
117.2	5.87642625006723\\
118	6.01918446079648\\
118.8	5.76900018102762\\
119.6	4.87184714958363\\
120.4	4.07890743694838\\
121.2	3.13639277134882\\
122	1.73443796070893\\
122.8	0.477041275727174\\
123.6	-0.43860756114367\\
124.4	-1.0236367156501\\
125.2	-1.52920134159893\\
126	-1.69659124218765\\
126.8	-1.96214003249446\\
127.6	-1.67889005990165\\
128.4	-1.16031755212194\\
129.2	-0.882911073168316\\
130	-0.545883606145367\\
130.8	-0.249634743025382\\
131.6	0.25076716704541\\
132.4	1.02452730828486\\
133.2	1.39631531554127\\
134	1.51335530272451\\
134.8	1.98047119300972\\
135.6	1.65204901728907\\
136.4	1.73385963578531\\
137.2	1.63800316014778\\
138	1.36782766373022\\
138.8	1.00784916027044\\
140.4	0.932350810506051\\
141.2	0.578402296866358\\
142	0.406242362916032\\
143.6	1.2460844509732\\
144.4	1.0940460408743\\
};
\end{axis}
\end{tikzpicture}%
		};
		\begin{scope}[x={(myplot.south east)}, y={(myplot.north west)}]
		\node at (0.34,0.85) {$v=1$};
		\node at (0.61,0.85) {$v=-0.8$};
		\node at (0.87,0.85) {$v=1$};
		\draw [dashed] (0.493,0.28) -- (0.493,0.92);
		\draw [dashed] (0.75,0.28) -- (0.75,0.92);  
		\end{scope}
		\end{tikzpicture}
		\label{j1:fig:T_turn_b2error}
	}
	\caption{Absolute position error (a), controlled curvature (b) and estimated path-following error (c)--(f), during the execution of the T-point turn maneuver in Figure~\ref{j1:fig:T-turn}.}
	\label{j1:fig:T_turn_error_states}
\end{figure}

Figure~\ref{j1:fig:T-turn} shows the optimal nominal path for the axle of the semitrailer $(x_{3,r}(\cdot),y_{3,r}(\cdot))$, which essentially is composed by two $90\degree$-turns in forward motion together with a parallel maneuver in backward motion. 
In this example, the impact of penalizing complex backward motions is clear, the advanced maneuvers are performed while driving forwards if allowed by the surrounding environment. 
In the same plot, the estimated path taken by the axle of the semitrailer $(\hat x_{3}(\cdot),\hat y_{3}(\cdot))$ and its ground truth path $(x_{3,GT}(\cdot),y_{3,GT}(\cdot))$ obtained from the external RTK-GPS are presented.

More detailed plots are provided in Figure~\ref{j1:fig:T_turn_error_states}, where the vehicle is changing from forward to backward motion at $t=50$ s and from backward to forward motion at $t=100$ s. 
In Figure~\ref{j1:fig:T_turn_poes_error}, the norm of the position estimation error for the axle of the semitrailer is plotted. 
In this experiment, the maximum estimation error was $0.20$ m and the mean absolute error was $0.12$ m. The path-following error states are presented in Figure~\ref{j1:fig:T_turn_z3}--\ref{j1:fig:T_turn_b2error}. In Figure~\ref{j1:fig:T_turn_z3}, the estimated lateral control error $\tilde z_3$ is plotted, where the maximum absolute error was $0.31$ m and the mean absolute error was $0.11$ m. In this experiment, both joint angle errors, $\tilde\beta_2$ and $\tilde \beta_3$, as well as the orientation error of the semitrailer $\tilde\theta_3$, lie within $\pm 5\degree$ for the majority of the path execution. The controlled curvature $\kappa$ of the car-like tractor is plotted in Figure~\ref{j1:fig:T_turn_kapp}. Similar to the two-point turn experiment, it can be seen that for large nominal curvature values $\kappa_r$, the feedback part in the path-following controller is compensating for lateral dynamical effects that are not captured by the kinematic vehicle model. 

\subsection{Discussion of lessons learned}
The proposed path planning and path-following control framework has been successfully deployed on a full-scale test platform. 
Since the full system is built upon several modules, an important key to fast deployment was to separately test and evaluate each module in simulations. 
By performing extensive simulations during realistic conditions, the functionality of each module as well as the communication between them could be verified before real-world experiments was performed. 

As illustrated in the real-world experiments, the performance of the system in terms of path-following capability is highly dependent on accurate estimates of the vehicle’s state. The tuning and calibration of the nonlinear observer was also the most time-consuming part of the process when the step from simulations to real-world experiments was taken. The main difficulty was to verify that the nonlinear observer was capable of tracking the true trajectory of the position and orientation of the axle of the semitrailer as well as the two joint angles, despite that their true state trajectories were partially or completely unknown. 
To resolve this, data was collected from manual tests with the vehicle. This data was then used offline to tune the covariance matrices in the EKF and to calibrate the position and orientation of the rear view LIDAR sensor. For the calibration of the LIDAR sensor, an accurately calibrated yaw angle was found to be very important. 

The deployment of the hybrid path-following controller and the lattice planner only required minor tuning compared to simulations. For the design of the path-following controller, the penalty for the lateral path-following error $\tilde z_3$ was found to be the most important tuning parameter which had the largest effects on the region of attraction for the closed-loop system. However, since the lattice planner is planning from the vehicle's current state, the initial error in $\tilde z_3$ will be small and a rather aggressive tuning of the path-following controller was possible.

\section{Conclusions and future work}
\label{j1:sec:conclusions}
A path planning and path-following control framework for a G2T with a car-like tractor is presented. 
The framework is targeting low-speed maneuvers in unstructured environments and has been successfully deployed on a full-scale test vehicle. 
A lattice-based path planner is used to generate kinematically feasible and optimal nominal paths in all vehicle states and controls, and the ARA$^*$ graph-search algorithm is used during online planning. 
To follow the planned path, a hybrid path-following controller is developed to stabilize the path-following error states of the vehicle and a nonlinear observer is proposed that is only utilizing information from sensors that are mounted on the car-like tractor, which makes the solution compatible with basically all of today's commercially available semitrailers that have a rectangular body. 
The framework is first evaluated in simulations and then in three different real-world experiments and results in terms of closed-loop performance and real-time planning capabilities are presented. 
In the experiments, the system shows that it is able to consistently solve challenging planning problems and that it is able to execute the resulting motion plans, despite no sensors on the dolly or semitrailer.

A drawback with the lattice-based path planning framework is the need of manually selecting the connectivity in the state lattice. Even though this procedure is done offline, it is both nontrivial and time-consuming. 
Future work includes automating this procedure to make the algorithm more user-friendly and compatible with different vehicle parameters. 
Moreover, the discretization of the vehicle's state-space, restricts the set of possible initial states the lattice planner can plan from and desired goal states that can be reached. 
As mentioned in the text, this is a general problem with sampling-based motion planning algorithms, which could for example be alleviated by the use of numerical optimal control as a post-processing step~\cite{lavalle2006planning,CirilloIROS2014,oliveira2018combining,andreasson2015fastsmoothing}. 
Thus, future work includes exploiting the structure of the path planning problem and develop an efficient and numerically stable online smoothing framework by, $e.g.$, the use of numerical optimal control as a backbone.

The main benefits of the proposed hybrid path-following controller include that it is computationally efficient and that it is tailored to follow nominal paths composed of full state and control information. 
A limitation of the proposed controller is that it does not explicitly handle constraints on states and controls, but since the lattice planner plans from the vehicle current location and provides a nominal path in all states and controls which satisfies these constraints, they can be neglected locally around the nominal path. 
Regarding scalability, as long as a nominal path is provided, it is possible to generalize the proposed path-following controller to handle more complex paths and tractor-trailer system with additional trailers. 
The region of attraction for the closed-loop system will however become smaller with increasing path complexity as well as increasing number of trailers.

At some parts of the figure-eight path-following experiments, the magnitude of the estimation error for the position of the axle of the semitrailer had a size which potentially could cause problems in narrow environments. Hence, reasonable future work also includes exploring alternative onboard sensors as well as using external sensors that can be placed at strategic locations where high-accuracy path tracking is critical, $e.g.$, when reversing to a loading bay.  

\section*{Appendix A}
\begin{lemma}
	\label{j1:L1}
	Denote $z(s),$ $s\in[0,s_G]$, as the solution to~\eqref{j1:driftless_system} that satisfies $|\alpha(\cdot)|\leq\alpha_{\text{max}}<\pi/2$, when the control signal $u_p(s)\in\mathbb U_p$, $s\in[0,s_G]$ is applied from the initial state $z(0)$ which ends at the final state $z(s_G)$. Moreover, denote $\bar z(\bar s)$, $\bar s\in[0,s_G]$ as the distance-reversed solution to~\eqref{j1:driftless_system} when the distance-reversed control signal 
	\begin{align}\label{j1:eq:reversed_controls}
	\bar u_p(\bar s) = \begin{bmatrix}
	-v(s_G-\bar s) & u_\omega(s_G-\bar s)
	\end{bmatrix}^T,\quad \bar s \in[0,s_G]
	\end{align}
	is applied from the initial state $\bar z(0) = \begin{bmatrix}x(s_G)^T & \alpha(s_G) & -\omega(s_G)\end{bmatrix}^T$. Then, $z(s),$ $s\in[0,s_G]$ and $\bar z(\bar s)$, $\bar s\in[0,s_G]$ are unique and they are related according to
	\begin{align}\label{j1:eq:reversed_states}
	\bar{z}(\bar s) = \begin{bmatrix} x(s_G-\bar s)^T & \alpha(s_G-\bar s) & -\omega(s_G-\bar s)\end{bmatrix}^T,\quad \bar s\in[0,s_G].
	\end{align} 
	In particular, the final state is $\bar z(s_G)=\begin{bmatrix}x(0)^T & \alpha(0) & -\omega(0)\end{bmatrix}^T$.
\end{lemma}
\begin{proof} 
	Given a piecewise continuous $u_p(s)=\begin{bmatrix}
	v(s) & u_{\omega}(s)
	\end{bmatrix}^T\in\mathbb U_p$, $s\in[0,s_G]$, define 
	$$\tilde f_z (s,z)\triangleq f_z(z,u_p(s))= \begin{bmatrix}
	v(s)f(x,\tan\alpha/L_1) \\ \omega \\ u_\omega(s) \end{bmatrix}.
	$$
	Direct calculations verify that $f(x,\tan\alpha/L_1)$ in~\eqref{j1:eq:model_global_coord} is continuous and continuously differentiable with respect to $z$ for all $z\in\mathbb Z_o\in \{z\in\mathbb R^7 \mid |\alpha| < \pi/2\}$. This is true since $f(x,\tan\alpha/L_1)$ is composed of sums and products of trigonometric functions which are continuous and continuously differentiable with respect to $z$ for all $z\in \mathbb Z_o$.
	Furthermore, $\tilde f_z (s,z)$ is piecewise continuous in $s$ since $f_z(z,u_p(s))$ is continuous in $u_p$ for all $z\in \mathbb Z_o$. Therefore, on any interval $[a,b]\subset [0,s_G]$ where $u_p(\cdot)$ is continuous, $\tilde f_z (s,z)$ and $[\partial\tilde f_z (s,z)/\partial z]$ are continuous on $[a,b]\times Z_o$. Then, from Lemma 3.2 in~\cite{khalil}, the vector field $\tilde f_z (s,z)$ is piecewise continuous in $s$ and locally Lipschitz in $z$, for all $s\in[0,s_G]$ and all $z\in \mathbb Z_o$. Define $\mathbb Z_c = \{z\in\mathbb R^7\mid|\alpha| \leq \alpha_{\text{max}}\}$ which is a compact subset of $\mathbb Z_o$. 
	Then, from Theorem 3.3 in~\cite{khalil}, every solution $z(s)$, $s\in[0,s_G]$ that lies entirely in $\mathbb Z_c$ is unique for all $s\in[0,s_G]$.   
	
	Now, let $z(s),$ $s\in[0,s_G]$, be the unique solution to~\eqref{j1:driftless_system} assumed to lie entirely in $\mathbb Z_c$, when the control signal $u_p(s)\in\mathbb U_p$, $s\in[0,s_G]$ is applied from the initial state $z(0)$ which ends at the final state $z(s_G)$. Introduce $(\bar z(\bar s),\bar u_p(\bar s)),$ $\bar s\in[0,s_G]$ with
	\begin{subequations}
		\label{j1:eq:sym:proof}
		\begin{align}
		\bar{z}(\bar s) &= \begin{bmatrix} x(s_G-\bar s)^T &  \alpha(s_G-\bar s) & -\omega(s_G-\bar s) \end{bmatrix}^T, \quad \bar s\in[0, s_G],\\
		\bar u_p(\bar s) &= \begin{bmatrix}-v(s_G-\bar s) & u_\omega(s_G-\bar s)\end{bmatrix}^T,\hspace{62pt} \bar s\in[0, s_G].
		\end{align}
	\end{subequations}
	Since
	\begin{align*}
	\frac{\text d}{\text{d}\bar s}\bar z(\bar s) &= \frac{\text d}{\text{d}\bar s} 
	\begin{bmatrix}
	x(s_G-\bar s) \\ \alpha(s_G-\bar s) \\ -\omega(s_G-\bar s) 
	\end{bmatrix} = 
	\{s = s_G- \bar s\} = 
	\frac{\text d}{\text{d}s}
	\left.\begin{bmatrix}
	x(s) \\ \alpha(s) \\ -\omega(s) 
	\end{bmatrix}\right|_{s=s_G-\bar s}\underbrace{\frac{\text{d}s}{\text{d}\bar s}}_{=-1}= \nonumber \\
	&=\left.\begin{bmatrix}
	-v(s)f(x(s),\tan\alpha(s)/L_1) \\ -\omega(s) \\ u_\omega(s) 
	\end{bmatrix}\right|_{s=s_G-\bar s} 
	= 
	\begin{bmatrix}
	\bar v(\bar s)f(\bar x(\bar s),\tan\bar \alpha(\bar s)/L_1) \\ \bar \omega(\bar s) \\ \bar u_\omega(\bar s) 
	\end{bmatrix} 
	=\nonumber\\&=f_z(\bar z(\bar s),\bar u_p(\bar s)), \hspace{4pt}\bar s\in[0,s_G],
	\end{align*}
	\eqref{j1:eq:sym:proof} is also a solution to~\eqref{j1:driftless_system} from the initial state $\bar z(0)=\begin{bmatrix} x(s_G)^T & \alpha(s_G) & -\omega(s_G)\end{bmatrix}^T$. Finally, since the solution $\bar z(\bar s)$, $\bar s\in[0,s_G]$ also lies entirely in $\mathbb Z_c$, this solution is also unique.  
	\subsection*{Proof of Theorem~\ref{j1:T-optimal-symmetry}}
	Let $(z(s),u_p(s)),$ $s\in[0,s_G]$ denote a feasible solution to the optimal path planning problem~\eqref{j1:eq:MotionPlanningOCP} with objective functional value $J$. 
	Now, consider the reverse optimal path planning problem
	\begin{subequations}
		\label{j1:eq:revMotionPlanningOCP_appendix}
		\begin{align} 
		\minimize_{\bar u_{p}(\cdot), \hspace{0.5ex}\bar s_{G} }\hspace{3.7ex}
		& \bar J = \int_{0}^{\bar s_{G}}L(\bar x(\bar s),\bar \alpha(\bar s), \bar \omega(\bar s), \bar u_\omega(\bar s))\,\text{d}\bar s	\label{j1:eq:revMotionPlanningOCP_obj_appendix}\\
		\subjectto\hspace{3ex}
		& \frac{\text d\bar z}{\text d\bar s} = f_z(\bar z(\bar s),\bar u_p(\bar s)), \label{j1:eq:revMotionPlanningOCP_syseq_appendix} \\
		& \bar z(0) = z_G, \quad \bar z(\bar s_{G}) = z_I, \label{j1:eq:revMotionPlanningOCP_initfinal_appendix} \\ 
		& \bar z(\bar s) \in \mathbb Z_{\text{free}}, \quad
		\bar u_{p}(\bar s) \in {\mathbb  U}_p. \label{j1:eq:revMotionPlanningOCP_constraints_appendix} 
		\end{align}
	\end{subequations}
	Then, using the invertible transformations~\eqref{j1:eq:reversed_controls}--\eqref{j1:eq:reversed_states}: 
	\begin{subequations}
		\label{j1:eq:invertible_transformation}
		\begin{align}
		\bar{z}(\bar s) &= \begin{bmatrix} x(s_G-\bar s)^T &  \alpha(s_G-\bar s) & -\omega(s_G-\bar s) \end{bmatrix}^T, \quad \bar s\in[0, s_G],\\
		\bar u_p(\bar s) &= \begin{bmatrix}-v(s_G-\bar s) & u_\omega(s_G-\bar s)\end{bmatrix}^T,\hspace{62pt} \bar s\in[0, s_G]
		\end{align}
	\end{subequations}
	and $\bar s_G=s_G$, the reverse optimal path planning problem~\eqref{j1:eq:revMotionPlanningOCP_appendix} becomes
	\begin{subequations}
		\label{j1:eq:revMotionPlanningOCP_1}
		\begin{align} 
		\minimize_{u_{p}(\cdot), \hspace{0.5ex}s_{G} }\hspace{3.7ex}
		& \bar J = \int_{0}^{s_{G}}L(x(s_G-\bar s),\alpha(s_G-\bar s), -\omega(s_G-\bar s) , u_\omega(s_G-\bar s))\,\text{d}\bar s	\label{j1:eq:revMotionPlanningOCP_obj_1}\\
		\subjectto\hspace{3ex}
		& \frac{\text d}{\text{d}\bar s} 
		\begin{bmatrix}
		x(s_G-\bar s) \\ \alpha(s_G-\bar s) \\ -\omega(s_G-\bar s) 
		\end{bmatrix} = \begin{bmatrix}
		-v(s_G-\bar s)f(x(s_G-\bar s),\tan\alpha(s_G-\bar s)/L_1) \\ -\omega(s_G-\bar s) \\ u_\omega(s_G-\bar s) 
		\end{bmatrix}, \label{j1:eq:revMotionPlanningOCP_syseq_1} \\
		& \begin{bmatrix} x(s_G)^T &  \alpha(s_G) & -\omega(s_G) \end{bmatrix}^T = z_G,\\ &\begin{bmatrix} x(0)^T &  \alpha(0) & -\omega(0) \end{bmatrix}^T = z_I, \label{j1:eq:revMotionPlanningOCP_initfinal_1} \\ 
		& \begin{bmatrix} x(s_G- \bar s)^T &  \alpha(s_G- \bar s) & -\omega(s_G-\bar s) \end{bmatrix}^T \in \mathbb Z_{\text{free}}, \\
		&\begin{bmatrix}-v(s_G-\bar s) & u_\omega(s_G-\bar s)\end{bmatrix}^T \in {\mathbb  U}_p. \label{j1:eq:revMotionPlanningOCP_constraints_1} 
		\end{align}
	\end{subequations} 
	Let $s=s_G-\bar s$, $s\in[0,s_G]$. It then follows from Lemma~\ref{j1:L1} that~\eqref{j1:eq:revMotionPlanningOCP_syseq_1} simplifies to $\frac{\mathrm{d}z}{\mathrm{d}s}=f_z(z(s),u_p(s))$. From Assumption~\ref{j1:A-optimal-symmetry} it follows that
	\begin{align}
	\bar J &=\int_{0}^{s_{G}}L(x(s_G-\bar s),\alpha(s_G-\bar s), -\omega(s_G-\bar s), u_\omega(s_G-\bar s))\,\text{d}\bar s=\{s = s_G - \bar s\} \nonumber  \\
	&= -\int_{s_G}^{0}L(x(s),\alpha(s), -\omega(s), u_\omega(s))\,\text{d}s \nonumber\\
	&= \int_{0}^{s_G}L(x(s),\alpha(s), -\omega(s), u_\omega(s))\,\text{d}s = \{L(x,\alpha, -\omega, u_\omega)=L(x,\alpha, \omega, u_\omega)\} \nonumber  \\
	&= \int_{0}^{s_G}L(x(s),\alpha(s), \omega(s), u_\omega(s))\,\text{d}s = J. \label{j1:proof:obj_1}
	\end{align}
	Hence, the problem in~\eqref{j1:eq:revMotionPlanningOCP_1} can equivalently be written as
	\begin{subequations}
		\label{j1:eq:revMotionPlanningOCP_2}
		\begin{align} 
		\minimize_{u_{p}(\cdot), \hspace{0.5ex}s_{G} }\hspace{3.7ex}
		& J = \int_{0}^{s_{G}}L(x(s),\alpha(s), \omega(s) , u_\omega(s))\,\text{d} s	\label{j1:eq:revMotionPlanningOCP_obj_2}\\
		\subjectto\hspace{3ex}
		& \frac{\text d z}{\text d s} = f_z(z(s),u_p(s)), \label{j1:eq:revMotionPlanningOCP_syseq_2} \\
		& \begin{bmatrix} x(s_G)^T &  \alpha(s_G) & -\omega(s_G) \end{bmatrix}^T = z_G,\\ &\begin{bmatrix} x(0)^T &  \alpha(0) & -\omega(0) \end{bmatrix}^T = z_I, \label{j1:eq:revMotionPlanningOCP_initfinal_2} \\ 
		& \begin{bmatrix} x(s)^T &  \alpha(s) & -\omega(s) \end{bmatrix}^T \in \mathbb Z_{\text{free}}, \label{j1:eq:revMotionPlanningOCP_constraints_state_2} \\
		&\begin{bmatrix}-v(s) & u_\omega(s)\end{bmatrix}^T \in {\mathbb  U}_p. \label{j1:eq:revMotionPlanningOCP_constraints_2} 
		\end{align}
	\end{subequations} 
	From the symmetry of the set $\mathbb U_p=\{-1,1\}\times[-u_{\omega,\text{max}}, u_{\omega,\text{max}}]$,~\eqref{j1:eq:revMotionPlanningOCP_constraints_2} is equivalent to $u_p(s)\in {\mathbb  U}_p$. From Assumption~\ref{j1:A-optimal-symmetry2}, \eqref{j1:eq:revMotionPlanningOCP_constraints_state_2} is equivalent to $z(s)=\begin{bmatrix} x(s)^T & \alpha(s) & \omega(s)\end{bmatrix}^T\in\mathbb Z_{\text{free}}$. Moreover, since $z_I = \begin{bmatrix} x_I^T & \alpha_I & 0 \end{bmatrix}^T$ and $z_G= \begin{bmatrix} x_G^T & \alpha_G & 0 \end{bmatrix}^T$, the problem in~\eqref{j1:eq:revMotionPlanningOCP_2} can equivalently be written as 
	\begin{subequations}
		\label{j1:eq:revMotionPlanningOCP_3}
		\begin{align} 
		\minimize_{u_{p}(\cdot), \hspace{0.5ex}s_{G} }\hspace{3.7ex}
		& J = \int_{0}^{s_{G}}L(x(s),\alpha(s), \omega(s) , u_\omega(s))\,\text{d} s	\label{j1:eq:revMotionPlanningOCP_obj_3}\\
		\subjectto\hspace{3ex}
		& \frac{\text dz}{\text ds} = f_z(z(s),u_p(s)), \label{j1:eq:revMotionPlanningOCP_syseq_3} \\
		& z(0) = z_I, \quad z(s_G) = z_G,  \label{j1:eq:revMotionPlanningOCP_initfinal_3}\\
		& z(s) \in \mathbb Z_{\text{free}}, \\
		&u_p(s) \in {\mathbb  U}_p, \label{j1:eq:revMotionPlanningOCP_constraints_3} 
		\end{align}
	\end{subequations}
	which is identical to the optimal path planning problem in~\eqref{j1:eq:MotionPlanningOCP}. Hence, the OCPs in~\eqref{j1:eq:MotionPlanningOCP} and~\eqref{j1:eq:revMotionPlanningOCP} are equivalent~\cite{boyd2004convex} and the invertible transformation relating the solutions to the two equivalent problems is given by~\eqref{j1:eq:invertible_transformation} and $\bar s_G=s_G$. Hence, if an optimal solution to one of the problems is known, an optimal solution to the other one can immediately be derived using~\eqref{j1:eq:invertible_transformation} and $\bar s_G=s_G$. Or more practically, given an optimal solution in one direction, an optimal solution in the other direction can be trivially found.
\end{proof}

\subsection*{Derivation of the path-following error model}
In this section, the details regarding the derivation of the model in~\eqref{j1:eq:model_s} are presented. First, the nominal path in $\eqref{j1:eq:tray:semitrailer}$ gives:
\begin{subequations}
	\begin{align}
	\frac{\text d\theta_{3,r} }{\text d\tilde s} &= v_r \kappa_{3,r},  \quad \tilde s \in[0,\tilde s_G], \label{j1:eq:s_a1}\\
	\frac{\text d\beta_{3,r}}{\text d\tilde s} &= v_r\left(\frac{\sin\beta_{2,r} - M_1\cos\beta_{2,r}\kappa_r}{L_2 \cos \beta_{3,r} C_1(\beta_{2,r},\kappa_r)} - \kappa_{3,r}\right),  \quad \tilde s \in[0,\tilde s_G], \label{j1:eq:s_a2}\\
	\frac{\text d\beta_{2,r}}{\text d\tilde s} &= v_r \left(\frac{\kappa_r - \frac{\sin \beta_{2,r}}{L_2} + \frac{M_1}{L_2}\cos\beta_{2,r}\kappa_r}{\cos \beta_{3,r} C_1(\beta_{2,r},\kappa_r)}\right),  \quad \tilde s \in[0,\tilde s_G]. \label{j1:eq:s_a3}
	\end{align}
	\label{j1:eq:model:S_a}
\end{subequations}
Moreover, the models of $\theta_3$, $\beta_3$ and $\beta_2$ in~\eqref{j1:eq:model_global_coord} can equivalently be represented as
\begin{subequations}
	\label{j1:eq:model:S_b}
	\begin{align} 
	\dot{\theta}_3 &= v_3 \frac{\tan \beta_3 }{L_3}, \label{j1:eq:s_a4}\\
	\dot{\beta}_3 &= v_3\left(\frac{\sin\beta_2 - M_1\cos\beta_2\kappa}{L_2 \cos \beta_3 C_1(\beta_2,\kappa)} - \frac{\tan\beta_3}{L_3}\right), \label{j1:eq:s_a5}\\
	\dot{\beta}_2 &= v_3 \left( \frac{\kappa - \frac{\sin \beta_2}{L_2} + \frac{M_1}{L_2}\cos \beta_2\kappa}{\cos \beta_3 C_1(\beta_2,\kappa)}\right), \label{j1:eq:s_a6}
	\end{align}
\end{subequations}
where $v$ has been replaced with $v_3$ using~\eqref{j1:relation_v_v3}. Now, since $\tilde\theta_3(t)=\theta_3(t)-\theta_{3,r}(\tilde s(t))$, the chain rule together with the equation for $\dot{\tilde s}$ in~\eqref{j1:eq:model_s1} yields
\begin{align} 
\dot{\tilde\theta}_3(t) &= \dot\theta_3 - \dot {\tilde s} \frac{\text d}{\text d\tilde s}\theta_{3,r}(\tilde s)  \nonumber \\ &= v_3 \left( \frac{\tan(\tilde{\beta}_3+\beta_{3,r})}{L_3} - \frac{\kappa_{3,r}\cos \tilde \theta_3}{1-\kappa_{3,r}\tilde z_3} \right) = v_3f_{\tilde\theta_3}(\tilde s,\tilde x_e,\tilde \kappa), \quad t\in\Pi(0,\tilde s_G).
\label{j1:eq:model:S_c}
\end{align}
In analogy, taking the time-derivative of $\tilde\beta_3(t)=\beta_3(t)-\beta_{3,r}(\tilde s(t))$ and applying the chain rule gives	
\begin{align}
\dot{\tilde \beta}_3 &= \dot \beta_3 + \dot{\tilde s}\frac{\text d}{\text d\tilde s}\beta_{3,r}(\tilde s) \nonumber\\
&=v_3\left(\frac{\sin(\tilde \beta_2+\beta_{2,r})-M_1\cos(\tilde \beta_2+\beta_{2,r}) (\tilde \kappa+ \kappa_r)}{L_2\cos(\tilde \beta_3+\beta_{3,r}) C_1(\tilde \beta_2+\beta_{2,r}, \tilde \kappa+ \kappa_r)} - \frac{\tan(\tilde \beta_3+\beta_{3,r})}{L_3} \nonumber \right. \\
&\left. -\frac{\cos{\tilde{\theta}_3}}{1-\kappa_{3,r}\tilde z_3}\left(\frac{\sin\beta_{2,r} -M_1 \cos\beta_{2,r}\kappa_r}{L_2\cos\beta_{3,r} C_1(\beta_{2,r},\kappa_r)}-\kappa_{3,r}\right)\right)=v_3f_{\tilde\beta_3}(\tilde s,\tilde x_e,\tilde \kappa), \quad t\in\Pi(0,\tilde s_G).
\label{j1:eq:model:S_d}
\end{align}
Finally, taking the time-derivative of $\tilde\beta_2(t)=\beta_2(t)-\beta_{2,r}(\tilde s(t))$ and applying the chain rule again yields
\begin{align}
\dot{\tilde \beta}_2 &=\dot \beta_2 + \dot {\tilde s}\frac{\text d}{\text d\tilde s}\beta_{2,r}(\tilde s)  \nonumber \\=&v_3\left( \left( \frac{\tilde \kappa+ \kappa_r - \frac{\sin(\tilde \beta_2+\beta_{2,r})}{L_2} + \frac{M_1}{L_2}\cos(\tilde \beta_2+\beta_{2,r})(\tilde \kappa+ \kappa_r)}{\cos(\tilde \beta_3+\beta_{3,r}) C_1(\tilde \beta_2+\beta_{2,r}, \tilde \kappa+ \kappa_r)}\right) \nonumber \right. \\
&\left. -\frac{\cos{\tilde{\theta}_3}}{1-\kappa_{3,r}\tilde z_3}\left( \frac{\kappa_r - \frac{\sin \beta_{2,r}}{L_2} + \frac{M_1}{L_2}\cos \beta_{2,r}\kappa_r}{\cos \beta_{3,r} C_1(\beta_{2,r}, \kappa_r)}\right)\right)=v_3f_{\tilde\beta_2}(\tilde s,\tilde x_e,\tilde \kappa), \quad t\in\Pi(0,\tilde s_G),
\label{j1:eq:model:S_e}
\end{align}
which finalizes the derivation. Moreover, by inserting $(\tilde x_e,\tilde\kappa) = (0,0)$ in~\eqref{j1:eq:model:S_c}--\eqref{j1:eq:model:S_e} yield $\dot{\tilde\theta}_3=\dot{\tilde \beta}_3=\dot{\tilde \beta}_2=0$, \mbox{$\forall t\in\Pi(0,\tilde s_G)$}, $i.e.$, the origin is an equilibrium point. Finally, from~\eqref{j1:relation_v_v3}, we have that $v_3 = vg_v(\beta_2,\beta_3,\kappa)$ and the models in~\eqref{j1:eq:model:S_c}--\eqref{j1:eq:model:S_e} can in a compact form also be represented as
\begin{subequations}
	\label{j1:eq:model:S_f}
	\begin{align}
	\dot{\tilde\theta}_3 &= vg_v(\tilde\beta_2+\beta_{2,r},\tilde\beta_3+\beta_{3,r},\tilde\kappa+\kappa_r)f_{\tilde\theta_3}(\tilde s,\tilde x_e,\tilde \kappa), \quad t\in\Pi(0,\tilde s_G), \\
	\dot{\tilde \beta}_3 &= vg_v(\tilde\beta_2+\beta_{2,r},\tilde\beta_3+\beta_{3,r},\tilde\kappa+\kappa_r)f_{\tilde\beta_3}(\tilde s,\tilde x_e,\tilde \kappa), \quad t\in\Pi(0,\tilde s_G), \\
	\dot{\tilde \beta}_2    &= vg_v(\tilde\beta_2+\beta_{2,r},\tilde\beta_3+\beta_{3,r},\tilde\kappa+\kappa_r)f_{\tilde\beta_2}(\tilde s,\tilde x_e,\tilde \kappa), \quad t\in\Pi(0,\tilde s_G),
	\end{align}
\end{subequations}
where the origin is still an equilibrium point since $f_{\tilde\theta_3}(\tilde s,0,0)=f_{\tilde\beta_3}(\tilde s,0,0)=f_{\tilde\beta_2}(\tilde s,0,0)=0$, $\forall \tilde s\in[0,\tilde s_G]$.

\subsubsection*{Acknowledgments}
The research leading to these results has been founded by Strategic vehicle research and innovation (FFI). 
The authors gratefully acknowledge the Royal Institute of Technology for providing the external RTK-GPS. 
The authors acknowledge the anonymous reviewers for their valuable suggestions and critical comments which have significantly improved the contents of this paper. 
The authors would also like to express their gratitude to Scania CV for providing necessary hardware, as well as software and technical support.

\bibliography{root}
\bibliographystyle{abbrv}

\end{document}